\documentclass{article}

\usepackage[preprint]{neurips_2025}

\usepackage{microtype}
\usepackage{graphicx}
\usepackage{booktabs} % for professional tables
\usepackage{url}
\usepackage{subcaption}
\usepackage[utf8]{inputenc} % allow utf-8 input
\usepackage[T1]{fontenc}    % use 8-bit T1 fonts
\usepackage{amsfonts}       % blackboard math symbols
\usepackage{nicefrac}       % compact symbols for 1/2, etc.
\usepackage{microtype}      % microtypography
\usepackage{xcolor}         % colors

\usepackage[font=normalsize,labelfont=bf]{caption}
\usepackage{natbib}
\usepackage{textcomp}
\usepackage{setspace}
\usepackage{multirow}
\usepackage{array}

\newcommand{\thickhat}[1]{\mathbf{\hat{\text{$#1$}}}}
\newcommand{\thickbar}[1]{\mathbf{\bar{\text{$#1$}}}}

\newcommand{\thicktilde}[1]{\mathbf{\tilde{\text{$#1$}}}}

\let\svthefootnote\thefootnote
\newcommand\blankfootnote[1]{%
  \let\thefootnote\relax\footnotetext{#1}%
  \let\thefootnote\svthefootnote%
}

\usepackage{hyperref}
\usepackage{subfiles}

\usepackage{amsmath}
\usepackage{amssymb}
\usepackage{mathtools}
\usepackage{amsthm}
\usepackage{algpseudocode}
\usepackage[linesnumbered,ruled,vlined]{algorithm2e}

\usepackage[capitalize,noabbrev]{cleveref}

\theoremstyle{plain}
\newtheorem{theorem}{Theorem}[section]

\theoremstyle{definition}
\newtheorem{definition}[theorem]{Definition}

\theoremstyle{remark}

\usepackage[textsize=tiny]{todonotes}

\newcommand{\ols}[1]{\mskip.5\thinmuskip\overline{\mskip-.5\thinmuskip {#1} \mskip-.5\thinmuskip}\mskip.5\thinmuskip} % overline short
\newcommand{\olsi}[1]{\,\overline{\!{#1}}} % overline short italic
\makeatletter
\newcommand\closure[1]{
  \tctestifnum{\count@stringtoks{#1}>1} %checks if number of chars in arg > 1 (including '\')
  {\ols{#1}} %if arg is longer than just one char, e.g. \mathbb{Q}, \mathbb{F},...
  {\olsi{#1}} %if arg is just one char, e.g. K, L,...
}
\long\def\count@stringtoks#1{\tc@earg\count@toks{\string#1}}
\long\def\count@toks#1{\the\numexpr-1\count@@toks#1.\tc@endcnt}
\long\def\count@@toks#1#2\tc@endcnt{+1\tc@ifempty{#2}{\relax}{\count@@toks#2\tc@endcnt}}
\def\tc@ifempty#1{\tc@testxifx{\expandafter\relax\detokenize{#1}\relax}}
\long\def\tc@earg#1#2{\expandafter#1\expandafter{#2}}
\long\def\tctestifnum#1{\tctestifcon{\ifnum#1\relax}}
\long\def\tctestifcon#1{#1\expandafter\tc@exfirst\else\expandafter\tc@exsecond\fi}
\long\def\tc@testxifx{\tc@earg\tctestifx}
\long\def\tctestifx#1{\tctestifcon{\ifx#1}}
\long\def\tc@exfirst#1#2{#1}
\long\def\tc@exsecond#1#2{#2}

\begin{document}

% Include the subfiles

\title{RASPNet: A Benchmark Dataset for Radar Adaptive Signal Processing Applications}

% The \author macro works with any number of authors. There are two commands
% used to separate the names and addresses of multiple authors: \And and \AND.
%
% Using \And between authors leaves it to LaTeX to determine where to break the
% lines. Using \AND forces a line break at that point. So, if LaTeX puts 3 of 4
% authors names on the first line, and the last on the second line, try using
% \AND instead of \And before the third author name.

\author{%
Shyam Venkatasubramanian \\
Duke University\\
\And
Bosung Kang \\
Air Force Research Laboratory \\
\And % Use \AND to break to the next line of authors
Ali Pezeshki \\
Colorado State University \\
\And
Muralidhar Rangaswamy \\
Air Force Research Laboratory \\
\And
Vahid Tarokh \\
Duke University \\
}

\maketitle
\blankfootnote{GitHub Repository: \url{https://github.com/shyamven/RASPNet}.}
\begin{abstract}
We present a large-scale dataset called \textit{RASPNet} for radar adaptive signal processing (RASP) applications to support the development of data-driven models within the adaptive radar community. RASPNet exceeds 16 TB in size and comprises 100 realistic scenarios compiled over a variety of topographies and land types across the contiguous United States. For each scenario, RASPNet comprises 10,000 clutter realizations from an airborne radar setting, which can be used to benchmark radar and complex-valued learning algorithms. RASPNet intends to fill a prominent gap in the availability of a large-scale, realistic dataset that standardizes the evaluation of RASP techniques and complex-valued neural networks. We outline its construction, organization, and several applications, including a transfer learning example to demonstrate how RASPNet can be used for real-world adaptive radar scenarios.
\end{abstract}

\addtocontents{toc}{\protect\setcounter{tocdepth}{0}}
\section{Introduction}
The success of radar technologies over the past century has been characterized by landmark advances in radar adaptive signal processing (RASP). Within RASP, Space-Time Adaptive Processing (STAP) has been instrumental in achieving order-of-magnitude improvements in detection and localization performance, especially in environments dominated by clutter and noise, which has revolutionized modern radar systems \citep{brennan1973adaptive}. For decades, STAP has been integral to advancing surveillance, defense, and navigation \citep{Ward1995stap, Li2008}. The foundations of RASP, realized using STAP, lies in the adaptive combination of spatial and temporal processing to optimize radar response to signals \citep{ward1998space, Guerci2000, Melvin2004, wicks2006space}.

Despite the theoretical advancements in RASP, the development, testing, and validation of RASP algorithms remains a significant challenge. This difficulty arises from the complexity of the real-world scenarios in which radar systems operate, which are characterized by dynamic and unpredictable environmental conditions. Theoretical models cannot capture the full range of operational scenarios, limiting the effectiveness of RASP algorithms developed in a vacuum \citep{Klemm2002, Melvin2004}.

Paralleling the evolution of adaptive radar, complex-valued neural networks (CVNNs) have emerged as promising tools for enhancing signal processing capabilities \citep{hirose2006complex, trabelsi2017deep, wu2023complex}. While CVNNs are well-suited for handling the inherently complex-valued data produced by radar systems, benchmarking CVNNs is limited by a lack of comprehensive datasets with naturally occurring complex-valued features \citep{asiyabi2023complex}. Practitioners often convert real-valued datasets to the frequency domain \citep{trabelsi2017deep, scardapane2020complex}, which fails to capture the nuanced characteristics of real-world complex-valued data.

The need for large-scale, comprehensive, realistic datasets in the adaptive radar community cannot be overstated. High-quality datasets serve critical functions in the development of RASP algorithms. Firstly, they provide a benchmark for evaluating and comparing the performance of algorithms under conditions that mirror real-world operational environments. Furthermore, these datasets facilitate the development of data-driven approaches that leverage machine learning \citep{hara1994application, jiang2022artificial, venkatasubramanian2024data}. The Knowledge Aided Sensor Signal Processing and Expert Reasoning (KASSPER) dataset represents a seminal effort in this direction \citep{guerci2006kassper}, offering a focused repository of radar data that has been significant in advancing the field \citep{page2004improving, blunt2006stap, stoica2008using, wang2009new}. 

In spite of the value provided by existing repositories like KASSPER, the adaptive radar community faces significant data-related challenges. KASSPER, for instance, comprises one 5 MB radar scenario, which is inadequate for data-driven algorithms. Thus, the primary issue is the absence of a publicly available dataset that encompasses the diversity of scenarios that adaptive radar systems encounter, as existing datasets are either too narrow in scope or have restricted access \citep{gogineni2022clutter}.

To address these challenges, we present \textit{RASPNet}, a large-scale, benchmark dataset for radar adaptive signal processing applications. Eclipsing $16$ TB in size, RASPNet comprises a comprehensive set of $100$ radar scenarios from across the United States, compiled over various environmental conditions and land types via the RFView\textsuperscript{\tiny\textregistered} modeling software. For each scenario, we provide $10{,}000$ realizations of clutter returns gathered with an airborne radar platform. RASPNet supports both traditional RASP algorithms and emerging data-driven approaches, facilitating a concrete benchmark to compare the performance of adaptive radar processing methods, and providing a critical testbed for CVNNs.

The outline of this paper is as follows. In Section \ref{sec:Dataset_properties}, we discuss the properties of RASPNet, including the radar parameters used to generate the dataset, the $100$ radar scenarios in the dataset, and validate RFView\textsuperscript{\tiny\textregistered} versus measured data. In Section \ref{sec:Dataset_organization}, we describe the organization of RASPNet via the energy distance metric. In Section \ref{sec:Dataset_applications}, we delineate several applications of RASPNet for benchmarking target localization accuracy, benchmarking complex-valued learning algorithms, and for transfer learning in realistic adaptive radar processing scenarios. In particular, we propose a convolutional transformer architecture, termed the \textit{Adaptive Radar Transformer}, which achieves benchmark target localization accuracy on RASPNet, and we extend the Steinmetz network construction from \citep{venkatasubramanian2024steinmetz} to propose the \textit{Convolutional Steinmetz Network}, which achieves benchmark performance on the RASPNet CVNN task. Finally, in Section \ref{sec:Conclusion}, we summarize RASPNet and discuss future work. RASPNet is accessible at: \href{https://www.sdms.afrl.af.mil/index.php?collection=raspnet}{https://www.sdms.afrl.af.mil/index.php?collection=raspnet}.

\section{Dataset Properties} \label{sec:Dataset_properties}
We first describe the procedure used to generate the $100$ scenarios comprising RASPNet and detail the intuition behind our scenario choices, subject to empirical validation.

\subsection{Radar Platform Parameters} \label{sec:Platform_parameters}
To obtain the data for each scenario comprising RASPNet, we make use of the RFView\textsuperscript{\tiny\textregistered} modeling and simulation tool. This physics-based, RF modeling and simulation platform integrates a global database of terrain and land cover for RF ray tracing simulations. RFView\textsuperscript{\tiny\textregistered} has been compared with measured datasets from VHF to K\textsubscript{a} band \citep{gogineni2022clutter}. We further validate RFView\textsuperscript{\tiny\textregistered} versus the MCARM measured dataset \citep{babu1996processing} in Section \ref{sec:Measured_data_validation}.

The scenarios we consider simulate an airborne radar surveying various regions across the United States. For simplicity, the radar was assumed as static during the simulation. The complete list of radar platform parameters is provided in Table \ref{tab:radar_parameters_complete} of the Supplementary Information.
% \begin{table*}[h!]
% \caption{Abbreviated List of Radar Platform Parameters}
% \label{tab:radar_parameters}
% \centering
% \begin{tabular}{l|l}
% \hline
% \textbf{Parameter} & \textbf{Value} \\
% \hline
% Carrier frequency $(f_c)$, Bandwidth $(B)$, PRF $(f_{PR})$ & $10\,\text{GHz}$, $5\,\text{MHz}$, $1100\,\text{Hz}$ \\
% Platform altitude $(h)$, Range resolution $(\Delta r)$ & $1000\,\text{m}$, $30\,\text{m}$\\
% Antenna configuration (horizontal $\times$ vertical) & $48 \times 5$ \\
% Antenna element spacing $(\Delta k)$ & $0.015\,\text{m}$ \\
% Number of pulses $(\Lambda)$, channels $(L)$, and range bins $(G)$ & $64$, $16$, and $1134$ \\
% \end{tabular}
% \end{table*}

We position the radar platform at a preset, scenario-specific, latitude and longitude location, utilizing the parameters from Table \ref{tab:radar_parameters_complete}. The radar platform aims $20$ km to the southwest and operates in spotlight mode. The clutter return is beamformed for each size $(\frac{48}{L} \times 5)$ receiver sub-array, condensing the array to size $(L \times 1)$. Formally, let $S = \{1,2,\ldots,M\}$ denote the set of scenario indices, where $M$ is the total number of scenarios within the dataset. We denote $Z^i \in \mathbb{C}^{\Lambda L G}$ as the random variable describing the clutter for scenario $i \in S$. Now, consider $z_j^i \sim Z^i$, which denotes the clutter return for the $j$th independent realization for scenario $i$. To obtain the clutter dataset for scenario $i$, we obtain $K$ i.i.d. realizations to form $\{z^i_j, \, \forall j \in D \}$, where $D = \{1,2,\ldots,K\}$.

\subsection{Radar Scenarios} \label{sec:Radar_scenarios}
\begin{figure}[h!]
    \centering
    \captionsetup{justification=centering}
    \includegraphics[width=0.9\linewidth]{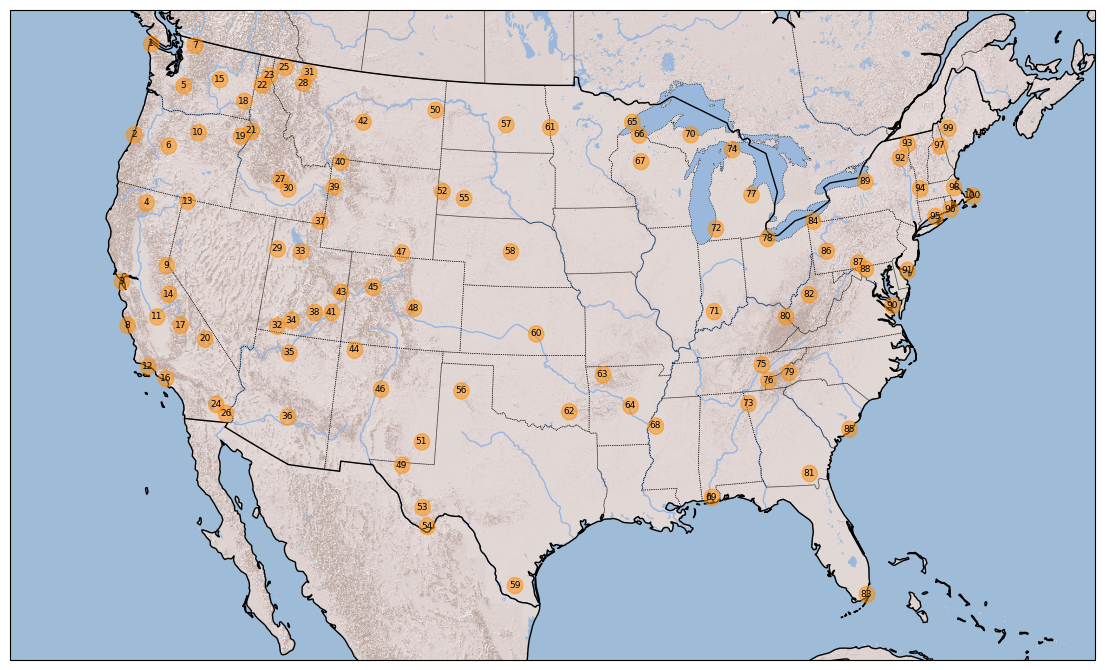}
    \caption{Mapping the $M = 100$ scenarios within RASPNet. Each scenario is indexed by $i \in S$.}
    \label{fig:dataset_scenarios}
\end{figure}

RASPNet comprises $M = 100$ hand-picked scenarios from the contiguous United States, where the clutter dataset for each scenario, $\{z^i_j, \, \forall j \in D\}$ with $i \in S$, is obtained via the procedure described in Section \ref{sec:Platform_parameters}. The selection criteria for these scenarios prioritize geographical diversity, focusing on varied landforms and topographies that present distinct challenges for radar systems. The geographic coverage of RASPNet scenarios across the contiguous United States is depicted in Figure \ref{fig:dataset_scenarios}, wherein the indices, $i \in S$, are ordered from the westernmost to easternmost scenario by default.

To evaluate the coverage of RASPNet scenarios, we empirically verify the exemplar nature of our $M$ hand-picked scenarios in Section \ref{sec:scenario_validation}. The complete tabulation of scenarios comprising RASPNet, along with short descriptions, is provided in Section \ref{sec:raspnet_scenario_discussion} of the Supplementary Information.

\subsection{Exemplar Scenario Validation} \label{sec:scenario_validation}
To determine whether our $100$ hand-picked scenarios are exemplar, we compare their geographic coverage with that of a collection of $100$ scenarios obtained through the Partitioning Around Medoids (PAM) algorithm \citep{kaufman1990PAM}, from $500$ total scenarios uniformly selected at random across the United States. We denote $\smash{\closure{S} = \{1,2,\ldots,\closure{M} \}}$ as the set of indices pertaining to the $\smash{\closure{M} = 500}$ scenarios, and $\smash{S^* \subset \closure{S}}$ as the set of indices yielded by PAM, where $\left|S^*\right| = M^* = 100$.

\begin{figure}[hbt!]
  \centering
    \begin{minipage}{.83\linewidth}
    \begin{algorithm}[H] \label{alg:pam}
    \SetAlgoLined
    \SetKwInput{KwInput}{Input}
    \SetKwInput{KwOutput}{Output}
    \SetKwInput{KwData}{Data}
    \SetKwFor{For}{for}{do}{end}
    \SetKwIF{If}{ElseIf}{Else}{if}{then}{else if}{else}{end}
    \SetKw{Return}{return}
    \KwInput{$\closure{S}$ (set of indices), $M^*$ (number of representative indices), $T$ (max steps)}
    \KwOutput{$S^*$ (set of representative indices obtained through PAM)}
    
    $E \gets \mathbf{0}_{\closure{M} \times \closure{M}}$ (Initialize the matrix of $\mathcal{E}$-statistics) \\
    \For{$i = 1$ \KwTo $\closure{M}$}{
        \lFor{$k = i$ \KwTo $\closure{M}$}{
            $E_{i,k} \gets \mathcal{E}(Z^i, Z^k)$; $E_{k,i} \gets \mathcal{E}(Z^i, Z^k)$
        }
    }
    $S^* \gets \text{SRSWOR}(\closure{S},M^*)$; $C \gets \sum_{i \in \closure{S}} \min_{k \in S^*} E_{i,k}$; \\
    \For{$t = 1$ \KwTo $T$}{
        \For{$m \in S^*$}{
            \For{$o \in \closure{S} \setminus S^*$}{
                $S'^* \gets (S^* \setminus \{m\}) \cup \{o\}$; $C' \gets \sum_{i \in \closure{S}} \min_{k \in S'^*} E_{i,k}$; \\
                \lIf{$C' < C$}{
                    $S^* \gets S'^*$; $C \gets C'$
                }
            }
        }
    }
    \Return{$S^*$};
    \caption{Partitioning Around Medoids (PAM) with the $\mathcal{E}$-statistic}
    \end{algorithm}
    \end{minipage}
\end{figure}

To determine $S^*$ using the PAM algorithm, we employ the energy statistic ($\mathcal{E}$-statistic) as our distance metric \citep{szekely2013energy}. The $\mathcal{E}$-statistic characterizes the statistical distance between the distributions of random variables by aggregating pairwise sample distances, capturing differences in their distributional characteristics. Rooted in potential energy principles, the $\mathcal{E}$-statistic facilitates robust comparisons of complex, multivariate datasets beyond location or scale. The formal definition of the $\mathcal{E}$-statistic for our analysis is provided in Definition \ref{def:energy_distance}.
\begin{definition} \label{def:energy_distance}
Consider clutter random variables $Z^i, Z^k \in \mathbb{C}^{\Lambda L G}$, for $i,k \in \closure{S}$, wherein we have the clutter datasets $\{z^i_j, \, \forall j \in D\}$, $\{z^k_l, \, \forall l \in D\}$. The $\mathcal{E}$-statistic is given by:
\begin{align}
    \mathcal{E}(Z^i,Z^k) = \frac{2}{K^2} \sum_{j=1}^{K}\sum_{l=1}^{K} \| z^i_j - z^k_l \| - \frac{1}{K^2} \sum_{j=1}^{K}\sum_{l=1}^{K} \| z^i_j - z^i_l \| - \frac{1}{K^2} \sum_{j=1}^{K}\sum_{l=1}^{K} \| z^k_j - z^k_l \|.
\end{align}
\end{definition}

We now outline the PAM algorithm for our analysis. We first construct the $\smash{\closure{M} \times \closure{M}}$ $\mathcal{E}$-statistic matrix, $E$, for $\smash{\closure{M}=500}$, where each element, $E_{i,k}$, is precomputed using $\mathcal{E}(Z^i, Z^k)$ for every pair $\smash{i, k \in \closure{S}}$. This precomputation step facilitates efficient distance lookups during the PAM algorithm's execution. Over $T$ steps, the PAM algorithm iteratively refines a subset of $M^* = 100$ representative scenarios, $S^*$, through a process of swapping current medoids with non-medoids to minimize the total cost, $C$, where $C$ is computed as the sum of distances from each scenario to its nearest medoid, based on the precomputed $\mathcal{E}$-statistic matrix, $E$. This procedure is summarized in Algorithm \ref{alg:pam}. The $500$ total scenarios and the $100$ representative scenarios from PAM are depicted in Section \ref{sec:scenario_validation_appendix} of the Appendix.

We observe that the geographic coverage of the $100$ hand-picked scenarios in RASPNet closely mirrors the coverage of the $100$ representative scenarios determined through PAM. A large concentration of these scenarios are in the Western United States --- this region is characterized by a wider variety of terrain types, which introduces greater variability in the distribution of geographic features, leading to a greater concentration of representative scenarios. We also note that the $100$ hand-picked scenarios in RASPNet are more dispersed across the contiguous United States by construction, as they encompass identifiable geographical landmarks representing a broad spectrum of environmental conditions. %A greater emphasis was also placed on incorporating land-to-water transition zones in the $100$ hand-picked scenarios, acknowledging these as inherently more varied scenarios.

\subsection{Validation Against Measured Data} \label{sec:Measured_data_validation}
To evaluate the realistic nature of the RFView\textsuperscript{\tiny\textregistered} simulator, we compare it with the MCARM measured dataset, focusing on Flight \#5 Acquisition 575 \citep{babu1996processing}. The MCARM radar parameters are tabulated in Section \ref{sec:mcarm_radar_platform_parameters} of the Appendix. Let $\smash{Z^i \in \mathbb{C}^{\tilde{\Lambda} \tilde{L} \tilde{G}}}$, for $\smash{i \in \hat{S} = \{\text{MCM, RF, BE} \}}$, denote the clutter random variables for the MCARM, RFView\textsuperscript{\tiny\textregistered}, and bald-earth (BE) models, wherein the bald-earth model is a more classical clutter model without terrain heights. For the RFView\textsuperscript{\tiny\textregistered} and BE models, we generate the clutter datasets $\smash{\{z^{\text{RF}}_j, \, \forall j \in D \}}$ and $\smash{\{z^{\text{BE}}_j, \, \forall j \in D \}}$, which we then reshape to obtain $\smash{\mathbf{Z}^{\text{RF}}, \mathbf{Z}^{\text{BE}} \in \mathbb{C}^{\tilde{\Lambda} \tilde{L} \tilde{G} \times K}}$. The MCARM Flight \#5 Acquisition 575 provides a single clutter realization, $\smash{z^{\text{MCM}}_1}$, which we reshape to form $\smash{\mathbf{Z}^{\text{MCM}} \in \mathbb{C}^{\tilde{\Lambda} \tilde{L} \tilde{G} \times 1}}$.

\begin{figure*}[t!]
    \centering
    \begin{subfigure}{0.32\textwidth}
        \includegraphics[width=\textwidth]{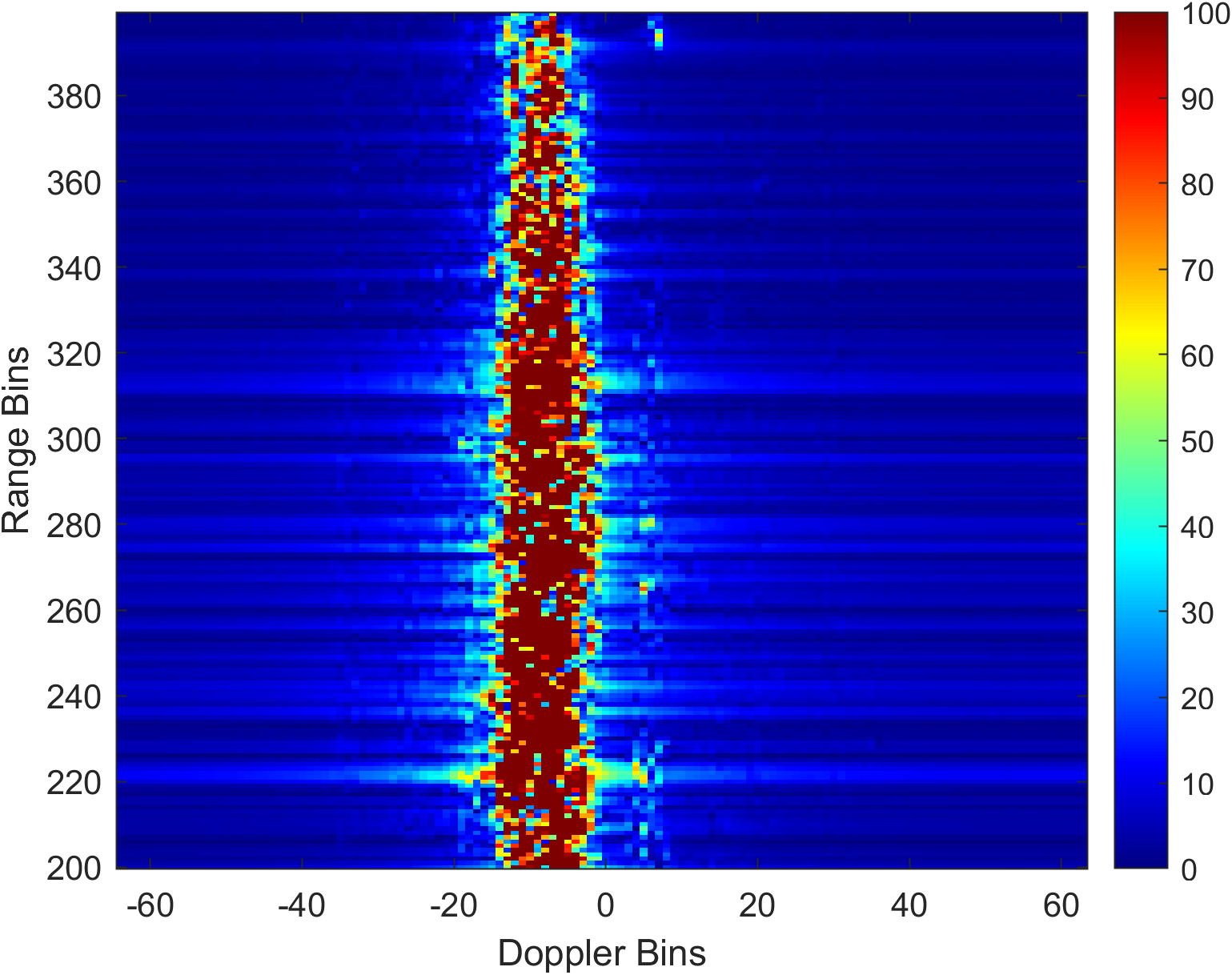}
        \caption{MCARM Measured Data}
    \end{subfigure}
    \begin{subfigure}{0.32\textwidth}
        \includegraphics[width=\textwidth]{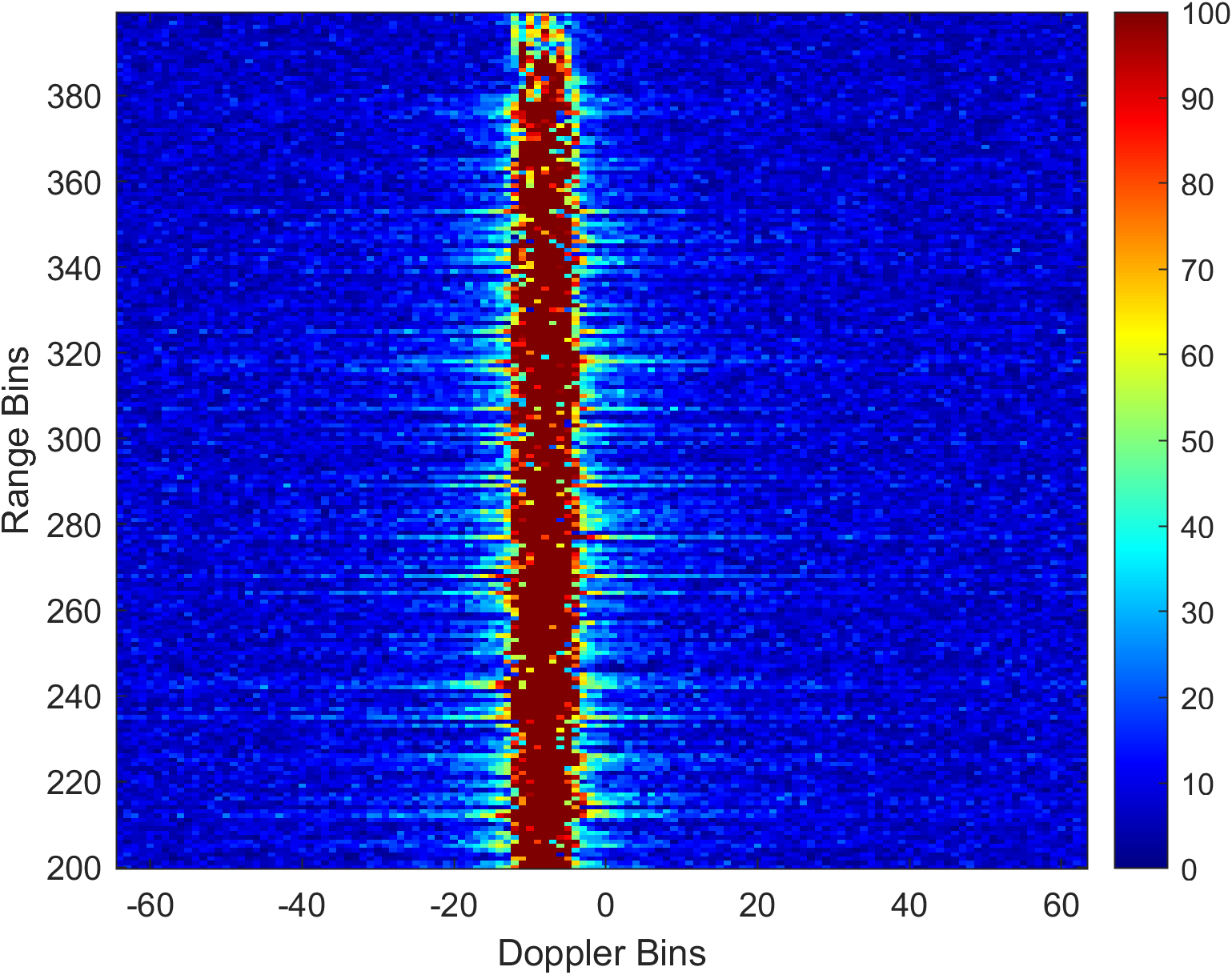}
        \caption{RFView\textsuperscript{\tiny\textregistered} Simulated Data}
    \end{subfigure}
    \begin{subfigure}{0.32\textwidth}
        \includegraphics[width=\textwidth]{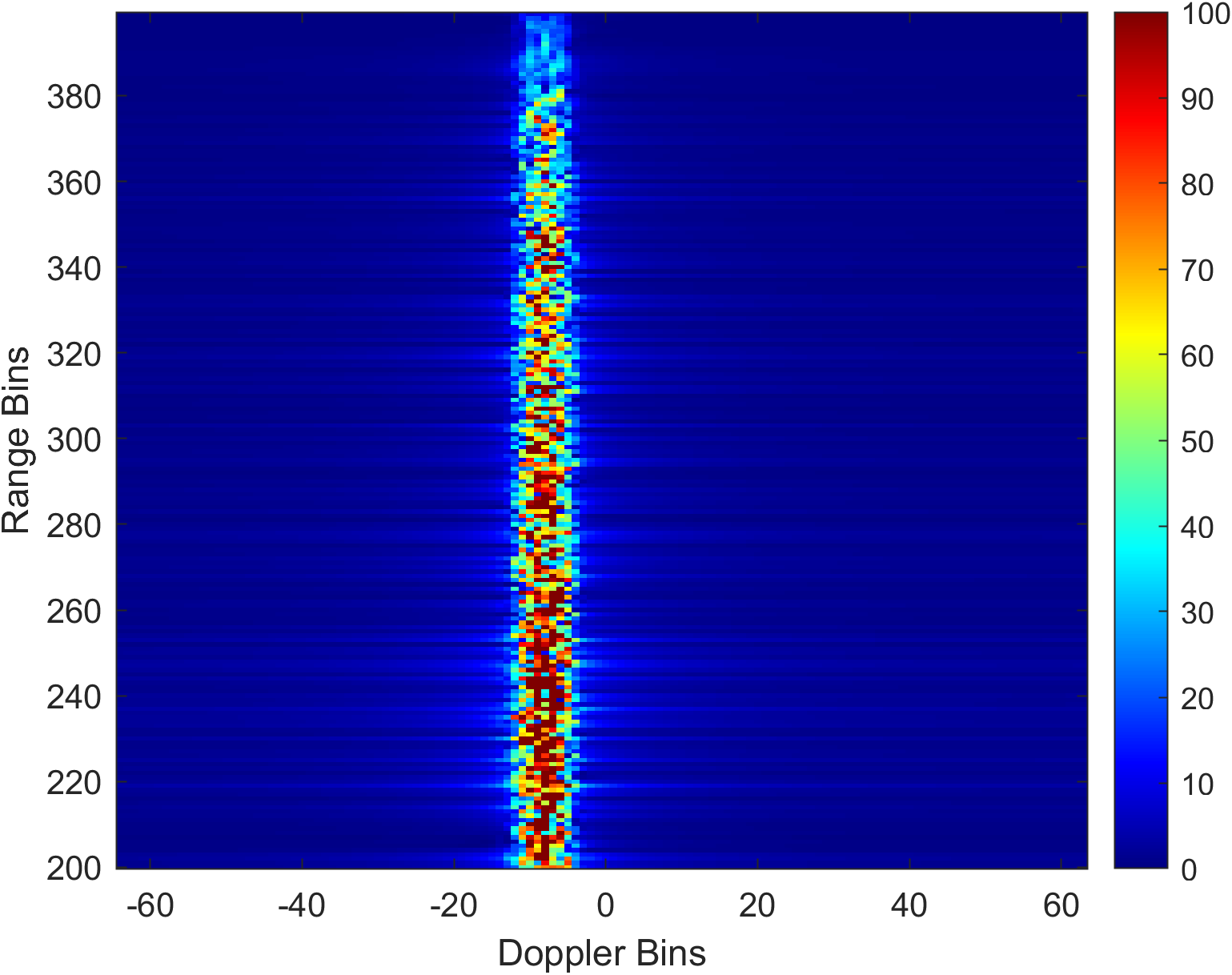}
        \caption{Bald-Earth Model}
    \end{subfigure}
    \caption{Range-Doppler plots for (a) Flight \#5 Acquisition 575 from the MCARM measured dataset, (b) RFView\textsuperscript{\tiny\textregistered} simulated data, and (c) classical bald-earth model without terrain heights. The colormap depicts the magnitude squared of the radar range-Doppler spectrum (red is higher and blue is lower).}
    \label{fig:range_doppler_plots}
\end{figure*}

We produce range-Doppler plots for both the RFView\textsuperscript{\tiny\textregistered} simulated data and the MCARM measured data by performing Doppler processing via the Fast Fourier Transform (FFT), coherently integrating across antenna elements. The range-Doppler plots for the MCARM, RFView\textsuperscript{\tiny\textregistered}, and BE models are shown in Figure \ref{fig:range_doppler_plots}. We clearly observe that the RFView-generated data exhibits a higher degree of similarity to the MCARM measured data than the classical BE model.

To validate the similarity between the RFView\textsuperscript{\tiny\textregistered} simulated data and the MCARM measured data, we perform a modified version of the likelihood-based test from \citep{johnson2010doa}. Let $\smash{\mathbf{Z}^{\text{MCM}}[l,g] \in \mathbb{C}^{\tilde{L} \times 1}}$, $\smash{\mathbf{Z}^{\text{RF}}[l,g]}$, and $\smash{\mathbf{Z}^{\text{BE}}[l,g] \in \mathbb{C}^{\tilde{L} \times K}}$ form the MCARM, RFView\textsuperscript{\tiny\textregistered}, and bald-earth clutter realizations, for the $l$-th Doppler bin and $g$-th range bin. The null hypothesis, $H_0$, asserts that $\mathbf{Z}^{\text{MCM}}[l,g]$ is from the same distribution as $\mathbf{Z}^{\text{RF}}[l,g]$ --- we repeat this procedure for $\mathbf{Z}^{\text{BE}}[l,g]$). Next, $\forall l, g$, we compare the likelihood of the MCARM clutter realization to the empirical distribution of likelihoods from the RFView\textsuperscript{\tiny\textregistered} clutter realizations. We let $\mathbf{\Sigma}_{\text{RF}}[l,g]$ denote the sample covariance matrix for the RFView case, such that the likelihood of the MCARM data given $\mathbf{\Sigma}_{\text{RF}}[l,g]$, and the likelihood of the $j^{\text{th}}$ RFView clutter realization given $\mathbf{\Sigma}_{\text{RF}}[l,g]$, are given by:
\begin{align}
    \mathcal{L}_{\text{MCM}}[l,g] = \frac{e^{\left(-\frac{1}{2} \mathbf{Z}^{\text{MCM}}[l,g]^H \boldsymbol{\Sigma}_{\text{RF}}[l,g]^{-1} \mathbf{Z}^{\text{MCM}}[l,g] \right)}}{\sqrt{(2\pi)^L \det(\boldsymbol{\Sigma}_{\text{RF}}[l,g])}}, \ \mathcal{L}_{\text{RF},j}[l,g] = \frac{e^{\left( -\frac{1}{2} \mathbf{Z}_j^{\text{RF}}[l,g]^H \boldsymbol{\Sigma}_{\text{RF}}[l,g]^{-1} \mathbf{Z}_j^{\text{RF}}[l,g] \right)}}{\sqrt{(2\pi)^L \det(\boldsymbol{\Sigma}_{\text{RF}}[l,g])}}.
\end{align}
To determine if the real-world data is consistent with the simulated data distribution, we compare $\mathcal{L}_{\text{MCM}}[l,g]$ to the empirical distribution of $\{ \mathcal{L}_{\text{RF},j}[l,g], \ \forall j \in D \}$. We form the $95\%$ confidence interval from the percentiles of $\mathcal{L}_{\text{RF},j}[l,g]$. If $\mathcal{L}_{\text{MCM}}[l,g]$ falls within this confidence interval, we fail to reject $H_0$. We compute the total fraction of how often the MCARM likelihood falls in the confidence interval over all Doppler and range bins, denoted $\rho \in [0,1]$:
\begin{align} \label{eq:fraction_failed_rejection}
    \rho = \frac{1}{\Lambda G} \sum_{l=1}^{\Lambda} \sum_{g=1}^{G} \mathbb{I}\big( \mathcal{L}_{\text{MCM}}[l,g] \in \big[ \text{\textperthousand}_{2.5}(\mathcal{L}_{\text{RF}}[l,g]), \text{\textperthousand}_{97.5}(\mathcal{L}_{\text{RF}}[l,g]) \big] \big).
\end{align}
In the RFView\textsuperscript{\tiny\textregistered} case ($\mathbf{Z}^{\text{RF}}[l,g]$), we obtain $\rho = 0.1969$ for the real part of $\mathcal{L}_{\text{MCM}}[l,g]$, $\rho = 0.436$ for the imaginary part of $\mathcal{L}_{\text{MCM}}[l,g]$. In the bald-earth case ($\mathbf{Z}^{\text{BE}}[l,g]$), we obtain $\rho = 0$ for both the real and imaginary parts of $\mathcal{L}_{\text{MCM}}[l,g]$. This study shows that while RFView\textsuperscript{\tiny\textregistered} does not offer a complete match with the real-world data, it is more realistic than the bald-earth model. Furthermore, this study suggests that for real-world applications, it is feasible to train networks on RASPNet scenarios, and fine tune them using real-world samples. We explore this further in Section \ref{sec:Transfer_learning}.

\section{Dataset Organization} \label{sec:Dataset_organization}
The organization of RASPNet is guided by the premise of ranking scenarios based on their difficulty for RASP algorithms, which we quantify using the $\mathcal{E}$-statistic. We employ this metric to systematically order the $M$ scenarios from the most similar to the baseline scenario, defined as index $i = 29$ from $S$ (the Bonneville Salt Flats), to the least similar, as a proxy for difficulty. This baseline scenario typifies an environment that has minimal variability in elevation and a uniform land type --- characteristics that typically simplify adaptive processing tasks. We denote $Q = (q_i)_{i=1}^{M}$ as the sequence of ordered indices, and let $F$ be an $M$-dimensional array with elements $\mathcal{E}(Z^i, Z^b)$, where $i \in S$, and $b$ is the index of the baseline scenario. Thus, the $i^{\text{th}}$ scenario corresponds to the $q_i^{\text{th}}$ `easiest' scenario, $Z^{q_i}$, in RASPNet. We summarize this ordering procedure in Algorithm \ref{alg:org}.
\begin{figure}[h!]
  \centering
    \begin{minipage}{.79\linewidth}
    \begin{algorithm}[H] \label{alg:org}
    \SetAlgoLined
    \SetKwInput{KwInput}{Input}
    \SetKwInput{KwOutput}{Output}
    \SetKwInput{KwData}{Data}
    \SetKwFor{For}{for}{do}{end}
    \SetKwIF{If}{ElseIf}{Else}{if}{then}{else if}{else}{end}
    \SetKw{Return}{return}
    \KwInput{$S$ (set of indices), $b$ (index of baseline scenario)}
    \KwOutput{$Q$ (sequence of indices ordered by $\mathcal{E}(Z^i, Z^b)$)}
    $F \gets \mathbf{0}_{M}$ (Initialize the array of $\mathcal{E}$-statistics) \\
    \lFor{$i \in S$}{
        $F_i \gets \mathcal{E}(Z^i, Z^b)$
    }
    $Q \gets \text{argsort}(F, \leq)$ (Sort the indices of $F$ in ascending order) \\
    \Return{$Q$}\
    \caption{Ordering RASPNet Scenarios by Difficulty with the $\mathcal{E}$-statistic}
    \end{algorithm}
    \end{minipage}
\end{figure}

The rationale behind using the $\mathcal{E}$-statistic with respect to the Bonneville Salt Flats as a benchmark is its inherent simplicity for RASP algorithms. Scenarios with lower $\mathcal{E}$-values relative to this baseline are deemed less challenging, as their clutter characteristics more closely resemble those of an environment with low variability and uniformity, simplifying the radar's task of distinguishing between targets and clutter. Conversely, scenarios with higher $\mathcal{E}$-values are deemed more challenging as they indicate nonuniform environments, requiring advanced radar processing algorithms. 

\begin{figure*}[h!]
    \centering
    \begin{subfigure}{0.4\textwidth}
        \includegraphics[width=\textwidth]{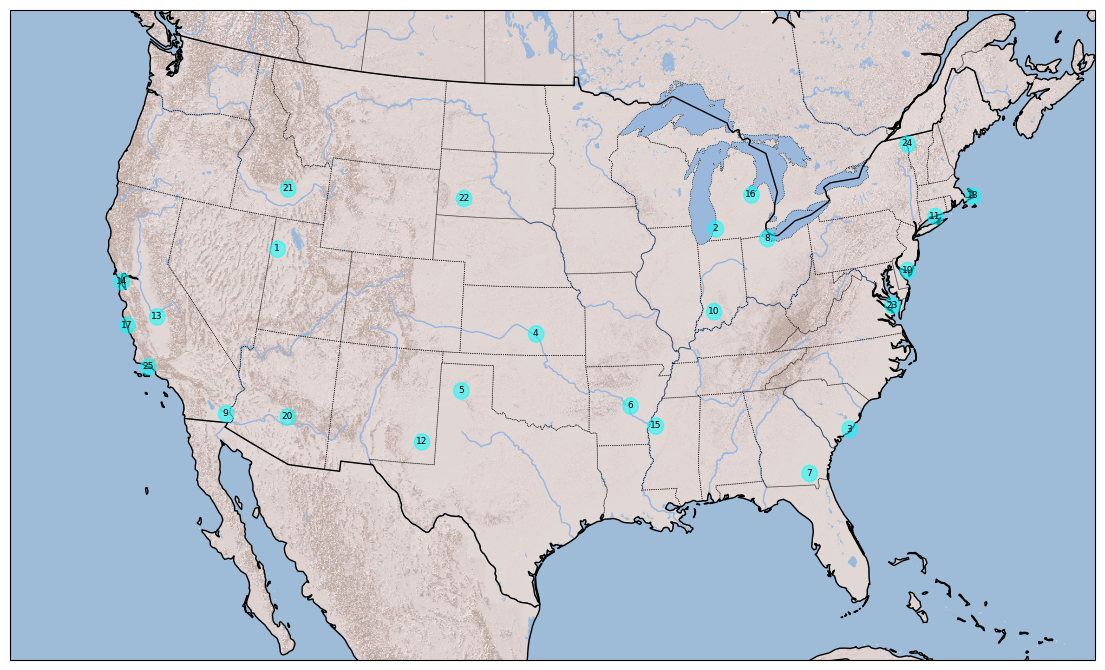}
        \caption{Basic Scenarios}
    \end{subfigure}
    \begin{subfigure}{0.4\textwidth}
        \includegraphics[width=\textwidth]{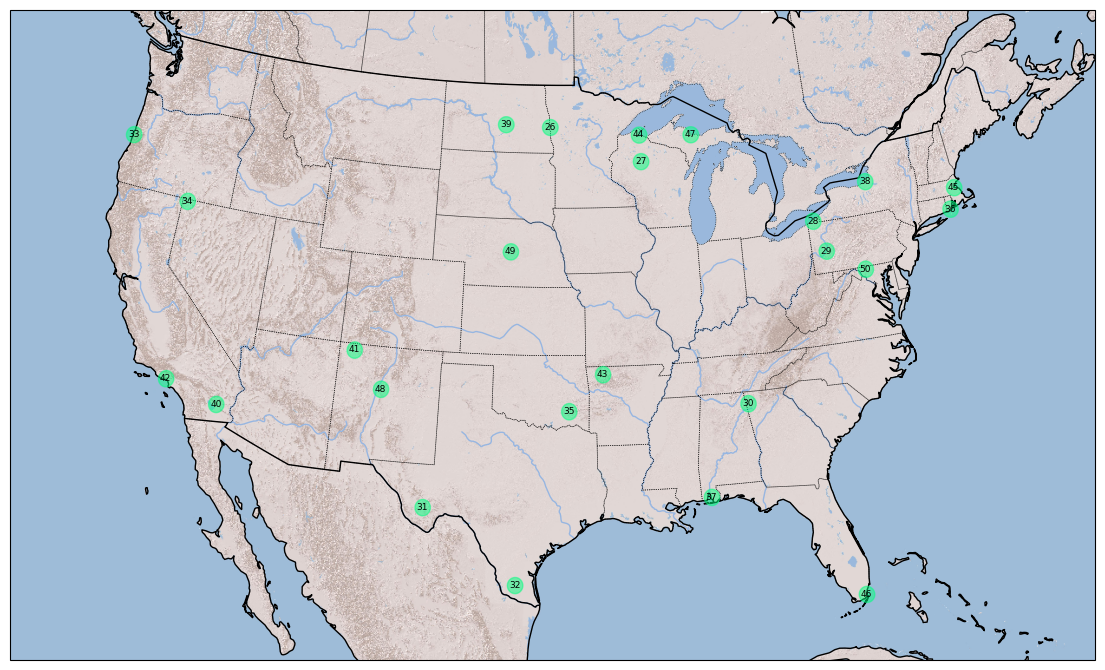}
        \caption{Intermediate Scenarios}
    \end{subfigure}
\\
    \begin{subfigure}{0.4\textwidth}
        \includegraphics[width=\textwidth]{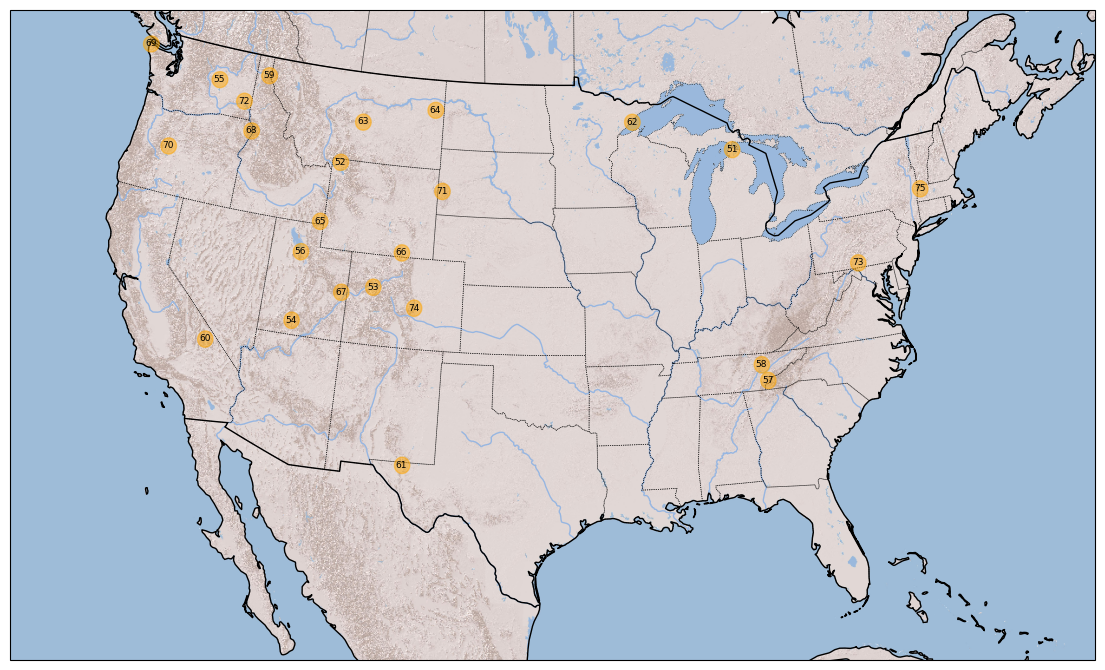}
        \caption{Challenging Scenarios}
    \end{subfigure}
    \begin{subfigure}{0.4\textwidth}
        \includegraphics[width=\textwidth]{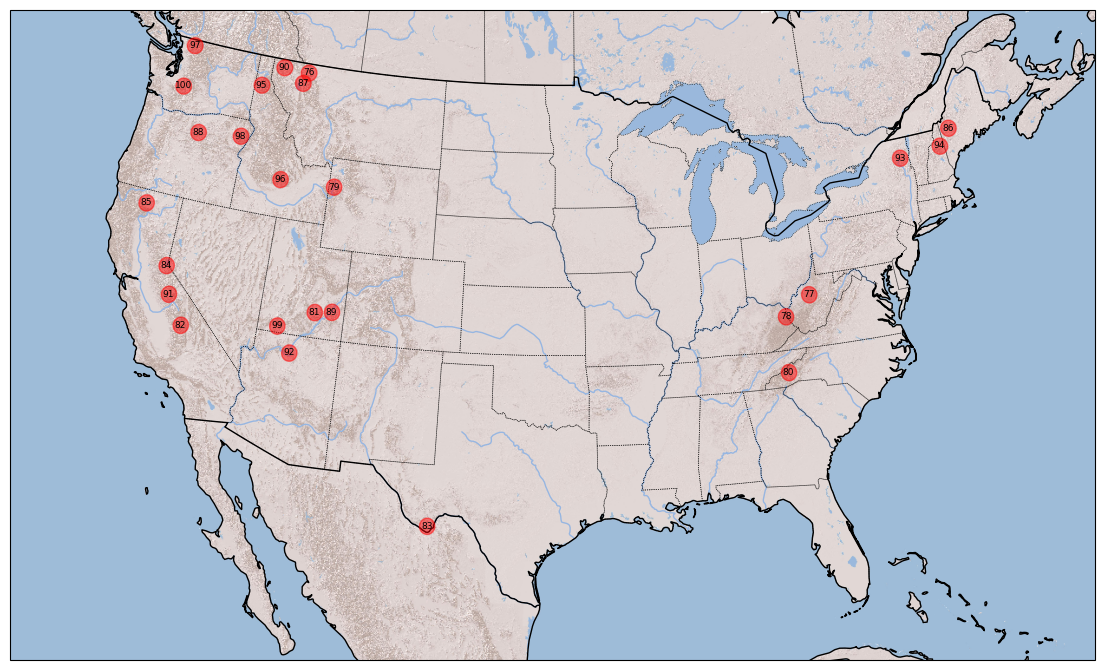}
        \caption{Eye-watering Scenarios}
    \end{subfigure}
    \caption{Categorizing RASPNet scenarios based on difficulty. Each scenario is indexed by $q_i \in Q$.} \label{fig:categorized_scenarios}
\end{figure*}

We categorize RASPNet scenarios into four subsets by difficulty: \textit{Basic} scenarios: $q_i \in \{1,2,\ldots,25 \}$, \textit{Intermediate} scenarios: $q_i \in \{26,27,\ldots,50 \}$, \textit{Challenging} scenarios: $q_i \in \{51,52,\ldots,75 \}$, and \textit{Eye-watering} scenarios: $q_i \in \{76,77,\ldots,100 \}$. These four distinct subsets are depicted in Figure \ref{fig:categorized_scenarios} --- the basic scenarios are largely concentrated within the Eastern United States and the plains regions, having simpler clutter characteristics (e.g., the Bonneville Salt Flats, UT). The intermediate scenarios have more uneven terrain than the basic scenarios, and encompass hillside regions, coastlines and valleys. The challenging scenarios are largely concentrated in the Western United States, comprising mountainous and variable terrain\iffalse--- this shift to more rugged and uneven terrain introduces clutter characteristics that deviate substantially from the baseline scenario\fi. Lastly, the eye-watering scenarios are composed almost exclusively of mountainous terrain, canyons, and other diversified land types.

\section{Dataset Applications} \label{sec:Dataset_applications}
Among the various applications of RASPNet to the radar and CVNN communities, we describe three key applications in this section: benchmarking data-driven algorithms for target localization accuracy versus existing approaches, benchmarking complex-valued learning algorithms, and transfer learning in realistic radar scenarios (which enables the adaptation of pre-trained models to new environments --- we demonstrate this via a toy example). Across all applications, we consider five example scenarios in RASPNet: Bonneville Salt Flats, UT ($q_i = 1, i = 29$) as the baseline scenario, Central Kansas, KS ($q_i = 4, i = 60$) from the Basic scenarios, the Ouachita Mountains, OK ($q_i = 35, i = 62$) from the Intermediate scenarios, the Great Smoky Mountains, TN ($q_i = 57, i = 76$) from the Challenging scenarios, and the Grand Canyon, AZ ($q_i = 92, i = 35$) from the Eye-watering scenarios, wherein $\smash{\tilde{S} = \{ 29, 35, 60, 62, 76 \}}$. We first preprocess the RASPNet data via conventional RASP techniques, per Section \ref{sec:Data_preprocessing}. We compiled all results on a NVIDIA GeForce RTX 3090 GPU.

\subsection{Data Preprocessing} \label{sec:Data_preprocessing}
To prepare RASPNet for the target localization and transfer learning tasks, we consider sub-regions from the scenarios, $\smash{i \in \tilde{S}}$, which we call \textit{radar processing regions}. The minimum and maximum range of each radar processing region from the airborne radar platform and the azimuth angle bounds of each radar processing region are provided in Table~\ref{tab:radar_processing_parameters} (see Appendix Section \ref{sec:radar_processing_parameters}), and the range bin indices are $g \in \{75,76,\ldots,95 \} \subset \{ 1,2,\ldots,G\}$, whereby $G' = 21$. We run a set of $N' = 1000$ independent experiments, where in each experiment, a point target is placed in range and azimuth, uniformly at random inside each radar processing region. We select the radar cross section (RCS) of each target independently at random from the uniform distribution, $\mathcal{U}[15,25]$ (dBsm), and we generate $K' = 100$ independent random realizations of the radar return for each target placement. We consider stationary targets in this example, with $\Lambda' = 1$, wherein there is no Doppler processing. For compatibility with our generated target data, we divide the $K = 10{,}000$ clutter realizations provided by RASPNet for scenario $\smash{i \in \tilde{S}}$ into $N^* = 100$ separate clutter datasets with $K' = 100$ realizations, where $K = N^*K'$. As $\Lambda' = 1$, we consider the first pulse from the clutter realizations. For the CVNN target localization benchmark task, we take the first $5$ of the $100$ realizations, where $K' = 5$. Per this construction, we now consider the signal model outlined in Section \ref{sec:signal_model} of the Appendix.

For both non-CVNN target localization and transfer learning tasks, we leverage datasets of \textit{heatmap matrices} \citep{venkatasubramanian2024data}, which comprise test statistics obtained via the signal model from Appendix Section \ref{sec:signal_model}, where $\phi = 0$. To produce each heatmap matrix, we sweep the radar steering vector, $\mathbf{\tilde{a}} (\theta, 0, 0)$, across azimuth with step size $\Delta\theta = 0.4^{\circ}$, from $\theta_{\text{min}}$ to $\theta_{\text{max}}$, which yields an NAMF azimuth spectrum at each range bin. Arranging these spectra across the $G'$ range bins produces a size $G' \times (\lfloor (\theta_{\text{max}} - \theta_{\text{min}})/\Delta\theta \rfloor + 1)$ matrix of NAMF test statistics. The values for $\theta_{\text{min}}$ and $\theta_{\text{max}}$ are in Table \ref{tab:radar_processing_parameters} for scenarios $\smash{i \in \tilde{S}}$. The cumulative dataset comprises the first $25{,}000$ of the $N = N'N^* = 100{,}000$ total heatmap matrices, formed using all combinations of $\mathbf{X}^i[g]$ and $(\mathbf{Z}^i[g] + \mathbf{A}^i[g])$. For the CVNN task, we consider $3$-dimensional feature vectors, which are formed from the radar return data matrices described in Section \ref{sec:signal_model} of the Appendix, where $\phi = 0$. The cumulative dataset comprises the first $25{,}000$ feature vectors of the $100{,}000$ total combinations.

\subsection{Target Localization Benchmark} \label{sec:Target_localization_benchmark}
Using the heatmap matrix datasets for scenarios $\smash{i \in \tilde{S}}$, we benchmark the target localization accuracy of two existing algorithms --- the classical peak-cell midpoint algorithm outlined in \citep{Shyam_STAP} and the regression CNN architecture introduced in \citep{venkatasubramanian2024data} --- versus a new specialized architecture we call \textit{Adaptive Radar Transformer} (ART). The ART applies lightweight convolutional filters to capture the local leakage pattern around the true target cell, then pools and projects these features into a compact token sequence. A transformer encoder then attends across all range–azimuth bins to resolve non‑local correlations in the leakage signature before a MLP head regresses the precise target coordinates. By adaptively weighting attention scores over leakage structures, ART can more precisely pinpoint the true target location than a purely local model can. We clarify that other classical algorithms exist for target localization, including gradient-based local search methods; these algorithms can also be benchmarked using RASPNet. Our considered model architectures are further described in Section \ref{sec:architectures} of the Appendix.

We now produce the training and test datasets for each scenario, $\smash{i \in \tilde{S}}$. We partition each dataset of $25{,}000$ heatmap matrices such that the first $N_{train} = 0.8N$ examples form the training dataset, and the remaining $N_{test} = 0.2N$ examples form the test dataset. Denote $(\thickbar{r}\vphantom{r}_t^i, \thickbar{\theta}\vphantom{\theta}_t^i)$ as the range and azimuth of the midpoint of the peak heatmap matrix cell, and denote $\smash{(\thicktilde{r}\vphantom{r}_t^i, \thicktilde{\theta}\vphantom{\theta}_t^i)}$ as the range and azimuth values predicted by each neural network, for example $t$ from our dataset for scenario $i$. While our heatmap matrices follow the polar coordinate system, we can transform their ground truth target locations into Cartesian coordinates to report the localization error in meters, whereby $\smash{(\thickbar{r}\vphantom{r}_t^i, \thickbar{\theta}\vphantom{\theta}_t^i) \rightarrow (\thickbar{x}\vphantom{x}_t^i, \thickbar{y}\vphantom{y}_t^i)}$ and $\smash{(\thicktilde{r}\vphantom{r}_t^i, \thicktilde{\theta}\vphantom{\theta}_t^i) \rightarrow (\thicktilde{x}\vphantom{x}_t^i, \thicktilde{y}\vphantom{y}_t^i)}$. Let $(r_t^{*i},\theta_t^{*i}) \rightarrow (x_t^{*i},y_t^{*i})$ be the ground truth target location for example $t$ from our dataset. For vectors $\mathbf{s}_t^{*i},\thickhat{\mathbf{s}}\vphantom{s}_t^i$, the localization error, $Err(\mathbf{s}_t^{*i}, \thickhat{\mathbf{s}}\vphantom{s}_t^i)$, over the $N_{test}$ test examples, is:
\begin{align}
    Err(\mathbf{s}_t^{*i}, \thickhat{\mathbf{s}}\vphantom{s}_t^i) = \Big(\sum\nolimits_{t = 1}^{N_{test}} \| \mathbf{s}_t^{*i} - \thickhat{\mathbf{s}}\vphantom{s}_t^i \|_2\Big) / N_{test} \label{eq_CNN_toc}.
\end{align}
In our benchmark, $(\mathbf{s}\vphantom{s}_t^{*i},\thickhat{\mathbf{s}}\vphantom{s}_t^i) = ([x_t^{*i},y_t^{*i}], [\thickhat{x}_t, \thickhat{y}_t])$, with $(\thickhat{x}\vphantom{x}_t^i, \thickhat{y}\vphantom{y}_t^i) = (\thickbar{x}\vphantom{x}_t^i, \thickbar{y}\vphantom{y}_t^i)$ for the peak cell midpoint $[Err_{\text{NAMF}}]$, and $(\thickhat{x}\vphantom{x}_t^i, \thickhat{y}\vphantom{y}_t^i) = (\thicktilde{x}\vphantom{x}_t^i, \thicktilde{y}\vphantom{y}_t^i)$ for each neural network $[Err_{\text{CNN/ART}}]$. For each scenario, $\smash{i \in \tilde{S}}$, we train each neural network until convergence on the training dataset using MSE as our loss function, and we leverage the Adam optimizer \citep{kingma2014adam} with a learning rate of $\alpha = 1\text{e-}3$. The localization errors on the test dataset are given for the peak cell midpoint and each neural network in Table \ref{tab:error_values}. We see that as the scenarios become more challenging, measured using the $\mathcal{E}$-statistic with respect to the baseline scenario, the localization performance of all methods worsen, matching our intuition from Section \ref{sec:Dataset_organization}. Our proposed ART achieves SOTA performance across all scenarios.
\begin{table}[h!]
\caption{Benchmarking target localization performance (average Euclidean distance) for $\smash{i \in \tilde{S}}$.}
\label{tab:error_values}
\centering
\begin{tabular}{c|c|c|c|c|c}
\hline
$\boldsymbol{q_i}$ & $\boldsymbol{i}$ & $\boldsymbol{Err_{\text{NAMF}}}$ \textbf{(m)} & $\boldsymbol{Err_{\text{CNN}}}$ \textbf{(m)} & $\boldsymbol{Err_{\text{ART}}}$ \textbf{(m)} \textbf{[ours]} & $\boldsymbol{\mathcal{E}}$\textbf{-statistic} \\ 
\hline
1 & 29 & 21.62 $\pm$ 0 & 9.88 $\pm$ 0.50 & \textbf{7.75 $\pm$ 0.36} & 0 \\ 
4 & 60 & 21.83 $\pm$ 0 & 14.60 $\pm$ 0.82 & \textbf{9.76 $\pm$ 0.44} & 3.51\text{e-}8 \\ 
35 & 62 & 187.31 $\pm$ 0 & 68.31 $\pm$ 2.23 & \textbf{51.86 $\pm$ 1.59} & 2.63\text{e-}7 \\ 
57 & 76 & 360.53 $\pm$ 0 & 145.34 $\pm$ 5.91 & \textbf{123.08 $\pm$ 2.40} & 7.91\text{e-}7 \\ 
92 & 35 & 1004.20 $\pm$ 0 & 220.45 $\pm$ 9.67 & \textbf{180.14 $\pm$ 5.50} & 4.04\text{e-}6 \\ 
\hline
\end{tabular}
\end{table}

\begin{figure*}[t!]
    \centering
    \begin{subfigure}{0.32\textwidth}
        \includegraphics[width=\textwidth]{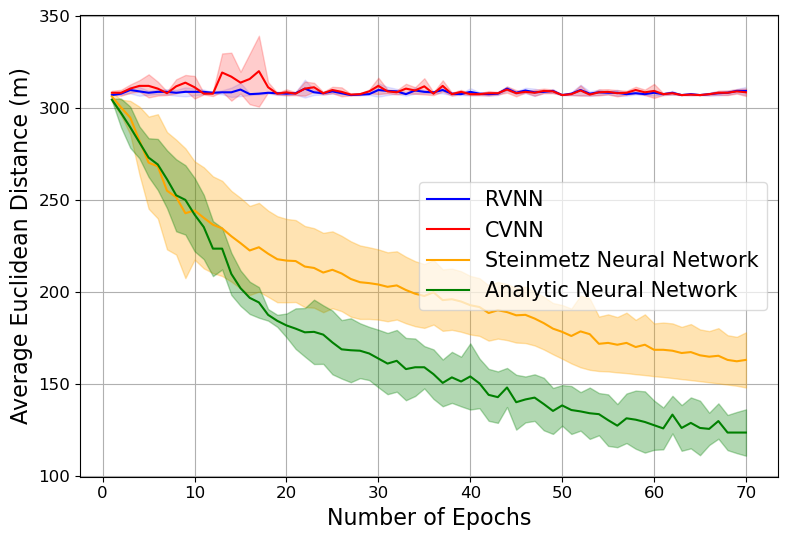}
        \caption{Bonneville Salt Flats $(i = 29)$}
    \end{subfigure}
    \begin{subfigure}{0.32\textwidth}
        \includegraphics[width=\textwidth]{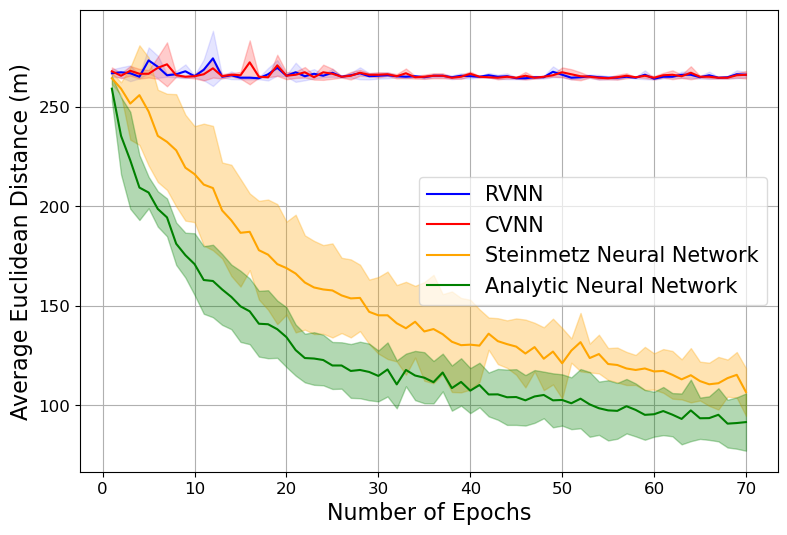}
        \caption{Central Kansas $(i = 60)$}
    \end{subfigure}
    \begin{subfigure}{0.32\textwidth}
        \includegraphics[width=\textwidth]{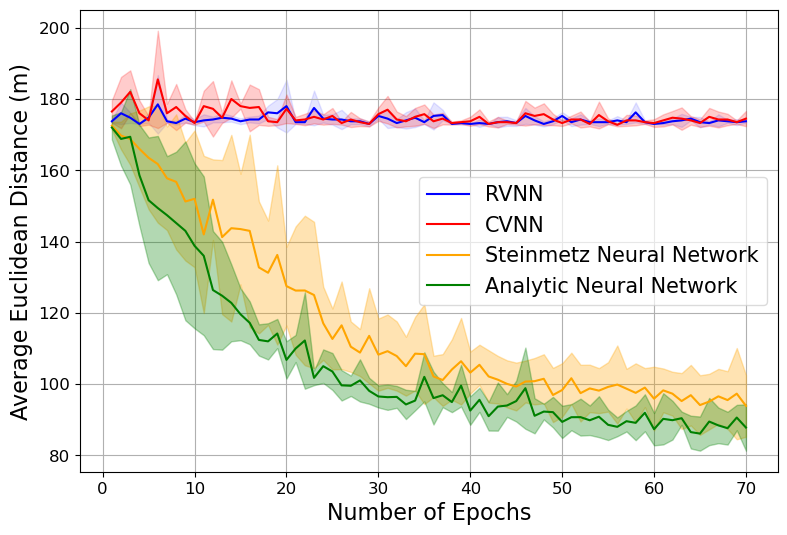}
        \caption{Ouachita Mountains $(i = 62)$}
    \end{subfigure}
    \\
    \begin{subfigure}{0.332\textwidth}
        \includegraphics[width=\textwidth]{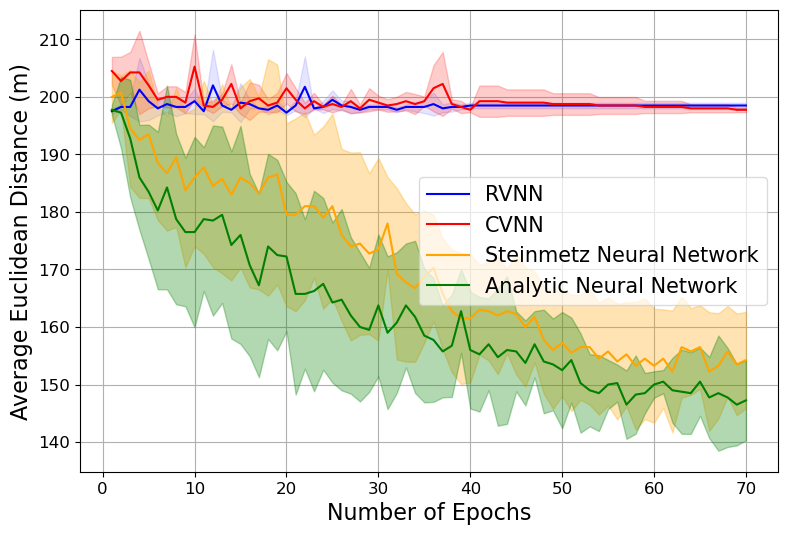}
        \caption{Great Smoky Mountains $(i = 76)$}
    \end{subfigure}
    \begin{subfigure}{0.332\textwidth}
        \includegraphics[width=\textwidth]{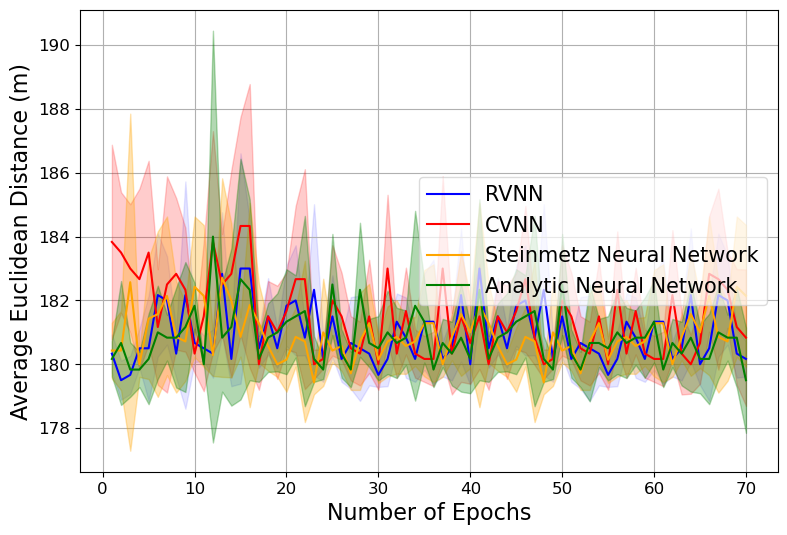}
        \caption{Grand Canyon $(i = 35)$}
    \end{subfigure}
    \caption{Test performance comparison on RASPNet scenarios $i \in \tilde{S}$ using convolutional CVNN, convolutional RVNN, and convolutional Steinmetz and analytic networks [ours]. The x-axis represents the training epochs, while the y-axis indicates the target localization performance.}
    \label{fig:epoch_performance}
\end{figure*}

\subsection{CVNN Target Localization Benchmark} \label{sec:CVNN_target_localization_benchmark}
We now benchmark the performance of CVNNs, recalling the datasets of feature vectors described in Section \ref{sec:Data_preprocessing} for scenarios $\smash{i \in \tilde{S}}$. Extending \citep{trabelsi2017deep, venkatasubramanian2024steinmetz}, we compare the performance of convolutional CVNNs and real-valued neural networks (RVNNs), versus Steinmetz and analytic neural networks (shown to be SOTA for complex-valued learning tasks) via a new Steinmetz variant called \textit{Convolutional Steinmetz Networks} for improved spatial processing of complex data. We use the partitioning scheme from Section \ref{sec:Target_localization_benchmark} and compute localization errors via Eq.~(\ref{eq_CNN_toc}). For each scenario, $\smash{i \in \tilde{S}}$, we train all networks until convergence on the training dataset using MSE Loss and the Adam optimizer \citep{kingma2014adam} with a learning rate of $\alpha = 5\text{e-}2$. Per Figure \ref{fig:epoch_performance}, while the convolutional RVNN and CVNN falter, the convolutional Steinmetz and analytic networks effectively learn the task across all scenarios $\smash{i \in \tilde{S} \setminus \{35\}}$, achieving SOTA performance.

\subsection{Transfer Learning} \label{sec:Transfer_learning}
Apart from benchmarking the performance of algorithms, RASPNet can also be leveraged for transfer learning applications in realistic adaptive radar processing scenarios. We recall that the RFView\textsuperscript{\tiny\textregistered}-generated data exhibits some degree of statistical similarity to real-world data, as examined in Section \ref{sec:Measured_data_validation}. Now, suppose we collect $K$ clutter realizations, $\{z_j^\omega, \ \forall j \in D \}$ from some real-world scenario, $\omega$, where $z_j^\omega \sim Z^\omega$, and train $100$ models for some predictive adaptive radar processing task on each of the $100$ RASPNet scenarios, $i \in S$. Our aim is to find the trained model that has the best performance for the same task on the real-world scenario, $\omega$. We can proceed by ranking the RASPNet scenarios, $i \in S$, by their similarity to scenario $b = \omega$, using Algorithm \ref{alg:org}. We hypothesize that the scenario with the lowest $\mathcal{E}$-statistic, relative to the real-world scenario, is the optimal choice for selecting the corresponding trained model. To improve the performance of this trained model on the real-world scenario, we can use a few samples from the real-world scenario to fine-tune it for the task at hand.
\begin{figure*}[t!]
    \centering
    \includegraphics[width=0.9\linewidth]{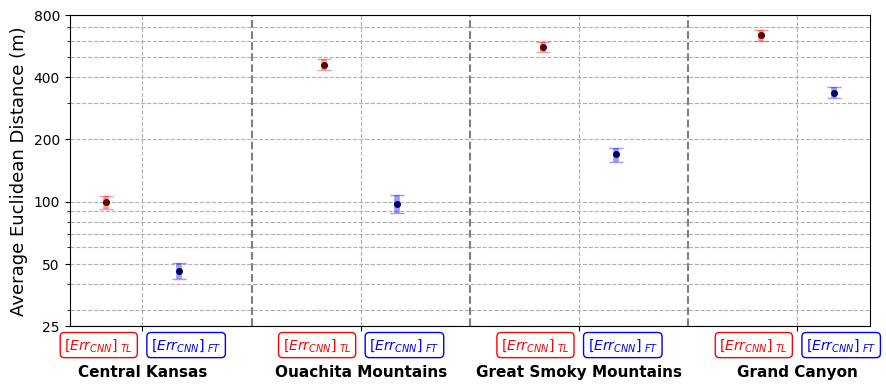}
    \caption{Regression CNN target localization performance for scenarios $\smash{i \in \tilde{S} \setminus \{29\}}$ after transfer learning $([Err_{\text{CNN}}]_{\text{TL}})$ and fine-tuning $([Err_{\text{CNN}}]_{\text{FT}})$.}
    \label{fig:transfer_learning}
\end{figure*}

As a proof-of-concept, we consider again the target localization task and train the regression CNN to localize targets in scenario $i = 29$ (Bonneville Salt Flats, UT). However, we evaluate this trained network on each scenario $\smash{i \in \tilde{S} \setminus \{29\}}$ to evaluate whether the scenario with the highest similarity to $i = 29$ from $\smash{\tilde{S} \setminus \{29\}}$ achieves the best localization performance. As in Section \ref{sec:Target_localization_benchmark}, our training dataset comprises the first $0.8N$ of the $25{,}000$ heatmap matrices generated for scenario $i = 29$. We consider four test datasets, each comprising the last $0.2N$ of the $25{,}000$ total heatmap matrices from scenario $\smash{i \in \tilde{S} \setminus \{29\}}$, and we consider four datasets for fine-tuning, each comprising the first $64$ of the $25{,}000$ total heatmap matrices from scenario $\smash{i \in \tilde{S} \setminus \{29\}}$. We train the regression CNN on the training dataset using MSE Loss, and we leverage the Adam optimizer \citep{kingma2014adam} with $\alpha = 1\text{e-}3$. The localization accuracies on the test datasets, $[Err_{\text{CNN}}]_{\text{TL}}$, $\smash{\forall i \in \tilde{S} \setminus \{29\}}$, are depicted in Figure \ref{fig:transfer_learning}. As expected, the best performance is given by the model evaluated on the scenario $i = 60$ (Central Kansas). This scenario has the lowest $\mathcal{E}$-statistic with respect to $i = 29$ from $\smash{\tilde{S} \setminus \{29\}}$.

We now consider the case of fine-tuning the trained regression CNN using a small set of $64$ heatmap matrices from the scenarios $\smash{i \in \tilde{S} \setminus \{29\}}$. We freeze the convolutional and batch normalization layers of our trained regression CNN, halving the number of parameters. Using the $64$ heatmap matrices, the weights and biases of the remaining layers are fine-tuned using MSE Loss and the Adam optimizer with $\alpha = 1\text{e-}3$. After fine-tuning the regression CNN for each scenario $\smash{i \in \tilde{S} \setminus \{29\}}$, we obtain the localization accuracies on the test datasets, $[Err_{\text{CNN}}]_{\text{FT}}$, $\smash{\forall i \in \tilde{S} \setminus \{29\}}$, depicted in Figure \ref{fig:transfer_learning}. We note that the improvements afforded by the regression CNN are largely recovered after fine-tuning.

\section{Limitations and Other Applications} \label{sec:Limitations}
Apart from the presented applications, RASPNet serves as a comprehensive testbed for benchmarking adaptive radar target detection and target classification algorithms. While it currently does not support target tracking capabilities, we plan to incorporate this feature in future work. A central limitation of RASPNet is the substantial size of its clutter datasets, with the entire dataset exceeding 16 TB and each scenario's clutter data surpassing 160 GB. These large sizes are necessary to ensure sufficient resolution in range and Doppler and to provide ample clutter realizations for data-driven methods.

\section{Conclusion} \label{sec:Conclusion}
In this work, we presented RASPNet, a large-scale dataset consisting of $100$ airborne radar scenarios from across the contiguous United States. RASPNet provides a standardized benchmark for evaluating and developing RASP algorithms and complex-valued learning algorithms, bridging a significant gap in the availability of a comprehensive, realistic dataset for adaptive radar processing applications. We outlined relevant properties of RASPNet, described its organization, and provided three applications of the dataset for benchmarking target localization and CVNNs, and transfer learning in realistic scenarios. Our experiments demonstrated RASPNet's capability to support data-driven approaches, highlighting its potential to accelerate research and development in the adaptive radar community. % As RASPNet is an ongoing initiative, additional work will also be devoted to its comprehensive documentation and improving its accessibility to practitioners in the field.

\ack{This work was supported in part by the U.S. Air Force Office of Scientific Research (AFOSR) under award FA9550-21-1-0235. Dr. Muralidhar Rangaswamy and Dr. Bosung Kang were supported by AFOSR under project 20RYCORO51. The authors would like to thank the Sensors Directorate of the U.S. Air Force Research Laboratory for providing the radar sample of the Multi-Channel Airborne Radar Measurement (MCARM) program, and Dr. Erik Blasch at AFOSR for his continued support, which has made this effort possible. The opinions within this paper are the authors’ own and do not constitute any explicit or implicit endorsement by the U.S. Department of Defense.}

% \section*{Accessibility}
% Authors are kindly asked to make their submissions as accessible as possible for everyone including people with disabilities and sensory or neurological differences.
% Tips of how to achieve this and what to pay attention to will be provided on the conference website \url{http://icml.cc/}.

% \section*{Software and Data}

% If a paper is accepted, we strongly encourage the publication of software and data with the
% camera-ready version of the paper whenever appropriate. This can be
% done by including a URL in the camera-ready copy. However, \textbf{do not}
% include URLs that reveal your institution or identity in your
% submission for review. Instead, provide an anonymous URL or upload
% the material as ``Supplementary Material'' into the OpenReview reviewing
% system. Note that reviewers are not required to look at this material
% when writing their review.

% % Acknowledgements should only appear in the accepted version.

% \textbf{Do not} include acknowledgements in the initial version of
% the paper submitted for blind review.

% If a paper is accepted, the final camera-ready version can (and
% usually should) include acknowledgements.  Such acknowledgements
% should be placed at the end of the section, in an unnumbered section
% that does not count towards the paper page limit. Typically, this will 
% include thanks to reviewers who gave useful comments, to colleagues 
% who contributed to the ideas, and to funding agencies and corporate 
% sponsors that provided financial support.

\bibliography{main.bib}
\bibliographystyle{plainnat}

%%%%%%%%%%%%%%%%%%%%%%%%%%%%%%%%%%%%%%%%%%%%%%%%%%%%%%%%%%%%%%%%%%%%%%%%%%%%%%%
%%%%%%%%%%%%%%%%%%%%%%%%%%%%%%%%%%%%%%%%%%%%%%%%%%%%%%%%%%%%%%%%%%%%%%%%%%%%%%%
% APPENDIX
%%%%%%%%%%%%%%%%%%%%%%%%%%%%%%%%%%%%%%%%%%%%%%%%%%%%%%%%%%%%%%%%%%%%%%%%%%%%%%%
%%%%%%%%%%%%%%%%%%%%%%%%%%%%%%%%%%%%%%%%%%%%%%%%%%%%%%%%%%%%%%%%%%%%%%%%%%%%%%%
\newpage
\appendix
\onecolumn
{\Huge \textbf{Appendix}}

\section{Exemplar Scenario Validation} \label{sec:scenario_validation_appendix}
Below, Figure \ref{fig:reconstruction_trainsize_100} pertains to the validation of the exemplary nature of RASPNet scenarios. We observe that the geographic coverage of the $M^* = 100$ representative scenarios yielded by PAM mirrors the geographic coverage of the $100$ scenarios in RASPNet, as a preliminary proof-of-concept.

\begin{figure*}[h!]
    \centering
    \begin{subfigure}{0.495\textwidth}
        \includegraphics[width=\textwidth]{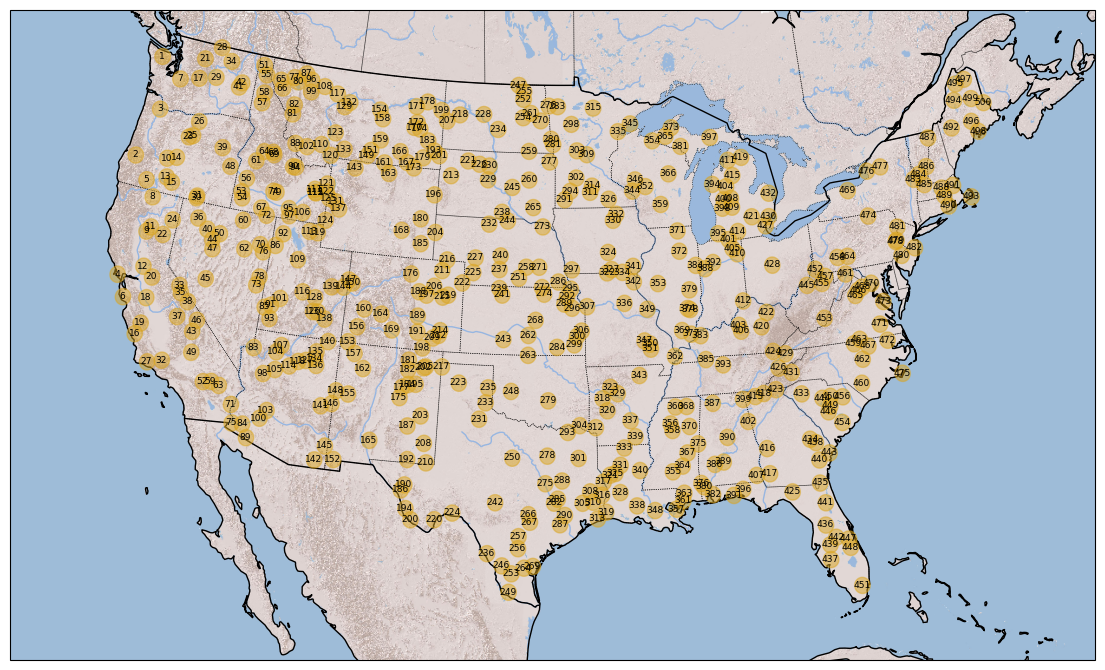}
        \caption{$\closure{M} = 500$ Total Scenarios}
    \end{subfigure}
    \begin{subfigure}{0.495\textwidth}
        \includegraphics[width=\textwidth]{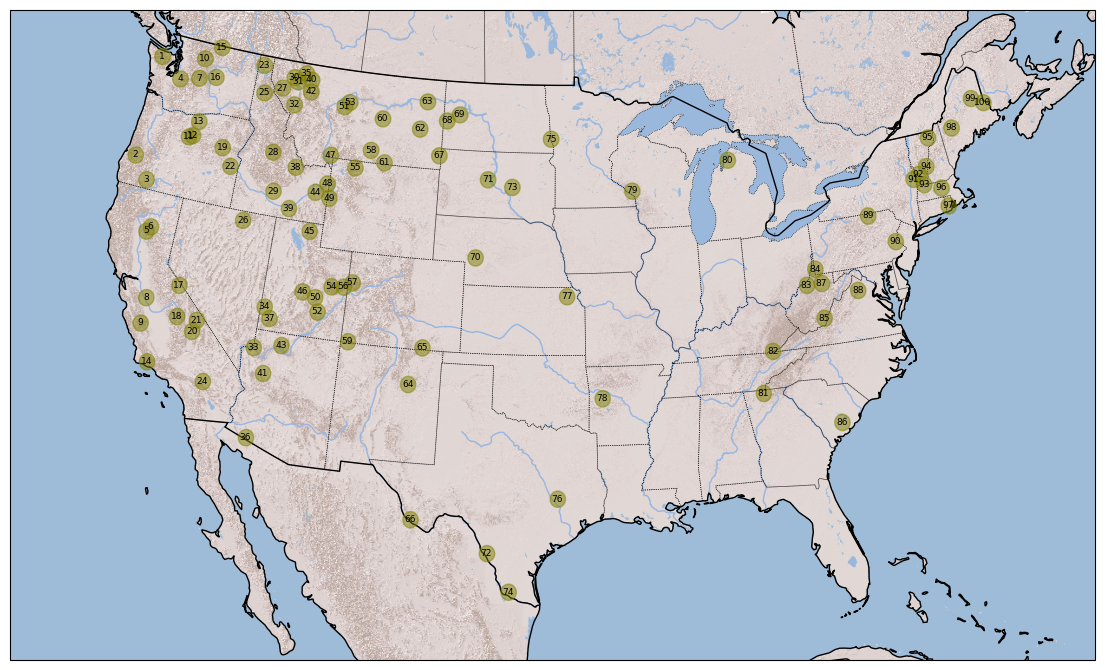}
        \caption{$M^* = 100$ Representative Scenarios}
    \end{subfigure}
    \caption{Geographic coverage of the $\closure{M} = 500$ total scenarios (left) and $M^* = 100$ representative scenarios (right) yielded by the PAM algorithm across the contiguous United States.}
    \label{fig:reconstruction_trainsize_100}
\end{figure*}

\section{MCARM Radar Platform Parameters} \label{sec:mcarm_radar_platform_parameters}
The list of parameters pertaining to the airborne radar platform from the MCARM measured dataset are provided in this section \citep{babu1996processing}. To simulate the clutter realizations for the RFView and bald-earth cases, we instantiate an airborne radar platform in RFView\textsuperscript{\tiny\textregistered} utilizing the parameters provided in Table \ref{tab:radar_parameters_MCARM}. The complete description of MCARM parameters is provided in \citep{himed1999mcarm}.
\begin{table}[h!]
\caption{MCARM Dataset Radar Platform Parameters}
\label{tab:radar_parameters_MCARM}
\centering
\begin{tabular}{l|l}
\hline
\textbf{Parameter} & \textbf{Value} \\
\hline
Carrier frequency $(\tilde{f}_c)$ & $1.25 \,\text{GHz}$ \\
Bandwidth $(\tilde{B})$ & $1.25\,\text{MHz}$ \\
Pulse Repetition Frequency $(\tilde{f}_{PR})$ & $1984\,\text{Hz}$ \\
Radar platform altitude $(\tilde{h})$ & $3072\,\text{m}$ \\
Radar platform position (lat, lon) & $(39.3805^{\circ}, -75.9746^{\circ})$ \\
Radar platform aim location (lat, lon) & $(39.3630^{\circ}, -74.7435^{\circ})$ \\
Radar platform speed & $100\, \text{m/s}$ \\
Range resolution $(\Delta \tilde{r})$ & $120\,\text{m}$ \\
Antenna element configuration \{horizontal $\times$ vertical\} & $16 \times 8$ \\
Number of receiver channels $(\tilde{L})$ \{horizontal $\times$ vertical\} & $24 \times 1$ \\
Antenna element horizontal spacing & $0.1092\,\text{m}$ \\
Antenna element vertical spacing & $0.1407\,\text{m}$ \\
Number of pulses $(\tilde{\Lambda})$ & $128$ \\
Transmitted waveform number of samples & $63$ \\
Total number of range bins $(\tilde{G})$ & $630$ \\
Radar range swath & $75{,}500\,\text{m}$
\end{tabular}
\end{table}

\section{Data Preprocessing Addendum}

\subsection{Radar Processing Region Parameters} \label{sec:radar_processing_parameters}
Provided below are the radar processing region parameters for the scenarios considered in Section \ref{sec:Dataset_applications} from the main text, where $\smash{i \in \tilde{S} = \{ 29, 35, 60, 62, 76 \}}$.
\begin{table}[h!]
\caption{Radar Processing Region Parameters}
\label{tab:radar_processing_parameters}
\centering
\begin{tabular}{p{0.65\linewidth}|p{0.3\linewidth}}
\hline
\textbf{Bonneville Salt Flats, UT ($q_i = 1, i = 29$)} -- Parameters & Values \\ 
\hline
Radar platform position (lat, lon) & $(40.6391^{\circ}, -113.6525^{\circ})$ \\
Radar platform aim location (lat, lon) & $(40.5161^{\circ}, -113.8025^{\circ})$ \\
Sub-region extrema range bin midpoints $(r_{\text{min}},r_{\text{max}})$ & $(10851 \ \text{m},11451 \ \text{m})$ \\
Sub-region extrema azimuths $(\theta_{\text{min}},\theta_{\text{max}})$ & $(215^{\circ},225^{\circ})$ \\
\hline
\textbf{Central Kansas, KS ($q_i = 4, i = 60$)} -- Parameters & Values \\ 
\hline
Radar platform position (lat, lon) & $(37.9689^{\circ}, -97.5750^{\circ})$ \\
Radar platform aim location (lat, lon) & $(37.8459^{\circ}, -97.7250^{\circ})$ \\
Sub-region extrema range bin midpoints $(r_{\text{min}},r_{\text{max}})$ & $(11073 \ \text{m},11673 \ \text{m})$ \\
Sub-region extrema azimuths $(\theta_{\text{min}},\theta_{\text{max}})$ & $(215^{\circ},225^{\circ})$ \\
\hline
\textbf{Ouachita Mountains, OK ($q_i = 35, i = 62$)} -- Parameters & Values \\
\hline
Radar platform position (lat, lon) & $(34.5560^{\circ}, -95.6221^{\circ})$ \\
Radar platform aim location (lat, lon) & $(34.4330^{\circ}, -95.7721^{\circ})$ \\
Sub-region extrema range bin midpoints $(r_{\text{min}},r_{\text{max}})$ & $(11471 \ \text{m},12071 \ \text{m})$ \\
Sub-region extrema azimuths $(\theta_{\text{min}},\theta_{\text{max}})$ & $(215^{\circ},225^{\circ})$ \\
\hline
\textbf{Great Smoky Mountains, TN ($q_i = 57, i = 76$)} -- Parameters & Values \\ 
\hline
Radar platform position (lat, lon) & $(35.4121^{\circ}, -84.1288^{\circ})$ \\
Radar platform aim location (lat, lon) & $(35.2891^{\circ}, -84.2788^{\circ})$ \\
Sub-region extrema range bin midpoints $(r_{\text{min}},r_{\text{max}})$ & $(11388 \ \text{m},11988 \ \text{m})$ \\
Sub-region extrema azimuths $(\theta_{\text{min}},\theta_{\text{max}})$ & $(215^{\circ},225^{\circ})$ \\
\hline
\textbf{Grand Canyon, AZ ($q_i = 92, i = 35$)} -- Parameters & Values \\ 
\hline
Radar platform position (lat, lon) & $(36.2069^{\circ}, -111.9628^{\circ})$ \\
Radar platform aim location (lat, lon) & $(36.0839^{\circ}, -112.1128^{\circ})$ \\
Sub-region extrema range bin midpoints $(r_{\text{min}},r_{\text{max}})$ & $(11381 \ \text{m},11981 \ \text{m})$ \\
Sub-region extrema azimuths $(\theta_{\text{min}},\theta_{\text{max}})$ & $(215^{\circ},225^{\circ})$ \\
\end{tabular}
\end{table}

\subsection{Signal Model} \label{sec:signal_model}
Suppose that $X^i, Z^i, A^i \in \mathbb{C}^{\Lambda' L G'}$ are the random variables describing the target, clutter, and noise for scenario $\smash{i \in \tilde{S}}$, and suppose that $x_j^i \sim X^i$, $z_j^i \sim Z^i$, and $a_j^i \sim A^i$ denote the target, clutter, and noise data for the $j^{\text{th}}$ independent random realization of the radar return for scenario $i$. Recording $\smash{x_j^i, z_j^i, a_j^i}$, $\forall j \in \{1,\ldots,K'\}$, we obtain $\smash{\mathbf{X}^i, \mathbf{Z}^i, \mathbf{A}^i \in \mathbb{C}^{\Lambda' L G' \times K'}}$. We denote $\smash{\mathbf{Y}^i[g]\in \mathbb{C}^{\Lambda' L \times K'}}$ as the matched filtered radar return data matrix associated with range bin $g$. The columns of $\mathbf{Y}^i[g]$ are $\Lambda' L \times 1$ dimensional vectors that are built by stacking $L$-dimensional array measurements, obtained from the $\Lambda'$ radar pulse transmissions --- for each pulse return, we have $K'$ independent realizations. We consider a fixed clutter-to-noise ratio (CNR) of $20$ dB. The radar return data matrix is given by: 
\begin{equation}
\mathbf{Y}^i[g]=\beta^i[g] \mathbf{X}^i[g] + \mathbf{Z}^i[g] + \mathbf{A}^i[g]. \label{eq_alternative}
\end{equation}
Where $\beta^i[g]=1$ if range bin $g$ contains a target and $\beta^i[g]=0$ otherwise, $\mathbf{X}^i[g]$ is the target response matrix, $\mathbf{Z}^i[g]$ is the clutter response matrix, and $\mathbf{A}^i[g]$ is the noise response matrix. For the CVNN benchmark target localization task, we permute and reshape this radar return data matrix to obtain a $3$-dimensional feature vector, $\mathbf{Y}^i_{\text{CVNN}} \in \mathbb{C}^{K' \times G' \times (\Lambda' L)}$.

Let $\mathbf{a}^i(\theta, \phi)\in \mathbb{C}^L$ denote the receiver array steering vector, where $\theta$ is the azimuth angle variable and $\phi$ is the elevation angle variable. Let $\mathbf{d}(v)$ be the Doppler processing linear phase vector across $\Lambda$ pulses; that is, $\mathbf{d}^i(v) = \left[ 1, e^{-i 2 \pi f_d/f_{PR}}, \ldots, e^{-i 2 \pi  (\Lambda - 1)f_{d}/f_{PR}} \right]\vphantom{]}^T$, where $f_{\text{d}}$ is the Doppler shift frequency. Accordingly, the effective space-time steering vector, $\smash{\mathbf{\tilde{a}}^i(\theta,\phi,v)}$, in azimuth, elevation, and Doppler for radar processing is:
\begin{equation}
\mathbf{\tilde{a}}^i (\theta, \phi, v) =  \mathbf{d}^i(v) \otimes \mathbf{a}^i(\theta, \phi).
\end{equation}
Where $\otimes$ denotes Kronecker product. Suppose now that $\mathbf{\Sigma}^i[g] = \mathbb{E}[B^i[\rho] B^i[\rho]^H]$ is the true clutter-plus-noise covariance matrix obtained from $B^i[g] = Z^i[g] + A^i[g]$, where $\mathbf{\Sigma}_j^i[g] \in \mathbb{C}^{(\Lambda' L) \times (\Lambda' L)}$. The sample clutter-plus-noise covariance matrix is computed as: $\mathbf{\hat{\Sigma}}\vphantom{\Sigma}^i[g] = (\mathbf{B}^i[g] \mathbf{B}^i[g]^H) / K'$. The NAMF test statistic, $\Gamma^i[g](\theta, \phi,v)$, at coordinates $(\theta, \phi, v)$ in range bin $g$, for scenario $i$, is given by:
\begin{align}
    \Gamma^i[g](\theta,\phi,v)= \frac{\| \mathbf{\Tilde{a}}^i(\theta, \phi, v)^H \mathbf{\hat{\Sigma}}\vphantom{\Sigma}^i[g]{\vphantom{\Sigma}}^{-1} \mathbf{Y}^i[g] \|_2^2}{\mathbf{\Tilde{a}}^i(\theta, \phi, v)^H \mathbf{\hat{\Sigma}}\vphantom{\sigma}^i[g]{\vphantom{\Sigma}}^{-1} \mathbf{\Tilde{a}}^i(\theta, \phi, v) \ \|\text{diag}(\mathbf{Y}^i[g]^H \mathbf{\hat{\Sigma}}\vphantom{\Sigma}^i[g]{\vphantom{\Sigma}}^{-1} \mathbf{Y}^i[g])\|_2 }.
\end{align}
When the radar platform is at a long range from the ground scene, the elevation angle, $\phi$, is close to zero. We consider this to be the case in our provided RASPNet empirical experiments.

\section{Neural Network Architectures} \label{sec:architectures}

\subsection{Regression CNN}
\begin{figure}[h!]
\centering
\captionsetup{justification=centering}
\includegraphics[width=0.84\linewidth]{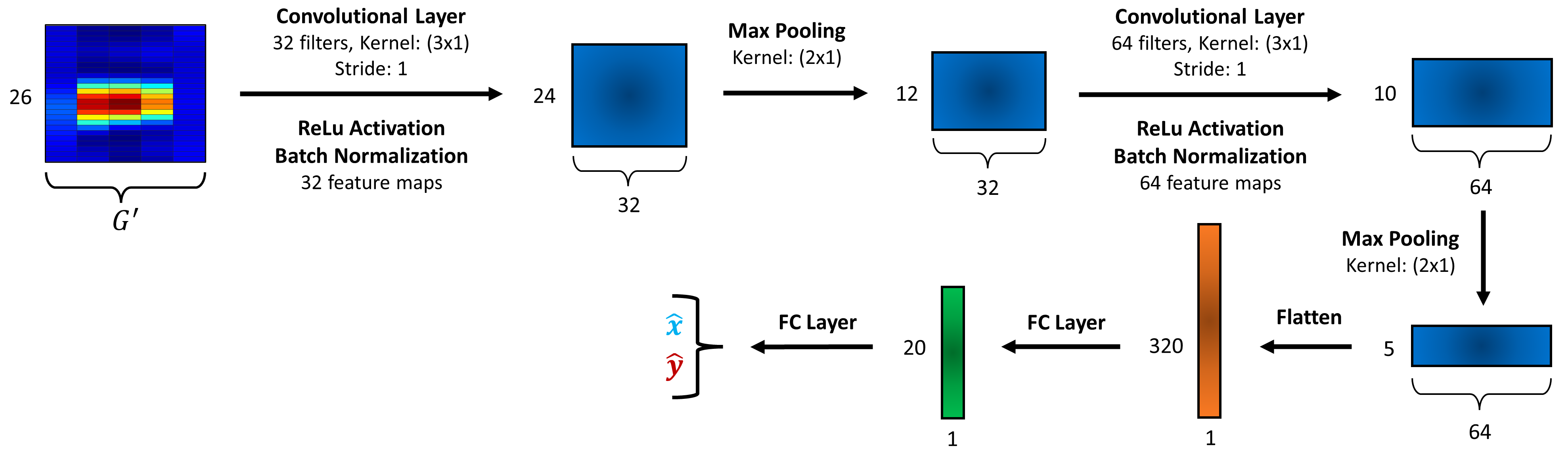}
\caption{Regression CNN architecture for azimuth step size $\Delta \theta = 0.4^{\circ}$ and $G'$ range bins.}
\label{CNN_default_parameters}
\end{figure}

The regression CNN architecture, shown in Figure \ref{CNN_default_parameters}, is designed for radar processing areas with an azimuth extent of $10^\circ$ and an azimuth step size of $\Delta\theta=0.4^\circ$. The radar processing areas consist of $G' = 21$ range bins. Thus, each input data sample to the CNN is a size $21 \times 26$ heatmap matrix.

The input data first passes through a convolutional layer with a $3 \times 1$ kernel and stride of 1, producing 32 feature maps, followed by ReLU activation and batch normalization. Max pooling is then applied with a $2 \times 1$ kernel. This output goes through another convolutional layer with a $3 \times 1$ kernel and stride of 1, producing 64 feature maps, again followed by ReLU activation, batch normalization, and max pooling utilizing a $2 \times 1$ kernel. The output is flattened and passed through two fully connected layers to estimate the target location $\smash{(\hat{r}, \hat{\theta})}$ in range-azimuth coordinates, which is then converted to Cartesian coordinates $\smash{(\hat{x}, \hat{y})}$. The regression CNN has $14{,}910$ parameters.

\subsection{Adaptive Radar Transformer} \label{sec:adaptive_radar_transformer}
The Adaptive Radar Transformer (ART) takes a single‑target heatmap matrix of size \(21\times26\) and predicts the target coordinates \((\hat{x},\hat{y})\).  ART first extracts local leakage features using convolutional blocks, then builds global context via a compact Transformer encoder. ART has $94{,}242$ parameters.
\begin{itemize}
  \item \textbf{Conv1d Block 1:}
    \begin{itemize}
      \item \texttt{conv1}: 1D convolution from 21 → 32 channels, kernel size 3, padding 1.
      \item \texttt{bn1}: BatchNorm1d over 32 channels.
      \item \texttt{relu1}: ReLU activation.
      \item \texttt{pool1}: MaxPool1d with kernel=2, stride=2 → length halved.
    \end{itemize}
  \item \textbf{Conv1d Block 2:}
    \begin{itemize}
      \item \texttt{conv2}: 1D convolution from 32 → 64 channels, kernel size 3, padding 1.
      \item \texttt{bn2}: BatchNorm1d over 64 channels.
      \item \texttt{relu2}: ReLU activation.
      \item \texttt{pool2}: MaxPool1d with kernel=2, stride=2 → length halved again.
    \end{itemize}
  \item \textbf{Token preparation:}
    \begin{itemize}
      \item \texttt{permute}: Permute to \((\text{batch},\,L,\,64)\) where \(L=26/2^2\).
      \item \texttt{token\_proj}: Linear layer mapping 64 → 64 to produce \(L\) patch tokens.
    \end{itemize}
  \item \textbf{CLS token \& positional embeddings:}
    \begin{itemize}
      \item \texttt{cls\_token}: Learned \((1\times1\times64)\) token prepended to tokens.
      \item \texttt{pos\_embed}: Learned \((1\times(L+1)\times64)\) positional embeddings added to sequence.
    \end{itemize}
  \item \textbf{Transformer Encoder:}
    \begin{itemize}
      \item \texttt{encoder\_layer}: Multi‑head self‑attention (\(d_{model}=64\), 4 heads) + feed‑forward (\(128\) hidden), dropout 0.1.
      \item \texttt{transformer}: Stack 2 encoder layers, producing \((\text{batch},\,L+1,\,64)\).
    \end{itemize}
  \item \textbf{MLP head:}
    \begin{itemize}
      \item \texttt{cls\_rep}: Extract token 0 → \((\text{batch},\,64)\).
      \item \texttt{ln}: LayerNorm over 64 dims.
      \item \texttt{linear1}: 64 → 32.
      \item \texttt{gelu}: GELU activation.
      \item \texttt{linear2}: 32 → \((\hat{r},\hat{\theta})\) → \((\hat{x},\hat{y})\).
    \end{itemize}
\end{itemize}

\subsection{Convolutional Steinmetz Network} \label{sec:steinmetz_network}
The convolutional Steinmetz network is designed to handle both real and imaginary components of the input data separately before combining them for the final prediction. This architecture is evaluated on the CVNN benchmark from Section \ref{sec:CVNN_target_localization_benchmark}. The convolutional analytic network is a Steinmetz network in which the separately processed real and imaginary components are related by the discrete Hilbert transform (DHT), weighted by a tradeoff parameter, $\beta$. Across all experiments in Section \ref{sec:CVNN_target_localization_benchmark}, we let $\beta = 1e{-}3$, and consider $lN = 128$, $k = 2$ (see below). The network has $368{,}002$ parameters.
\begin{itemize}
    \item \textbf{Convolutional Layer (\texttt{conv\_real1})}: Applies a 2D convolution to the real input features, transforming them from $K'$-dimensional inputs to 32 feature maps using a \(3 \times 3\) kernel with padding 1, followed by a \(2 \times 2\) average pooling with stride 2 applied after activation.
    \item \textbf{ReLU Activation (\texttt{real\_relu1})}: Applies the ReLU activation function element-wise to the output of \texttt{conv\_real1}.
    \item \textbf{Convolutional Layer (\texttt{conv\_real2})}: Applies a 2D convolution to the output of \texttt{real\_relu1}, increasing feature maps from 32 to 64 using a \(3 \times 3\) kernel with padding 1, followed by a \(2 \times 2\) average pooling with stride 2 applied after activation.
    \item \textbf{ReLU Activation (\texttt{real\_relu2})}: Applies the ReLU activation function element-wise to the output of \texttt{conv\_real2}.
    \item \textbf{Convolutional Layer (\texttt{conv\_imag1})}: Applies a 2D convolution to the imaginary input features, transforming them from $K'$-dimensional inputs to 32 feature maps using a \(3 \times 3\) kernel with padding 1, followed by a \(2 \times 2\) average pooling with stride 2 applied after activation.
    \item \textbf{ReLU Activation (\texttt{imag\_relu1})}: Applies the ReLU activation function element-wise to the output of \texttt{conv\_imag1}.
    \item \textbf{Convolutional Layer (\texttt{conv\_imag2})}: Applies a 2D convolution to the output of \texttt{imag\_relu1}, increasing feature maps from 32 to 64 using a \(3 \times 3\) kernel with padding 1, followed by a \(2 \times 2\) average pooling with stride 2 applied after activation.
    \item \textbf{ReLU Activation (\texttt{imag\_relu2})}: Applies the ReLU activation function element-wise to the output of \texttt{conv\_imag2}.
    \item \textbf{Fully Connected Layer (\texttt{fc1})}: Concatenates the flattened real and imaginary feature maps into a single vector and transforms it to an \(lN\)-dimensional latent space.
    \item \textbf{ReLU Activation (\texttt{fcrelu1})}: Applies the ReLU activation function element-wise to the output of \texttt{fc1}.
    \item \textbf{Fully Connected Layer (\texttt{fc2})}: Projects the \(lN\)-dimensional latent space to the final output dimension \(k\), producing the network's prediction.
\end{itemize}

\subsection{Convolutional Real-Valued Neural Network}
The real-valued neural network (RVNN) architecture is an approach for handling both real and imaginary components by concatenating them and processing them together. This architecture is evaluated on the CVNN benchmark from Section \ref{sec:CVNN_target_localization_benchmark}. Across all experiments in Section \ref{sec:CVNN_target_localization_benchmark}, we consider $lN = 128$, $k = 2$ (see below). The network has $185{,}634$ parameters.
\begin{itemize}
    \item \textbf{Convolutional Layer (\texttt{conv1})}: Applies a 2D convolution to the concatenated real and imaginary input features, transforming them from \(2K'\)-dimensional inputs to 32 feature maps using a \(3 \times 3\) kernel with padding 1, followed by a \(2 \times 2\) average pooling with stride 2 applied after activation.
    \item \textbf{ReLU Activation (\texttt{relu1})}: Applies the ReLU activation function element-wise to the output of \texttt{conv1}.
    \item \textbf{Convolutional Layer (\texttt{conv2})}: Applies a 2D convolution to the output of \texttt{relu1}, increasing feature maps from 32 to 64 using a \(3 \times 3\) kernel with padding 1, followed by a \(2 \times 2\) average pooling with stride 2 applied after activation. 
    \item \textbf{ReLU Activation (\texttt{relu2})}: Applies the ReLU activation function element-wise to the output of \texttt{conv2}.
    \item \textbf{Fully Connected Layer (\texttt{fc1})}: Flattens the pooled feature maps into a single vector, transforming it to an \(lN\)-dimensional latent space.
    \item \textbf{ReLU Activation (\texttt{fc\_relu1})}: Applies the ReLU activation function element-wise to the output of \texttt{fc1}.
    \item \textbf{Fully Connected Layer (\texttt{fc2})}: Projects the \(lN\)-dimensional latent space to the final output dimension \(k\), producing the network's prediction.
\end{itemize}

\subsection{Convolutional Complex-Valued Neural Network}
The Complex-Valued Neural Network (CVNN) is designed to handle complex-valued data by treating the real and imaginary parts jointly as complex numbers. To implement this network, we use the Complex PyTorch library \citep{maxime2021disorder}. This architecture is evaluated on the CVNN benchmark from Section \ref{sec:CVNN_target_localization_benchmark}. Across all experiments in Section \ref{sec:CVNN_target_localization_benchmark}, we consider $lN = 128$, $k = 2$ (see below). The network has $368{,}388$ parameters.
\begin{itemize}
    \item \textbf{Convolutional Layer (\texttt{complex\_conv1})}: Applies a 2D complex convolution to the complex-valued input features, transforming them from $K'$-dimensional inputs to 32 feature maps using a \(3 \times 3\) kernel with padding 1, followed by a \(2 \times 2\) average pooling with stride 2 applied after activation.
    \item \textbf{ReLU Activation (\texttt{complex\_relu1})}: Applies the complex ReLU activation function element-wise to the output of \texttt{complex\_conv1}.
    \item \textbf{Convolutional Layer (\texttt{complex\_conv2})}: Applies a 2D complex convolution to the output of \texttt{complex\_relu1}, increasing feature maps from 32 to 64 using a \(3 \times 3\) kernel with padding 1, followed by a \(2 \times 2\) complex average pooling with stride 2 applied after activation. 
    \item \textbf{ReLU Activation (\texttt{complex\_relu2})}: Applies the complex ReLU activation function element-wise to the output of \texttt{complex\_conv2}.
    \item \textbf{Fully Connected Layer (\texttt{fc1})}: Flattens the pooled feature maps into a single vector, transforming it to an \(lN\)-dimensional latent space.
    \item \textbf{ReLU Activation (\texttt{fc\_relu1})}: Applies the complex ReLU activation function element-wise to the output of \texttt{fc1}.
    \item \textbf{Fully Connected Layer (\texttt{fc2})}: Projects the \(lN\)-dimensional latent space to the final output dimension \(k\), producing the network's prediction.
\end{itemize}

\addtocontents{toc}{\protect\setcounter{tocdepth}{2}}

\newpage

\title{RASPNet: Supplementary Information}

% The \author macro works with any number of authors. There are two commands
% used to separate the names and addresses of multiple authors: \And and \AND.
%
% Using \And between authors leaves it to LaTeX to determine where to break the
% lines. Using \AND forces a line break at that point. So, if LaTeX puts 3 of 4
% authors names on the first line, and the last on the second line, try using
% \AND instead of \And before the third author name.

\author{%
Shyam Venkatasubramanian \\
Duke University\\
\And
Bosung Kang \\
Air Force Research Laboratory \\
\And % Use \AND to break to the next line of authors
Ali Pezeshki \\
Colorado State University \\
\And
Muralidhar Rangaswamy \\
Air Force Research Laboratory \\
\And
Vahid Tarokh \\
Duke University \\
}

\maketitle

\tableofcontents
\clearpage

\section{Dataset and Code Access}

\subsection{Dataset Access and Organization}
RASPNet is hosted by the United States Air Force and can be downloaded from the \href{https://www.sdms.afrl.af.mil/index.php?collection=raspnet}{AFRL Sensor Data Management System}. The dataset comprises $100$ radar scenarios from across the contiguous United States, each comprising $10{,}000$ clutter realizations. The main webpage for the dataset can be accessed at: \url{https://www.sdms.afrl.af.mil/index.php?collection=raspnet}.

The base folder in RASPNet consists of three folders: $\textbf{EXAMPLES}$, $\textbf{CVNN}$, and $\textbf{CLUTTER}$. The $\textbf{EXAMPLES}$ folder consists of data files used to generate the target localization and transfer learning results presented in the main text, the $\textbf{CVNN}$ folder consists of data files used to generate the CVNN results presented in the main text, and the $\textbf{CLUTTER}$ folder consists of $100$ subfolders, each of which pertain to a different radar scenario. These scenarios are ordered from westernmost to easternmost, such that the folder $\textbf{CLUTTER/num1\_lat\_lon/}$ is the westernmost scenario in RASPNet, and the folder $\textbf{CLUTTER/num100\_lat\_lon/}$ is the easternmost scenario in RASPNet ($\textbf{lat}$ and $\textbf{lon}$ denote the platform latitude and longitude for each scenario). This nomenclature follows from the main text, where $M = 100, S = \{1,2,\ldots,M\}$, and $i \in S$.

Each of the $100$ scenario subfolders in RASPNet contain $10{,}000$ i.i.d. clutter realizations, where each realization is stored within $\textbf{real}\{j\}\textbf{.mat}$. The realization $\textbf{CLUTTER/num}\{i\}\textbf{\_lat\_lon/real}\{j\}\textbf{.mat}$ parallels $z^i_j$ from the main text, wherein $K = 10{,}000, D = \{1,2,\ldots,K\}$, and $j \in D$. Importing the realization $\textbf{real}\{j\}\textbf{.mat}$ yields the variable \textbf{X\_clut}, which is a size $L \times \Lambda \times G$ matrix.

The $\textbf{EXAMPLES}$ folder consists of feature-labels pairs for five scenarios: $i \in \{ 29, 35, 60, 62, 76 \}$. The features for scenario $i$ are stored in $\textbf{num}\{i\}\textbf{\_NAMF\_DATA\_25k.zip}$ and the labels are stored in $\textbf{num}\{i\}\textbf{\_Ground\_Truth\_25k.csv}$. These feature-label pairs were used to generate the target localization and transfer learning results from the main text, using the code files in $\textbf{Localization\_Examples}$ and $\textbf{Transfer\_Learning\_Examples}$ (see Section \ref{sec:Code_Access}). % The Croissant metadata for the EXAMPLES dataset can be accessed at: \url{https://duke.is/j/99gb}. 

The $\textbf{CVNN}$ folder consists of feature-labels pairs from five scenarios: $i \in \{ 29, 35, 60, 62, 76 \}$. The train and test features for scenario $i$ are stored within $\textbf{num}\{i\}\textbf{.zip}$ in the folders $\textbf{num}\{i\}/\textbf{train}$ and $\textbf{num}\{i\}/\textbf{test}$, and the labels are stored in $\textbf{num}\{i\}/\textbf{train.csv}$ and $\textbf{num}\{i\}/\textbf{test.csv}$. These feature-label pairs were used to generate the CVNN target localization results from the main text using the code file $\textbf{RASPNet\_CVNN.ipynb}$ in $\textbf{Localization\_Examples}$ (see Section \ref{sec:Code_Access}).

\subsection{Code Access} \label{sec:Code_Access}
The code files for generating the empirical results provided within the main text can be found at the url: \url{https://github.com/shyamven/RASPNet}. We organize these files into two separate subfolders: $\textbf{Localization\_Examples}$ and $\textbf{Transfer\_Learning\_Examples}$. The $\textbf{Localization\_Examples}$ consist of code files for benchmarking target localization accuracy in the scenarios $i \in \{ 29, 35, 60, 62, 76 \}$. The $\textbf{Transfer\_Learning\_Examples}$ consist of example transfer learning code files, where we train a regression neural network on scenario $i = 29$, and evaluate the network for $i \in \{ 35, 60, 62, 76\}$.

\section{Datasheet} \label{sec:Datasheet}
% ===============================
\subsection{Motivation}
% ===============================

\begin{enumerate}
    \item \textbf{For what purpose was the dataset created?}
        \textit{RASPNet was created to fill a noteworthy gap in the availability of a large-scale, realistic dataset to standardize the evaluation of adaptive radar processing techniques and support the development of data-driven complex-valued learning algorithms.}
    \item \textbf{Who created the dataset and on behalf of which entity?}
        \textit{The dataset was created by a team of radar scientists and machine learning researchers provided in the author list.}
    \item \textbf{Who funded the creation of the dataset?}
        \textit{The main funding body is the United States Air Force Office of Scientific Research. Other funding sources of individual authors are listed in the Acknowledgments section.}
\end{enumerate}

% ===============================
\subsection{Distribution}
% ===============================

\begin{enumerate}
    \item \textbf{Will the dataset be distributed to third parties outside of the entity (e.g., company) on behalf of which the dataset was created?}
        \textit{Yes, the dataset is open to the public.}
    \item\textbf{How will the dataset will be distributed (e.g., tarball on website, API, GitHub)?}
        \textit{The dataset is hosted by the United States Air Force, and can be downloaded from the following link: (\href{https://www.sdms.afrl.af.mil/index.php?collection=raspnet}{https://www.sdms.afrl.af.mil/index.php?collection=raspnet}).}
    \item \textbf{Have any third parties imposed IP-based or other restrictions on the data associated with the instances?}
        \textit{No.}
    \item \textbf{Do any export controls or other regulatory restrictions apply to the dataset or to individual instances?} 
        \textit{No.}
\end{enumerate}

% ===============================
\subsection{Maintenance}
% ===============================

\begin{enumerate}
    \item \textbf{Who will be supporting/hosting/maintaining the dataset?}
        \textit{The authors, with the United States Air Force, will support, host, and maintain the dataset.}
    \item \textbf{How can the owner/curator/manager of the dataset be contacted (e.g., email address)?}
        \textit{The owner/curator/manager of the dataset can be contacted via the following emails: Shyam Venkatasubramanian (sv222@duke.edu), and Vahid Tarokh (vahid.tarokh@duke.edu).}
    \item \textbf{Is there an erratum?}
        \textit{No. If errors are found in the future, we will release errata on the main webpage for the dataset: \href{https://www.sdms.afrl.af.mil/index.php?collection=raspnet}{https://www.sdms.afrl.af.mil/index.php?collection=raspnet}.}
    \item \textbf{Will the dataset be updated (e.g., to correct labeling errors, add new instances, delete instances)?}
        \textit{Yes, the datasets will be updated whenever necessary to ensure accuracy, and announcements will be made accordingly. These updates will be posted on the main webpage for the dataset: \href{https://www.sdms.afrl.af.mil/index.php?collection=raspnet}{https://www.sdms.afrl.af.mil/index.php?collection=raspnet}.}
    \item \textbf{Do the authors bear all responsibility in case of violation of rights, and confirm the data license?)}
        \textit{Yes, the authors bear all responsibility in case of violation of rights. The dataset, RASPNet \textcopyright \ 2024 by Shyam Venkatasubramanian, Bosung Kang, Ali Pezeshki, Muralidhar Rangaswamy, Vahid Tarokh is licensed under CC BY 4.0. }
    \item \textbf{Will older version of the dataset continue to be supported/hosted/maintained?}
        \textit{Yes, older versions of the dataset will continue to be maintained and hosted.}
    \item \textbf{If others want to extend/augment/build on/contribute to the dataset, is there a mechanisms for them to do so?}
        \textit{No.}
\end{enumerate}

% ===============================
\subsection{Composition}
% ===============================

\begin{enumerate}
    \item \textbf{What do the instance that comprise the dataset represent (e.g., documents, photos, people, countries?)}
        \textit{RASPNet comprises $100$ airborne radar scenarios from across the contiguous United States. Each scenario consists of $10{,}000$ clutter realizations, which are stored as multidimensional matrices (.mat files).}
    \item \textbf{How many instances are there in total (of each type, if appropriate)?}
        \textit{There are $1{,}000{,}000$ total clutter realizations present within RASPNet, amounting to more than $16$ TB of data.}
    \item \textbf{Does the dataset contain all possible instances or is it a sample of instances from a larger set?}
        \textit{The dataset contains all possible instances of clutter realizations.}
    \item \textbf{Is there a label or target associated with each instance?}
        \textit{The example code files provided alongside RASPNet have associated data files, which are stored in separate folders in RASPNet. These data files include feature-label pairs.}
    \item \textbf{Is any information missing from individual instances?}
        \textit{No.}
    \item \textbf{Are there recommended data splits (e.g., training, development/validation, testing)?} 
        \textit{We do not have specific recommendations on the split within the training/validation set.}
    \item \textbf{Are there any errors, sources of noise, or redundancies in the dataset?}
        \textit{RASPNet was generated using the RFView\textsuperscript{\tiny\textregistered} modeling \& simulation software, which incorporates thermal noise and other sources into the data generation process.}
    \item \textbf{Is the dataset self-contained, or does it link to or otherwise rely on external resources (e.g., websites, tweets, other datasets)?}
        \textit{The dataset is self-contained.}
    \item \textbf{Does the dataset contain data that might be considered confidential?}
        \textit{No.}
\end{enumerate}

% ===============================
\subsection{Collection Process}
% ===============================

\begin{enumerate}
    \item \textbf{How was the data associated with each instance acquired?}
        \textit{The data associated with each realization was produced using the RFView\textsuperscript{\tiny\textregistered} modeling and simulation software.}
    \item \textbf{What mechanisms or procedures were used to collect the data (e.g., hardware apparatus or sensor, manual human curation, software program, software API)? }
        \textit{We used NVIDIA GeForce RTX 2080 and NVIDIA GeForce RTX 3090 GPU nodes in a high-performance computing cluster to run the RFView\textsuperscript{\tiny\textregistered} simulations.}
    \item \textbf{Who was involved in the data collection process (e.g., students, crowdworkers, contractors) and how were they compensated (e.g., how much were crowdworkers paid)?}
        \textit{Researchers at Duke University were involved in the data collection process. No crowdworkers were involved during the data collection process.}
    \item \textbf{Did you collect the data from the individuals in questions directly, or obtain it via third parties or other sources (e.g., websites)?}
        \textit{We obtained the dataset using the RFView\textsuperscript{\tiny\textregistered} modeling and simulation software.}
\end{enumerate}

% ===============================
\subsection{Uses}
% ===============================

\begin{enumerate}
    \item \textbf{Has the dataset been used for any tasks already?}
        \textit{No, this dataset has not been used for any tasks yet, apart from the examples provided in the main text.}
    % \item \textbf{What (other) tasks could be the dataset be used for?}
    %     \textit{Apart from target localization, RASPNet also provides a testbed-like environment for the benchmarking and validation of candidate adaptive radar target detection and classification methodologies.}
    \item \textbf{Is there anything about the composition of the dataset or the way it was collected and preprocessed/cleaned/labeled that might impact future uses?}
        \textit{The current composition of the dataset is self-sufficient, and any changes and updates made to the dataset will be documented via the dataset webpage.} %: \url{https://shyamven.github.io/RASPNet}.}
    \item \textbf{Are there tasks for which the dataset should not be used?}
        \textit{No.}
\end{enumerate}

\section{Discussion of RASPNet Scenarios} \label{sec:raspnet_scenario_discussion}
We provide a complete tabulation of the scenarios comprising RASPNet, along with short descriptions, in this section. The scenarios are indexed by $q_i \in Q$ (difficulty), $i \in S$ (westernmost to easternmost), and the airborne radar platform parameters are provided in Table \ref{tab:radar_parameters_complete}.

\subsection{Radar Platform Parameters} \label{sec:radar_platform_parameters}
The full list of parameters pertaining to the airborne radar platform from the main text is provided below. This radar platform is used to obtain the clutter dataset for each scenario comprising RASPNet. We note that the radar processing region for each scenario is configured such that the aim location is halfway between the near range and far range, which is half of the radar range swath.
\begin{table}[h!]
\caption{Complete List of Radar Platform Parameters}
\label{tab:radar_parameters_complete}
\centering
\begin{tabular}{l|l}
\hline
\textbf{Parameter} & \textbf{Value} \\
\hline
Carrier frequency $(f_c)$ & $10\,\text{GHz}$ \\
Bandwidth $(B)$ & $5\,\text{MHz}$ \\
Pulse Repetition Frequency $(f_{PR})$ & $1100\,\text{Hz}$ \\
Radar platform altitude $(h)$ & $1000\,\text{m}$ \\
Range resolution $(\Delta r)$ & $30\,\text{m}$ \\
Antenna element configuration \{horizontal $\times$ vertical\} & $48 \times 5$ \\
Number of receiver channels $(L)$ \{horizontal $\times$ vertical\} & $16 \times 1$ \\
Antenna element spacing $(\Delta k)$ & $0.015\,\text{m}$ \\
Number of pulses $(\Lambda)$ & $64$ \\
Transmitted waveform number of samples & $455$ \\
Total number of range bins $(G)$ & $1134$ \\
Radar range swath & $20372\,\text{m}$
\end{tabular}
\end{table}

\subsection{Basic Scenarios}
We first outline the Basic RASPNet scenarios, where $q_i \in \{1,2,\ldots,25\}$.

\paragraph{\underline{Scenario $q_i = 1$}: Bonneville Salt Flats, UT ($i = 29$)} \vphantom{ }
\begin{table}[h!]
\centering
\setlength{\fboxsep}{0pt}
\setlength{\fboxrule}{2pt}
\fbox{%
\begin{tabular}{c|>{\arraybackslash}m{0.615\textwidth}}
\hline
\multirow{3}{*}{\raisebox{-\totalheight}{\includegraphics[width=0.5\textwidth,height=5.1cm,keepaspectratio]{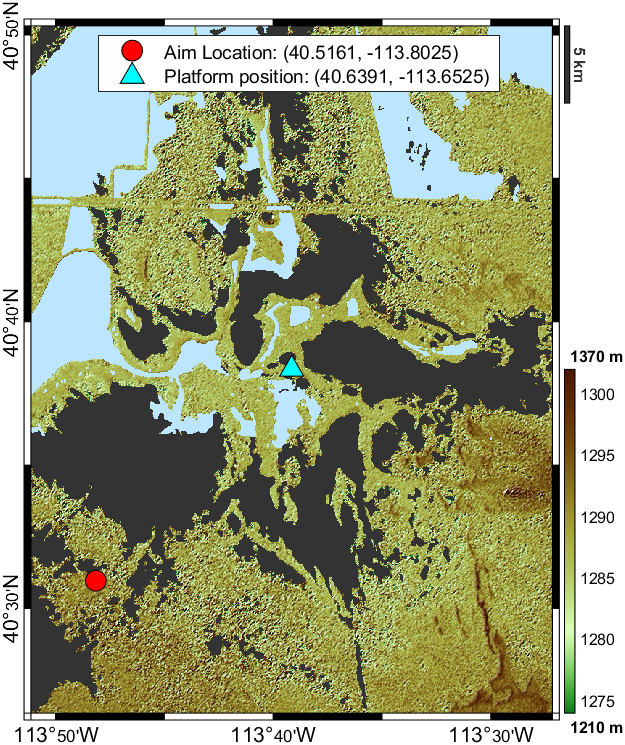}}} \\[-0.1cm] & \textbf{Platform Aim Location} (lat, lon): $(40.5161^{\circ}, -113.8025^{\circ})$ \\[0.2cm]
\cline{2-2} \\[-0.1cm] & 
\textbf{Platform Position} (lat, lon): $(40.6391^{\circ}, -113.6525^{\circ})$ \\[0.2cm]
\cline{2-2} \\[-0.1cm] & 
\textbf{Folder}: \textbf{/CLUTTER/num29\_lat40.6391\_lon-113.6525/} \\[0.2cm]
\cline{2-2} \\[-0.1cm] &
\parbox[t][2.6cm][t]{0.62\textwidth}{\textbf{Description}: Scenario $q_i = 1 \ (i = 29)$ in RASPNet is in the Bonneville Salt Flats region of Utah. The radar platform is positioned at latitude $40.6391^{\circ}$ N and longitude $113.6525^{\circ}$ W. It aims towards a location at latitude $40.5161^{\circ}$ N and longitude $113.8025^{\circ}$ W. This region is known for its minimal topographic relief, clear line of sight, and flat crust of salty soil.} \\
\hline
\end{tabular}
}
\end{table}

\paragraph{\underline{Scenario $q_i = 2$}: Lake Michigan Coast, MI ($i = 72$)} \vphantom{ }
\begin{table}[h!]
\centering
\setlength{\fboxsep}{0pt}
\setlength{\fboxrule}{2pt}
\fbox{%
\begin{tabular}{c|>{\arraybackslash}m{0.615\textwidth}}
\hline
\multirow{3}{*}{\raisebox{-\totalheight}{\includegraphics[width=0.5\textwidth,height=5.1cm,keepaspectratio]{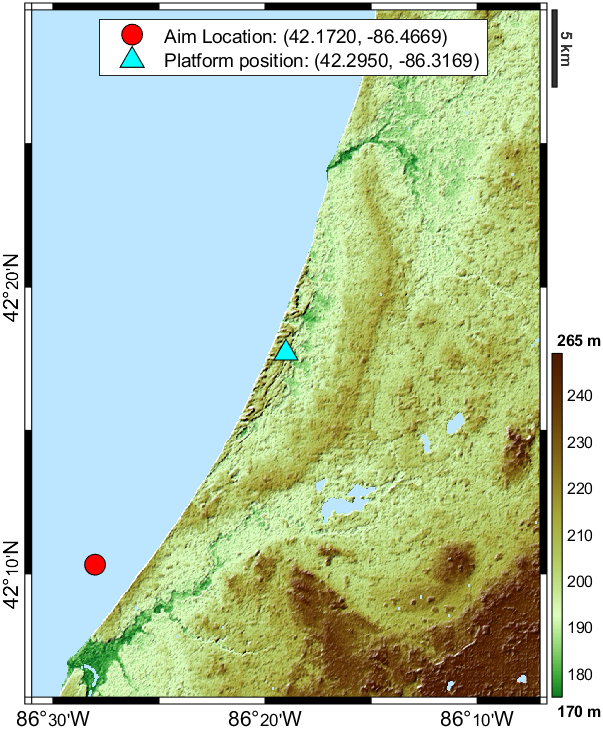}}} \\[-0.1cm] & \textbf{Platform Aim Location} (lat, lon): $(42.1720^{\circ}, -86.4669^{\circ})$ \\[0.2cm]
\cline{2-2} \\[-0.1cm] & 
\textbf{Platform Position} (lat, lon): $(42.2950^{\circ}, -86.3169^{\circ})$ \\[0.2cm]
\cline{2-2} \\[-0.1cm] & 
\textbf{Folder}: \textbf{/CLUTTER/num72\_lat42.295\_lon-86.3169/} \\[0.2cm]
\cline{2-2} \\[-0.1cm] &
\parbox[t][2.6cm][t]{0.62\textwidth}{\textbf{Description}: Scenario $q_i = 2 \ (i = 72)$ in RASPNet is on the Lake Michigan coastline in Michigan. The radar platform is positioned at latitude $42.2950^{\circ}$ N and longitude $86.3169^{\circ}$ W. It aims towards a location at latitude $42.1720^{\circ}$ N and longitude $86.4669^{\circ}$ W. This is a land-to-water transition region with a flat coastline, where the aim location is on the water surface.} \\
\hline
\end{tabular}
}
\end{table}

\paragraph{\underline{Scenario $q_i = 3$}: Charleston Harbor, SC ($i = 85$)} \vphantom{ }
\begin{table}[h!]
\centering
\setlength{\fboxsep}{0pt}
\setlength{\fboxrule}{2pt}
\fbox{%
\begin{tabular}{c|>{\arraybackslash}m{0.615\textwidth}}
\hline
\multirow{3}{*}{\raisebox{-\totalheight}{\includegraphics[width=0.5\textwidth,height=5.1cm,keepaspectratio]{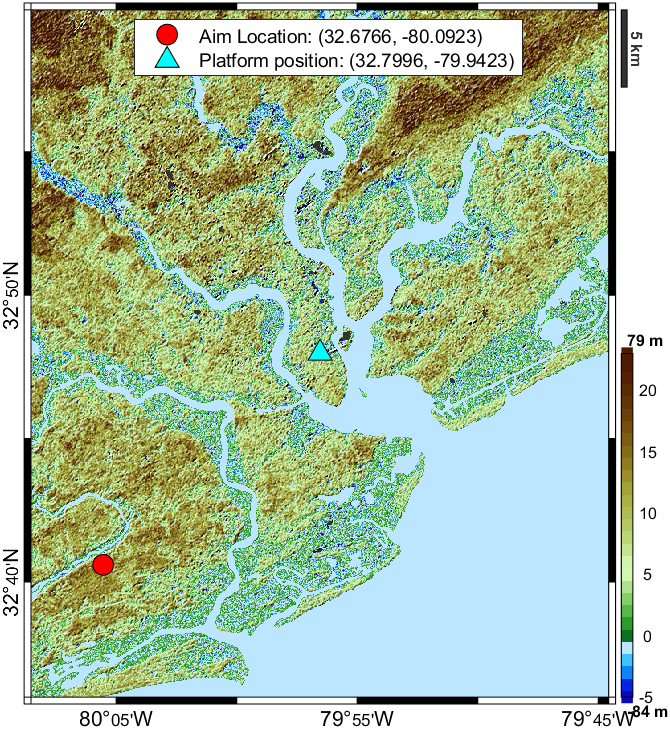}}} \\[-0.1cm] & \textbf{Platform Aim Location} (lat, lon): $(32.6766^{\circ}, -80.0932^{\circ})$ \\[0.2cm]
\cline{2-2} \\[-0.1cm] & 
\textbf{Platform Position} (lat, lon): $(32.7996^{\circ}, -79.9423^{\circ})$ \\[0.2cm]
\cline{2-2} \\[-0.1cm] & 
\textbf{Folder}: \textbf{/CLUTTER/num85\_lat32.7996\_lon-79.9423/} \\[0.2cm]
\cline{2-2} \\[-0.1cm] &
\parbox[t][2.6cm][t]{0.62\textwidth}{\textbf{Description}: Scenario $q_i = 3 \ (i = 85)$ in RASPNet is in the Charleston Harbor in South Carolina. The radar platform is positioned at latitude $32.7996^{\circ}$ N and longitude $79.9423^{\circ}$ W. It aims towards a location at latitude $32.6766^{\circ}$ N and longitude $80.0932^{\circ}$ W. This is a land-to-water transition region with a flat coastline, surrounding the Cooper River estuary.} \\
\hline
\end{tabular}
}
\end{table}

\paragraph{\underline{Scenario $q_i = 4$}: Central Kansas, KS ($i = 60$)} \vphantom{ }
\begin{table}[h!]
\centering
\setlength{\fboxsep}{0pt}
\setlength{\fboxrule}{2pt}
\fbox{%
\begin{tabular}{c|>{\arraybackslash}m{0.615\textwidth}}
\hline
\multirow{3}{*}{\raisebox{-\totalheight}{\includegraphics[width=0.5\textwidth,height=5.3cm,keepaspectratio]{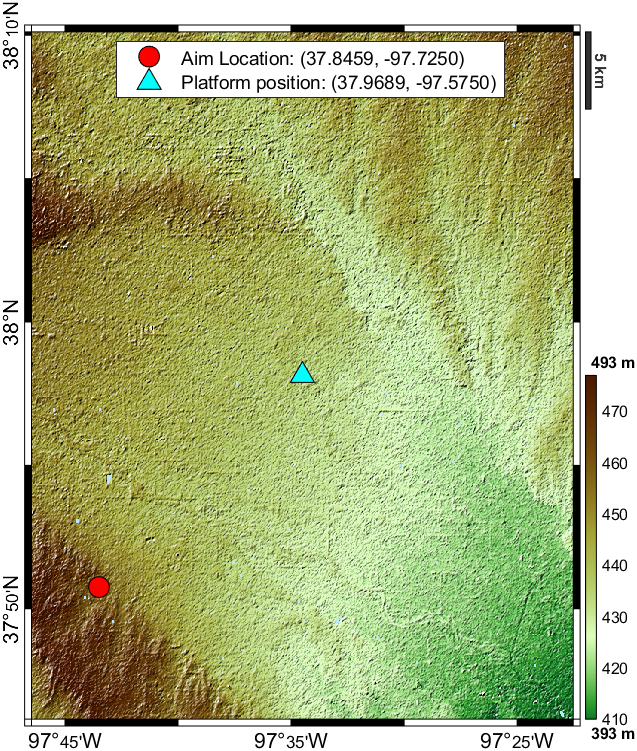}}} \\[-0.1cm] & \textbf{Platform Aim Location} (lat, lon): $(37.8459^{\circ}, -97.7250^{\circ})$ \\[0.2cm]
\cline{2-2} \\[-0.1cm] & 
\textbf{Platform Position} (lat, lon): $(37.9689^{\circ}, -97.5750^{\circ})$ \\[0.2cm]
\cline{2-2} \\[-0.1cm] & 
\textbf{Folder}: \textbf{/CLUTTER/num60\_lat37.9689\_lon-97.575/} \\[0.2cm]
\cline{2-2} \\[-0.1cm] &
\parbox[t][3.0cm][t]{0.62\textwidth}{\textbf{Description}: Scenario $q_i = 4 \ (i = 60)$ in RASPNet is in the Great Plains region of Central Kansas. The radar platform is positioned at latitude $37.9689^{\circ}$ N and longitude $-97.5750^{\circ}$ W. It aims towards a location at latitude $37.8459^{\circ}$ N and longitude $-97.7250^{\circ}$ W. This region is characterized by an expansive and flat grassland environment, with low terrain variability.} \\
\hline
\end{tabular}
}
\end{table}

\paragraph{\underline{Scenario $q_i = 5$}: Texas Panhandle, TX ($i = 56$)} \vphantom{ }
\begin{table}[h!]
\centering
\setlength{\fboxsep}{0pt}
\setlength{\fboxrule}{2pt}
\fbox{%
\begin{tabular}{c|>{\arraybackslash}m{0.615\textwidth}}
\hline
\multirow{3}{*}{\raisebox{-\totalheight}{\includegraphics[width=0.5\textwidth,height=5.3cm,keepaspectratio]{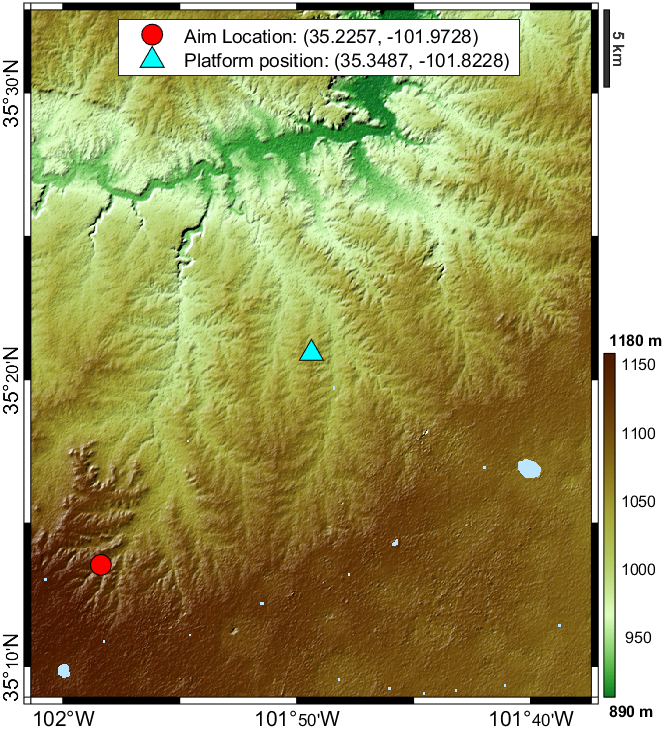}}} \\[-0.1cm] & \textbf{Platform Aim Location} (lat, lon): $(35.2257^{\circ}, -101.9728^{\circ})$ \\[0.2cm]
\cline{2-2} \\[-0.1cm] & 
\textbf{Platform Position} (lat, lon): $(35.3487^{\circ}, -101.8228^{\circ})$ \\[0.2cm]
\cline{2-2} \\[-0.1cm] & 
\textbf{Folder}: \textbf{/CLUTTER/num56\_lat35.3487\_lon-101.8228/} \\[0.2cm]
\cline{2-2} \\[-0.1cm] &
\parbox[t][3.0cm][t]{0.62\textwidth}{\textbf{Description}: Scenario $q_i = 5 \ (i = 56)$ in RASPNet is in the Great Plains region of panhandle Texas. The radar platform is positioned at latitude $35.3487^{\circ}$ N and longitude $101.8228^{\circ}$ W. It aims towards a location at latitude $35.2257^{\circ}$ N and longitude $101.9728^{\circ}$ W. This region is characterized by an expansive and flat grassland environment, with low terrain variability.} \\
\hline
\end{tabular}
}
\end{table}

\paragraph{\underline{Scenario $q_i = 6$}: Central Arkansas, AR ($i = 64$)} \vphantom{ }
\begin{table}[h!]
\centering
\setlength{\fboxsep}{0pt}
\setlength{\fboxrule}{2pt}
\fbox{%
\begin{tabular}{c|>{\arraybackslash}m{0.615\textwidth}}
\hline
\multirow{3}{*}{\raisebox{-\totalheight}{\includegraphics[width=0.5\textwidth,height=5.3cm,keepaspectratio]{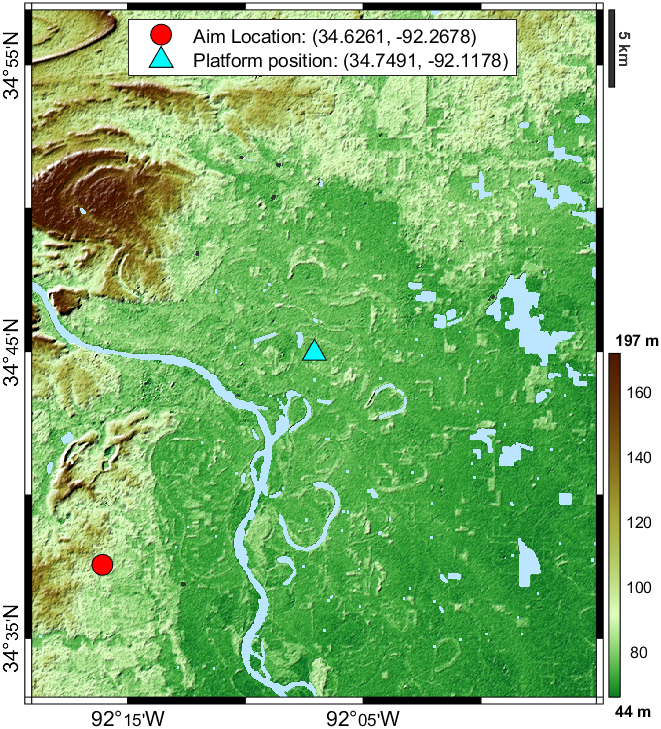}}} \\[-0.1cm] & \textbf{Platform Aim Location} (lat, lon): $(34.6261^{\circ}, -92.2678^{\circ})$ \\[0.2cm]
\cline{2-2} \\[-0.1cm] & 
\textbf{Platform Position} (lat, lon): $(34.7491^{\circ}, -92.1178^{\circ})$ \\[0.2cm]
\cline{2-2} \\[-0.1cm] & 
\textbf{Folder}: \textbf{/CLUTTER/num64\_lat34.7491\_lon-92.1178/} \\[0.2cm]
\cline{2-2} \\[-0.1cm] &
\parbox[t][3.0cm][t]{0.62\textwidth}{\textbf{Description}: Scenario $q_i = 6 \ (i = 64)$ in RASPNet is in the Arkansas River Valley of Central Arkansas. The radar platform is positioned at latitude $34.7491^{\circ}$ N and longitude $92.1178^{\circ}$ W. It aims towards a location at latitude $34.6261^{\circ}$ N and longitude $92.2678^{\circ}$ W. This region is characterized by flat lowlands and fertile farmlands, with low terrain variability.} \\
\hline
\end{tabular}
}
\end{table}

\paragraph{\underline{Scenario $q_i = 7$}: Okefenokee Swamp, GA ($i = 81$)} \vphantom{ }
\begin{table}[h!]
\centering
\setlength{\fboxsep}{0pt}
\setlength{\fboxrule}{2pt}
\fbox{%
\begin{tabular}{c|>{\arraybackslash}m{0.615\textwidth}}
\hline
\multirow{3}{*}{\raisebox{-\totalheight}{\includegraphics[width=0.5\textwidth,height=5.3cm,keepaspectratio]{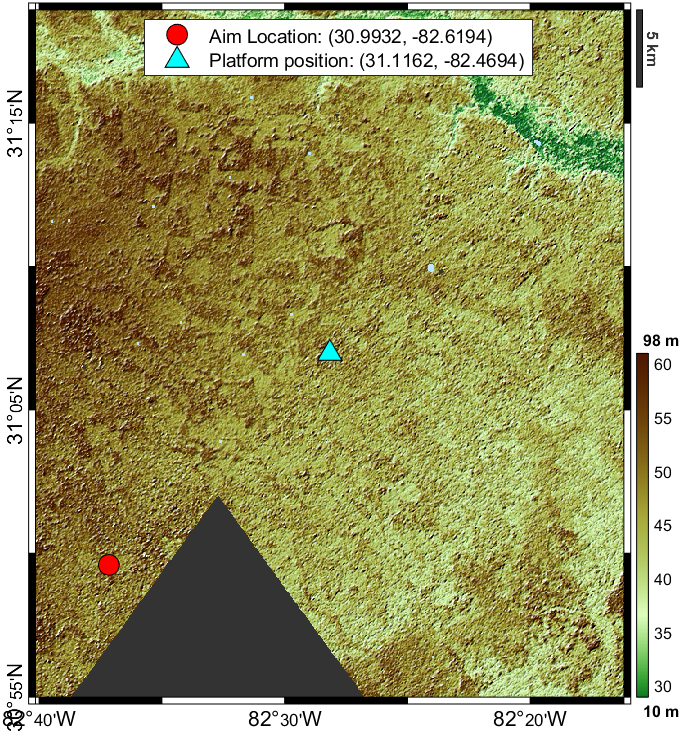}}} \\[-0.1cm] & \textbf{Platform Aim Location} (lat, lon): $(30.9932^{\circ}, -82.6194^{\circ})$ \\[0.2cm]
\cline{2-2} \\[-0.1cm] & 
\textbf{Platform Position} (lat, lon): $(31.1162^{\circ}, -82.4694^{\circ})$ \\[0.2cm]
\cline{2-2} \\[-0.1cm] & 
\textbf{Folder}: \textbf{/CLUTTER/num81\_lat31.1162\_lon-82.4694/} \\[0.2cm]
\cline{2-2} \\[-0.1cm] &
\parbox[t][3.0cm][t]{0.62\textwidth}{\textbf{Description}: Scenario $q_i = 7 \ (i = 81)$ in RASPNet is in the Okefenokee Swamp of Southern Georgia. The radar platform is positioned at latitude $31.1162^{\circ}$ N and longitude $82.4694^{\circ}$ W. It aims towards a location at latitude $30.9932^{\circ}$ N and longitude $82.6194^{\circ}$ W. This region is characterized by a shallow and flat wetland environment, with low terrain variability.} \\
\hline
\end{tabular}
}
\end{table}

\paragraph{\underline{Scenario $q_i = 8$}: Lake Erie Coast, OH ($i = 78$)} \vphantom{ }
\begin{table}[h!]
\centering
\setlength{\fboxsep}{0pt}
\setlength{\fboxrule}{2pt}
\fbox{%
\begin{tabular}{c|>{\arraybackslash}m{0.615\textwidth}}
\hline
\multirow{3}{*}{\raisebox{-\totalheight}{\includegraphics[width=0.5\textwidth,height=5.3cm,keepaspectratio]{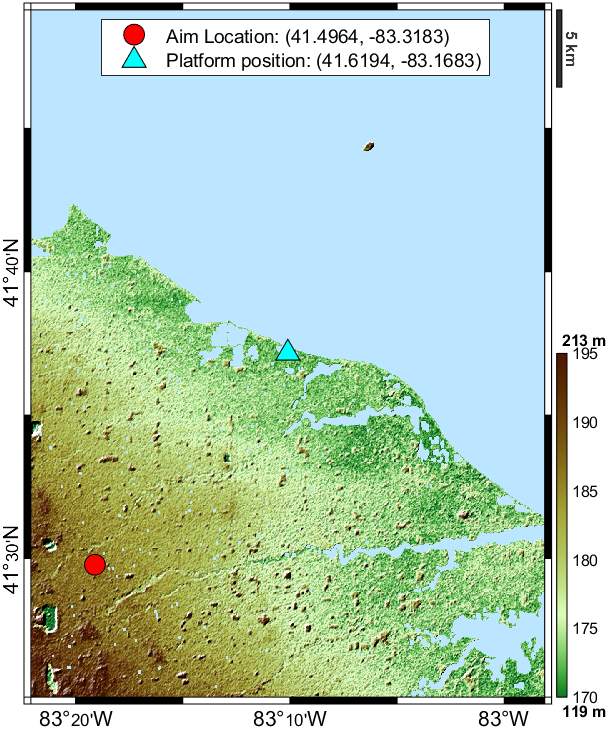}}} \\[-0.1cm] & \textbf{Platform Aim Location} (lat, lon): $(41.4964^{\circ}, -83.3183^{\circ})$ \\[0.2cm]
\cline{2-2} \\[-0.1cm] & 
\textbf{Platform Position} (lat, lon): $(41.6194^{\circ}, -83.1683^{\circ})$ \\[0.2cm]
\cline{2-2} \\[-0.1cm] & 
\textbf{Folder}: \textbf{/CLUTTER/num78\_lat41.6194\_lon-83.1683/} \\[0.2cm]
\cline{2-2} \\[-0.1cm] &
\parbox[t][3.0cm][t]{0.62\textwidth}{\textbf{Description}: Scenario $q_i = 8 \ (i = 78)$ in RASPNet is on the Lake Erie coastline in Northern Ohio. The radar platform is positioned at latitude $41.6194^{\circ}$ N and longitude $83.1683^{\circ}$ W. It aims towards a location at latitude $41.4964^{\circ}$ N and longitude $83.3183^{\circ}$ W. This is a land-to-water transition region with a flat coastline, where the aim location is on the land surface.} \\
\hline
\end{tabular}
}
\end{table}

\paragraph{\underline{Scenario $q_i = 9$}: Algodones Dunes, CA ($i = 26$)} \vphantom{ }
\begin{table}[h!]
\centering
\setlength{\fboxsep}{0pt}
\setlength{\fboxrule}{2pt}
\fbox{%
\begin{tabular}{c|>{\arraybackslash}m{0.615\textwidth}}
\hline
\multirow{3}{*}{\raisebox{-\totalheight}{\includegraphics[width=0.5\textwidth,height=5.3cm,keepaspectratio]{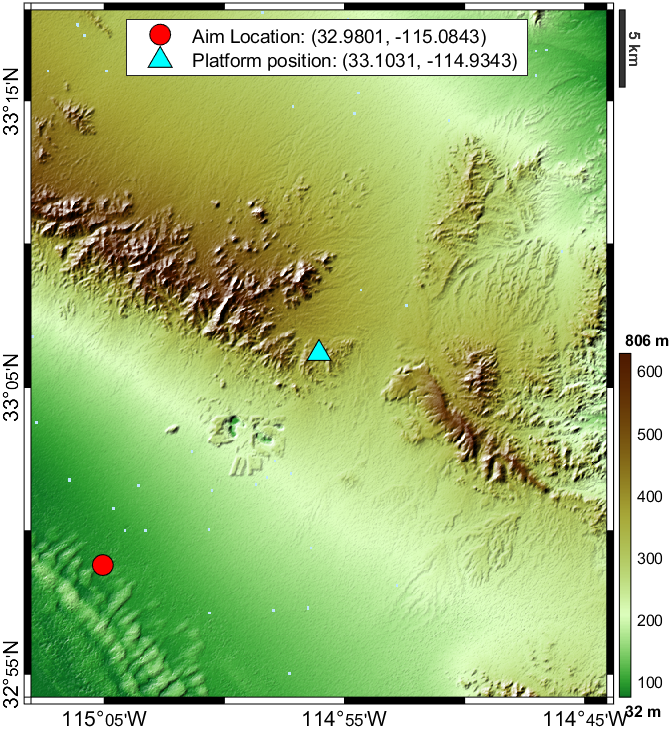}}} \\[-0.1cm] & \textbf{Platform Aim Location} (lat, lon): $(32.9801^{\circ}, -115.0843^{\circ})$ \\[0.2cm]
\cline{2-2} \\[-0.1cm] & 
\textbf{Platform Position} (lat, lon): $(33.1031^{\circ}, -114.9343^{\circ})$ \\[0.2cm]
\cline{2-2} \\[-0.1cm] & 
\textbf{Folder}: \textbf{/CLUTTER/num26\_lat33.1031\_lon-114.9343/} \\[0.2cm]
\cline{2-2} \\[-0.1cm] &
\parbox[t][3.0cm][t]{0.62\textwidth}{\textbf{Description}: Scenario $q_i = 9 \ (i = 26)$ in RASPNet is in the Algodones Dunes region of California. The radar platform is positioned at latitude $33.1031^{\circ}$ N and longitude $114.9343^{\circ}$ W. It aims towards a location at latitude $32.9801^{\circ}$ N and longitude $115.0843^{\circ}$ W. This region is characterized by a uniform desert environment, where the aim location is at the dune foothills.} \\
\hline
\end{tabular}
}
\end{table}

\paragraph{\underline{Scenario $q_i = 10$}: Southern Indiana, IN ($i = 71$)} \vphantom{ }
\begin{table}[h!]
\centering
\setlength{\fboxsep}{0pt}
\setlength{\fboxrule}{2pt}
\fbox{%
\begin{tabular}{c|>{\arraybackslash}m{0.615\textwidth}}
\hline
\multirow{3}{*}{\raisebox{-\totalheight}{\includegraphics[width=0.5\textwidth,height=5.3cm,keepaspectratio]{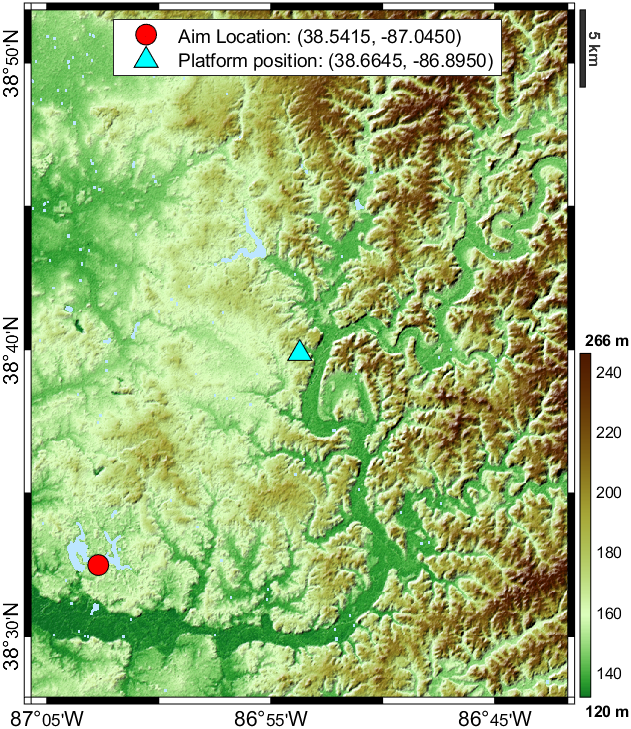}}} \\[-0.1cm] & \textbf{Platform Aim Location} (lat, lon): $(38.5415^{\circ}, -87.0450^{\circ})$ \\[0.2cm]
\cline{2-2} \\[-0.1cm] & 
\textbf{Platform Position} (lat, lon): $(38.6645^{\circ}, -86.8950^{\circ})$ \\[0.2cm]
\cline{2-2} \\[-0.1cm] & 
\textbf{Folder}: \textbf{/CLUTTER/num71\_lat38.6645\_lon-86.895/} \\[0.2cm]
\cline{2-2} \\[-0.1cm] &
\parbox[t][3.0cm][t]{0.62\textwidth}{\textbf{Description}: Scenario $q_i = 10 \ (i = 71)$ in RASPNet is in the Wabash Valley of Southern Indiana. The radar platform is positioned at latitude $38.6645^{\circ}$ N and longitude $86.8950^{\circ}$ W. It aims towards a location at latitude $38.5415^{\circ}$ N and longitude $87.0450^{\circ}$ W. This region is characterized by a uniform river valley environment with low elevation variability.} \\
\hline
\end{tabular}
}
\end{table}

\paragraph{\underline{Scenario $q_i = 11$}: Connecticut Coast, CT ($i = 95$)} \vphantom{ }
\begin{table}[h!]
\centering
\setlength{\fboxsep}{0pt}
\setlength{\fboxrule}{2pt}
\fbox{%
\begin{tabular}{c|>{\arraybackslash}m{0.615\textwidth}}
\hline
\multirow{3}{*}{\raisebox{-\totalheight}{\includegraphics[width=0.5\textwidth,height=5.3cm,keepaspectratio]{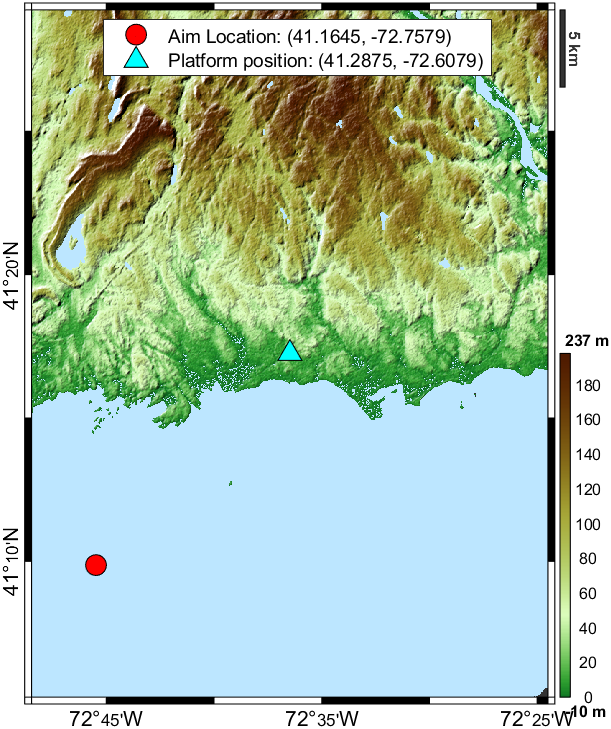}}} \\[-0.1cm] & \textbf{Platform Aim Location} (lat, lon): $(41.1645^{\circ}, -72.7579^{\circ})$ \\[0.2cm]
\cline{2-2} \\[-0.1cm] & 
\textbf{Platform Position} (lat, lon): $(41.2875^{\circ}, -72.6079^{\circ})$ \\[0.2cm]
\cline{2-2} \\[-0.1cm] & 
\textbf{Folder}: \textbf{/CLUTTER/num95\_lat41.2875\_lon-72.6079/} \\[0.2cm]
\cline{2-2} \\[-0.1cm] &
\parbox[t][3.0cm][t]{0.62\textwidth}{\textbf{Description}: Scenario $q_i = 11 \ (i = 95)$ in RASPNet is on the Atlantic coastline of Connecticut. The radar platform is positioned at latitude $41.2875^{\circ}$ N and longitude $72.6079^{\circ}$ W. It aims towards a location at latitude $41.1645^{\circ}$ N and longitude $72.7579^{\circ}$ W. This is a land-to-water transition region with a flat coastline, where the aim location is on the water surface. } \\
\hline
\end{tabular}
}
\end{table}

\paragraph{\underline{Scenario $q_i = 12$}: Southeastern New Mexico, NM ($i = 51$)} \vphantom{ }
\begin{table}[h!]
\centering
\setlength{\fboxsep}{0pt}
\setlength{\fboxrule}{2pt}
\fbox{%
\begin{tabular}{c|>{\arraybackslash}m{0.615\textwidth}}
\hline
\multirow{3}{*}{\raisebox{-\totalheight}{\includegraphics[width=0.5\textwidth,height=5.3cm,keepaspectratio]{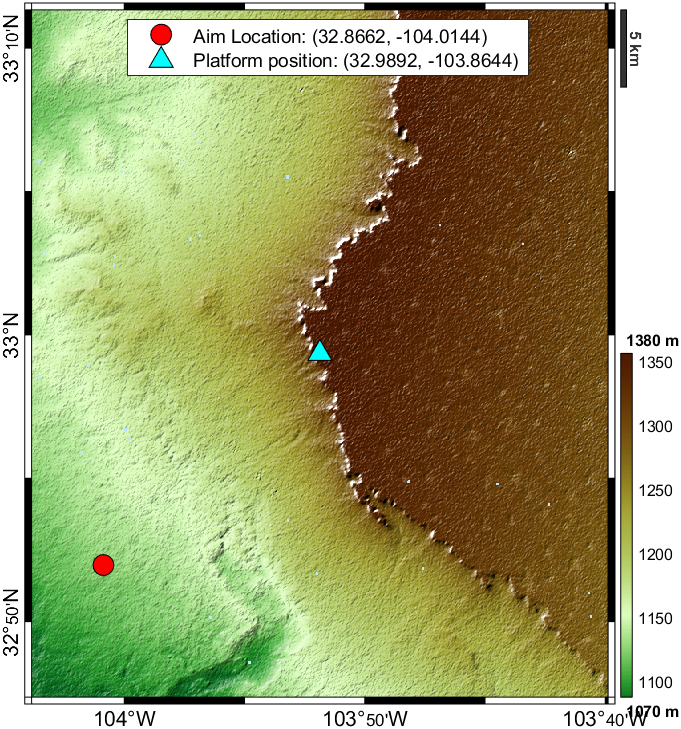}}} \\[-0.1cm] & \textbf{Platform Aim Location} (lat, lon): $(32.8662^{\circ}, -104.0144^{\circ})$ \\[0.2cm]
\cline{2-2} \\[-0.1cm] & 
\textbf{Platform Position} (lat, lon): $(32.9892^{\circ}, -103.8644^{\circ})$ \\[0.2cm]
\cline{2-2} \\[-0.1cm] & 
\textbf{Folder}: \textbf{/CLUTTER/num51\_lat32.9892\_lon-103.8644/} \\[0.2cm]
\cline{2-2} \\[-0.1cm] &
\parbox[t][3.0cm][t]{0.62\textwidth}{\textbf{Description}: Scenario $q_i = 12 \ (i = 51)$ in RASPNet is in the plains of Southeastern New Mexico. The radar platform is positioned at latitude $32.9892^{\circ}$ N and longitude $103.8644^{\circ}$ W. It aims towards a location at latitude $32.8662^{\circ}$ N and longitude $104.0144^{\circ}$ W. The radar platform is located above an escarpment overlooking and aiming towards a flat steppe region. } \\
\hline
\end{tabular}
}
\end{table}

\paragraph{\underline{Scenario $q_i = 13$}: San Joaquin Valley, CA ($i = 11$)} \vphantom{ }
\begin{table}[h!]
\centering
\setlength{\fboxsep}{0pt}
\setlength{\fboxrule}{2pt}
\fbox{%
\begin{tabular}{c|>{\arraybackslash}m{0.615\textwidth}}
\hline
\multirow{3}{*}{\raisebox{-\totalheight}{\includegraphics[width=0.5\textwidth,height=5.3cm,keepaspectratio]{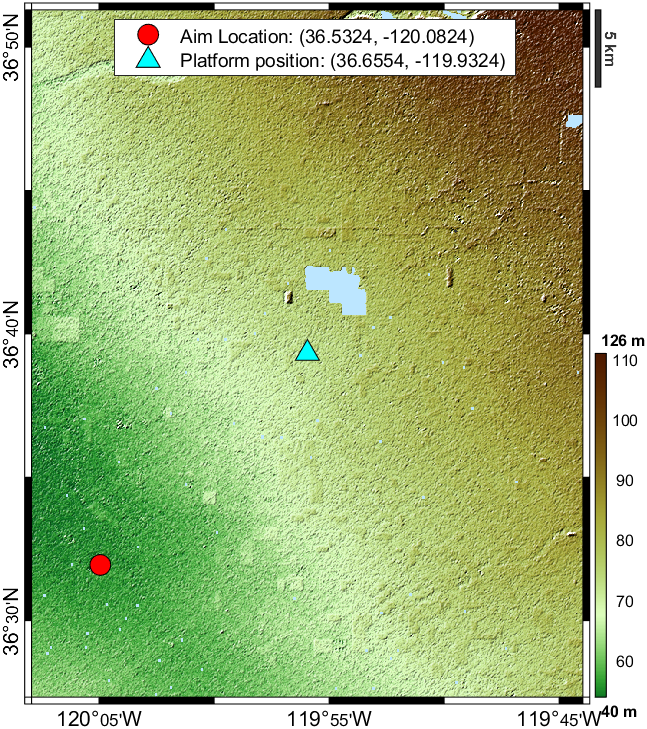}}} \\[-0.1cm] & \textbf{Platform Aim Location} (lat, lon): $(36.5324^{\circ}, -120.0824^{\circ})$ \\[0.2cm]
\cline{2-2} \\[-0.1cm] & 
\textbf{Platform Position} (lat, lon): $(36.6554^{\circ}, -119.9324^{\circ})$ \\[0.2cm]
\cline{2-2} \\[-0.1cm] & 
\textbf{Folder}: \textbf{/CLUTTER/num11\_lat36.6554\_lon-119.9324/} \\[0.2cm]
\cline{2-2} \\[-0.1cm] &
\parbox[t][3.0cm][t]{0.62\textwidth}{\textbf{Description}: Scenario $q_i = 13 \ (i = 11)$ in RASPNet is in the San Joaquin Valley of Central California. The radar platform is positioned at latitude $36.6554^{\circ}$ N and longitude $119.9324^{\circ}$ W. It aims towards a location at latitude $36.5324^{\circ}$ N and longitude $120.0824^{\circ}$ W. This region consists of a flat river valley, with fertile farmlands and low terrain variability. } \\
\hline
\end{tabular}
}
\end{table}

\paragraph{\underline{Scenario $q_i = 14$}: San Francisco Bay, CA ($i = 3$)} \vphantom{ }
\begin{table}[h!]
\centering
\setlength{\fboxsep}{0pt}
\setlength{\fboxrule}{2pt}
\fbox{%
\begin{tabular}{c|>{\arraybackslash}m{0.615\textwidth}}
\hline
\multirow{3}{*}{\raisebox{-\totalheight}{\includegraphics[width=0.5\textwidth,height=5.3cm,keepaspectratio]{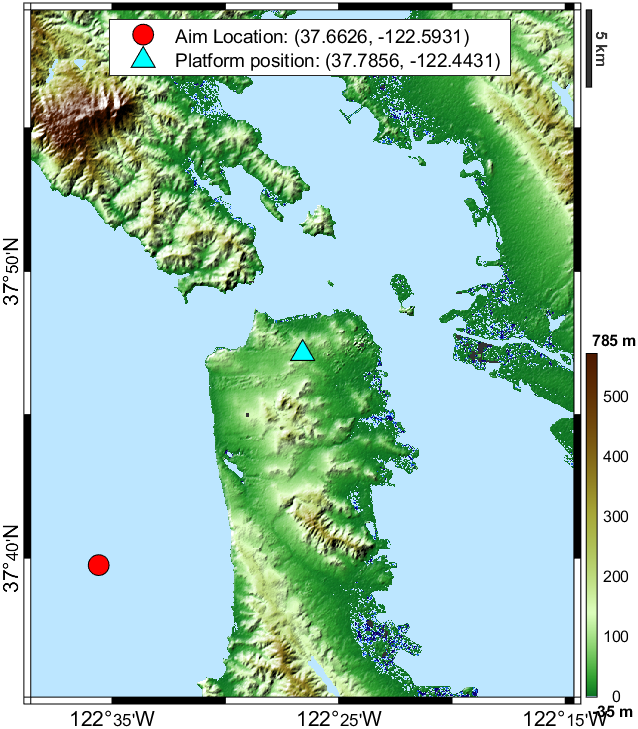}}} \\[-0.1cm] & \textbf{Platform Aim Location} (lat, lon): $(37.6626^{\circ}, -122.5931^{\circ})$ \\[0.2cm]
\cline{2-2} \\[-0.1cm] & 
\textbf{Platform Position} (lat, lon): $(37.7856^{\circ}, -122.4431^{\circ})$ \\[0.2cm]
\cline{2-2} \\[-0.1cm] & 
\textbf{Folder}: \textbf{/CLUTTER/num3\_lat37.7856\_lon-122.4431/} \\[0.2cm]
\cline{2-2} \\[-0.1cm] &
\parbox[t][3.0cm][t]{0.62\textwidth}{\textbf{Description}: Scenario $q_i = 14 \ (i = 3)$ in RASPNet is in the San Francisco Bay area of California. The radar platform is positioned at latitude $37.7856^{\circ}$ N and longitude $122.4431^{\circ}$ W. It aims towards a location at latitude $37.6626^{\circ}$ N and longitude $122.5931^{\circ}$ W. This is a land-to-water transition region, where the aim direction is over a flatter section of coastline. } \\
\hline
\end{tabular}
}
\end{table}

\paragraph{\underline{Scenario $q_i = 15$}: Lower Mississippi River, MS ($i = 68$)} \vphantom{ }
\begin{table}[h!]
\centering
\setlength{\fboxsep}{0pt}
\setlength{\fboxrule}{2pt}
\fbox{%
\begin{tabular}{c|>{\arraybackslash}m{0.615\textwidth}}
\hline
\multirow{3}{*}{\raisebox{-\totalheight}{\includegraphics[width=0.5\textwidth,height=5.3cm,keepaspectratio]{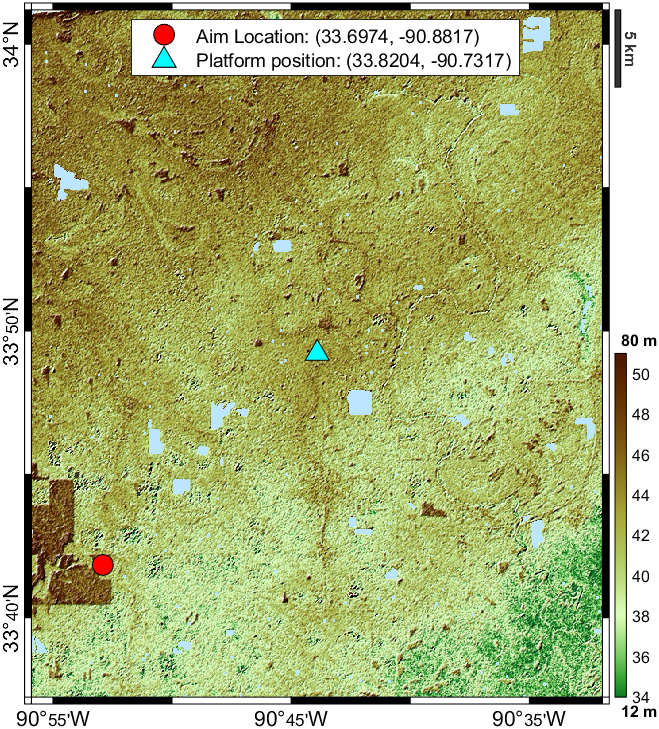}}} \\[-0.1cm] & \textbf{Platform Aim Location} (lat, lon): $(33.6974^{\circ}, -90.8817^{\circ})$ \\[0.2cm]
\cline{2-2} \\[-0.1cm] & 
\textbf{Platform Position} (lat, lon): $(33.8204^{\circ}, -90.7317^{\circ})$ \\[0.2cm]
\cline{2-2} \\[-0.1cm] & 
\textbf{Folder}: \textbf{/CLUTTER/num68\_lat33.8204\_lon-90.7317/} \\[0.2cm]
\cline{2-2} \\[-0.1cm] &
\parbox[t][3.0cm][t]{0.62\textwidth}{\textbf{Description}: Scenario $q_i = 15 \ (i = 68)$ in RASPNet is on the Lower Mississippi River in Mississippi. The radar platform is positioned at latitude $33.8204^{\circ}$ N and longitude $90.7317^{\circ}$ W. It aims towards a location at latitude $33.6974^{\circ}$ N and longitude $90.8817^{\circ}$ W. This region is characteried by a flat river valley, with fertile farmlands and low terrain variability. } \\
\hline
\end{tabular}
}
\end{table}

\paragraph{\underline{Scenario $q_i = 16$}: Saginaw Bay, MI ($i = 77$)} \vphantom{ }
\begin{table}[h!]
\centering
\setlength{\fboxsep}{0pt}
\setlength{\fboxrule}{2pt}
\fbox{%
\begin{tabular}{c|>{\arraybackslash}m{0.615\textwidth}}
\hline
\multirow{3}{*}{\raisebox{-\totalheight}{\includegraphics[width=0.5\textwidth,height=5.3cm,keepaspectratio]{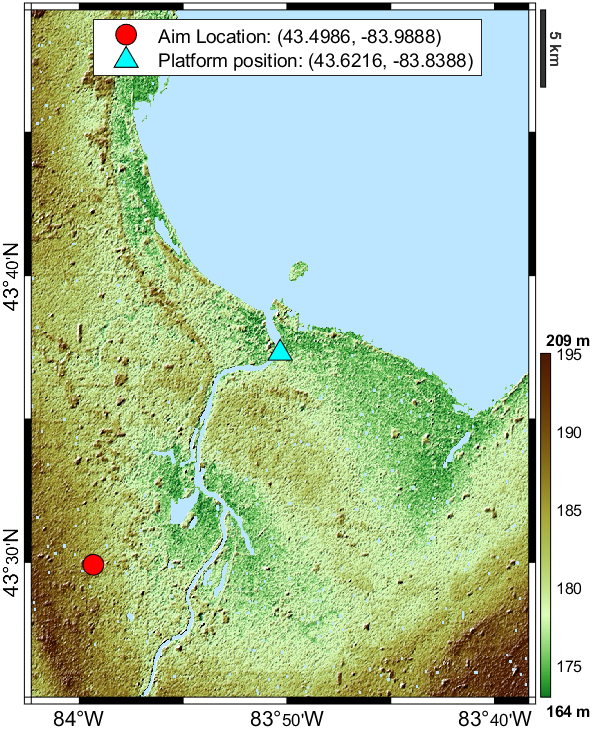}}} \\[-0.1cm] & \textbf{Platform Aim Location} (lat, lon): $(43.4986^{\circ}, -83.9888^{\circ})$ \\[0.2cm]
\cline{2-2} \\[-0.1cm] & 
\textbf{Platform Position} (lat, lon): $(43.6216^{\circ}, -83.8388^{\circ})$ \\[0.2cm]
\cline{2-2} \\[-0.1cm] & 
\textbf{Folder}: \textbf{/CLUTTER/num77\_lat43.6216\_lon-83.8388/} \\[0.2cm]
\cline{2-2} \\[-0.1cm] &
\parbox[t][3.0cm][t]{0.62\textwidth}{\textbf{Description}: Scenario $q_i = 16 \ (i = 77)$ in RASPNet is on the Saginaw Bay coastline of Michigan. The radar platform is positioned at latitude $43.6216^{\circ}$ N and longitude $83.8388^{\circ}$ W. It aims towards a location at latitude $43.4986^{\circ}$ N and longitude $83.9888^{\circ}$ W. The radar is positioned near the coastline, the aim direction is inland, and the surrounding terrain is flat. } \\
\hline
\end{tabular}
}
\end{table}

\paragraph{\underline{Scenario $q_i = 17$}: Big Sur, CA ($i = 8$)} \vphantom{ }
\begin{table}[h!]
\centering
\setlength{\fboxsep}{0pt}
\setlength{\fboxrule}{2pt}
\fbox{%
\begin{tabular}{c|>{\arraybackslash}m{0.615\textwidth}}
\hline
\multirow{3}{*}{\raisebox{-\totalheight}{\includegraphics[width=0.5\textwidth,height=5.3cm,keepaspectratio]{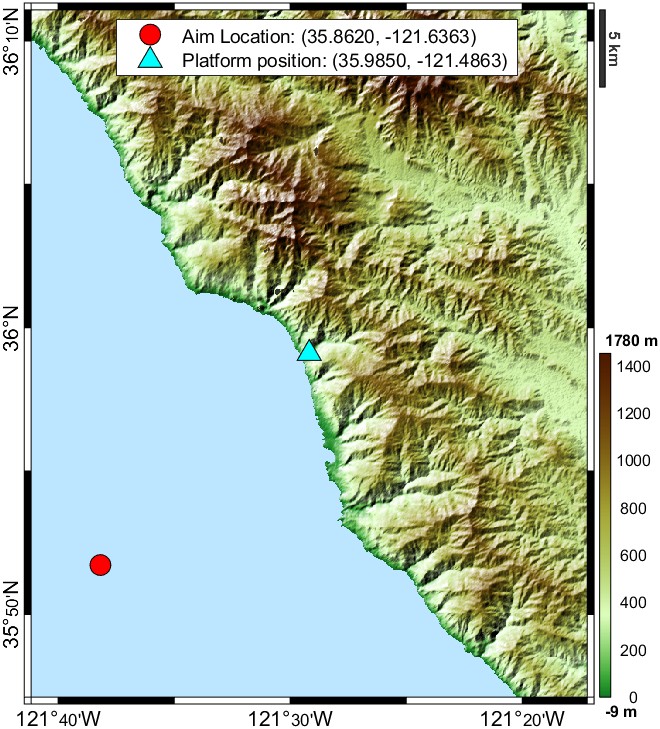}}} \\[-0.1cm] & \textbf{Platform Aim Location} (lat, lon): $(35.8620^{\circ}, -121.6363^{\circ})$ \\[0.2cm]
\cline{2-2} \\[-0.1cm] & 
\textbf{Platform Position} (lat, lon): $(35.9850^{\circ}, -121.4863^{\circ})$ \\[0.2cm]
\cline{2-2} \\[-0.1cm] & 
\textbf{Folder}: \textbf{/CLUTTER/num8\_lat35.985\_lon-121.4863/} \\[0.2cm]
\cline{2-2} \\[-0.1cm] &
\parbox[t][3.0cm][t]{0.62\textwidth}{\textbf{Description}: Scenario $q_i = 17 \ (i = 8)$ in RASPNet is in the Big Sur region of Coastal California. The radar platform is positioned at latitude $35.9850^{\circ}$ N and longitude $121.4863^{\circ}$ W. It aims towards a location at latitude $35.8620^{\circ}$ N and longitude $121.6363^{\circ}$ W. This is a land-to-water transition region, where the aim location is on the water surface and has low variability.} \\
\hline
\end{tabular}
}
\end{table}

\paragraph{\underline{Scenario $q_i = 18$}: Cape Cod, MA ($i = 100$)} \vphantom{ }
\begin{table}[h!]
\centering
\setlength{\fboxsep}{0pt}
\setlength{\fboxrule}{2pt}
\fbox{%
\begin{tabular}{c|>{\arraybackslash}m{0.615\textwidth}}
\hline
\multirow{3}{*}{\raisebox{-\totalheight}{\includegraphics[width=0.5\textwidth,height=5.3cm,keepaspectratio]{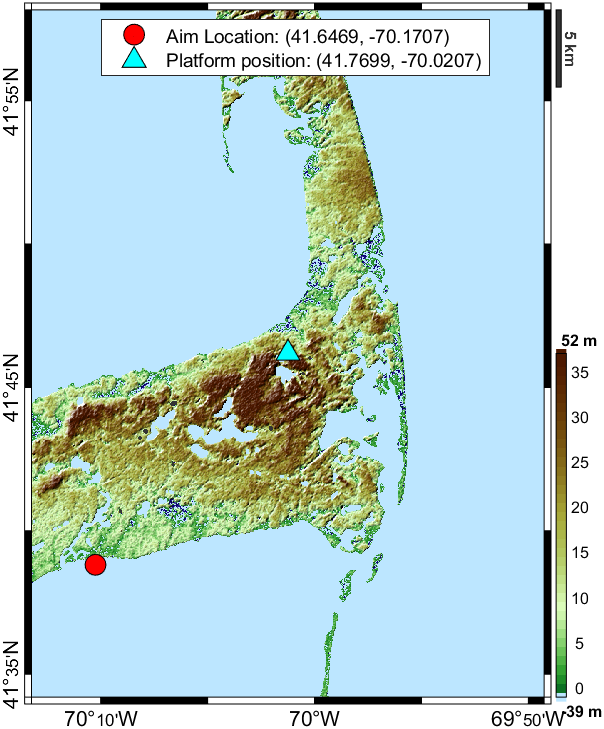}}} \\[-0.1cm] & \textbf{Platform Aim Location} (lat, lon): $(41.6469^{\circ}, -70.1707^{\circ})$ \\[0.2cm]
\cline{2-2} \\[-0.1cm] & 
\textbf{Platform Position} (lat, lon): $(41.7699^{\circ}, -70.0207^{\circ})$ \\[0.2cm]
\cline{2-2} \\[-0.1cm] & 
\textbf{Folder}: \textbf{/CLUTTER/num100\_lat41.7699\_lon-70.0207/} \\[0.2cm]
\cline{2-2} \\[-0.1cm] &
\parbox[t][3.0cm][t]{0.62\textwidth}{\textbf{Description}: Scenario $q_i = 18 \ (i = 100)$ in RASPNet is in the Cape Cod region of Coastal Massachusetts. The radar platform is positioned at latitude $41.7699^{\circ}$ N and longitude $70.0207^{\circ}$ W. It aims towards a location at latitude $41.6469^{\circ}$ N and longitude $70.1707^{\circ}$ W. This is a land-to-water transition region, where the aim direction is over flat land with low variability.} \\
\hline
\end{tabular}
}
\end{table}

\paragraph{\underline{Scenario $q_i = 19$}: Delaware Bay, NJ ($i = 91$)}  \vphantom{ }
\begin{table}[h!]
\centering
\setlength{\fboxsep}{0pt}
\setlength{\fboxrule}{2pt}
\fbox{%
\begin{tabular}{c|>{\arraybackslash}m{0.615\textwidth}}
\hline
\multirow{3}{*}{\raisebox{-\totalheight}{\includegraphics[width=0.5\textwidth,height=5.3cm,keepaspectratio]{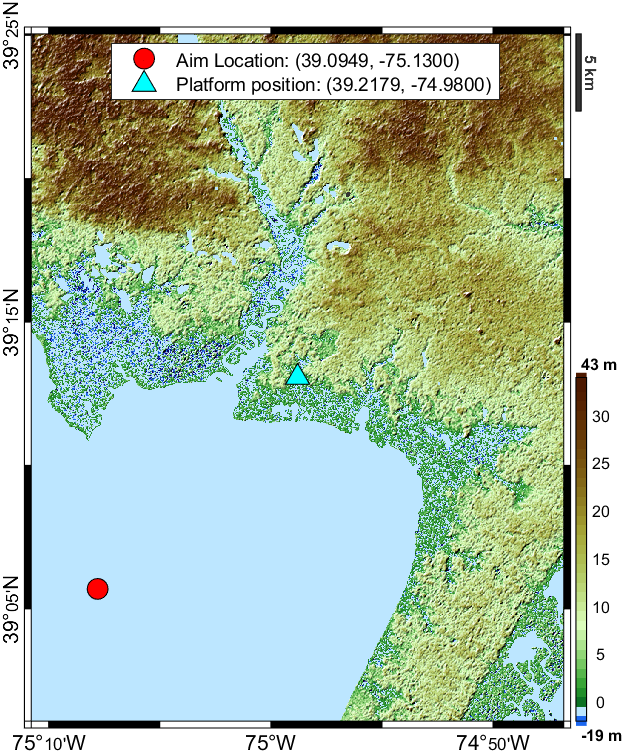}}} \\[-0.1cm] & \textbf{Platform Aim Location} (lat, lon): $(39.0949^{\circ}, -75.1300^{\circ})$ \\[0.2cm]
\cline{2-2} \\[-0.1cm] & 
\textbf{Platform Position} (lat, lon): $(39.2179^{\circ}, -74.9800^{\circ})$ \\[0.2cm]
\cline{2-2} \\[-0.1cm] & 
\textbf{Folder}: \textbf{/CLUTTER/num91\_lat39.2179\_lon-74.98/} \\[0.2cm]
\cline{2-2} \\[-0.1cm] &
\parbox[t][3.0cm][t]{0.62\textwidth}{\textbf{Description}: Scenario $q_i = 19 \ (i = 91)$ in RASPNet is on Delaware Bay in Southern New Jersey. The radar platform is positioned at latitude $39.2179^{\circ}$ N and longitude $74.9800^{\circ}$ W. It aims towards a location at latitude $39.0949^{\circ}$ N and longitude $75.1300^{\circ}$ W. This is a land-to-water transition region with a flat coastline, where the aim location is on the water surface.} \\
\hline
\end{tabular}
}
\end{table}

\paragraph{\underline{Scenario $q_i = 20$}: Salt River Valley, AZ ($i = 36$)}  \vphantom{ }
\begin{table}[h!]
\centering
\setlength{\fboxsep}{0pt}
\setlength{\fboxrule}{2pt}
\fbox{%
\begin{tabular}{c|>{\arraybackslash}m{0.615\textwidth}}
\hline
\multirow{3}{*}{\raisebox{-\totalheight}{\includegraphics[width=0.5\textwidth,height=5.3cm,keepaspectratio]{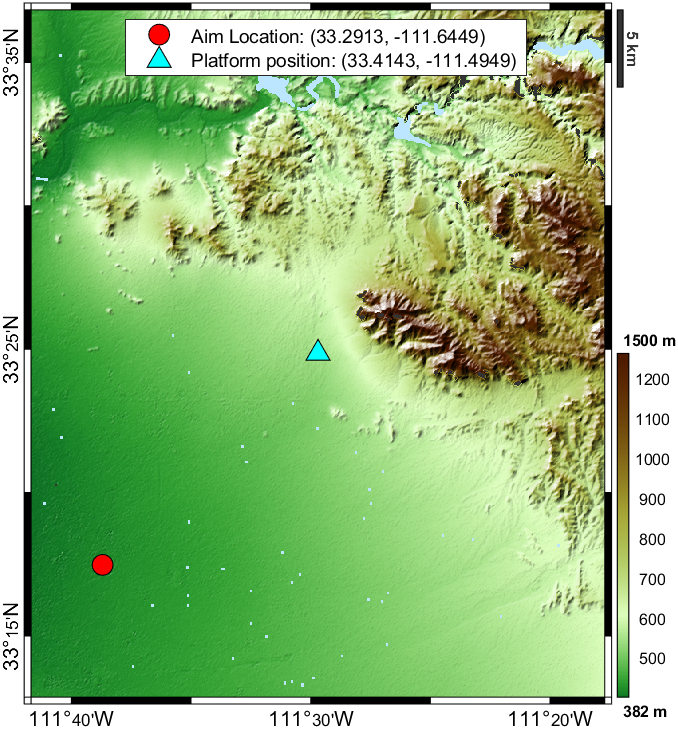}}} \\[-0.1cm] & \textbf{Platform Aim Location} (lat, lon): $(33.2913^{\circ}, -111.6449^{\circ})$ \\[0.2cm]
\cline{2-2} \\[-0.1cm] & 
\textbf{Platform Position} (lat, lon): $(33.4143^{\circ}, -111.4949^{\circ})$ \\[0.2cm]
\cline{2-2} \\[-0.1cm] & 
\textbf{Folder}: \textbf{/CLUTTER/num36\_lat33.4143\_lon-111.4949/} \\[0.2cm]
\cline{2-2} \\[-0.1cm] &
\parbox[t][3.0cm][t]{0.62\textwidth}{\textbf{Description}: Scenario $q_i = 20 \ (i = 36)$ in RASPNet is in the Salt River Valley in Central Arizona. The radar platform is positioned at latitude $33.4143^{\circ}$ N and longitude $111.4949^{\circ}$ W. It aims towards a location at latitude $33.2913^{\circ}$ N and longitude $111.6449^{\circ}$ W. This region is characterized by an expansive and flat desert environment, with low terrain variability.} \\
\hline
\end{tabular}
}
\end{table}

\paragraph{\underline{Scenario $q_i = 21$}: Snake River Plain, ID ($i = 30$)}  \vphantom{ }
\begin{table}[h!]
\centering
\setlength{\fboxsep}{0pt}
\setlength{\fboxrule}{2pt}
\fbox{%
\begin{tabular}{c|>{\arraybackslash}m{0.615\textwidth}}
\hline
\multirow{3}{*}{\raisebox{-\totalheight}{\includegraphics[width=0.5\textwidth,height=5.3cm,keepaspectratio]{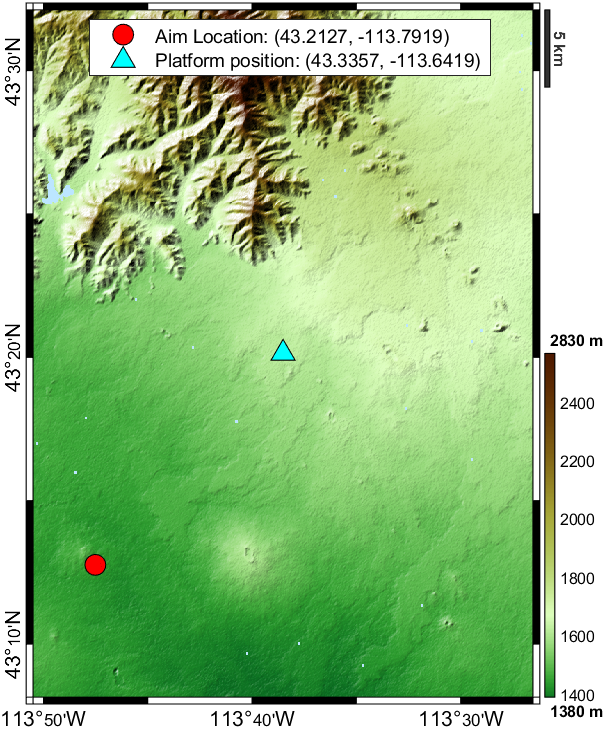}}} \\[-0.1cm] & \textbf{Platform Aim Location} (lat, lon): $(43.2127^{\circ}, -113.7919^{\circ})$ \\[0.2cm]
\cline{2-2} \\[-0.1cm] & 
\textbf{Platform Position} (lat, lon): $(43.3357^{\circ}, -113.6419^{\circ})$ \\[0.2cm]
\cline{2-2} \\[-0.1cm] & 
\textbf{Folder}: \textbf{/CLUTTER/num30\_lat43.3357\_lon-113.6419/} \\[0.2cm]
\cline{2-2} \\[-0.1cm] &
\parbox[t][3.0cm][t]{0.62\textwidth}{\textbf{Description}: Scenario $q_i = 21 \ (i = 30)$ in RASPNet is in the Snake River Plain in Central Idaho. The radar platform is positioned at latitude $43.3357^{\circ}$ N and longitude $113.6419^{\circ}$ W. It aims towards a location at latitude $43.2127^{\circ}$ N and longitude $113.7919^{\circ}$ W. This region is characterized by a uniform and flat grassland environment, with low terrain variability.} \\
\hline
\end{tabular}
}
\end{table}

\paragraph{\underline{Scenario $q_i = 22$}: Badlands National Park, SD ($i = 55$)}  \vphantom{ }
\begin{table}[h!]
\centering
\setlength{\fboxsep}{0pt}
\setlength{\fboxrule}{2pt}
\fbox{%
\begin{tabular}{c|>{\arraybackslash}m{0.615\textwidth}}
\hline
\multirow{3}{*}{\raisebox{-\totalheight}{\includegraphics[width=0.5\textwidth,height=5.3cm,keepaspectratio]{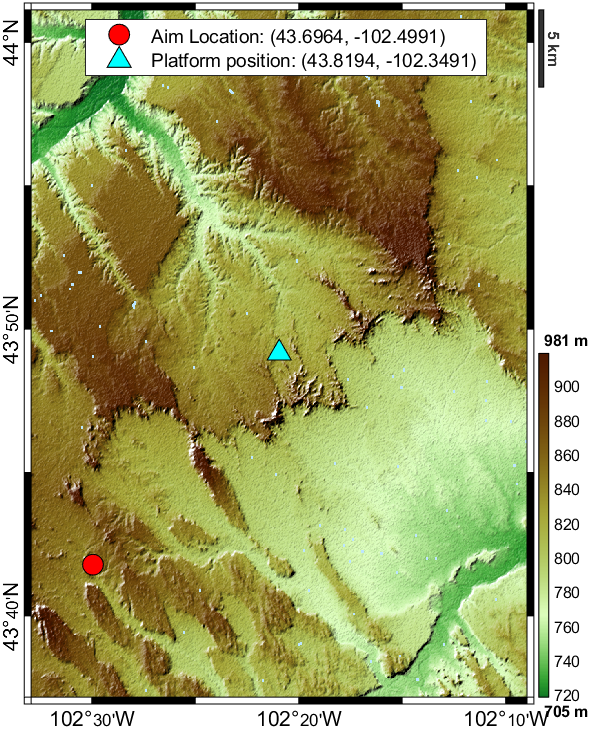}}} \\[-0.1cm] & \textbf{Platform Aim Location} (lat, lon): $(43.6964^{\circ}, -102.4991^{\circ})$ \\[0.2cm]
\cline{2-2} \\[-0.1cm] & 
\textbf{Platform Position} (lat, lon): $(43.8194^{\circ}, -102.3491^{\circ})$ \\[0.2cm]
\cline{2-2} \\[-0.1cm] & 
\textbf{Folder}: \textbf{/CLUTTER/num55\_lat43.8194\_lon-102.3491/} \\[0.2cm]
\cline{2-2} \\[-0.1cm] &
\parbox[t][3.0cm][t]{0.62\textwidth}{\textbf{Description}: Scenario $q_i = 22 \ (i = 55)$ in RASPNet is in Badlands National Park in South Dakota. The radar platform is located at latitude $43.8194^{\circ}$ N and longitude $102.3491^{\circ}$ W. It aims towards a location at latitude $43.6964^{\circ}$ N and longitude $102.4991^{\circ}$ W. While this region is characterized by rough terrain, the aim location has low elevation variability.} \\
\hline
\end{tabular}
}
\end{table}

\paragraph{\underline{Scenario $q_i = 23$}: Chesapeake Bay, VA ($i = 90$)}  \vphantom{ }
\begin{table}[h!]
\centering
\setlength{\fboxsep}{0pt}
\setlength{\fboxrule}{2pt}
\fbox{%
\begin{tabular}{c|>{\arraybackslash}m{0.615\textwidth}}
\hline
\multirow{3}{*}{\raisebox{-\totalheight}{\includegraphics[width=0.5\textwidth,height=5.3cm,keepaspectratio]{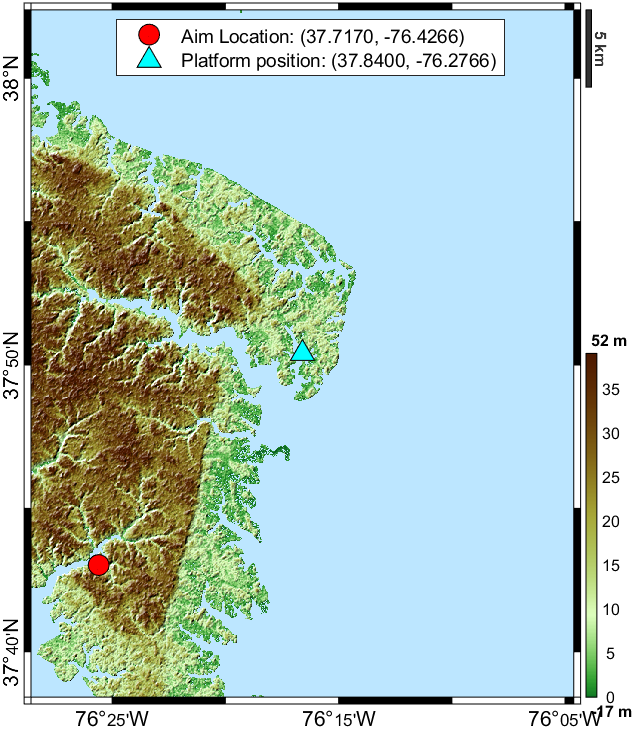}}} \\[-0.1cm] & \textbf{Platform Aim Location} (lat, lon): $(37.7170^{\circ}, -76.4266^{\circ})$ \\[0.2cm]
\cline{2-2} \\[-0.1cm] & 
\textbf{Platform Position} (lat, lon): $(37.8400^{\circ}, -76.2766^{\circ})$ \\[0.2cm]
\cline{2-2} \\[-0.1cm] & 
\textbf{Folder}: \textbf{/CLUTTER/num90\_lat37.84\_lon-76.2766/} \\[0.2cm]
\cline{2-2} \\[-0.1cm] &
\parbox[t][3.0cm][t]{0.62\textwidth}{\textbf{Description}: Scenario $q_i = 23 \ (i = 90)$ in RASPNet is in the Chesapeake Bay region of Coastal Virginia. The radar platform is positioned at latitude $37.8400^{\circ}$ N and longitude $76.2766^{\circ}$ W. It aims towards a location at latitude $37.7170^{\circ}$ N and longitude $76.4266^{\circ}$ W. This is a land-to-water transition region, where the aim direction is over flat land with low variability.} \\
\hline
\end{tabular}
}
\end{table}

\paragraph{\underline{Scenario $q_i = 24$}: Lake Champlain, VT ($i = 93$)}  \vphantom{ }
\begin{table}[h!]
\centering
\setlength{\fboxsep}{0pt}
\setlength{\fboxrule}{2pt}
\fbox{%
\begin{tabular}{c|>{\arraybackslash}m{0.615\textwidth}}
\hline
\multirow{3}{*}{\raisebox{-\totalheight}{\includegraphics[width=0.5\textwidth,height=5.3cm,keepaspectratio]{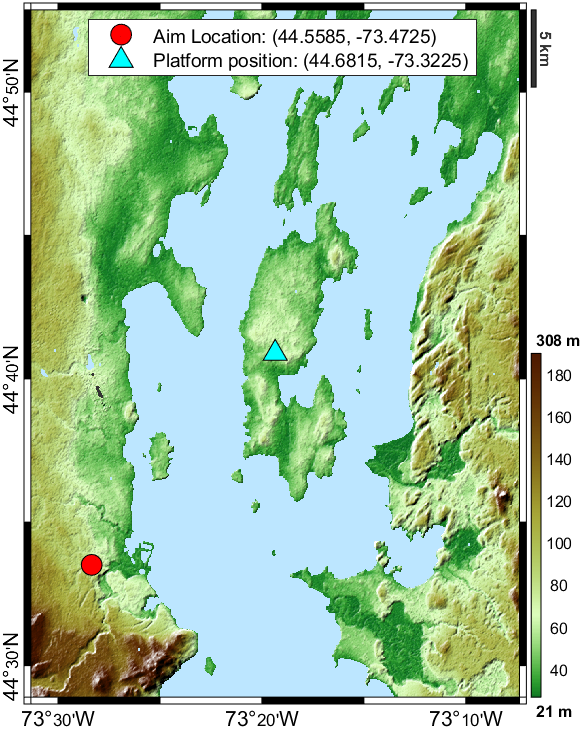}}} \\[-0.1cm] & \textbf{Platform Aim Location} (lat, lon): $(44.5585^{\circ}, -73.4725^{\circ})$ \\[0.2cm]
\cline{2-2} \\[-0.1cm] & 
\textbf{Platform Position} (lat, lon): $(44.6815^{\circ}, -73.3225^{\circ})$ \\[0.2cm]
\cline{2-2} \\[-0.1cm] & 
\textbf{Folder}: \textbf{/CLUTTER/num93\_lat44.6815\_lon-73.3225/} \\[0.2cm]
\cline{2-2} \\[-0.1cm] &
\parbox[t][3.0cm][t]{0.62\textwidth}{\textbf{Description}: Scenario $q_i = 24 \ (i = 93)$ in RASPNet is in the Lake Champlain region of Vermont. The radar platform is positioned at latitude $44.6815^{\circ}$ N and longitude $73.3225^{\circ}$ W. It aims towards a location at latitude $44.5585^{\circ}$ N and longitude $73.4725^{\circ}$ W. This is a land-to-water transition region, where the aim direction is over water and has low variability.} \\
\hline
\end{tabular}
}
\end{table}

\paragraph{\underline{Scenario $q_i = 25$}: Santa Barbara Coast, CA ($i = 12$)}  \vphantom{ }
\begin{table}[h!]
\centering
\setlength{\fboxsep}{0pt}
\setlength{\fboxrule}{2pt}
\fbox{%
\begin{tabular}{c|>{\arraybackslash}m{0.615\textwidth}}
\hline
\multirow{3}{*}{\raisebox{-\totalheight}{\includegraphics[width=0.5\textwidth,height=5.1cm,keepaspectratio]{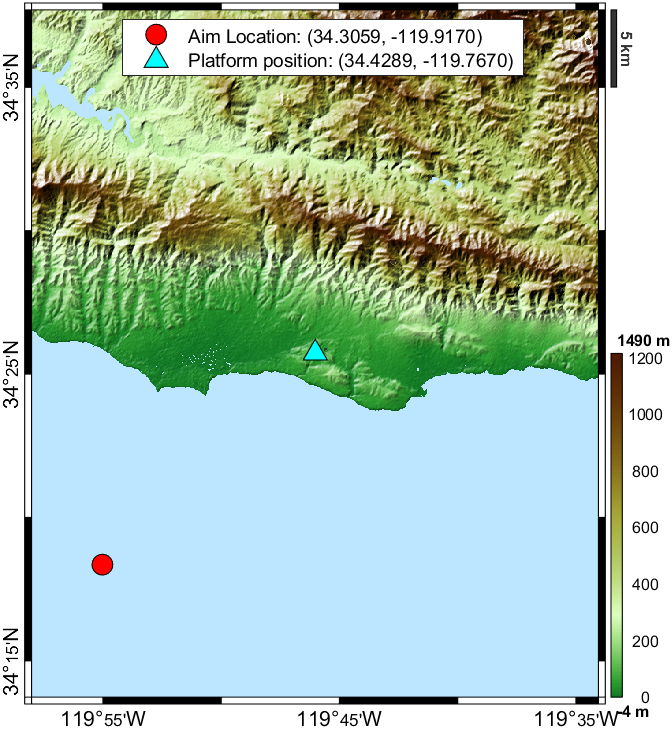}}} \\[-0.1cm] & \textbf{Platform Aim Location} (lat, lon): $(34.3059^{\circ}, -119.9170^{\circ})$ \\[0.2cm]
\cline{2-2} \\[-0.1cm] & 
\textbf{Platform Position} (lat, lon): $(34.4289^{\circ}, -119.7670^{\circ})$ \\[0.2cm]
\cline{2-2} \\[-0.1cm] & 
\textbf{Folder}: \textbf{/CLUTTER/num12\_lat34.4289\_lon-119.767/} \\[0.2cm]
\cline{2-2} \\[-0.1cm] &
\parbox[t][2.6cm][t]{0.62\textwidth}{\textbf{Description}: Scenario $q_i = 25 \ (i = 12)$ in RASPNet is on the Santa Barbara Coast in California. The radar platform is positioned at latitude $34.4289^{\circ}$ N and longitude $119.7670^{\circ}$ W. It aims towards a location at latitude $34.3059^{\circ}$ N and longitude $119.9170^{\circ}$ W. This is a land-to-water transition region, where the aim location is on the water surface.} \\
\hline
\end{tabular}
}
\end{table}

\subsection{Intermediate Scenarios}
We now outline the Intermediate RASPNet scenarios, where $q_i \in \{26,27,\ldots,50\}$.

\paragraph{\underline{Scenario $q_i = 26$}: Red River Valley, ND ($i = 61$)}  \vphantom{ }
\begin{table}[h!]
\centering
\setlength{\fboxsep}{0pt}
\setlength{\fboxrule}{2pt}
\fbox{%
\begin{tabular}{c|>{\arraybackslash}m{0.615\textwidth}}
\hline
\multirow{3}{*}{\raisebox{-\totalheight}{\includegraphics[width=0.5\textwidth,height=5.1cm,keepaspectratio]{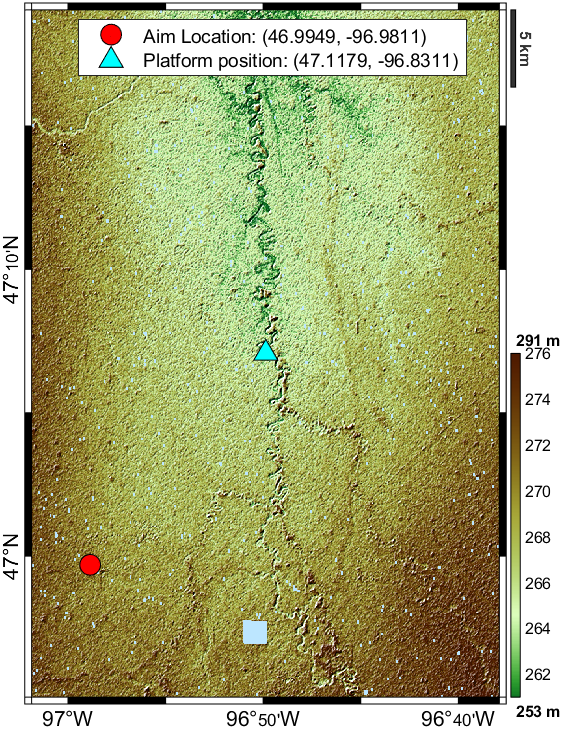}}} \\[-0.1cm] & \textbf{Platform Aim Location} (lat, lon): $(46.9949^{\circ}, -96.9811^{\circ})$ \\[0.2cm]
\cline{2-2} \\[-0.1cm] & 
\textbf{Platform Position} (lat, lon): $(47.1179^{\circ}, -96.8311^{\circ})$ \\[0.2cm]
\cline{2-2} \\[-0.1cm] & 
\textbf{Folder}: \textbf{/CLUTTER/num61\_lat47.1179\_lon-96.8311/} \\[0.2cm]
\cline{2-2} \\[-0.1cm] &
\parbox[t][2.6cm][t]{0.62\textwidth}{\textbf{Description}: Scenario $q_i = 26 \ (i = 61)$ in RASPNet is in the Red River Valley in Eastern North Dakota. The radar platform is positioned at latitude $47.1179^{\circ}$ N and longitude $96.8311^{\circ}$ W. It aims towards a location at latitude $46.9949^{\circ}$ N and longitude $96.9811^{\circ}$ W. This region is characterized by a uniform and flat grassland environment, with low terrain variability.} \\
\hline
\end{tabular}
}
\end{table}

\paragraph{\underline{Scenario $q_i = 27$}: Superior Upland, WI ($i = 67$)}  \vphantom{ }
\begin{table}[h!]
\centering
\setlength{\fboxsep}{0pt}
\setlength{\fboxrule}{2pt}
\fbox{%
\begin{tabular}{c|>{\arraybackslash}m{0.615\textwidth}}
\hline
\multirow{3}{*}{\raisebox{-\totalheight}{\includegraphics[width=0.5\textwidth,height=5.1cm,keepaspectratio]{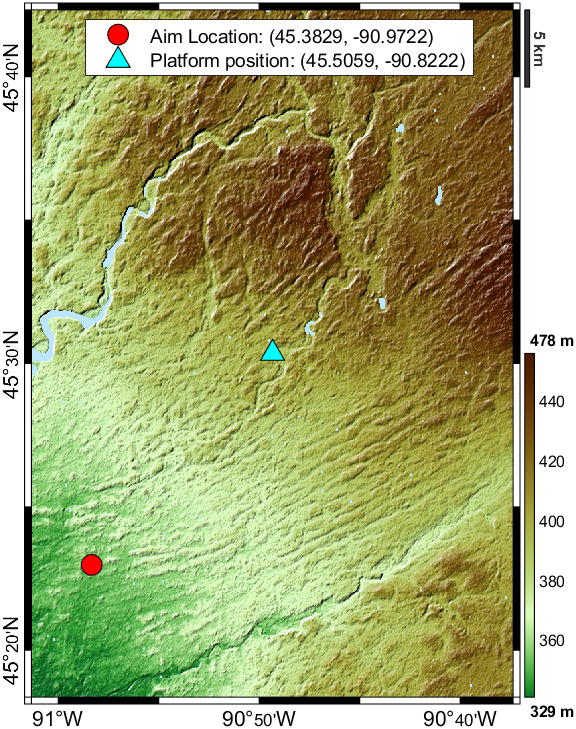}}} \\[-0.1cm] & \textbf{Platform Aim Location} (lat, lon): $(45.3829^{\circ}, -90.9722^{\circ})$ \\[0.2cm]
\cline{2-2} \\[-0.1cm] & 
\textbf{Platform Position} (lat, lon): $(45.5059^{\circ}, -90.8222^{\circ})$ \\[0.2cm]
\cline{2-2} \\[-0.1cm] & 
\textbf{Folder}: \textbf{/CLUTTER/num67\_lat45.5059\_lon-90.8222/} \\[0.2cm]
\cline{2-2} \\[-0.1cm] &
\parbox[t][2.6cm][t]{0.62\textwidth}{\textbf{Description}: Scenario $q_i = 27 \ (i = 67)$ in RASPNet is in the Superior Upland region of Wisconsin. The radar platform is located at latitude $45.5059^{\circ}$ N and longitude $90.8222^{\circ}$ W. It aims towards a location at latitude $45.3829^{\circ}$ N and longitude $90.9722^{\circ}$ W. While this region is characterized by nonuniform terrain, the aim location has low elevation variability.} \\
\hline
\end{tabular}
}
\end{table}

\paragraph{\underline{Scenario $q_i = 28$}: Lake Erie Coast, PA ($i = 84$)}  \vphantom{ }
\begin{table}[h!]
\centering
\setlength{\fboxsep}{0pt}
\setlength{\fboxrule}{2pt}
\fbox{%
\begin{tabular}{c|>{\arraybackslash}m{0.615\textwidth}}
\hline
\multirow{3}{*}{\raisebox{-\totalheight}{\includegraphics[width=0.5\textwidth,height=5.3cm,keepaspectratio]{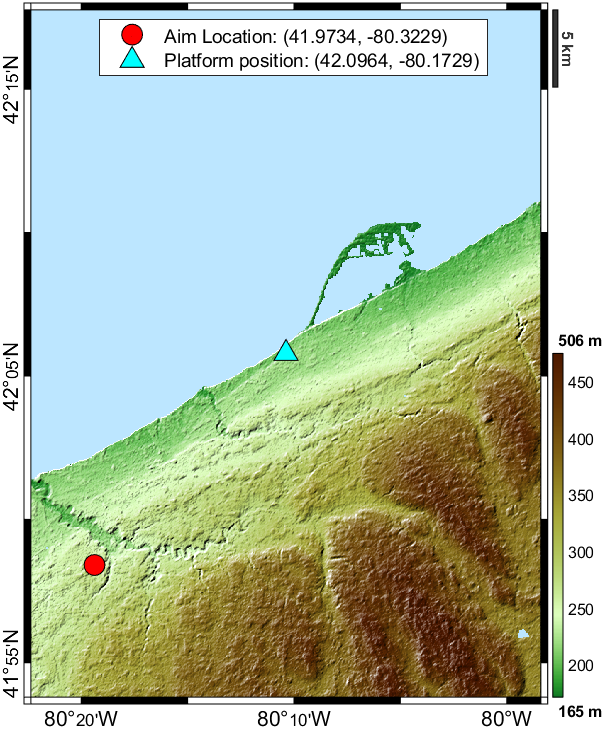}}} \\[-0.1cm] & \textbf{Platform Aim Location} (lat, lon): $(41.9734^{\circ}, -80.3229^{\circ})$ \\[0.2cm]
\cline{2-2} \\[-0.1cm] & 
\textbf{Platform Position} (lat, lon): $(42.0964^{\circ}, -80.1729^{\circ})$ \\[0.2cm]
\cline{2-2} \\[-0.1cm] & 
\textbf{Folder}: \textbf{/CLUTTER/num84\_lat42.0964\_lon-80.1729/} \\[0.2cm]
\cline{2-2} \\[-0.1cm] &
\parbox[t][3.0cm][t]{0.62\textwidth}{\textbf{Description}: Scenario $q_i = 28 \ (i = 84)$ in RASPNet is on the Lake Erie coastline of Pennsylvania. The radar platform is positioned at latitude $42.0964^{\circ}$ N and longitude $80.1729^{\circ}$ W. It aims towards a location at latitude $41.9734^{\circ}$ N and longitude $80.3229^{\circ}$ W. The radar is positioned near the coastline, the aim direction is inland, and the surrounding terrain is nonuniform.} \\
\hline
\end{tabular}
}
\end{table}

\paragraph{\underline{Scenario $q_i = 29$}: Allegheny Plateau, PA ($i = 86$)}  \vphantom{ }
\begin{table}[h!]
\centering
\setlength{\fboxsep}{0pt}
\setlength{\fboxrule}{2pt}
\fbox{%
\begin{tabular}{c|>{\arraybackslash}m{0.615\textwidth}}
\hline
\multirow{3}{*}{\raisebox{-\totalheight}{\includegraphics[width=0.5\textwidth,height=5.3cm,keepaspectratio]{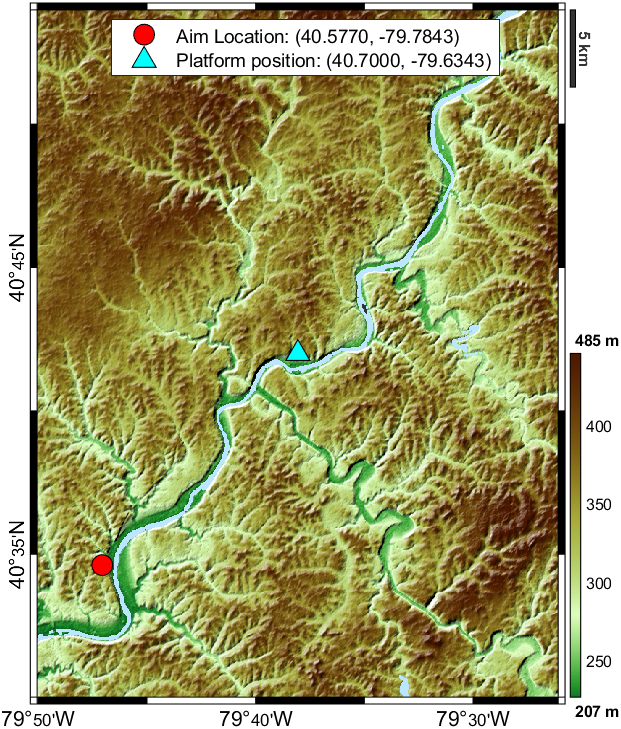}}} \\[-0.1cm] & \textbf{Platform Aim Location} (lat, lon): $(40.5770^{\circ}, -79.7843^{\circ})$ \\[0.2cm]
\cline{2-2} \\[-0.1cm] & 
\textbf{Platform Position} (lat, lon): $(40.7000^{\circ}, -79.6343^{\circ})$ \\[0.2cm]
\cline{2-2} \\[-0.1cm] & 
\textbf{Folder}: \textbf{/CLUTTER/num86\_lat40.7\_lon-79.6343/} \\[0.2cm]
\cline{2-2} \\[-0.1cm] &
\parbox[t][3.0cm][t]{0.62\textwidth}{\textbf{Description}: Scenario $q_i = 29 \ (i = 86)$ in RASPNet is in the Allegheny Plateau of Pennsylvania. The radar platform is positioned at latitude $40.7000^{\circ}$ N and longitude $79.6343^{\circ}$ W. It aims towards a location at latitude $40.5770^{\circ}$ N and longitude $79.7843^{\circ}$ W. While this region is characterized by nonuniform terrain, the aim location has low elevation variability.} \\
\hline
\end{tabular}
}
\end{table}

\paragraph{\underline{Scenario $q_i = 30$}: North Georgia Mountains, GA ($i = 73$)} \vphantom{ }
\begin{table}[h!]
\centering
\setlength{\fboxsep}{0pt}
\setlength{\fboxrule}{2pt}
\fbox{%
\begin{tabular}{c|>{\arraybackslash}m{0.615\textwidth}}
\hline
\multirow{3}{*}{\raisebox{-\totalheight}{\includegraphics[width=0.5\textwidth,height=5.3cm,keepaspectratio]{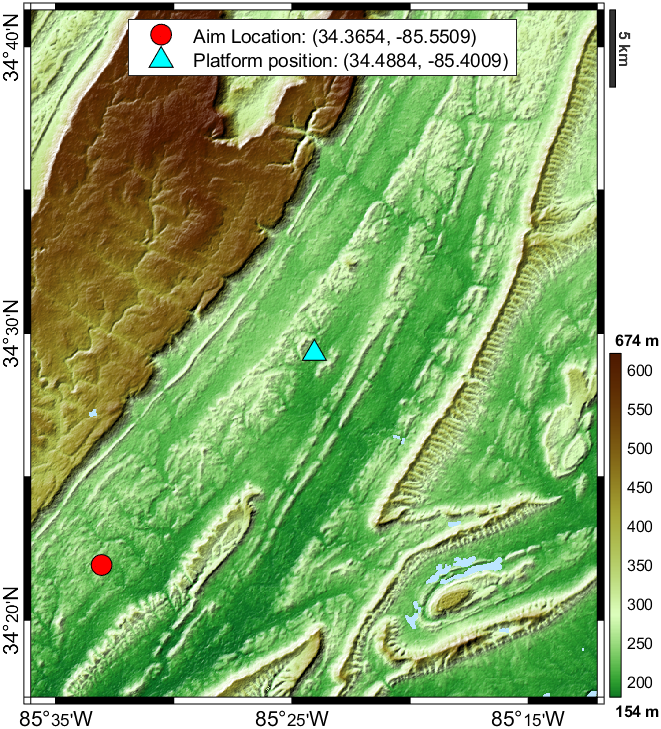}}} \\[-0.1cm] & \textbf{Platform Aim Location} (lat, lon): $(34.3654^{\circ}, -85.5509^{\circ})$ \\[0.2cm]
\cline{2-2} \\[-0.1cm] & 
\textbf{Platform Position} (lat, lon): $(34.4884^{\circ}, -85.4009^{\circ})$ \\[0.2cm]
\cline{2-2} \\[-0.1cm] & 
\textbf{Folder}: \textbf{/CLUTTER/num73\_lat34.4884\_lon-85.4009/} \\[0.2cm]
\cline{2-2} \\[-0.1cm] &
\parbox[t][3.0cm][t]{0.62\textwidth}{\textbf{Description}: Scenario $q_i = 30 \ (i = 73)$ in RASPNet is in mountainous Northern Georgia. The radar platform is located at latitude $34.4884^{\circ}$ N and longitude $85.4009^{\circ}$ W. It aims towards a location at latitude $34.3654^{\circ}$ N and longitude $85.5509^{\circ}$ W. While this region consists of rough terrain, the radar position and aim location is in a flat valley between ridges.} \\
\hline
\end{tabular}
}
\end{table}

\paragraph{\underline{Scenario $q_i = 31$}: Southwestern Texas, TX ($i = 53$)} \vphantom{ }
\begin{table}[h!]
\centering
\setlength{\fboxsep}{0pt}
\setlength{\fboxrule}{2pt}
\fbox{%
\begin{tabular}{c|>{\arraybackslash}m{0.615\textwidth}}
\hline
\multirow{3}{*}{\raisebox{-\totalheight}{\includegraphics[width=0.5\textwidth,height=5.3cm,keepaspectratio]{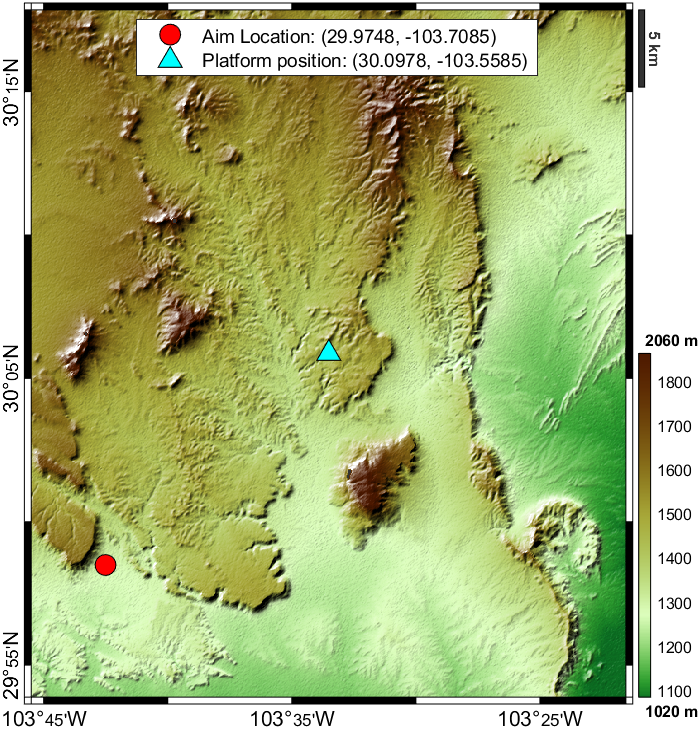}}} \\[-0.1cm] & \textbf{Platform Aim Location} (lat, lon): $(29.9748^{\circ}, -103.7085^{\circ})$ \\[0.2cm]
\cline{2-2} \\[-0.1cm] & 
\textbf{Platform Position} (lat, lon): $(30.0978^{\circ}, -103.5585^{\circ})$ \\[0.2cm]
\cline{2-2} \\[-0.1cm] & 
\textbf{Folder}: \textbf{/CLUTTER/num53\_lat30.0978\_lon-103.5585/} \\[0.2cm]
\cline{2-2} \\[-0.1cm] &
\parbox[t][3.0cm][t]{0.62\textwidth}{\textbf{Description}: Scenario $q_i = 31 \ (i = 53)$ in RASPNet is in the hills region of Southwestern Texas. The radar platform is positioned at latitude $30.0978^{\circ}$ N and longitude $103.5585^{\circ}$ W. It aims towards a location at latitude $29.9748^{\circ}$ N and longitude $103.7085^{\circ}$ W. This region consists of hilly terrain with more pronounced nonuniformities.} \\
\hline
\end{tabular}
}
\end{table}

\paragraph{\underline{Scenario $q_i = 32$}: Southern Texas, TX ($i = 59$)} \vphantom{ }
\begin{table}[h!]
\centering
\setlength{\fboxsep}{0pt}
\setlength{\fboxrule}{2pt}
\fbox{%
\begin{tabular}{c|>{\arraybackslash}m{0.615\textwidth}}
\hline
\multirow{3}{*}{\raisebox{-\totalheight}{\includegraphics[width=0.5\textwidth,height=5.3cm,keepaspectratio]{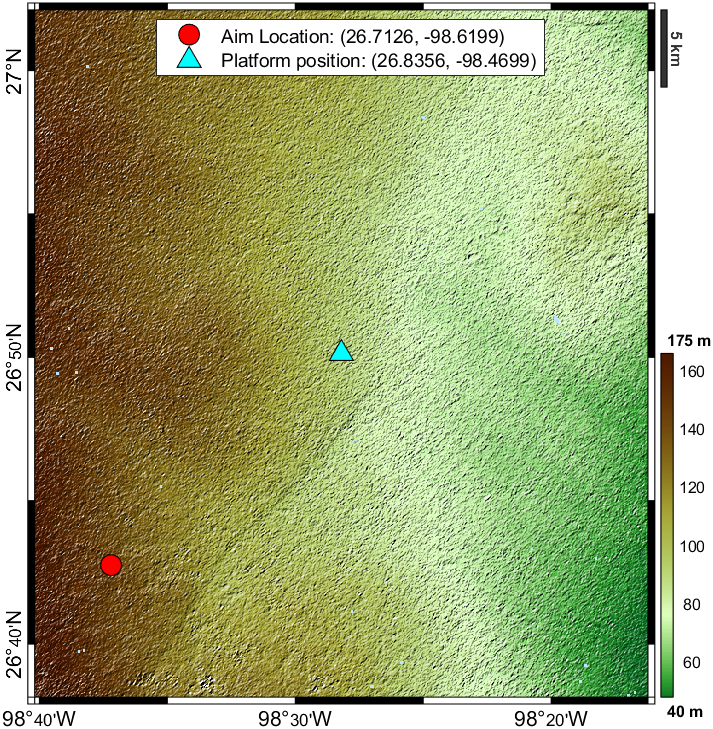}}} \\[-0.1cm] & \textbf{Platform Aim Location} (lat, lon): $(26.7126^{\circ}, -98.6199^{\circ})$ \\[0.2cm]
\cline{2-2} \\[-0.1cm] & 
\textbf{Platform Position} (lat, lon): $(26.8356^{\circ}, -98.4699^{\circ})$ \\[0.2cm]
\cline{2-2} \\[-0.1cm] & 
\textbf{Folder}: \textbf{/CLUTTER/num59\_lat26.8356\_lon-98.4699/} \\[0.2cm]
\cline{2-2} \\[-0.1cm] &
\parbox[t][3.0cm][t]{0.62\textwidth}{\textbf{Description}: Scenario $q_i = 32 \ (i = 59)$ in RASPNet is in the brush and plains region of Southern Texas. The radar platform is positioned at latitude $26.8356^{\circ}$ N and longitude $98.4699^{\circ}$ W. It aims towards a location at latitude $26.7126^{\circ}$ N and longitude $98.6199^{\circ}$ W. This region is characterized by a uniform steppe environment, with minimal terrain variability.} \\
\hline
\end{tabular}
}
\end{table}

\paragraph{\underline{Scenario $q_i = 33$}: Oregon Coast, OR ($i = 2$)} \vphantom{ }
\begin{table}[h!]
\centering
\setlength{\fboxsep}{0pt}
\setlength{\fboxrule}{2pt}
\fbox{%
\begin{tabular}{c|>{\arraybackslash}m{0.615\textwidth}}
\hline
\multirow{3}{*}{\raisebox{-\totalheight}{\includegraphics[width=0.5\textwidth,height=5.3cm,keepaspectratio]{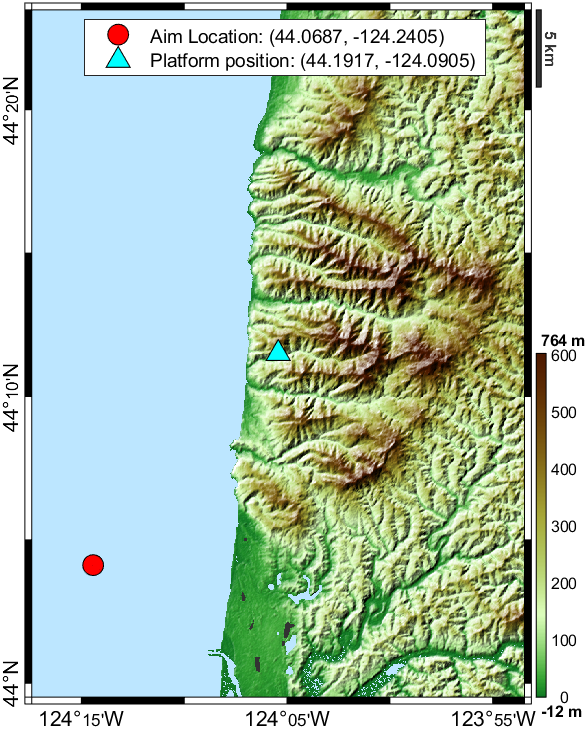}}} \\[-0.1cm] & \textbf{Platform Aim Location} (lat, lon): $(44.0687^{\circ}, -124.2405^{\circ})$ \\[0.2cm]
\cline{2-2} \\[-0.1cm] & 
\textbf{Platform Position} (lat, lon): $(44.1917^{\circ}, -124.0905^{\circ})$ \\[0.2cm]
\cline{2-2} \\[-0.1cm] & 
\textbf{Folder}: \textbf{/CLUTTER/num2\_lat44.1917\_lon-124.0905/} \\[0.2cm]
\cline{2-2} \\[-0.1cm] &
\parbox[t][3.0cm][t]{0.62\textwidth}{\textbf{Description}: Scenario $q_i = 33 \ (i = 2)$ in RASPNet is in the Coast Mountains of the Oregon Coast. The radar platform is positioned at latitude $44.1917^{\circ}$ N and longitude $124.0905^{\circ}$ W. It aims towards a location at latitude $44.0687^{\circ}$ N and longitude $124.2405^{\circ}$ W. This is a land-to-water transition region, where the aim location is on the water surface.} \\
\hline
\end{tabular}
}
\end{table}

\paragraph{\underline{Scenario $q_i = 34$}: Great Basin, NV ($i = 13$)} \vphantom{ }
\begin{table}[h!]
\centering
\setlength{\fboxsep}{0pt}
\setlength{\fboxrule}{2pt}
\fbox{%
\begin{tabular}{c|>{\arraybackslash}m{0.615\textwidth}}
\hline
\multirow{3}{*}{\raisebox{-\totalheight}{\includegraphics[width=0.5\textwidth,height=5.3cm,keepaspectratio]{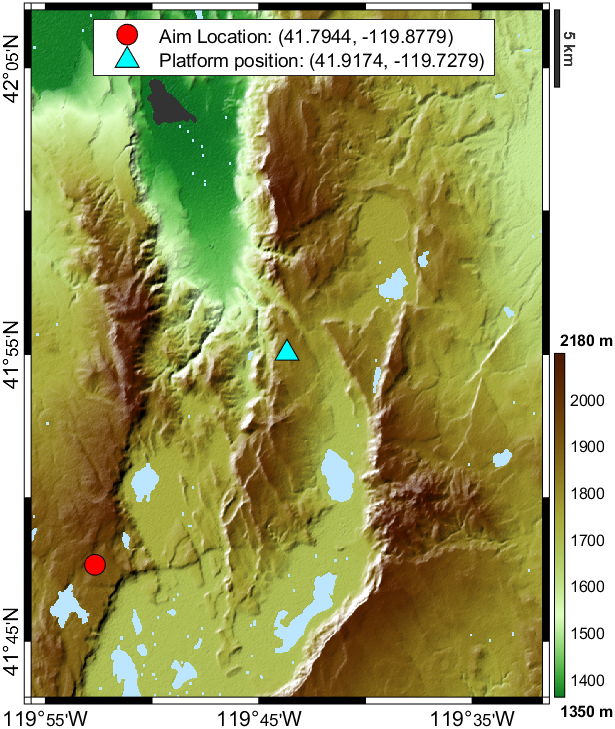}}} \\[-0.1cm] & \textbf{Platform Aim Location} (lat, lon): $(41.7944^{\circ}, -119.8779^{\circ})$ \\[0.2cm]
\cline{2-2} \\[-0.1cm] & 
\textbf{Platform Position} (lat, lon): $(41.9174^{\circ}, -119.7279^{\circ})$ \\[0.2cm]
\cline{2-2} \\[-0.1cm] & 
\textbf{Folder}: \textbf{/CLUTTER/num13\_lat41.9174\_lon-119.7279/} \\[0.2cm]
\cline{2-2} \\[-0.1cm] &
\parbox[t][3.0cm][t]{0.62\textwidth}{\textbf{Description}: Scenario $q_i = 34 \ (i = 13)$ in RASPNet is in the Great Basin region of Northern Nevada. The radar platform is positioned at latitude $41.9174^{\circ}$ N and longitude $119.7279^{\circ}$ W. It aims towards a location at latitude $41.9174^{\circ}$ N and longitude $119.7279^{\circ}$ W. This region is known for its basin and range topography and has more pronounced nonuniformities.} \\
\hline
\end{tabular}
}
\end{table}

\paragraph{\underline{Scenario $q_i = 35$}: Ouachita Mountains, OK ($i = 62$)} \vphantom{ }
\begin{table}[h!]
\centering
\setlength{\fboxsep}{0pt}
\setlength{\fboxrule}{2pt}
\fbox{%
\begin{tabular}{c|>{\arraybackslash}m{0.615\textwidth}}
\hline
\multirow{3}{*}{\raisebox{-\totalheight}{\includegraphics[width=0.5\textwidth,height=5.3cm,keepaspectratio]{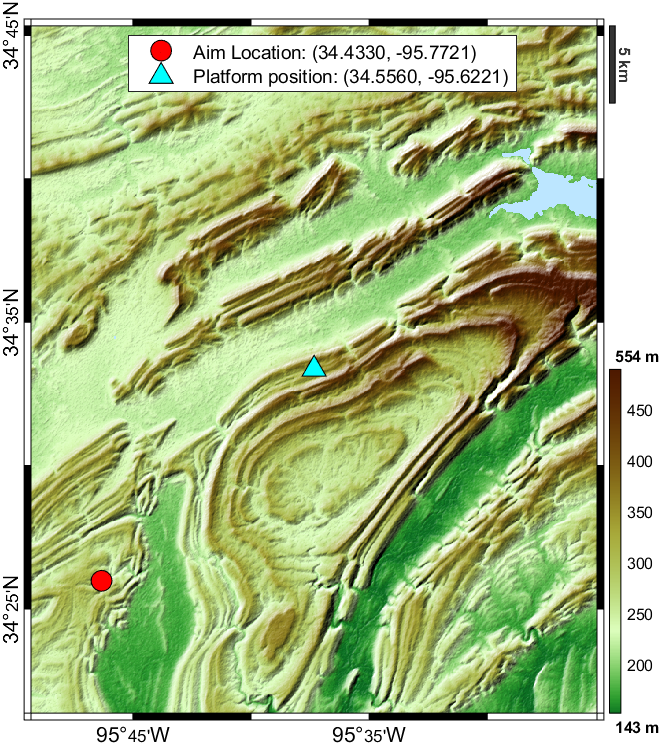}}} \\[-0.1cm] & \textbf{Platform Aim Location} (lat, lon): $(34.4330^{\circ}, -95.7721^{\circ})$ \\[0.2cm]
\cline{2-2} \\[-0.1cm] & 
\textbf{Platform Position} (lat, lon): $(34.5560^{\circ}, -95.6221^{\circ})$ \\[0.2cm]
\cline{2-2} \\[-0.1cm] & 
\textbf{Folder}: \textbf{/CLUTTER/num62\_lat34.556\_lon-95.6221/} \\[0.2cm]
\cline{2-2} \\[-0.1cm] &
\parbox[t][3.0cm][t]{0.62\textwidth}{\textbf{Description}: Scenario $q_i = 35 \ (i = 62)$ in RASPNet is in the Ouachita Mountains of Oklahoma. The radar platform is positioned at latitude $34.5560^{\circ}$ N and longitude $95.6221^{\circ}$ W. It aims towards a location at latitude $34.4330^{\circ}$ N and longitude $95.7721^{\circ}$ W. This region is characterized by small mountains and hills, and has some terrain variability.} \\
\hline
\end{tabular}
}
\end{table}

\paragraph{\underline{Scenario $q_i = 36$}: Narragansett Bay, RI ($i = 96$)} \vphantom{ }
\begin{table}[h!]
\centering
\setlength{\fboxsep}{0pt}
\setlength{\fboxrule}{2pt}
\fbox{%
\begin{tabular}{c|>{\arraybackslash}m{0.615\textwidth}}
\hline
\multirow{3}{*}{\raisebox{-\totalheight}{\includegraphics[width=0.5\textwidth,height=5.3cm,keepaspectratio]{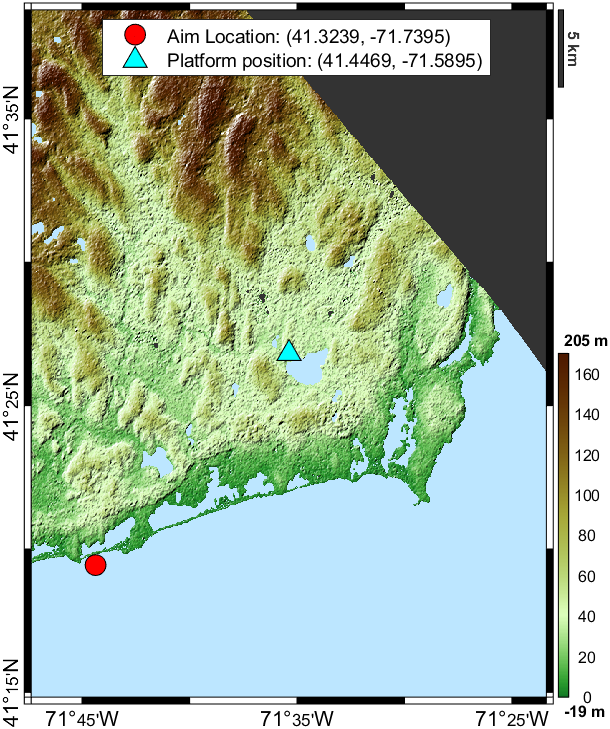}}} \\[-0.1cm] & \textbf{Platform Aim Location} (lat, lon): $(41.3239^{\circ}, -71.7395^{\circ})$ \\[0.2cm]
\cline{2-2} \\[-0.1cm] & 
\textbf{Platform Position} (lat, lon): $(41.4469^{\circ}, -71.5895^{\circ})$ \\[0.2cm]
\cline{2-2} \\[-0.1cm] & 
\textbf{Folder}: \textbf{/CLUTTER/num96\_lat41.4469\_lon-71.5895/} \\[0.2cm]
\cline{2-2} \\[-0.1cm] &
\parbox[t][3.0cm][t]{0.62\textwidth}{\textbf{Description}: Scenario $q_i = 36 \ (i = 96)$ in RASPNet is near Narragansett Bay in Rhode Island. The radar platform is positioned at latitude $41.4469^{\circ}$ N and longitude $71.5895^{\circ}$ W. It aims towards a location at latitude $41.3239^{\circ}$ N and longitude $71.7395^{\circ}$ W. This is a land-to-water transition region, where the aim location is on the coastline.} \\
\hline
\end{tabular}
}
\end{table}

\paragraph{\underline{Scenario $q_i = 37$}: Mobile Bay, AL ($i = 69$)} \vphantom{ }
\begin{table}[h!]
\centering
\setlength{\fboxsep}{0pt}
\setlength{\fboxrule}{2pt}
\fbox{%
\begin{tabular}{c|>{\arraybackslash}m{0.615\textwidth}}
\hline
\multirow{3}{*}{\raisebox{-\totalheight}{\includegraphics[width=0.5\textwidth,height=5.3cm,keepaspectratio]{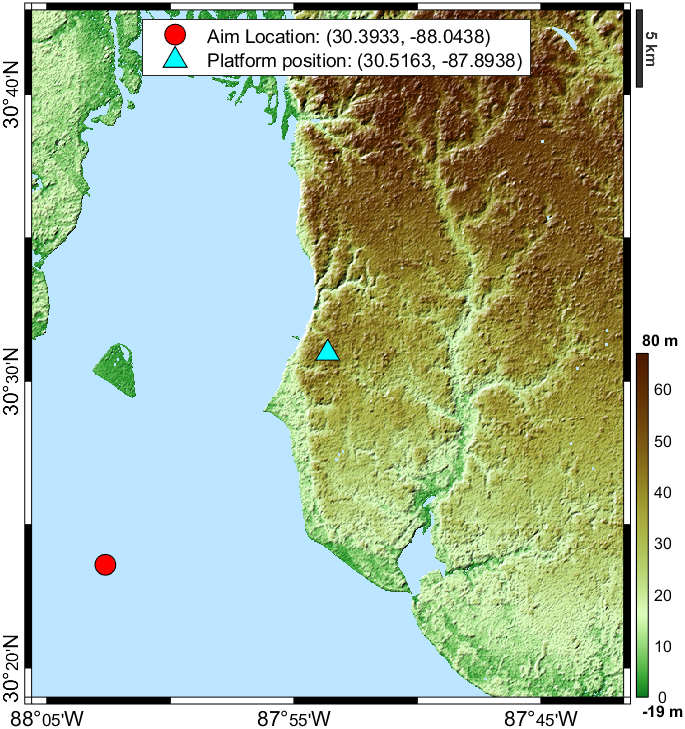}}} \\[-0.1cm] & \textbf{Platform Aim Location} (lat, lon): $(30.3933^{\circ}, -88.0438^{\circ})$ \\[0.2cm]
\cline{2-2} \\[-0.1cm] & 
\textbf{Platform Position} (lat, lon): $(30.5163^{\circ}, -87.8938^{\circ})$ \\[0.2cm]
\cline{2-2} \\[-0.1cm] & 
\textbf{Folder}: \textbf{/CLUTTER/num69\_lat30.5163\_lon-87.8938/} \\[0.2cm]
\cline{2-2} \\[-0.1cm] &
\parbox[t][3.0cm][t]{0.62\textwidth}{\textbf{Description}: Scenario $q_i = 37 \ (i = 69)$ in RASPNet is on the Mobile Bay coastline of Alabama. The radar platform is positioned at latitude $30.5163^{\circ}$ N and longitude $87.8938^{\circ}$ W. It aims towards a location at latitude $30.3933^{\circ}$ N and longitude $88.0438^{\circ}$ W. This is a land-to-water transition region, where the aim location is on the water surface.} \\
\hline
\end{tabular}
}
\end{table}

\paragraph{\underline{Scenario $q_i = 38$}: Lake Ontario Coast, NY ($i = 89$)} \vphantom{ }
\begin{table}[h!]
\centering
\setlength{\fboxsep}{0pt}
\setlength{\fboxrule}{2pt}
\fbox{%
\begin{tabular}{c|>{\arraybackslash}m{0.615\textwidth}}
\hline
\multirow{3}{*}{\raisebox{-\totalheight}{\includegraphics[width=0.5\textwidth,height=5.3cm,keepaspectratio]{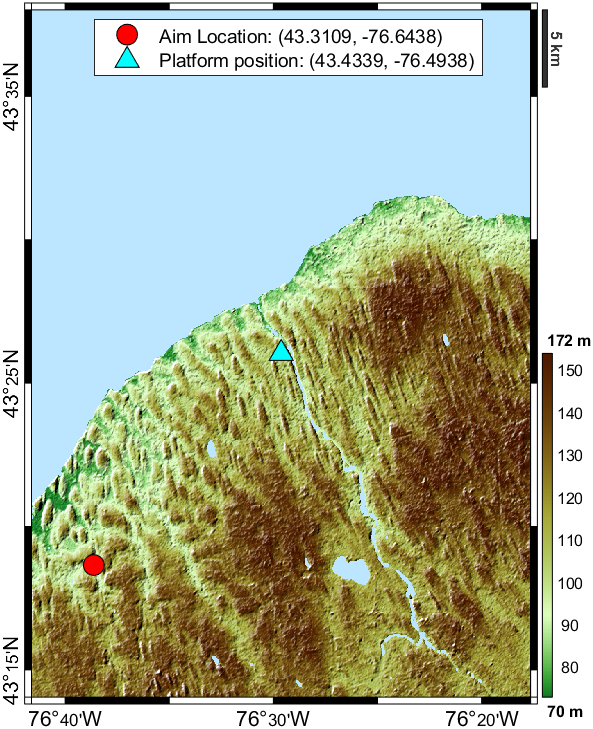}}} \\[-0.1cm] & \textbf{Platform Aim Location} (lat, lon): $(43.3109^{\circ}, -76.6438^{\circ})$ \\[0.2cm]
\cline{2-2} \\[-0.1cm] & 
\textbf{Platform Position} (lat, lon): $(43.4339^{\circ}, -76.4938^{\circ})$ \\[0.2cm]
\cline{2-2} \\[-0.1cm] & 
\textbf{Folder}: \textbf{/CLUTTER/num89\_lat43.4339\_lon-76.4938/} \\[0.2cm]
\cline{2-2} \\[-0.1cm] &
\parbox[t][3.0cm][t]{0.62\textwidth}{\textbf{Description}: Scenario $q_i = 38 \ (i = 89)$ in RASPNet is on the Lake Ontario coastline of New York. The radar platform is positioned at latitude $43.4339^{\circ}$ N and longitude $76.4938^{\circ}$ W. It aims towards a location at latitude $43.3109^{\circ}$ N and longitude $76.6438^{\circ}$ W. The radar is positioned near the coastline, the aim direction is inland, and the terrain has some variability.} \\
\hline
\end{tabular}
}
\end{table}

\paragraph{\underline{Scenario $q_i = 39$}: Central North Dakota, ND ($i = 57$)} \vphantom{ }
\begin{table}[h!]
\centering
\setlength{\fboxsep}{0pt}
\setlength{\fboxrule}{2pt}
\fbox{%
\begin{tabular}{c|>{\arraybackslash}m{0.615\textwidth}}
\hline
\multirow{3}{*}{\raisebox{-\totalheight}{\includegraphics[width=0.5\textwidth,height=5.3cm,keepaspectratio]{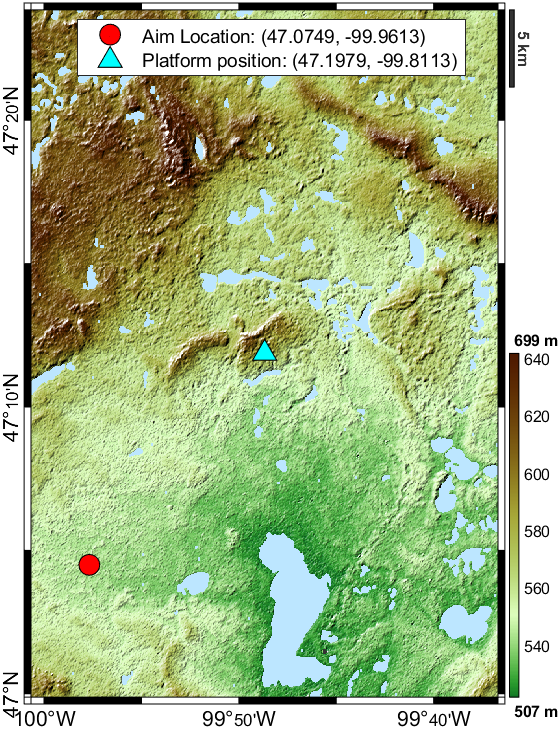}}} \\[-0.1cm] & \textbf{Platform Aim Location} (lat, lon): $(47.0749^{\circ}, -99.9613^{\circ})$ \\[0.2cm]
\cline{2-2} \\[-0.1cm] & 
\textbf{Platform Position} (lat, lon): $(47.1979^{\circ}, -99.8113^{\circ})$ \\[0.2cm]
\cline{2-2} \\[-0.1cm] & 
\textbf{Folder}: \textbf{/CLUTTER/num57\_lat47.1979\_lon-99.8113/} \\[0.2cm]
\cline{2-2} \\[-0.1cm] &
\parbox[t][3.0cm][t]{0.62\textwidth}{\textbf{Description}: Scenario $q_i = 39 \ (i = 57)$ in RASPNet is in the central prairie region of North Dakota. The radar platform is positioned at latitude $47.1979^{\circ}$ N and longitude $99.8113^{\circ}$ W. It aims towards a location at latitude $47.0749^{\circ}$ N and longitude $99.9613^{\circ}$ W. This region is known for its grassland topography and lakes, wherein the terrain has some variability.} \\
\hline
\end{tabular}
}
\end{table}

\paragraph{\underline{Scenario $q_i = 40$}: Salton Sea Coast, CA ($i = 24$)} \vphantom{ }
\begin{table}[h!]
\centering
\setlength{\fboxsep}{0pt}
\setlength{\fboxrule}{2pt}
\fbox{%
\begin{tabular}{c|>{\arraybackslash}m{0.615\textwidth}}
\hline
\multirow{3}{*}{\raisebox{-\totalheight}{\includegraphics[width=0.5\textwidth,height=5.3cm,keepaspectratio]{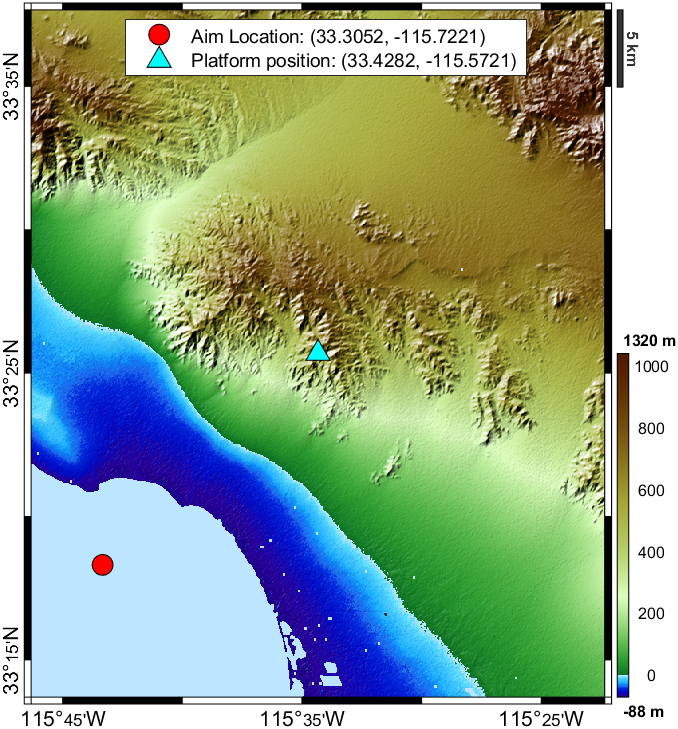}}} \\[-0.1cm] & \textbf{Platform Aim Location} (lat, lon): $(33.3052^{\circ}, -115.7221^{\circ})$ \\[0.2cm]
\cline{2-2} \\[-0.1cm] & 
\textbf{Platform Position} (lat, lon): $(33.4282^{\circ}, -115.5721^{\circ})$ \\[0.2cm]
\cline{2-2} \\[-0.1cm] & 
\textbf{Folder}: \textbf{/CLUTTER/num24\_lat33.4282\_lon-115.5721/} \\[0.2cm]
\cline{2-2} \\[-0.1cm] &
\parbox[t][3.0cm][t]{0.62\textwidth}{\textbf{Description}: Scenario $q_i = 40 \ (i = 24)$ in RASPNet is on the Salton Sea coastline in California. The radar platform is positioned at latitude $33.4282^{\circ}$ N and longitude $115.5721^{\circ}$ W. It aims towards a location at latitude $33.3052^{\circ}$ N and longitude $115.7221^{\circ}$ W. The radar is located above a mountain ridge, and the aim location is on the water surface.} \\
\hline
\end{tabular}
}
\end{table}

\paragraph{\underline{Scenario $q_i = 41$}: Colorado Plateau, NM ($i = 44$)} \vphantom{ }
\begin{table}[h!]
\centering
\setlength{\fboxsep}{0pt}
\setlength{\fboxrule}{2pt}
\fbox{%
\begin{tabular}{c|>{\arraybackslash}m{0.615\textwidth}}
\hline
\multirow{3}{*}{\raisebox{-\totalheight}{\includegraphics[width=0.5\textwidth,height=5.3cm,keepaspectratio]{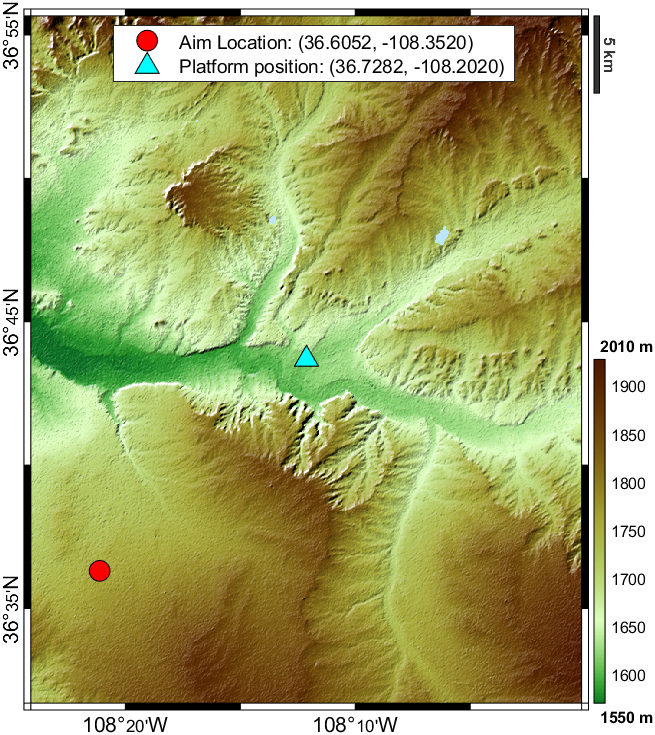}}} \\[-0.1cm] & \textbf{Platform Aim Location} (lat, lon): $(36.6052^{\circ}, -108.3520^{\circ})$ \\[0.2cm]
\cline{2-2} \\[-0.1cm] & 
\textbf{Platform Position} (lat, lon): $(36.7282^{\circ}, -108.2020^{\circ})$ \\[0.2cm]
\cline{2-2} \\[-0.1cm] & 
\textbf{Folder}: \textbf{/CLUTTER/num44\_lat36.7282\_lon-108.202/} \\[0.2cm]
\cline{2-2} \\[-0.1cm] &
\parbox[t][3.0cm][t]{0.62\textwidth}{\textbf{Description}: Scenario $q_i = 41 \ (i = 44)$ in RASPNet is in the Colorado Plateau in New Mexico. The radar platform is positioned at latitude $36.7282^{\circ}$ N and longitude $108.2020^{\circ}$ W. It aims towards a location at latitude $36.6052^{\circ}$ N and longitude $108.3520^{\circ}$ W. This region is known for its nonuniform topography, wherein the terrain has some variability.} \\
\hline
\end{tabular}
}
\end{table}

\paragraph{\underline{Scenario $q_i = 42$}: Santa Monica Coast, CA ($i = 16$)} \vphantom{ }
\begin{table}[h!]
\centering
\setlength{\fboxsep}{0pt}
\setlength{\fboxrule}{2pt}
\fbox{%
\begin{tabular}{c|>{\arraybackslash}m{0.615\textwidth}}
\hline
\multirow{3}{*}{\raisebox{-\totalheight}{\includegraphics[width=0.5\textwidth,height=5.3cm,keepaspectratio]{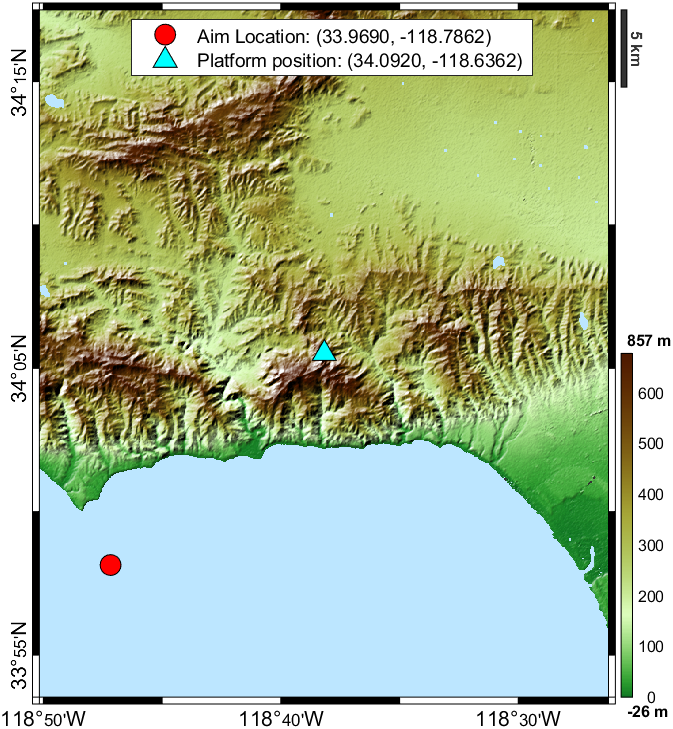}}} \\[-0.1cm] & \textbf{Platform Aim Location} (lat, lon): $(33.9690^{\circ}, -118.7862^{\circ})$ \\[0.2cm]
\cline{2-2} \\[-0.1cm] & 
\textbf{Platform Position} (lat, lon): $(34.0920^{\circ}, -118.6362^{\circ})$ \\[0.2cm]
\cline{2-2} \\[-0.1cm] & 
\textbf{Folder}: \textbf{/CLUTTER/num16\_lat34.092\_lon-118.6362/} \\[0.2cm]
\cline{2-2} \\[-0.1cm] &
\parbox[t][3.0cm][t]{0.62\textwidth}{\textbf{Description}: Scenario $q_i = 42 \ (i = 16)$ in RASPNet is on the Santa Monica coastline in California. The radar platform is positioned at latitude $34.0920^{\circ}$ N and longitude $118.6362^{\circ}$ W. It aims towards a location at latitude $33.9690^{\circ}$ N and longitude $118.7862^{\circ}$ W. The radar is located above a mountain ridge, and the aim location is on the water surface.} \\
\hline
\end{tabular}
}
\end{table}

\paragraph{\underline{Scenario $q_i = 43$}: Boston Mountains, AR ($i = 63$)} \vphantom{ }
\begin{table}[h!]
\centering
\setlength{\fboxsep}{0pt}
\setlength{\fboxrule}{2pt}
\fbox{%
\begin{tabular}{c|>{\arraybackslash}m{0.615\textwidth}}
\hline
\multirow{3}{*}{\raisebox{-\totalheight}{\includegraphics[width=0.5\textwidth,height=5.3cm,keepaspectratio]{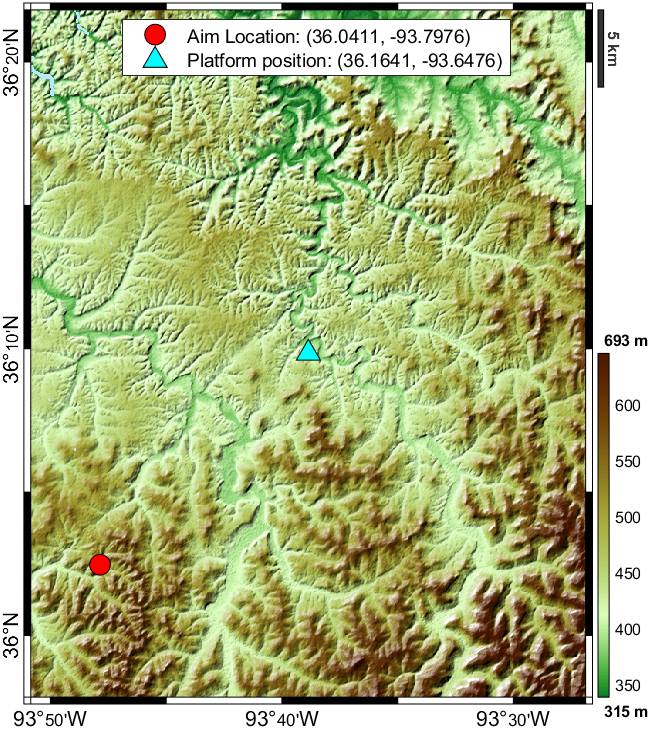}}} \\[-0.1cm] & \textbf{Platform Aim Location} (lat, lon): $(36.0411^{\circ}, -93.7976^{\circ})$ \\[0.2cm]
\cline{2-2} \\[-0.1cm] & 
\textbf{Platform Position} (lat, lon): $(36.1641^{\circ}, -93.6476^{\circ})$ \\[0.2cm]
\cline{2-2} \\[-0.1cm] & 
\textbf{Folder}: \textbf{/CLUTTER/num63\_lat36.1641\_lon-93.6476/} \\[0.2cm]
\cline{2-2} \\[-0.1cm] &
\parbox[t][3.0cm][t]{0.62\textwidth}{\textbf{Description}: Scenario $q_i = 43 \ (i = 63)$ in RASPNet is in the Boston Mountains region of Arkansas. The radar platform is positioned at latitude $36.1641^{\circ}$ N and longitude $93.6476^{\circ}$ W. It aims towards a location at latitude $36.0411^{\circ}$ N and longitude $93.7976^{\circ}$ W. This region is characterized by small mountains and hills, and has some terrain variability.} \\
\hline
\end{tabular}
}
\end{table}

\paragraph{\underline{Scenario $q_i = 44$}: Lake Superior Coast, WI ($i = 66$)} \vphantom{ }
\begin{table}[h!]
\centering
\setlength{\fboxsep}{0pt}
\setlength{\fboxrule}{2pt}
\fbox{%
\begin{tabular}{c|>{\arraybackslash}m{0.615\textwidth}}
\hline
\multirow{3}{*}{\raisebox{-\totalheight}{\includegraphics[width=0.5\textwidth,height=5.3cm,keepaspectratio]{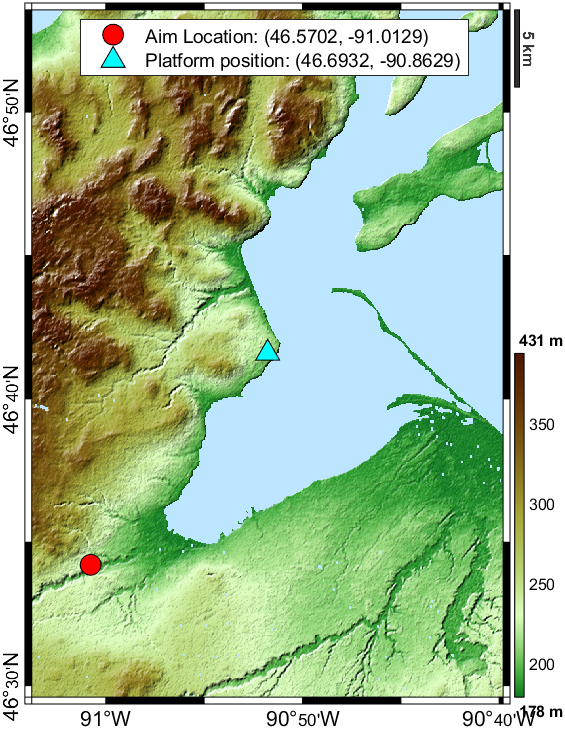}}} \\[-0.1cm] & \textbf{Platform Aim Location} (lat, lon): $(46.5702^{\circ}, -91.0129^{\circ})$ \\[0.2cm]
\cline{2-2} \\[-0.1cm] & 
\textbf{Platform Position} (lat, lon): $(46.6932^{\circ}, -90.8629^{\circ})$ \\[0.2cm]
\cline{2-2} \\[-0.1cm] & 
\textbf{Folder}: \textbf{/CLUTTER/num66\_lat46.6932\_lon-90.8629/} \\[0.2cm]
\cline{2-2} \\[-0.1cm] &
\parbox[t][3.0cm][t]{0.62\textwidth}{\textbf{Description}: Scenario $q_i = 44 \ (i = 66)$ in RASPNet is on the Lake Superior coastline in Wisconsin. The radar platform is positioned at latitude $46.6932^{\circ}$ N and longitude $90.8629^{\circ}$ W. It aims towards a location at latitude $46.5702^{\circ}$ N and longitude $91.0129^{\circ}$ W. The radar is positioned near the coastline, the aim direction is inland, and the terrain has some variability.} \\
\hline
\end{tabular}
}
\end{table}

\paragraph{\underline{Scenario $q_i = 45$}: Massachusetts Bay, MA ($i = 98$)} \vphantom{ }
\begin{table}[h!]
\centering
\setlength{\fboxsep}{0pt}
\setlength{\fboxrule}{2pt}
\fbox{%
\begin{tabular}{c|>{\arraybackslash}m{0.615\textwidth}}
\hline
\multirow{3}{*}{\raisebox{-\totalheight}{\includegraphics[width=0.5\textwidth,height=5.3cm,keepaspectratio]{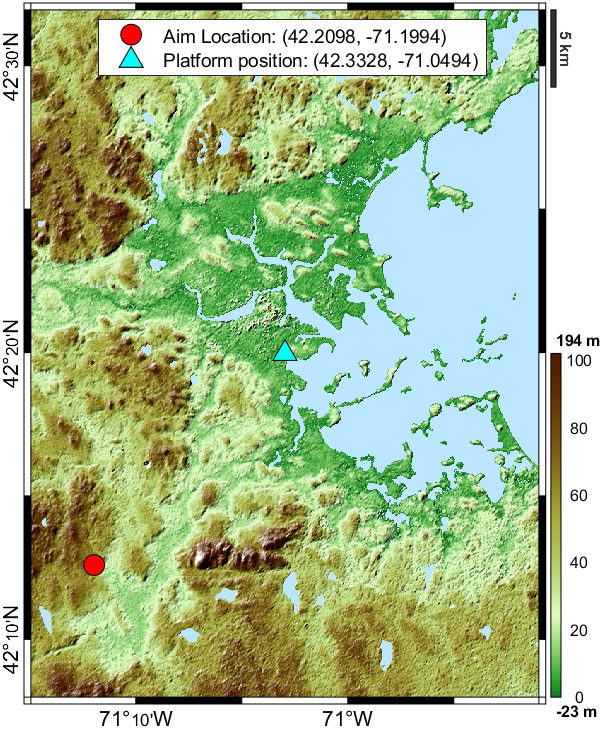}}} \\[-0.1cm] & \textbf{Platform Aim Location} (lat, lon): $(42.2098^{\circ}, -71.1994^{\circ})$ \\[0.2cm]
\cline{2-2} \\[-0.1cm] & 
\textbf{Platform Position} (lat, lon): $(42.3328^{\circ}, -71.0494^{\circ})$ \\[0.2cm]
\cline{2-2} \\[-0.1cm] & 
\textbf{Folder}: \textbf{/CLUTTER/num98\_lat42.3328\_lon-71.0494/} \\[0.2cm]
\cline{2-2} \\[-0.1cm] &
\parbox[t][3.0cm][t]{0.62\textwidth}{\textbf{Description}: Scenario $q_i = 45 \ (i = 98)$ in RASPNet is on the Massachusetts Bay coastline near Boston. The radar platform is positioned at latitude $42.3328^{\circ}$ N and longitude $71.0494^{\circ}$ W. It aims towards a location at latitude $42.2098^{\circ}$ N and longitude $71.1994^{\circ}$ W. The radar is positioned near the coastline, the aim direction is inland, and the terrain has some variability.} \\
\hline
\end{tabular}
}
\end{table}

\paragraph{\underline{Scenario $q_i = 46$}: The Everglades, FL ($i = 83$)} \vphantom{ }
\begin{table}[h!]
\centering
\setlength{\fboxsep}{0pt}
\setlength{\fboxrule}{2pt}
\fbox{%
\begin{tabular}{c|>{\arraybackslash}m{0.615\textwidth}}
\hline
\multirow{3}{*}{\raisebox{-\totalheight}{\includegraphics[width=0.5\textwidth,height=5.3cm,keepaspectratio]{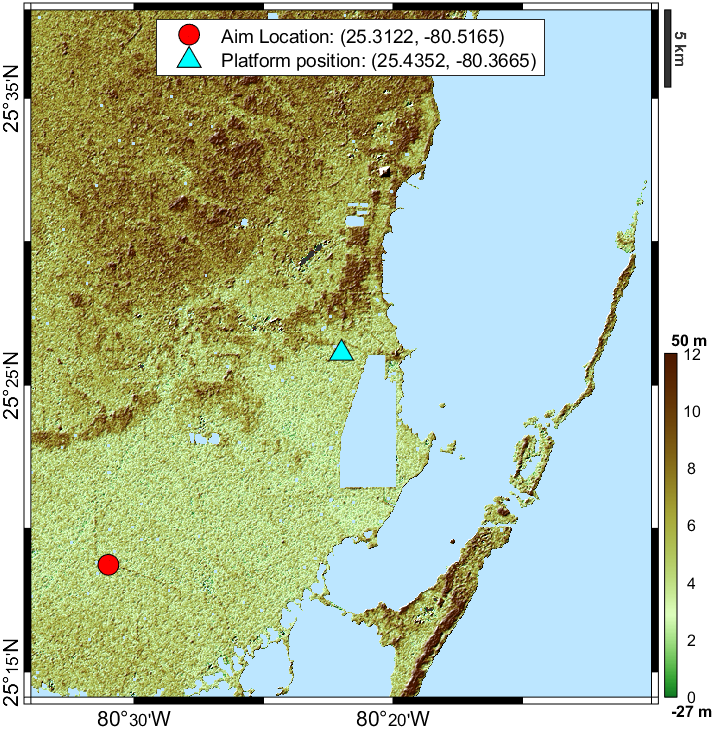}}} \\[-0.1cm] & \textbf{Platform Aim Location} (lat, lon): $(25.3122^{\circ}, -80.5165^{\circ})$ \\[0.2cm]
\cline{2-2} \\[-0.1cm] & 
\textbf{Platform Position} (lat, lon): $(25.4352^{\circ}, -80.3665^{\circ})$ \\[0.2cm]
\cline{2-2} \\[-0.1cm] & 
\textbf{Folder}: \textbf{/CLUTTER/num83\_lat25.4352\_lon-80.3665/} \\[0.2cm]
\cline{2-2} \\[-0.1cm] &
\parbox[t][3.0cm][t]{0.62\textwidth}{\textbf{Description}: Scenario $q_i = 46 \ (i = 83)$ in RASPNet is in the Everglades region of South Florida. The radar platform is positioned at latitude $25.4352^{\circ}$ N and longitude $80.3665^{\circ}$ W. It aims towards a location at latitude $25.3122^{\circ}$ N and longitude $80.5165^{\circ}$ W. This region is characterized by a shallow and flat wetland environment, with low terrain variability.} \\
\hline
\end{tabular}
}
\end{table}

\paragraph{\underline{Scenario $q_i = 47$}: Upper Peninsula, MI ($i = 70$)} \vphantom{ }
\begin{table}[h!]
\centering
\setlength{\fboxsep}{0pt}
\setlength{\fboxrule}{2pt}
\fbox{%
\begin{tabular}{c|>{\arraybackslash}m{0.615\textwidth}}
\hline
\multirow{3}{*}{\raisebox{-\totalheight}{\includegraphics[width=0.5\textwidth,height=5.3cm,keepaspectratio]{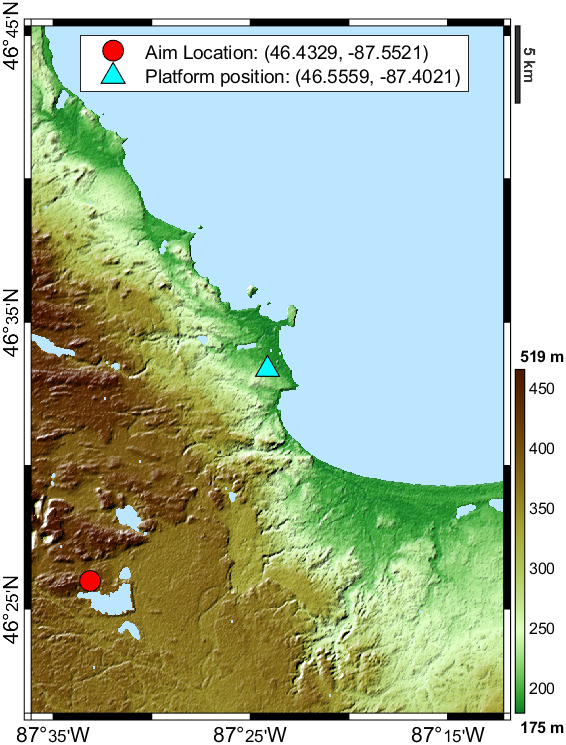}}} \\[-0.1cm] & \textbf{Platform Aim Location} (lat, lon): $(46.4329^{\circ}, -87.5521^{\circ})$ \\[0.2cm]
\cline{2-2} \\[-0.1cm] & 
\textbf{Platform Position} (lat, lon): $(46.5559^{\circ}, -87.4021^{\circ})$ \\[0.2cm]
\cline{2-2} \\[-0.1cm] & 
\textbf{Folder}: \textbf{/CLUTTER/num70\_lat46.5559\_lon-87.4021/} \\[0.2cm]
\cline{2-2} \\[-0.1cm] &
\parbox[t][3.0cm][t]{0.62\textwidth}{\textbf{Description}: Scenario $q_i = 47 \ (i = 70)$ in RASPNet is in the Upper Peninsula region of Michigan. The radar platform is positioned at latitude $46.5559^{\circ}$ N and longitude $87.4021^{\circ}$ W. It aims towards a location at latitude $46.4329^{\circ}$ N and longitude $87.5521^{\circ}$ W. The radar is positioned near the coastline, the aim direction is inland, and the terrain has some variability.} \\
\hline
\end{tabular}
}
\end{table}

\paragraph{\underline{Scenario $q_i = 48$}: Sandia Crest, NM ($i = 46)$} \vphantom{ }
\begin{table}[h!]
\centering
\setlength{\fboxsep}{0pt}
\setlength{\fboxrule}{2pt}
\fbox{%
\begin{tabular}{c|>{\arraybackslash}m{0.615\textwidth}}
\hline
\multirow{3}{*}{\raisebox{-\totalheight}{\includegraphics[width=0.5\textwidth,height=5.3cm,keepaspectratio]{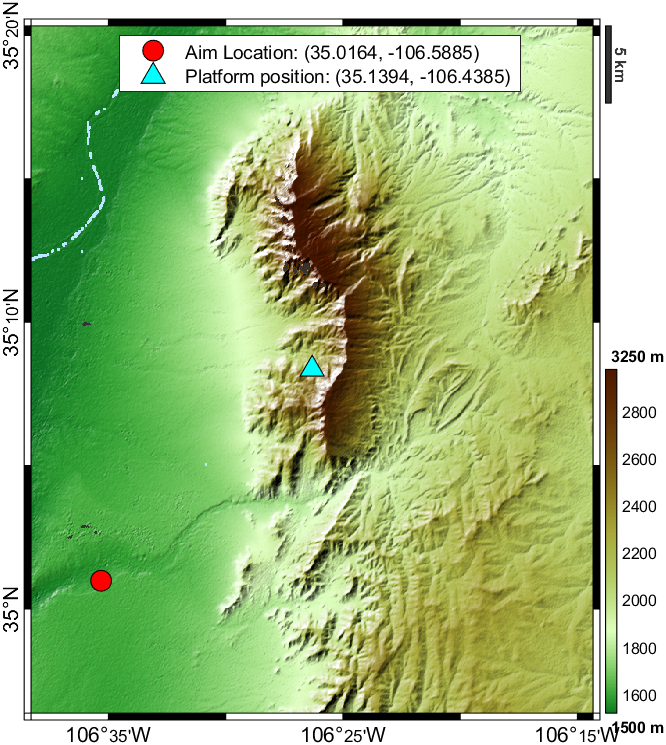}}} \\[-0.1cm] & \textbf{Platform Aim Location} (lat, lon): $(35.0164^{\circ}, -106.5885^{\circ})$ \\[0.2cm]
\cline{2-2} \\[-0.1cm] & 
\textbf{Platform Position} (lat, lon): $(35.1394^{\circ}, -106.4385^{\circ})$ \\[0.2cm]
\cline{2-2} \\[-0.1cm] & 
\textbf{Folder}: \textbf{/CLUTTER/num46\_lat35.1394\_lon-106.4385/} \\[0.2cm]
\cline{2-2} \\[-0.1cm] &
\parbox[t][3.0cm][t]{0.62\textwidth}{\textbf{Description}: Scenario $q_i = 48 \ (i = 46)$ in RASPNet is in the Sandia Mountains of New Mexico. The radar platform is positioned at latitude $35.1394^{\circ}$ N and longitude $106.4385^{\circ}$ W. It aims towards a location at latitude $35.0164^{\circ}$ N and longitude $106.5885^{\circ}$ W. This region is noted for its variable terrain, and the radar is positioned behind a mountain crest.} \\
\hline
\end{tabular}
}
\end{table}

\paragraph{\underline{Scenario $q_i = 49$}: Central Nebraska, NE ($i = 58$)} \vphantom{ }
\begin{table}[h!]
\centering
\setlength{\fboxsep}{0pt}
\setlength{\fboxrule}{2pt}
\fbox{%
\begin{tabular}{c|>{\arraybackslash}m{0.615\textwidth}}
\hline
\multirow{3}{*}{\raisebox{-\totalheight}{\includegraphics[width=0.5\textwidth,height=5.1cm,keepaspectratio]{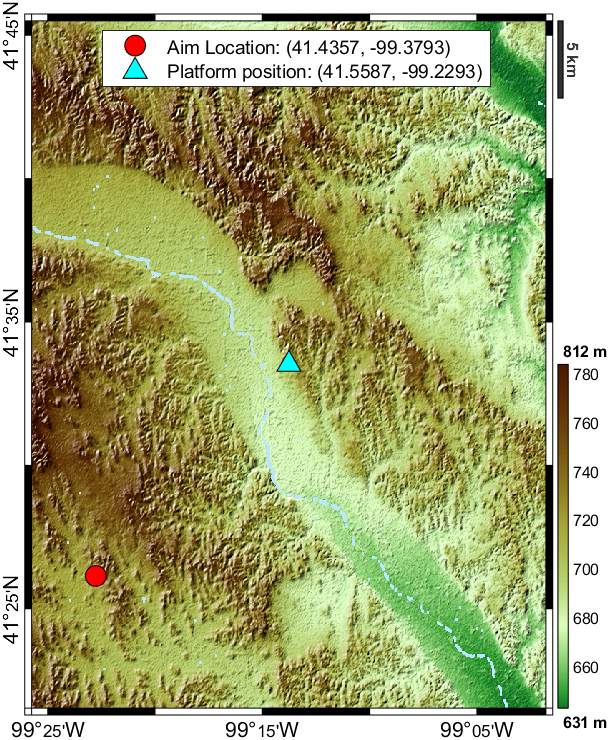}}} \\[-0.1cm] & \textbf{Platform Aim Location} (lat, lon): $(41.4357^{\circ}, -99.3793^{\circ})$ \\[0.2cm]
\cline{2-2} \\[-0.1cm] & 
\textbf{Platform Position} (lat, lon): $(41.5587^{\circ}, -99.2293^{\circ})$ \\[0.2cm]
\cline{2-2} \\[-0.1cm] &
\textbf{Folder}: \textbf{/CLUTTER/num58\_lat41.5587\_lon-99.2293/} \\[0.2cm]
\cline{2-2} \\[-0.1cm] &
\parbox[t][2.6cm][t]{0.62\textwidth}{\textbf{Description}: Scenario $q_i = 49 \ (i = 58)$ in RASPNet is in the Great Plains of Central Nebraska. The radar platform is positioned at latitude $41.5587^{\circ}$ N and longitude $99.2293^{\circ}$ W. It aims towards a location at latitude $41.4357^{\circ}$ N and longitude $99.3793^{\circ}$ W. This region is characterized by an expansive steppe environment, with low terrain variability.} \\
\hline
\end{tabular}
}
\end{table}

\paragraph{\underline{Scenario $q_i = 50$}: Catoctin Mountains, MD ($i = 88$)} \vphantom{ }
\begin{table}[h!]
\centering
\setlength{\fboxsep}{0pt}
\setlength{\fboxrule}{2pt}
\fbox{%
\begin{tabular}{c|>{\arraybackslash}m{0.615\textwidth}}
\hline
\multirow{3}{*}{\raisebox{-\totalheight}{\includegraphics[width=0.5\textwidth,height=5.1cm,keepaspectratio]{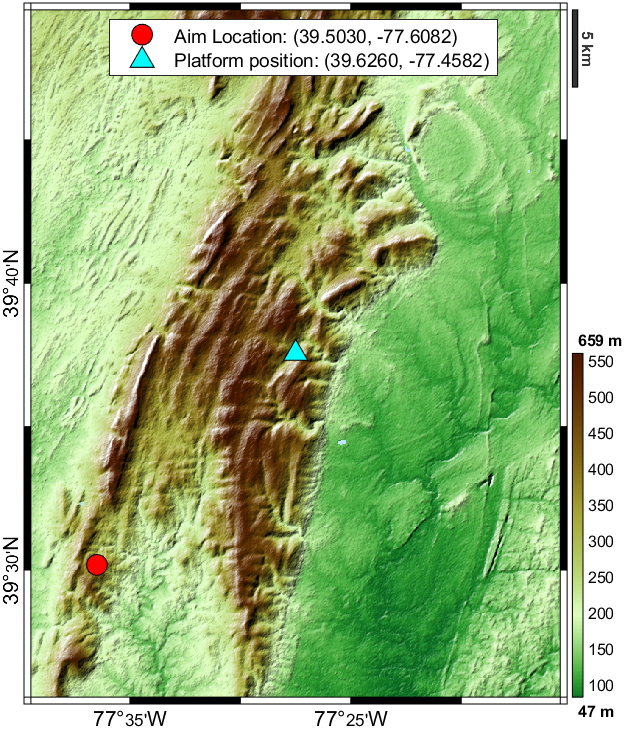}}} \\[-0.1cm] & \textbf{Platform Aim Location} (lat, lon): $(39.5030^{\circ}, -77.6082^{\circ})$ \\[0.2cm]
\cline{2-2} \\[-0.1cm] & 
\textbf{Platform Position} (lat, lon): $(39.6260^{\circ}, -77.4582^{\circ})$ \\[0.2cm]
\cline{2-2} \\[-0.1cm] &
\textbf{Folder}: \textbf{/CLUTTER/num88\_lat39.626\_lon-77.4582/} \\[0.2cm]
\cline{2-2} \\[-0.1cm] &
\parbox[t][2.6cm][t]{0.62\textwidth}{\textbf{Description}: Scenario $q_i = 50 \ (i = 88)$ in RASPNet is in the Catoctin Mountains of Maryland. The radar platform is positioned at latitude $39.6260^{\circ}$ N and longitude $77.4582^{\circ}$ W. It aims towards a location at latitude $39.5030^{\circ}$ N and longitude $77.6082^{\circ}$ W. This region is characterized by mountainous terrain, with increased nonuniformity.} \\
\hline
\end{tabular}
}
\end{table}

\subsection{Challenging Scenarios}
We now outline the Challenging RASPNet scenarios, where $q_i \in \{51,52,\ldots,75\}$.

\paragraph{\underline{Scenario $q_i = 51$}: Straits of Mackinac, MI ($i = 74$)} \vphantom{ }
\begin{table}[h!]
\centering
\setlength{\fboxsep}{0pt}
\setlength{\fboxrule}{2pt}
\fbox{%
\begin{tabular}{c|>{\arraybackslash}m{0.615\textwidth}}
\hline
\multirow{3}{*}{\raisebox{-\totalheight}{\includegraphics[width=0.5\textwidth,height=5.1cm,keepaspectratio]{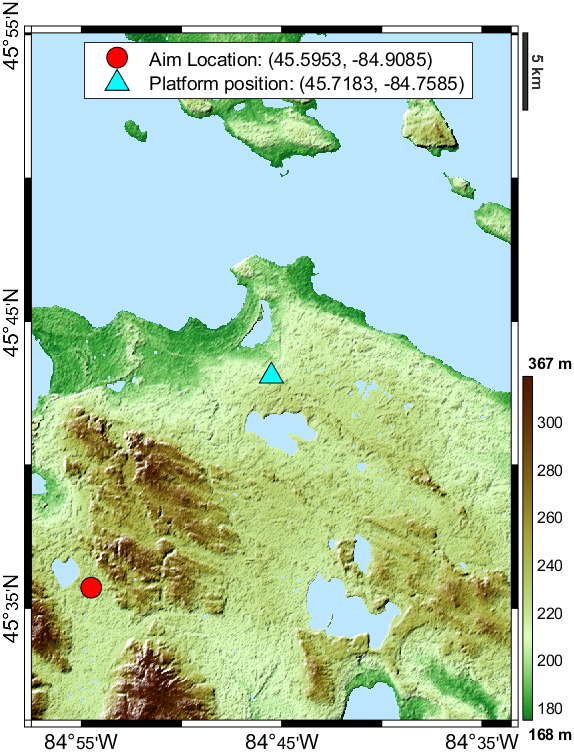}}} \\[-0.1cm] & \textbf{Platform Aim Location} (lat, lon): $(45.5953^{\circ}, -84.9085^{\circ})$ \\[0.2cm]
\cline{2-2} \\[-0.1cm] & 
\textbf{Platform Position} (lat, lon): $(45.7183^{\circ}, -87.7585^{\circ})$ \\[0.2cm]
\cline{2-2} \\[-0.1cm] & 
\textbf{Folder}: \textbf{/CLUTTER/num74\_lat45.7183\_lon-87.7585/} \\[0.2cm]
\cline{2-2} \\[-0.1cm] &
\parbox[t][2.6cm][t]{0.62\textwidth}{\textbf{Description}: Scenario $q_i = 51 \ (i = 74)$ in RASPNet is near the Straits of Mackinac in Michigan. The radar platform is positioned at latitude $45.7183^{\circ}$ N and longitude $-87.7585^{\circ}$ W. It aims towards a location at latitude $45.5953^{\circ}$ N and longitude $-84.9085^{\circ}$ W. The radar is positioned near the coastline, the aim direction is inland, and the terrain has moderate variability.} \\
\hline
\end{tabular}
}
\end{table}

\paragraph{\underline{Scenario $q_i = 52$}: Washburn Range, WY ($i = 40$)}
 \vphantom{ }
\begin{table}[h!]
\centering
\setlength{\fboxsep}{0pt}
\setlength{\fboxrule}{2pt}
\fbox{%
\begin{tabular}{c|>{\arraybackslash}m{0.615\textwidth}}
\hline
\multirow{3}{*}{\raisebox{-\totalheight}{\includegraphics[width=0.5\textwidth,height=5.3cm,keepaspectratio]{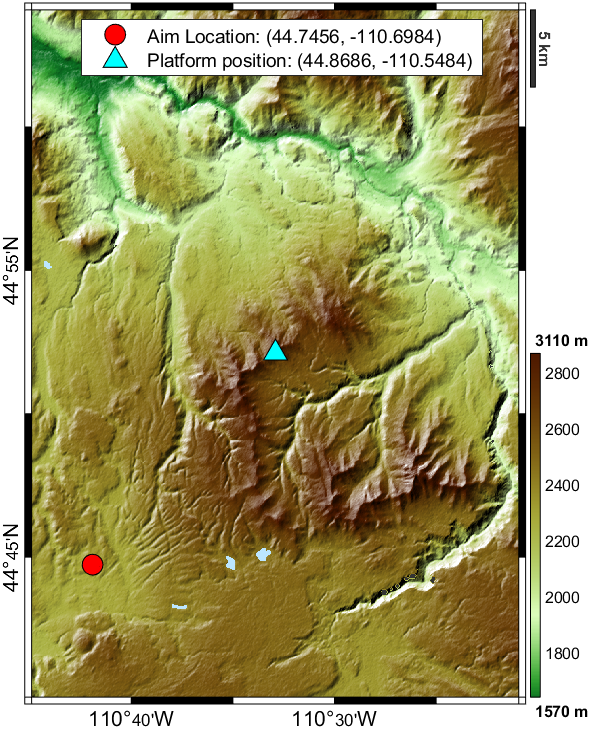}}} \\[-0.1cm] & \textbf{Platform Aim Location} (lat, lon): $(44.7456^{\circ}, -110.6984^{\circ})$ \\[0.2cm]
\cline{2-2} \\[-0.1cm] & 
\textbf{Platform Position} (lat, lon): $(44.8686^{\circ}, -110.5484^{\circ})$ \\[0.2cm]
\cline{2-2} \\[-0.1cm] & 
\textbf{Folder}: \textbf{/CLUTTER/num40\_lat44.8686\_lon-110.5484/} \\[0.2cm]
\cline{2-2} \\[-0.1cm] &
\parbox[t][3.0cm][t]{0.62\textwidth}{\textbf{Description}: Scenario $q_i = 52 \ (i = 40)$ in RASPNet is in the Washburn Range of Wyoming. The radar platform is positioned at latitude $44.8686^{\circ}$ N and longitude $110.5484^{\circ}$ W. It aims towards a location at latitude $44.7456^{\circ}$ N and longitude $110.6984^{\circ}$ W. The radar is positioned above a mountain summit, and the aim location has moderate terrain variability.} \\
\hline
\end{tabular}
}
\end{table}

\paragraph{\underline{Scenario $q_i = 53$}:  Colorado Plateau, CO ($i = 45$)} \vphantom{ }
\begin{table}[h!]
\centering
\setlength{\fboxsep}{0pt}
\setlength{\fboxrule}{2pt}
\fbox{%
\begin{tabular}{c|>{\arraybackslash}m{0.615\textwidth}}
\hline
\multirow{3}{*}{\raisebox{-\totalheight}{\includegraphics[width=0.5\textwidth,height=5.3cm,keepaspectratio]{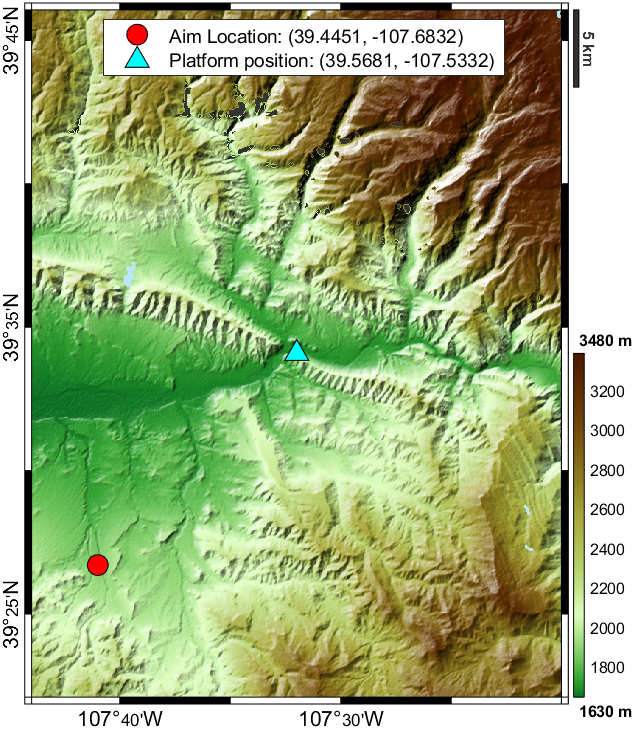}}} \\[-0.1cm] & \textbf{Platform Aim Location} (lat, lon): $(39.4451^{\circ}, -107.6832^{\circ})$ \\[0.2cm]
\cline{2-2} \\[-0.1cm] & 
\textbf{Platform Position} (lat, lon): $(39.5681^{\circ}, -107.5332^{\circ})$ \\[0.2cm]
\cline{2-2} \\[-0.1cm] & 
\textbf{Folder}: \textbf{/CLUTTER/num45\_lat39.5681\_lon-107.5332/} \\[0.2cm]
\cline{2-2} \\[-0.1cm] &
\parbox[t][3.0cm][t]{0.62\textwidth}{\textbf{Description}: Scenario $q_i = 53 \ (i = 45)$ in RASPNet is in the Colorado Plateau of Western Colorado. The radar platform is positioned at latitude $39.5681^{\circ}$ N and longitude $107.5332^{\circ}$ W. It aims towards a location at latitude $39.4451^{\circ}$ N and longitude $107.6832^{\circ}$ W. The radar is positioned above a ridge system, and the aim location has moderate terrain variability.} \\
\hline
\end{tabular}
}
\end{table}

\paragraph{\underline{Scenario $q_i = 54$}: Bryce Canyon National Park, UT ($i = 34$)} \vphantom{ }
\begin{table}[h!]
\centering
\setlength{\fboxsep}{0pt}
\setlength{\fboxrule}{2pt}
\fbox{%
\begin{tabular}{c|>{\arraybackslash}m{0.615\textwidth}}
\hline
\multirow{3}{*}{\raisebox{-\totalheight}{\includegraphics[width=0.5\textwidth,height=5.3cm,keepaspectratio]{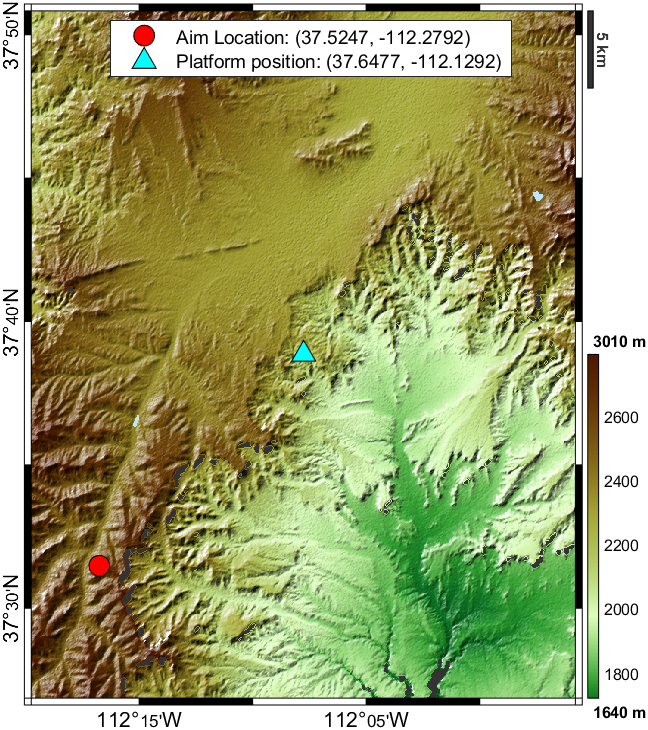}}} \\[-0.1cm] & \textbf{Platform Aim Location} (lat, lon): $(37.5247^{\circ}, -112.2792^{\circ})$ \\[0.2cm]
\cline{2-2} \\[-0.1cm] & 
\textbf{Platform Position} (lat, lon): $(37.6477^{\circ}, -112.1292^{\circ})$ \\[0.2cm]
\cline{2-2} \\[-0.1cm] & 
\textbf{Folder}: \textbf{/CLUTTER/num34\_lat37.6477\_lon-112.1292/} \\[0.2cm]
\cline{2-2} \\[-0.1cm] &
\parbox[t][3.0cm][t]{0.62\textwidth}{\textbf{Description}: Scenario $q_i = 54 \ (i = 34)$ in RASPNet is in Bryce Canyon National Park in Utah. The radar platform is positioned at latitude $37.6477^{\circ}$ N and longitude $112.1292^{\circ}$ W. It aims towards a location at latitude $37.5247^{\circ}$ N and longitude $112.2792^{\circ}$ W. The radar is positioned above a canyon system, and the aim location has moderate terrain variability.} \\
\hline
\end{tabular}
}
\end{table}

\paragraph{\underline{Scenario $q_i = 55$}: Dry Falls, WA ($i = 15$)} \vphantom{ }
\begin{table}[h!]
\centering
\setlength{\fboxsep}{0pt}
\setlength{\fboxrule}{2pt}
\fbox{%
\begin{tabular}{c|>{\arraybackslash}m{0.615\textwidth}}
\hline
\multirow{3}{*}{\raisebox{-\totalheight}{\includegraphics[width=0.5\textwidth,height=5.3cm,keepaspectratio]{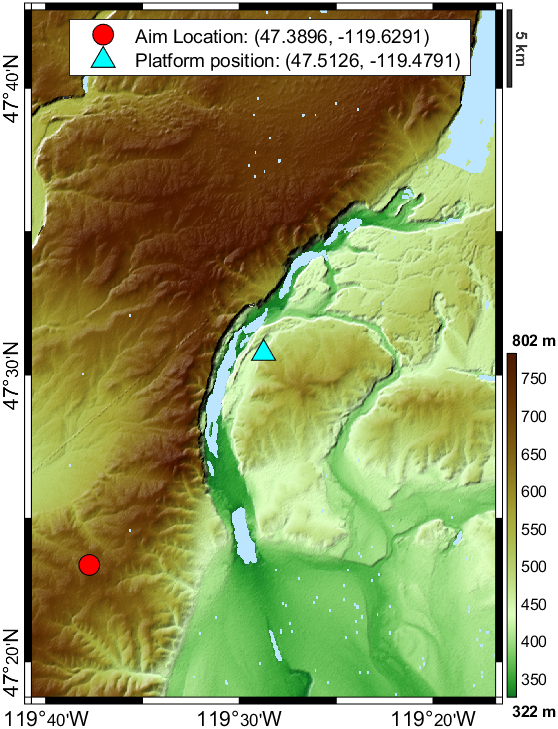}}} \\[-0.1cm] & \textbf{Platform Aim Location} (lat, lon): $(47.3896^{\circ}, -119.6291^{\circ})$ \\[0.2cm]
\cline{2-2} \\[-0.1cm] & 
\textbf{Platform Position} (lat, lon): $(47.5126^{\circ}, -119.4791^{\circ})$ \\[0.2cm]
\cline{2-2} \\[-0.1cm] & 
\textbf{Folder}: \textbf{/CLUTTER/num15\_lat47.5126\_lon-119.4791/} \\[0.2cm]
\cline{2-2} \\[-0.1cm] &
\parbox[t][3.0cm][t]{0.62\textwidth}{\textbf{Description}: Scenario $q_i = 55 \ (i = 15)$ in RASPNet is in Dry Falls State Park in Central Washington. The radar platform is positioned at latitude $47.5126^{\circ}$ N and longitude $119.4791^{\circ}$ W. It aims towards a location at latitude $47.3896^{\circ}$ N and longitude $119.6291^{\circ}$ W. This scenario comprises the rugged Channeled Scablands region, and has moderate terrain variability.} \\
\hline
\end{tabular}
}
\end{table}

\paragraph{\underline{Scenario $q_i = 56$}: Great Salt Lake Coast, UT ($i = 33$)} \vphantom{ }
\begin{table}[h!]
\centering
\setlength{\fboxsep}{0pt}
\setlength{\fboxrule}{2pt}
\fbox{%
\begin{tabular}{c|>{\arraybackslash}m{0.615\textwidth}}
\hline
\multirow{3}{*}{\raisebox{-\totalheight}{\includegraphics[width=0.5\textwidth,height=5.3cm,keepaspectratio]{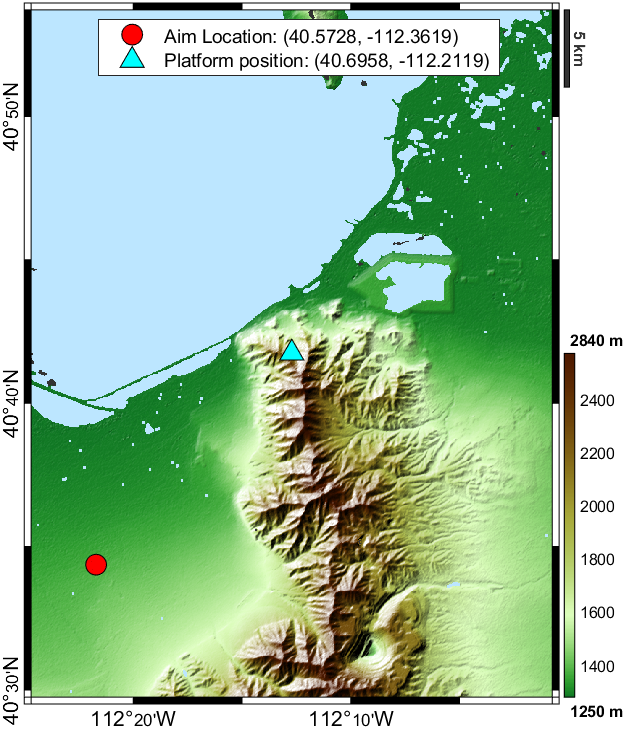}}} \\[-0.1cm] & \textbf{Platform Aim Location} (lat, lon): $(40.5728^{\circ}, -112.3619^{\circ})$ \\[0.2cm]
\cline{2-2} \\[-0.1cm] & 
\textbf{Platform Position} (lat, lon): $(40.6958^{\circ}, -112.2119^{\circ})$ \\[0.2cm]
\cline{2-2} \\[-0.1cm] & 
\textbf{Folder}: \textbf{/CLUTTER/num33\_lat40.6958\_lon-112.2119/} \\[0.2cm]
\cline{2-2} \\[-0.1cm] &
\parbox[t][3.0cm][t]{0.62\textwidth}{\textbf{Description}: Scenario $q_i = 56 \ (i = 33)$ in RASPNet is near the Great Salt Lake coastline in Utah. The radar platform is positioned at latitude $40.6958^{\circ}$ N and longitude $112.2119^{\circ}$ W. It aims towards a location at latitude $40.5728^{\circ}$ N and longitude $112.3619^{\circ}$ W. The radar is positioned near the coastline above a mountain ridge, and the aim direction is inland.} \\
\hline
\end{tabular}
}
\end{table}

\paragraph{\underline{Scenario $q_i = 57$}: Great Smoky Mountains, TN ($i = 76$)}  \vphantom{ }
\begin{table}[h!]
\centering
\setlength{\fboxsep}{0pt}
\setlength{\fboxrule}{2pt}
\fbox{%
\begin{tabular}{c|>{\arraybackslash}m{0.615\textwidth}}
\hline
\multirow{3}{*}{\raisebox{-\totalheight}{\includegraphics[width=0.5\textwidth,height=5.3cm,keepaspectratio]{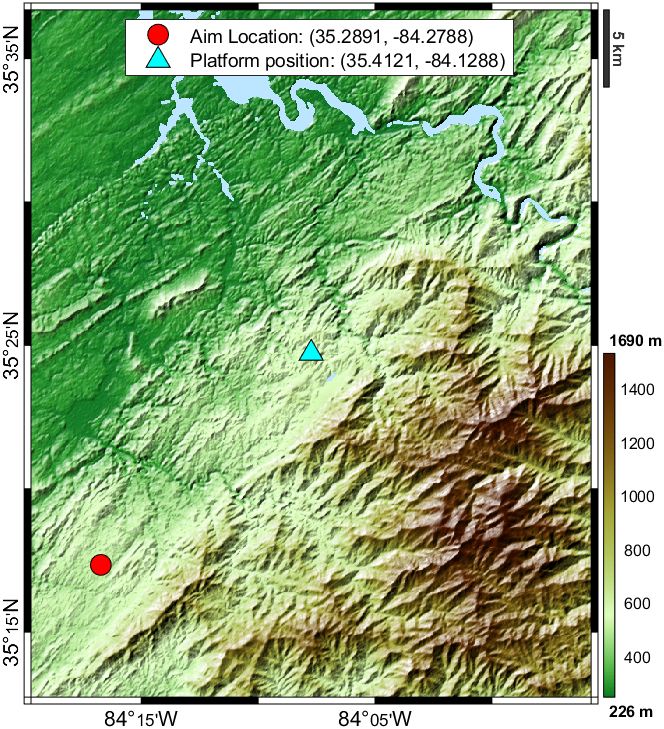}}} \\[-0.1cm] & \textbf{Platform Aim Location} (lat, lon): $(35.2891^{\circ}, -84.2788^{\circ})$ \\[0.2cm]
\cline{2-2} \\[-0.1cm] & 
\textbf{Platform Position} (lat, lon): $(35.4121^{\circ}, -84.1288^{\circ})$ \\[0.2cm]
\cline{2-2} \\[-0.1cm] & 
\textbf{Folder}: \textbf{/CLUTTER/num76\_lat35.4121\_lon-84.1288/} \\[0.2cm]
\cline{2-2} \\[-0.1cm] &
\parbox[t][3.0cm][t]{0.62\textwidth}{\textbf{Description}: Scenario $q_i = 57 \ (i = 76)$ in RASPNet is near Tennessee's Great Smoky Mountains. The radar platform is positioned at latitude $35.4121^{\circ}$ N and longitude $84.1288^{\circ}$ W. It aims towards a location at latitude $35.2891^{\circ}$ N and longitude $84.2788^{\circ}$ W. The radar is located above a series of hills, and the aim location has moderate terrain variability.} \\
\hline
\end{tabular}
}
\end{table}

\paragraph{\underline{Scenario $q_i = 58$}: Cumberland Mountains, TN ($i = 75$)} \vphantom{ }
\begin{table}[h!]
\centering
\setlength{\fboxsep}{0pt}
\setlength{\fboxrule}{2pt}
\fbox{%
\begin{tabular}{c|>{\arraybackslash}m{0.615\textwidth}}
\hline
\multirow{3}{*}{\raisebox{-\totalheight}{\includegraphics[width=0.5\textwidth,height=5.3cm,keepaspectratio]{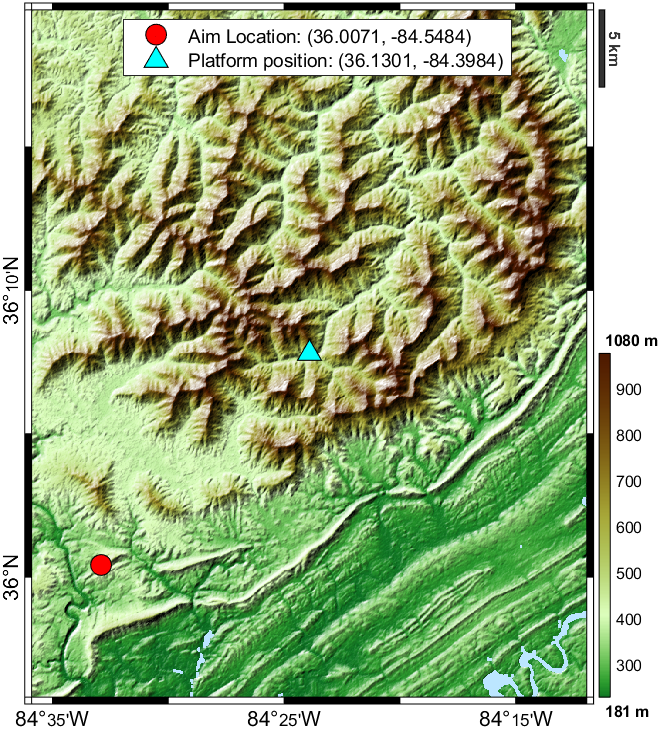}}} \\[-0.1cm] & \textbf{Platform Aim Location} (lat, lon): $(36.0071^{\circ}, -84.5484^{\circ})$ \\[0.2cm]
\cline{2-2} \\[-0.1cm] & 
\textbf{Platform Position} (lat, lon): $(36.1301^{\circ}, -84.3984^{\circ})$ \\[0.2cm]
\cline{2-2} \\[-0.1cm] & 
\textbf{Folder}: \textbf{/CLUTTER/num75\_lat36.1301\_lon-84.3984/} \\[0.2cm]
\cline{2-2} \\[-0.1cm] &
\parbox[t][3.0cm][t]{0.62\textwidth}{\textbf{Description}: Scenario $q_i = 58 \ (i = 75)$ in RASPNet is in the Cumberland Mountains of Tennessee. The radar platform is located at latitude $36.1301^{\circ}$ N and longitude $84.3984^{\circ}$ W. It aims towards a location at latitude $36.0071^{\circ}$ N and longitude $84.5484^{\circ}$ W. The radar is located above a mountain ridge, and the aim location has moderate terrain variability.} \\
\hline
\end{tabular}
}
\end{table}

\paragraph{\underline{Scenario $q_i = 59$}: Lake Pend Orielle Coast, ID ($i = 23$)} \vphantom{ }
\begin{table}[h!]
\centering
\setlength{\fboxsep}{0pt}
\setlength{\fboxrule}{2pt}
\fbox{%
\begin{tabular}{c|>{\arraybackslash}m{0.615\textwidth}}
\hline
\multirow{3}{*}{\raisebox{-\totalheight}{\includegraphics[width=0.5\textwidth,height=5.3cm,keepaspectratio]{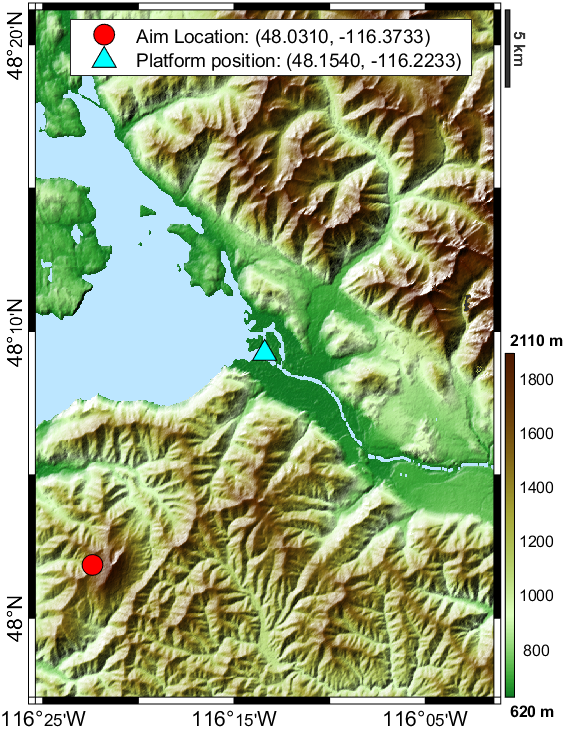}}} \\[-0.1cm] & \textbf{Platform Aim Location} (lat, lon): $(48.0310^{\circ}, -116.3733^{\circ})$ \\[0.2cm]
\cline{2-2} \\[-0.1cm] & 
\textbf{Platform Position} (lat, lon): $(48.1540^{\circ}, -116.2233^{\circ})$ \\[0.2cm]
\cline{2-2} \\[-0.1cm] & 
\textbf{Folder}: \textbf{/CLUTTER/num23\_lat48.154\_lon-116.2233/} \\[0.2cm]
\cline{2-2} \\[-0.1cm] &
\parbox[t][3.0cm][t]{0.62\textwidth}{\textbf{Description}: Scenario $q_i = 59 \ (i = 23)$ in RASPNet is near the Lake Pend Orielle coastline in Idaho. The radar platform is positioned at latitude $48.1540^{\circ}$ N and longitude $116.2233^{\circ}$ W. It aims towards a location at latitude $40.0310^{\circ}$ N and longitude $116.3733^{\circ}$ W. The radar is positioned near the coastline, and the aim direction is inland, located on a mountain ridge.} \\
\hline
\end{tabular}
}
\end{table}

\paragraph{\underline{Scenario $q_i = 60$}: Death Valley National Park ($i = 20$)} \vphantom{ }
\begin{table}[h!]
\centering
\setlength{\fboxsep}{0pt}
\setlength{\fboxrule}{2pt}
\fbox{%
\begin{tabular}{c|>{\arraybackslash}m{0.615\textwidth}}
\hline
\multirow{3}{*}{\raisebox{-\totalheight}{\includegraphics[width=0.5\textwidth,height=5.3cm,keepaspectratio]{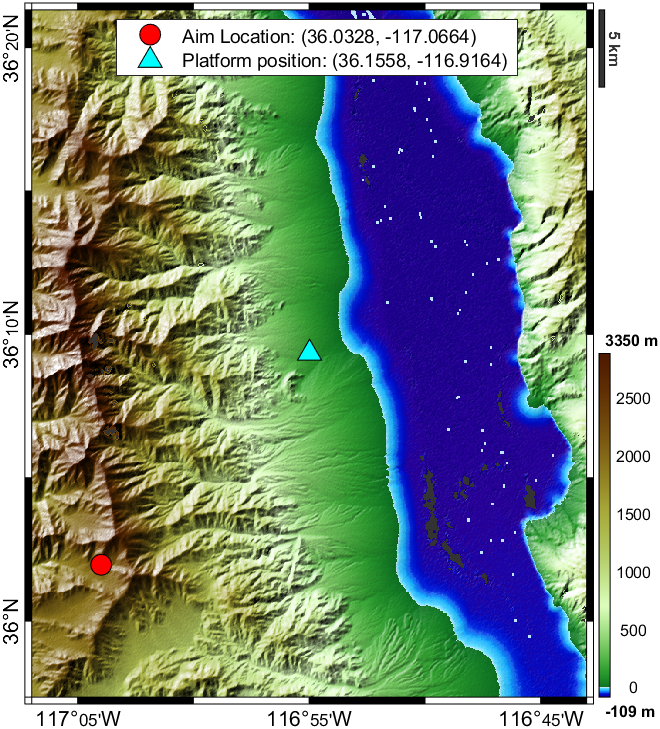}}} \\[-0.1cm] & \textbf{Platform Aim Location} (lat, lon): $(36.0328^{\circ}, -117.0664^{\circ})$ \\[0.2cm]
\cline{2-2} \\[-0.1cm] & 
\textbf{Platform Position} (lat, lon): $(36.1558^{\circ}, -116.9164^{\circ})$ \\[0.2cm]
\cline{2-2} \\[-0.1cm] & 
\textbf{Folder}: \textbf{/CLUTTER/num20\_lat36.1558\_lon-116.9164/} \\[0.2cm]
\cline{2-2} \\[-0.1cm] &
\parbox[t][3.0cm][t]{0.62\textwidth}{\textbf{Description}: Scenario $q_i = 60 \ (i = 20)$ in RASPNet is in California's Death Valley National Park. The radar platform is located at latitude $36.1558^{\circ}$ N and longitude $116.9164^{\circ}$ W. It aims towards a location at latitude $36.0328^{\circ}$ N and longitude $117.0664^{\circ}$ W. The radar is positioned above the valley floor, and the aim location is the summit of a tall mountain ridge.} \\
\hline
\end{tabular}
}
\end{table}

\paragraph{\underline{Scenario $q_i = 61$}: Guadalupe Mountains, TX ($i = 49$)} \vphantom{ }
\begin{table}[h!]
\centering
\setlength{\fboxsep}{0pt}
\setlength{\fboxrule}{2pt}
\fbox{%
\begin{tabular}{c|>{\arraybackslash}m{0.615\textwidth}}
\hline
\multirow{3}{*}{\raisebox{-\totalheight}{\includegraphics[width=0.5\textwidth,height=5.3cm,keepaspectratio]{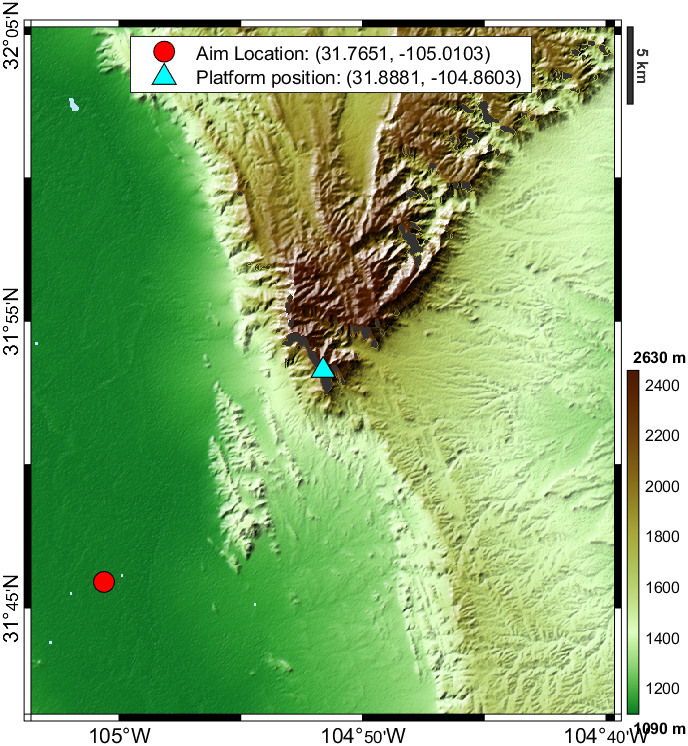}}} \\[-0.1cm] & \textbf{Platform Aim Location} (lat, lon): $(31.7651^{\circ}, -105.0103^{\circ})$ \\[0.2cm]
\cline{2-2} \\[-0.1cm] & 
\textbf{Platform Position} (lat, lon): $(31.8881^{\circ}, -104.8603^{\circ})$ \\[0.2cm]
\cline{2-2} \\[-0.1cm] & 
\textbf{Folder}: \textbf{/CLUTTER/num49\_lat31.8881\_lon-104.8603/} \\[0.2cm]
\cline{2-2} \\[-0.1cm] &
\parbox[t][3.0cm][t]{0.62\textwidth}{\textbf{Description}: Scenario $q_i = 61 \ (i = 49)$ in RASPNet is in the Guadalupe Mountains of Western Texas. The radar platform is located at latitude $31.8881^{\circ}$ N and longitude $104.8603^{\circ}$ W. It aims towards a location at latitude $31.7651^{\circ}$ N and longitude $105.0103^{\circ}$ W. The radar is positioned above a mountain summit, with the aim location far below within a flat basin.} \\
\hline
\end{tabular}
}
\end{table}

\paragraph{\underline{Scenario $q_i = 62$}: Misquah Hills, MN ($i = 65$)} \vphantom{ }
\begin{table}[h!]
\centering
\setlength{\fboxsep}{0pt}
\setlength{\fboxrule}{2pt}
\fbox{%
\begin{tabular}{c|>{\arraybackslash}m{0.615\textwidth}}
\hline
\multirow{3}{*}{\raisebox{-\totalheight}{\includegraphics[width=0.5\textwidth,height=5.3cm,keepaspectratio]{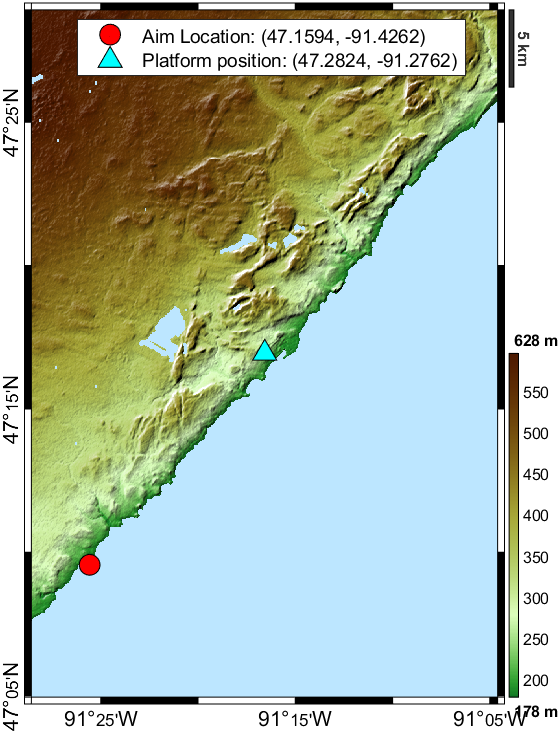}}} \\[-0.1cm] & \textbf{Platform Aim Location} (lat, lon): $(47.1594^{\circ}, -91.4262^{\circ})$ \\[0.2cm]
\cline{2-2} \\[-0.1cm] & 
\textbf{Platform Position} (lat, lon): $(47.2824^{\circ}, -91.2762^{\circ})$ \\[0.2cm]
\cline{2-2} \\[-0.1cm] & 
\textbf{Folder}: \textbf{/CLUTTER/num65\_lat47.2824\_lon-91.2762/} \\[0.2cm]
\cline{2-2} \\[-0.1cm] &
\parbox[t][3.0cm][t]{0.62\textwidth}{\textbf{Description}: Scenario $q_i = 62 \ (i = 65)$ in RASPNet is near the Misquah Hills of Upper Minnesota. The radar platform is located at latitude $47.2824^{\circ}$ N and longitude $91.2762^{\circ}$ W. It aims towards a location at latitude $47.1594^{\circ}$ N and longitude $91.2762^{\circ}$ W. The radar is positioned above a rugged coastline, and the aim location is along the Lake Superior Coast.} \\
\hline
\end{tabular}
}
\end{table}

\paragraph{\underline{Scenario $q_i = 63$}: Big Snowy Mountains, MT ($i = 42$)} \vphantom{ }
\begin{table}[h!]
\centering
\setlength{\fboxsep}{0pt}
\setlength{\fboxrule}{2pt}
\fbox{%
\begin{tabular}{c|>{\arraybackslash}m{0.615\textwidth}}
\hline
\multirow{3}{*}{\raisebox{-\totalheight}{\includegraphics[width=0.5\textwidth,height=5.3cm,keepaspectratio]{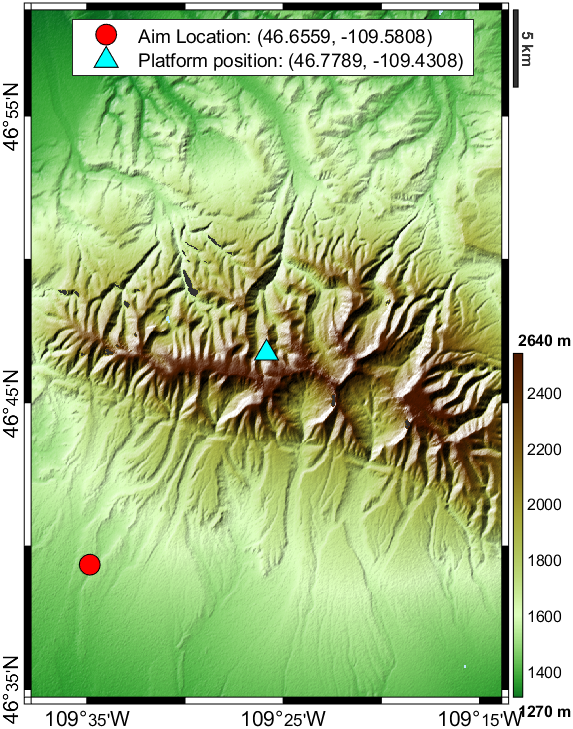}}} \\[-0.1cm] & \textbf{Platform Aim Location} (lat, lon): $(46.6559^{\circ}, -109.5808^{\circ})$ \\[0.2cm]
\cline{2-2} \\[-0.1cm] & 
\textbf{Platform Position} (lat, lon): $(46.7789^{\circ}, -109.4308^{\circ})$ \\[0.2cm]
\cline{2-2} \\[-0.1cm] & 
\textbf{Folder}: \textbf{/CLUTTER/num42\_lat46.7789\_lon-109.4308/} \\[0.2cm]
\cline{2-2} \\[-0.1cm] &
\parbox[t][3.0cm][t]{0.62\textwidth}{\textbf{Description}: Scenario $q_i = 63 \ (i = 42)$ in RASPNet is in the Big Snowy Mountains of Montana. The radar platform is positioned at latitude $46.7789^{\circ}$ N and longitude $109.4308^{\circ}$ W. It aims towards a location at latitude $46.6559^{\circ}$ N and longitude $109.5808^{\circ}$ W. The radar is positioned above a mountain summit, and the aim location has moderate terrain variability.} \\
\hline
\end{tabular}
}
\end{table}

\paragraph{\underline{Scenario $q_i = 64$}: Missouri River Country, MT ($i = 50$)} \vphantom{ }
\begin{table}[h!]
\centering
\setlength{\fboxsep}{0pt}
\setlength{\fboxrule}{2pt}
\fbox{%
\begin{tabular}{c|>{\arraybackslash}m{0.615\textwidth}}
\hline
\multirow{3}{*}{\raisebox{-\totalheight}{\includegraphics[width=0.5\textwidth,height=5.3cm,keepaspectratio]{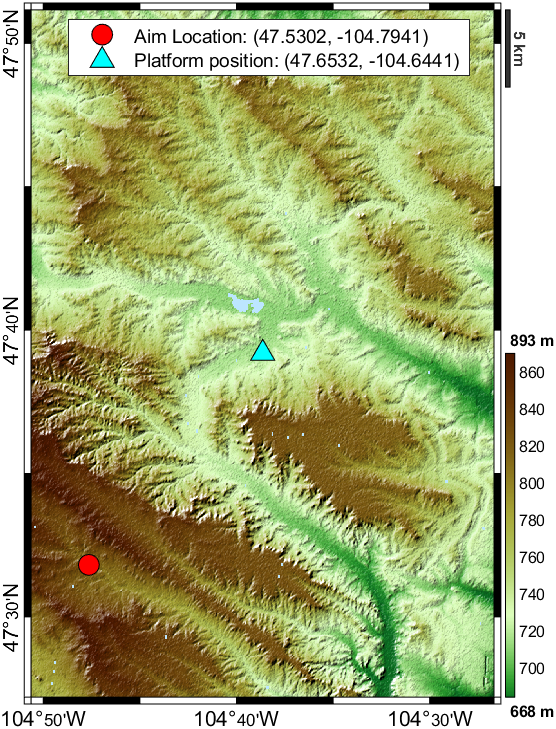}}} \\[-0.1cm] & \textbf{Platform Aim Location} (lat, lon): $(47.5302^{\circ}, -104.7941^{\circ})$ \\[0.2cm]
\cline{2-2} \\[-0.1cm] & 
\textbf{Platform Position} (lat, lon): $(47.6532^{\circ}, -104.6441^{\circ})$ \\[0.2cm]
\cline{2-2} \\[-0.1cm] & 
\textbf{Folder}: \textbf{/CLUTTER/num50\_lat47.6532\_lon-104.6441/} \\[0.2cm]
\cline{2-2} \\[-0.1cm] &
\parbox[t][3.0cm][t]{0.62\textwidth}{\textbf{Description}: Scenario $q_i = 64 \ (i = 50)$ in RASPNet is in the Missouri River Country of Montana. The radar platform is located at latitude $47.6532^{\circ}$ N and longitude $104.6441^{\circ}$ W. It aims towards a location at latitude $47.5302^{\circ}$ N and longitude $104.7941^{\circ}$ W. This region is characterized by a mixture of rugged terrain and nonuniform topography.} \\
\hline
\end{tabular}
}
\end{table}

\paragraph{\underline{Scenario $q_i = 65$}: Bear Lake, ID ($i = 37$)} \vphantom{ }
\begin{table}[h!]
\centering
\setlength{\fboxsep}{0pt}
\setlength{\fboxrule}{2pt}
\fbox{%
\begin{tabular}{c|>{\arraybackslash}m{0.615\textwidth}}
\hline
\multirow{3}{*}{\raisebox{-\totalheight}{\includegraphics[width=0.5\textwidth,height=5.3cm,keepaspectratio]{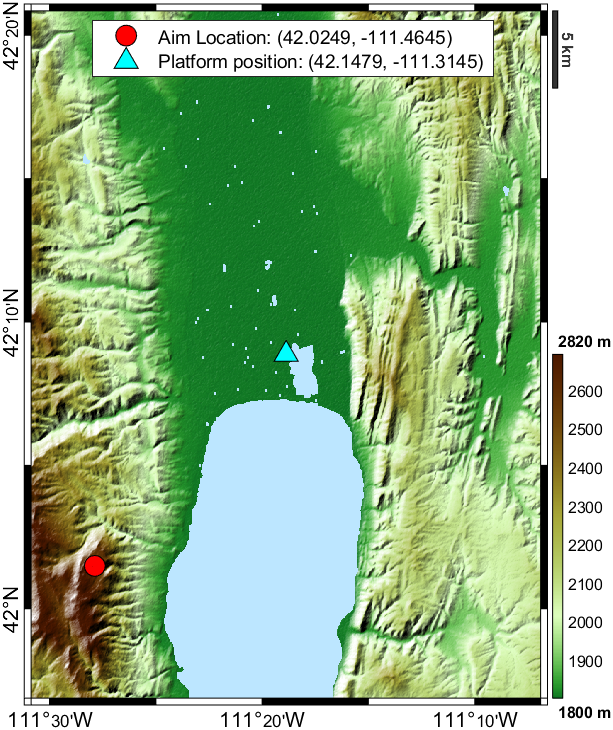}}} \\[-0.1cm] & \textbf{Platform Aim Location} (lat, lon): $(42.0249^{\circ}, -111.4645^{\circ})$ \\[0.2cm]
\cline{2-2} \\[-0.1cm] & 
\textbf{Platform Position} (lat, lon): $(42.1479^{\circ}, -111.3145^{\circ})$ \\[0.2cm]
\cline{2-2} \\[-0.1cm] & 
\textbf{Folder}: \textbf{/CLUTTER/num37\_lat42.1479\_lon-111.3145/} \\[0.2cm]
\cline{2-2} \\[-0.1cm] &
\parbox[t][3.0cm][t]{0.6\textwidth}{\textbf{Description}: Scenario $q_i = 65 \ (i = 37)$ in RASPNet is near Bear Lake in Southeastern Idaho. The radar platform is located at latitude $42.1479^{\circ}$ N and longitude $111.3145^{\circ}$ W. It aims towards a location at latitude $42.0249^{\circ}$ N and longitude $111.4645^{\circ}$ W. The radar is positioned above the coastline, and the aim location is the summit of a tall mountain ridge.} \\
\hline
\end{tabular}
}
\end{table}

\paragraph{\underline{Scenario $q_i = 66$}: Sheep Mountain, WY ($i = 47$)} \vphantom{ }
\begin{table}[h!]
\centering
\setlength{\fboxsep}{0pt}
\setlength{\fboxrule}{2pt}
\fbox{%
\begin{tabular}{c|>{\arraybackslash}m{0.615\textwidth}}
\hline
\multirow{3}{*}{\raisebox{-\totalheight}{\includegraphics[width=0.5\textwidth,height=5.3cm,keepaspectratio]{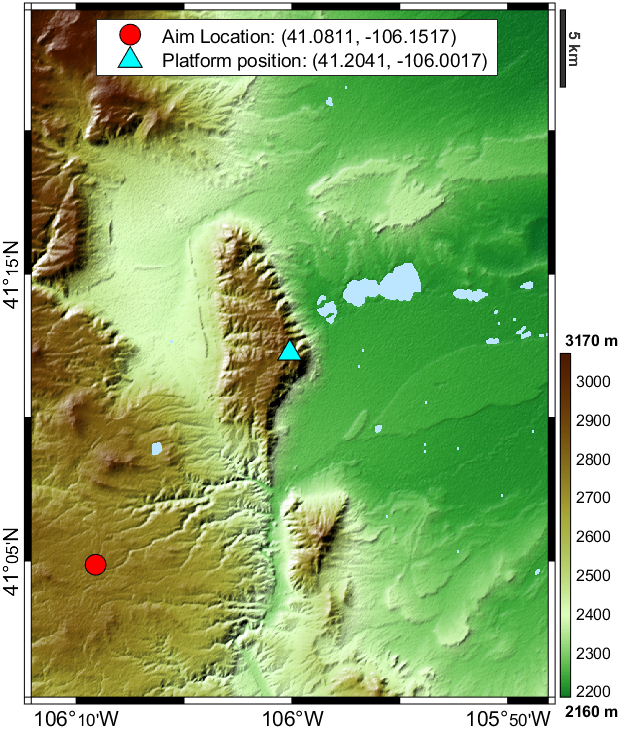}}} \\[-0.1cm] & \textbf{Platform Aim Location} (lat, lon): $(41.0811^{\circ}, -106.1517^{\circ})$ \\[0.2cm]
\cline{2-2} \\[-0.1cm] & 
\textbf{Platform Position} (lat, lon): $(41.2041^{\circ}, -106.0017^{\circ})$ \\[0.2cm]
\cline{2-2} \\[-0.1cm] & 
\textbf{Folder}: \textbf{/CLUTTER/num47\_lat41.2041\_lon-106.0017/} \\[0.2cm]
\cline{2-2} \\[-0.1cm] &
\parbox[t][3.0cm][t]{0.62\textwidth}{\textbf{Description}: Scenario $q_i = 66 \ (i = 47)$ in RASPNet is near Sheep Mountain in Southern Wyoming. The radar platform is positioned at latitude $41.2041^{\circ}$ N and longitude $106.0017^{\circ}$ W. It aims towards a location at latitude $41.0811^{\circ}$ N and longitude $106.1517^{\circ}$ W. The radar is positioned above a mountain summit, and the aim location has moderate terrain variability.} \\
\hline
\end{tabular}
}
\end{table}

\paragraph{\underline{Scenario $q_i = 67$}: Flume Canyon, UT ($i = 43$)} \vphantom{ }
\begin{table}[h!]
\centering
\setlength{\fboxsep}{0pt}
\setlength{\fboxrule}{2pt}
\fbox{%
\begin{tabular}{c|>{\arraybackslash}m{0.615\textwidth}}
\hline
\multirow{3}{*}{\raisebox{-\totalheight}{\includegraphics[width=0.5\textwidth,height=5.3cm,keepaspectratio]{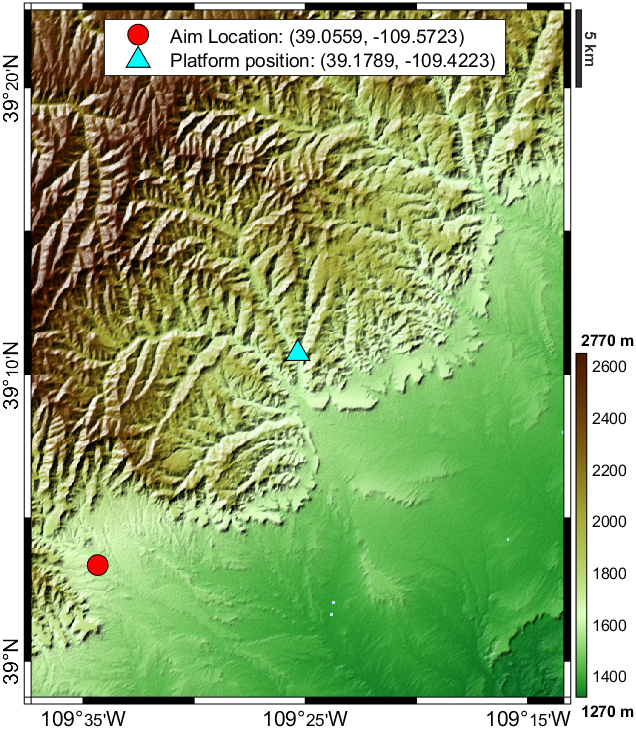}}} \\[-0.1cm] & \textbf{Platform Aim Location} (lat, lon): $(39.0559^{\circ}, -109.5723^{\circ})$ \\[0.2cm]
\cline{2-2} \\[-0.1cm] & 
\textbf{Platform Position} (lat, lon): $(39.1789^{\circ}, -109.4223^{\circ})$ \\[0.2cm]
\cline{2-2} \\[-0.1cm] & 
\textbf{Folder}: \textbf{/CLUTTER/num43\_lat39.1789\_lon-109.4223/} \\[0.2cm]
\cline{2-2} \\[-0.1cm] &
\parbox[t][3.0cm][t]{0.62\textwidth}{\textbf{Description}: Scenario $q_i = 67 \ (i = 43)$ in RASPNet is near Flume Canyon in Southeastern Utah. The radar platform is positioned at latitude $39.1789^{\circ}$ N and longitude $109.4223^{\circ}$ W. It aims towards a location at latitude $39.0559^{\circ}$ N and longitude $109.5723^{\circ}$ W. This region is characterized by rugged canyonlands, and the aim location has moderate terrain variability.} \\
\hline
\end{tabular}
}
\end{table}

\paragraph{\underline{Scenario $q_i = 68$}: Hells Canyon, OR ($i = 21$)} \vphantom{ }
\begin{table}[h!]
\centering
\setlength{\fboxsep}{0pt}
\setlength{\fboxrule}{2pt}
\fbox{%
\begin{tabular}{c|>{\arraybackslash}m{0.615\textwidth}}
\hline
\multirow{3}{*}{\raisebox{-\totalheight}{\includegraphics[width=0.5\textwidth,height=5.3cm,keepaspectratio]{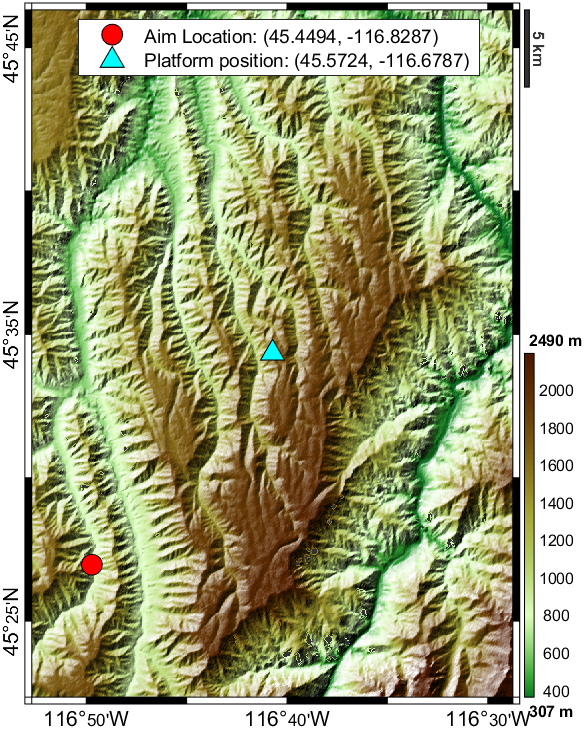}}} \\[-0.1cm] & \textbf{Platform Aim Location} (lat, lon): $(45.4494^{\circ}, -116.8287^{\circ})$ \\[0.2cm]
\cline{2-2} \\[-0.1cm] & 
\textbf{Platform Position} (lat, lon): $(45.5724^{\circ}, -116.6787^{\circ})$ \\[0.2cm]
\cline{2-2} \\[-0.1cm] & 
\textbf{Folder}: \textbf{/CLUTTER/num21\_lat45.5724\_lon-116.6787/} \\[0.2cm]
\cline{2-2} \\[-0.1cm] &
\parbox[t][3.0cm][t]{0.62\textwidth}{\textbf{Description}: Scenario $q_i = 68 \ (i = 21)$ in RASPNet is in Hells Canyon in Eastern Oregon. The radar platform is positioned at latitude $45.5724^{\circ}$ N and longitude $116.6787^{\circ}$ W. It aims towards a location at latitude $45.4494^{\circ}$ N and longitude $116.8287^{\circ}$ W. This region is characterized by rugged canyonlands, and the aim location has high terrain variability.} \\
\hline
\end{tabular}
}
\end{table}

\paragraph{\underline{Scenario $q_i = 69$}: Cape Flattery, WA ($i = 1$)} \vphantom{ }
\begin{table}[h!]
\centering
\setlength{\fboxsep}{0pt}
\setlength{\fboxrule}{2pt}
\fbox{%
\begin{tabular}{c|>{\arraybackslash}m{0.615\textwidth}}
\hline
\multirow{3}{*}{\raisebox{-\totalheight}{\includegraphics[width=0.5\textwidth,height=5.3cm,keepaspectratio]{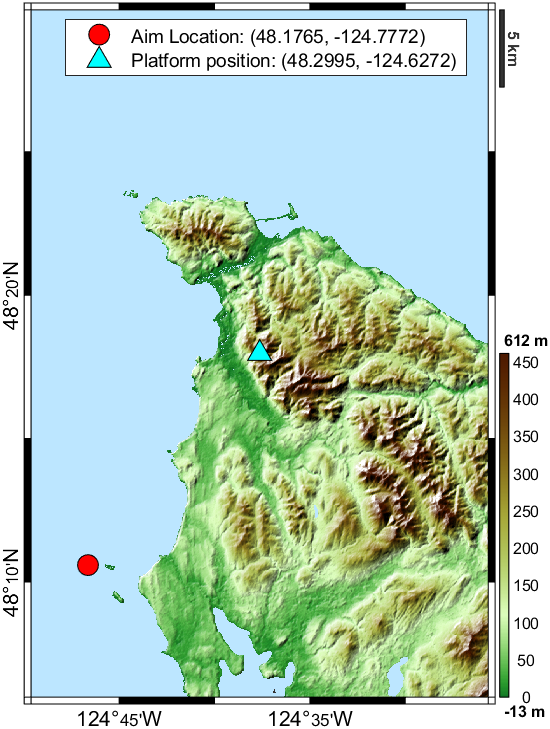}}} \\[-0.1cm] & \textbf{Platform Aim Location} (lat, lon): $(48.1765^{\circ}, -124.7772^{\circ})$ \\[0.2cm]
\cline{2-2} \\[-0.1cm] & 
\textbf{Platform Position} (lat, lon): $(48.2995^{\circ}, -124.6272^{\circ})$ \\[0.2cm]
\cline{2-2} \\[-0.1cm] & 
\textbf{Folder}: \textbf{/CLUTTER/num1\_lat48.2995\_lon-124.6272/} \\[0.2cm]
\cline{2-2} \\[-0.1cm] &
\parbox[t][3.0cm][t]{0.62\textwidth}{\textbf{Description}: Scenario $q_i = 69 \ (i = 1)$ in RASPNet is near Cape Flattery in Western Washington. The radar platform is positioned at latitude $48.2995^{\circ}$ N and longitude $124.6272^{\circ}$ W. It aims towards a location at latitude $48.1765^{\circ}$ N and longitude $124.7772^{\circ}$ W. This is a land-to-water transition region, where the aim location is near the Olympic Peninsula coastline.} \\
\hline
\end{tabular}
}
\end{table}

\paragraph{\underline{Scenario $q_i = 70$}: Three Sisters, OR ($i = 6$)} \vphantom{ }
\begin{table}[h!]
\centering
\setlength{\fboxsep}{0pt}
\setlength{\fboxrule}{2pt}
\fbox{%
\begin{tabular}{c|>{\arraybackslash}m{0.615\textwidth}}
\hline
\multirow{3}{*}{\raisebox{-\totalheight}{\includegraphics[width=0.5\textwidth,height=5.3cm,keepaspectratio]{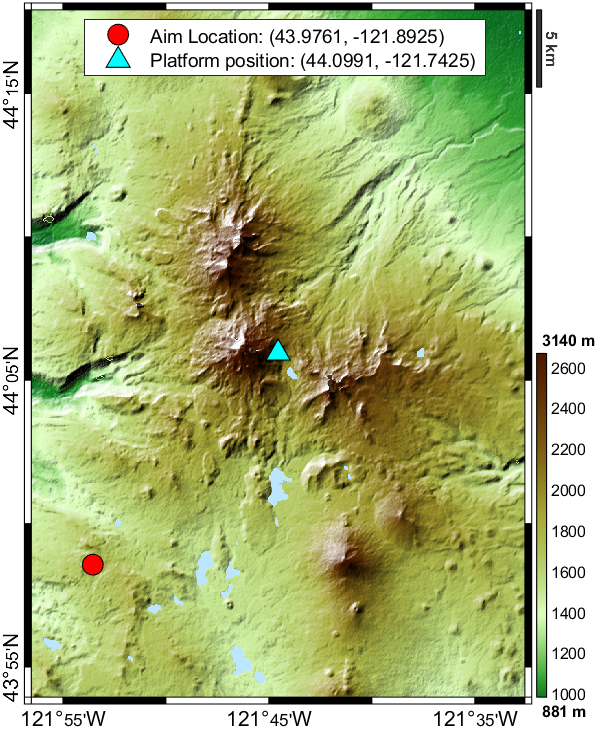}}} \\[-0.1cm] & \textbf{Platform Aim Location} (lat, lon): $(43.9761^{\circ}, -121.8925^{\circ})$ \\[0.2cm]
\cline{2-2} \\[-0.1cm] & 
\textbf{Platform Position} (lat, lon): $(44.0991^{\circ}, -121.7425^{\circ})$ \\[0.2cm]
\cline{2-2} \\[-0.1cm] & 
\textbf{Folder}: \textbf{/CLUTTER/num6\_lat44.0991\_lon-121.7425/} \\[0.2cm]
\cline{2-2} \\[-0.1cm] &
\parbox[t][3.0cm][t]{0.62\textwidth}{\textbf{Description}: Scenario $q_i = 70 \ (i = 6)$ in RASPNet is near the Three Sisters peaks of Central Oregon. The radar platform is positioned at latitude $44.0991^{\circ}$ N and longitude $121.7425^{\circ}$ W. It aims towards a location at latitude $43.9761^{\circ}$ N and longitude $121.8925^{\circ}$ W. The radar is positioned above a mountain ridge, and the aim location has high terrain variability.} \\
\hline
\end{tabular}
}
\end{table}

\paragraph{\underline{Scenario $q_i = 71$}: Black Hills, SD ($i = 52$)} \vphantom{ }
\begin{table}[h!]
\centering
\setlength{\fboxsep}{0pt}
\setlength{\fboxrule}{2pt}
\fbox{%
\begin{tabular}{c|>{\arraybackslash}m{0.615\textwidth}}
\hline
\multirow{3}{*}{\raisebox{-\totalheight}{\includegraphics[width=0.5\textwidth,height=5.3cm,keepaspectratio]{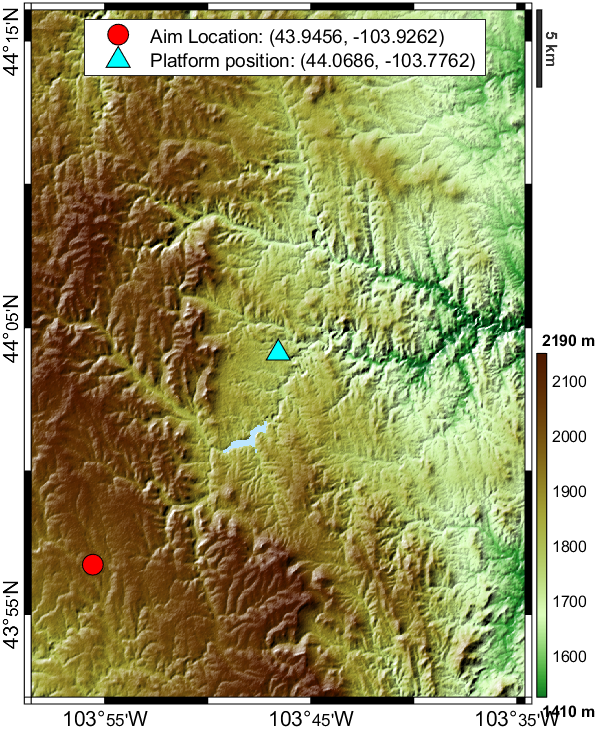}}} \\[-0.1cm] & \textbf{Platform Aim Location} (lat, lon): $(43.9456^{\circ}, -103.9262^{\circ})$ \\[0.2cm]
\cline{2-2} \\[-0.1cm] & 
\textbf{Platform Position} (lat, lon): $(44.0686^{\circ}, -103.7762^{\circ})$ \\[0.2cm]
\cline{2-2} \\[-0.1cm] & 
\textbf{Folder}: \textbf{/CLUTTER/num52\_lat44.0686\_lon-103.7762/} \\[0.2cm]
\cline{2-2} \\[-0.1cm] &
\parbox[t][3.0cm][t]{0.62\textwidth}{\textbf{Description}: Scenario $q_i = 71 \ (i = 52)$ in RASPNet is in the Black Hills region of South Dakota. The radar platform is positioned at latitude $44.0686^{\circ}$ N and longitude $103.7762^{\circ}$ W. It aims towards a location at latitude $43.9456^{\circ}$ N and longitude $103.9262^{\circ}$ W. The radar is positioned above rugged mountains, and the aim location has high terrain variability.} \\
\hline
\end{tabular}
}
\end{table}

\paragraph{\underline{Scenario $q_i = 72$}: Palouse Hills, WA ($i = 18$)} \vphantom{ }
\begin{table}[h!]
\centering
\setlength{\fboxsep}{0pt}
\setlength{\fboxrule}{2pt}
\fbox{%
\begin{tabular}{c|>{\arraybackslash}m{0.615\textwidth}}
\hline
\multirow{3}{*}{\raisebox{-\totalheight}{\includegraphics[width=0.5\textwidth,height=5.3cm,keepaspectratio]{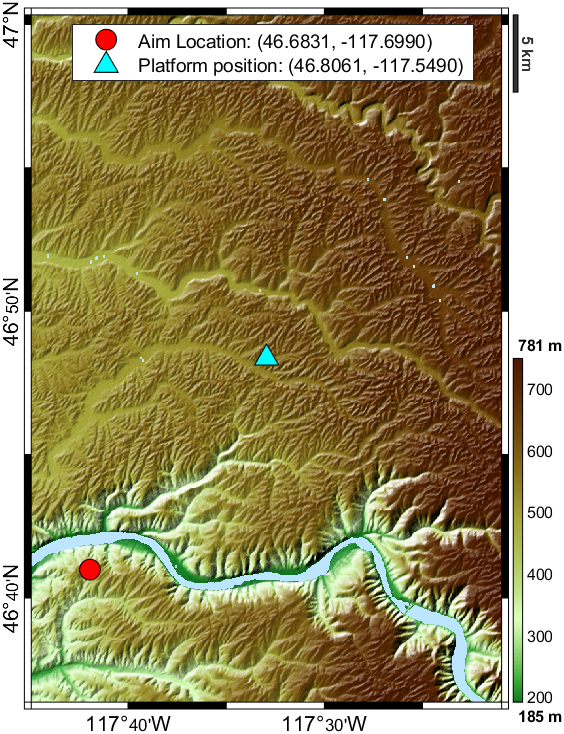}}} \\[-0.1cm] & \textbf{Platform Aim Location} (lat, lon): $(46.6831^{\circ}, -117.6990^{\circ})$ \\[0.2cm]
\cline{2-2} \\[-0.1cm] & 
\textbf{Platform Position} (lat, lon): $(46.8061^{\circ}, -117.5490^{\circ})$ \\[0.2cm]
\cline{2-2} \\[-0.1cm] & 
\textbf{Folder}: \textbf{/CLUTTER/num18\_lat46.8061\_lon-117.549/} \\[0.2cm]
\cline{2-2} \\[-0.1cm] &
\parbox[t][3.0cm][t]{0.62\textwidth}{\textbf{Description}: Scenario $q_i = 72 \ (i = 18)$ in RASPNet is in the Palouse Hills of Eastern Washington. The radar platform is positioned at latitude $46.8061^{\circ}$ N and longitude $117.5490^{\circ}$ W. It aims towards a location at latitude $46.6831^{\circ}$ N and longitude $117.6990^{\circ}$ W. The radar is positioned above rolling hills, and the aim location has high terrain variability.} \\
\hline
\end{tabular}
}
\end{table}

\paragraph{\underline{Scenario $q_i = 73$}: Appalachian Mountains, PA ($i = 87$)} \vphantom{ }
\begin{table}[h!]
\centering
\setlength{\fboxsep}{0pt}
\setlength{\fboxrule}{2pt}
\fbox{%
\begin{tabular}{c|>{\arraybackslash}m{0.615\textwidth}}
\hline
\multirow{3}{*}{\raisebox{-\totalheight}{\includegraphics[width=0.5\textwidth,height=5.3cm,keepaspectratio]{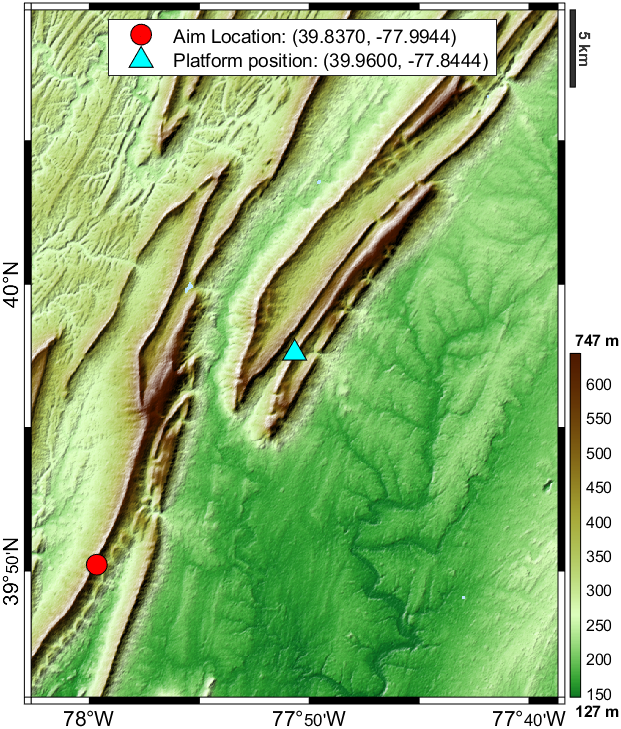}}} \\[-0.1cm] & \textbf{Platform Aim Location} (lat, lon): $(39.8370^{\circ}, -77.9944^{\circ})$ \\[0.2cm]
\cline{2-2} \\[-0.1cm] & 
\textbf{Platform Position} (lat, lon): $(39.9600^{\circ}, -77.8444^{\circ})$ \\[0.2cm]
\cline{2-2} \\[-0.1cm] & 
\textbf{Folder}: \textbf{/CLUTTER/num87\_lat39.96\_lon-77.8444/} \\[0.2cm]
\cline{2-2} \\[-0.1cm] &
\parbox[t][3.0cm][t]{0.62\textwidth}{\textbf{Description}: Scenario $q_i = 73 \ (i = 87)$ in RASPNet is in the Appalachian Mountains in Pennsylvania. The radar platform is positioned at latitude $39.9600^{\circ}$ N and longitude $77.8444^{\circ}$ W. It aims towards a location at latitude $39.8370^{\circ}$ N and longitude $77.9944^{\circ}$ W. The radar is positioned above a mountain ridge, and the aim location has high terrain variability.} \\
\hline
\end{tabular}
}
\end{table}

\paragraph{\underline{Scenario $q_i = 74$}: Pikes Peak, CO ($i = 48$)} \vphantom{ }
\begin{table}[h!]
\centering
\setlength{\fboxsep}{0pt}
\setlength{\fboxrule}{2pt}
\fbox{%
\begin{tabular}{c|>{\arraybackslash}m{0.615\textwidth}}
\hline
\multirow{3}{*}{\raisebox{-\totalheight}{\includegraphics[width=0.5\textwidth,height=5.3cm,keepaspectratio]{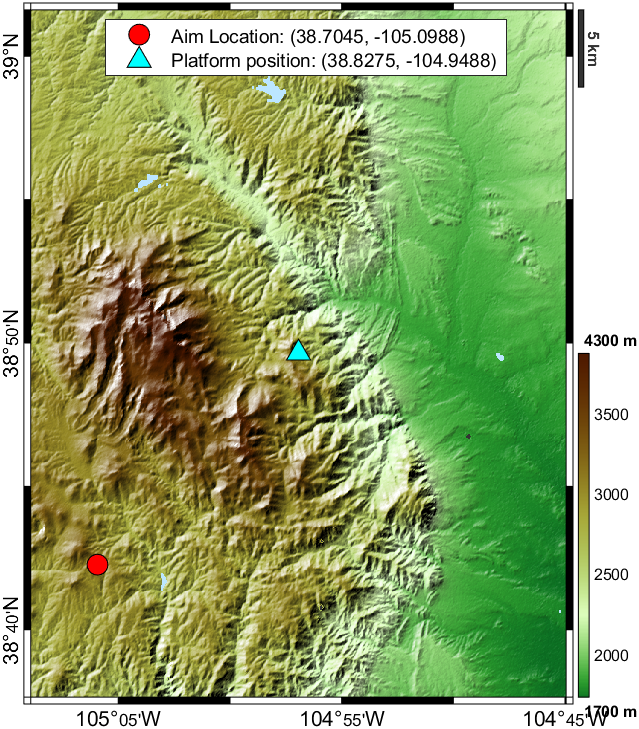}}} \\[-0.1cm] & \textbf{Platform Aim Location} (lat, lon): $(38.7045^{\circ}, -105.0988^{\circ})$ \\[0.2cm]
\cline{2-2} \\[-0.1cm] & 
\textbf{Platform Position} (lat, lon): $(38.8275^{\circ}, -104.9488^{\circ})$ \\[0.2cm]
\cline{2-2} \\[-0.1cm] & 
\textbf{Folder}: \textbf{/CLUTTER/num48\_lat38.8275\_lon-104.9488/} \\[0.2cm]
\cline{2-2} \\[-0.1cm] &
\parbox[t][3.0cm][t]{0.62\textwidth}{\textbf{Description}: Scenario $q_i = 74 \ (i = 48)$ in RASPNet is near Pikes Peak in the Colorado Front Range. The radar platform is positioned at latitude $38.8275^{\circ}$ N and longitude $104.9488^{\circ}$ W. It aims towards a location at latitude $38.7045^{\circ}$ N and longitude $105.0988^{\circ}$ W. The radar is positioned behind a mountain ridge, and the aim location has high terrain variability.} \\
\hline
\end{tabular}
}
\end{table}

\paragraph{\underline{Scenario $q_i = 75$}: Mount Greylock, MA ($i = 94$)} \vphantom{ }
\begin{table}[h!]
\centering
\setlength{\fboxsep}{0pt}
\setlength{\fboxrule}{2pt}
\fbox{%
\begin{tabular}{c|>{\arraybackslash}m{0.615\textwidth}}
\hline
\multirow{3}{*}{\raisebox{-\totalheight}{\includegraphics[width=0.5\textwidth,height=5.3cm,keepaspectratio]{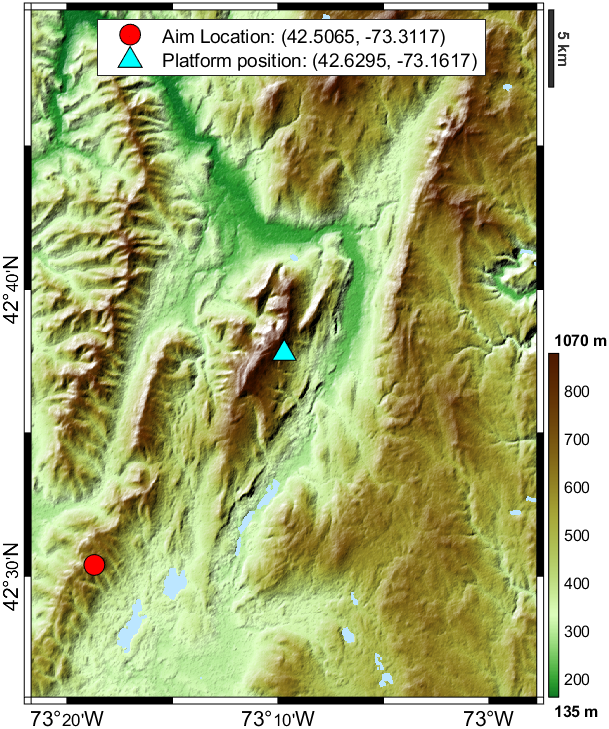}}} \\[-0.1cm] & \textbf{Platform Aim Location} (lat, lon): $(42.5065^{\circ}, -73.3117^{\circ})$ \\[0.2cm]
\cline{2-2} \\[-0.1cm] & 
\textbf{Platform Position} (lat, lon): $(42.6295^{\circ}, -73.1617^{\circ})$ \\[0.2cm]
\cline{2-2} \\[-0.1cm] & 
\textbf{Folder}: \textbf{/CLUTTER/num94\_lat42.6295\_lon-73.1617/} \\[0.2cm]
\cline{2-2} \\[-0.1cm] &
\parbox[t][3.0cm][t]{0.62\textwidth}{\textbf{Description}: Scenario $q_i = 75 \ (i = 94)$ in RASPNet is near Mount Greylock in Western Massachusetts. The radar platform is positioned at latitude $42.6295^{\circ}$ N and longitude $73.1617^{\circ}$ W. It aims towards a location at latitude $42.5065^{\circ}$ N and longitude $73.3117^{\circ}$ W. The radar is positioned above a mountain summit, and the aim location has high terrain variability.} \\
\hline
\end{tabular}
}
\end{table}

\subsection{Eye-watering Scenarios}
Lastly, we outline the Eye-watering RASPNet scenarios, where $q_i \in \{76,77,\ldots,100\}$.

\paragraph{\underline{Scenario $q_i = 76$}: Glacier National Park, MT ($i = 31$)} \vphantom{ }
\begin{table}[h!]
\centering
\setlength{\fboxsep}{0pt}
\setlength{\fboxrule}{2pt}
\fbox{%
\begin{tabular}{c|>{\arraybackslash}m{0.615\textwidth}}
\hline
\multirow{3}{*}{\raisebox{-\totalheight}{\includegraphics[width=0.5\textwidth,height=5.1cm,keepaspectratio]{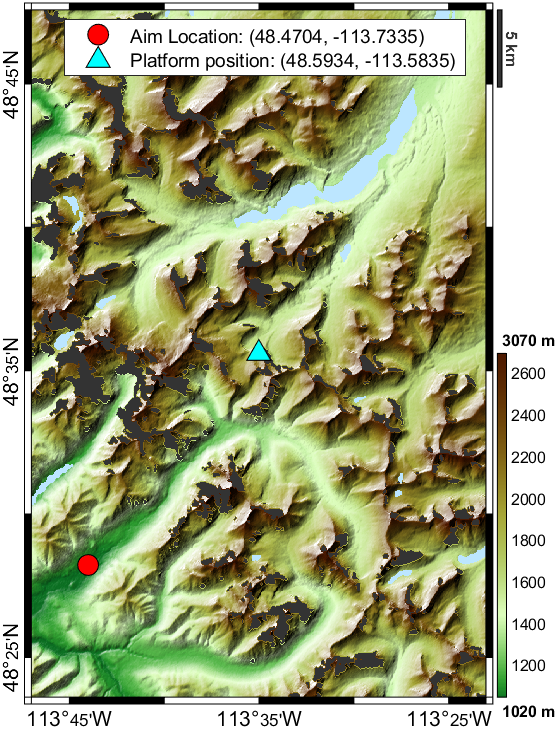}}} \\[-0.1cm] & \textbf{Platform Aim Location} (lat, lon): $(48.4704^{\circ}, -113.7335^{\circ})$ \\[0.2cm]
\cline{2-2} \\[-0.1cm] & 
\textbf{Platform Position} (lat, lon): $(48.5934^{\circ}, -113.5835^{\circ})$ \\[0.2cm]
\cline{2-2} \\[-0.1cm] & 
\textbf{Folder}: \textbf{/CLUTTER/num31\_lat48.5934\_lon-113.5835/} \\[0.2cm]
\cline{2-2} \\[-0.1cm] &
\parbox[t][2.6cm][t]{0.62\textwidth}{\textbf{Description}: Scenario $q_i = 76 \ (i = 31)$ in RASPNet is in Glacier National Park in Western Montana. The radar platform is located at latitude $48.5934^{\circ}$ N and longitude $113.5835^{\circ}$ W. It aims towards a location at latitude $48.4704^{\circ}$ N and longitude $113.7335^{\circ}$ W. The radar is located above a mountain ridge, and the aim location is a canyon with high terrain variability.} \\
\hline
\end{tabular}
}
\end{table}

\paragraph{\underline{Scenario $q_i = 77$}: Allegheny Plateau, WV ($i = 82$)} \vphantom{ }
\begin{table}[h!]
\centering
\setlength{\fboxsep}{0pt}
\setlength{\fboxrule}{2pt}
\fbox{%
\begin{tabular}{c|>{\arraybackslash}m{0.615\textwidth}}
\hline
\multirow{3}{*}{\raisebox{-\totalheight}{\includegraphics[width=0.5\textwidth,height=5.1cm,keepaspectratio]{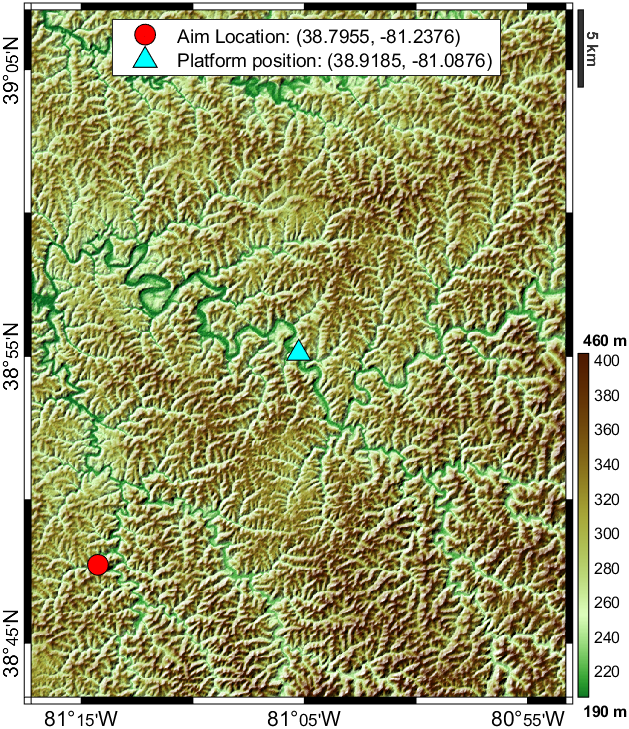}}} \\[-0.1cm] & \textbf{Platform Aim Location} (lat, lon): $(38.7955^{\circ}, -81.2376^{\circ})$ \\[0.2cm]
\cline{2-2} \\[-0.1cm] & 
\textbf{Platform Position} (lat, lon): $(38.9185^{\circ}, -81.0876^{\circ})$ \\[0.2cm]
\cline{2-2} \\[-0.1cm] & 
\textbf{Folder}: \textbf{/CLUTTER/num82\_lat38.9185\_lon-81.0876/} \\[0.2cm]
\cline{2-2} \\[-0.1cm] &
\parbox[t][2.6cm][t]{0.62\textwidth}{\textbf{Description}: Scenario $q_i = 77 \ (i = 82)$ in RASPNet is in the Allegheny Plateau region of West Virginia. The radar platform is located at latitude $38.9185^{\circ}$ N and longitude $81.0876^{\circ}$ W. It aims towards a location at latitude $38.7955^{\circ}$ N and longitude $81.2376^{\circ}$ W. This region is characterized by a mixture of rugged terrain and nonuniform topography.} \\
\hline
\end{tabular}
}
\end{table}

\paragraph{\underline{Scenario $q_i = 78$}: Cumberland Plateau, KY ($i = 80$)} \vphantom{ }
\begin{table}[h!]
\centering
\setlength{\fboxsep}{0pt}
\setlength{\fboxrule}{2pt}
\fbox{%
\begin{tabular}{c|>{\arraybackslash}m{0.615\textwidth}}
\hline
\multirow{3}{*}{\raisebox{-\totalheight}{\includegraphics[width=0.5\textwidth,height=5.1cm,keepaspectratio]{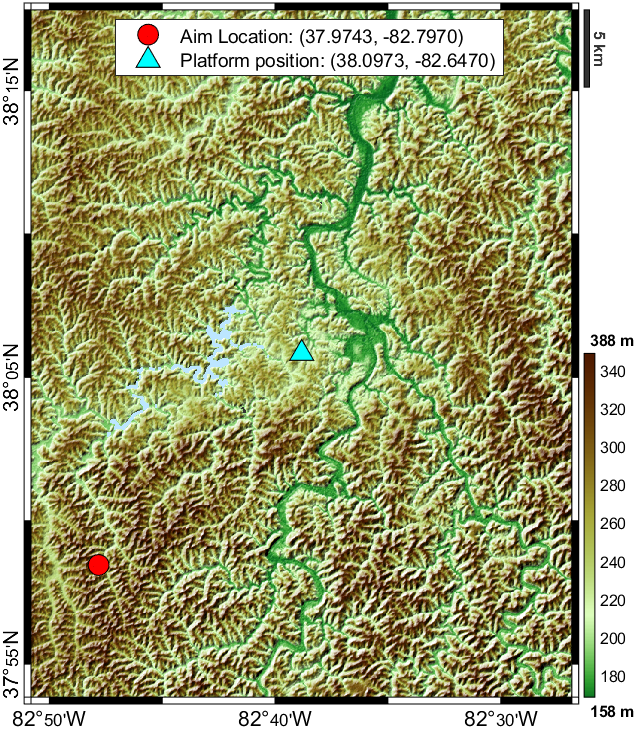}}} \\[-0.1cm] & \textbf{Platform Aim Location} (lat, lon): $(37.9743^{\circ}, -82.7970^{\circ})$ \\[0.2cm]
\cline{2-2} \\[-0.1cm] & 
\textbf{Platform Position} (lat, lon): $(38.0973^{\circ}, -82.6470^{\circ})$ \\[0.2cm]
\cline{2-2} \\[-0.1cm] & 
\textbf{Folder}: \textbf{/CLUTTER/num80\_lat38.0973\_lon-82.647/} \\[0.2cm]
\cline{2-2} \\[-0.1cm] &
\parbox[t][2.6cm][t]{0.62\textwidth}{\textbf{Description}: Scenario $q_i = 78 \ (i = 80)$ in RASPNet is in the Cumberland Plateau region of Kentucky. The radar platform is located at latitude $38.0973^{\circ}$ N and longitude $82.6470^{\circ}$ W. It aims towards a location at latitude $37.9743^{\circ}$ N and longitude $82.7970^{\circ}$ W. This region is characterized by a mixture of rugged terrain and nonuniform topography.} \\
\hline
\end{tabular}
}
\end{table}

\paragraph{\underline{Scenario $q_i = 79$}: Grand Teton National Park, WY ($i = 39$)} \vphantom{ }
\begin{table}[h!]
\centering
\setlength{\fboxsep}{0pt}
\setlength{\fboxrule}{2pt}
\fbox{%
\begin{tabular}{c|>{\arraybackslash}m{0.615\textwidth}}
\hline
\multirow{3}{*}{\raisebox{-\totalheight}{\includegraphics[width=0.5\textwidth,height=5.3cm,keepaspectratio]{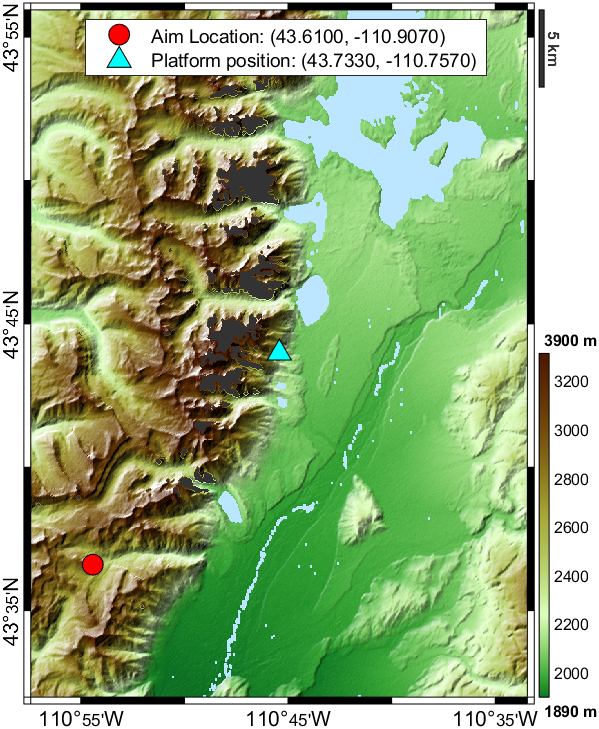}}} \\[-0.1cm] & \textbf{Platform Aim Location} (lat, lon): $(43.6100^{\circ}, -110.9070^{\circ})$ \\[0.2cm]
\cline{2-2} \\[-0.1cm] & 
\textbf{Platform Position} (lat, lon): $(43.7330^{\circ}, -110.7570^{\circ})$ \\[0.2cm]
\cline{2-2} \\[-0.1cm] & 
\textbf{Folder}: \textbf{/CLUTTER/num39\_lat43.733\_lon-110.757/} \\[0.2cm]
\cline{2-2} \\[-0.1cm] &
\parbox[t][3.0cm][t]{0.62\textwidth}{\textbf{Description}: Scenario $q_i = 79 \ (i = 39)$ in RASPNet is in Grand Teton National Park in Wyoming. The radar platform is located at latitude $43.7330^{\circ}$ N and longitude $110.7570^{\circ}$ W. It aims towards a location at latitude $43.6100^{\circ}$ N and longitude $110.9070^{\circ}$ W. The radar is positioned at the base of a mountain, and the aim location has high terrain variability.} \\
\hline
\end{tabular}
}
\end{table}

\paragraph{\underline{Scenario $q_i = 80$}: Blue Ridge Mountains, NC ($i = 79$)} \vphantom{ }
\begin{table}[h!]
\centering
\setlength{\fboxsep}{0pt}
\setlength{\fboxrule}{2pt}
\fbox{%
\begin{tabular}{c|>{\arraybackslash}m{0.615\textwidth}}
\hline
\multirow{3}{*}{\raisebox{-\totalheight}{\includegraphics[width=0.5\textwidth,height=5.3cm,keepaspectratio]{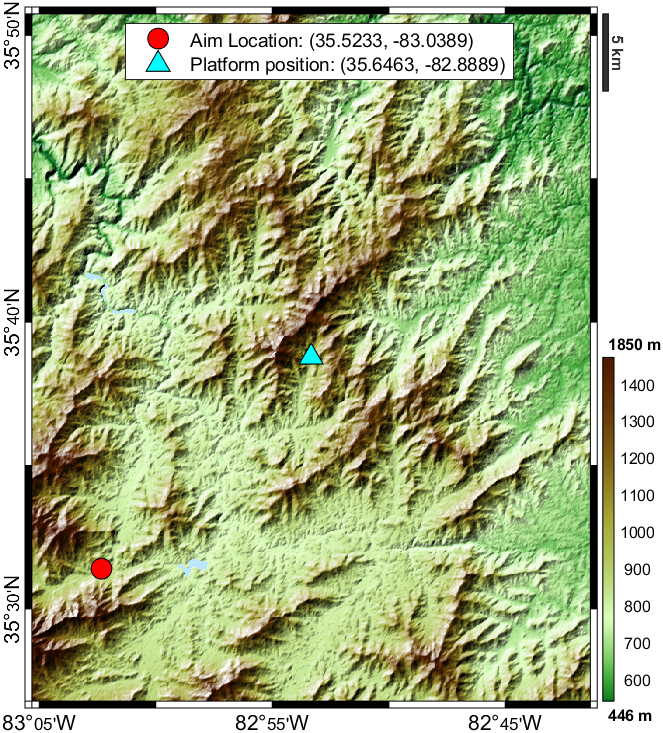}}} \\[-0.1cm] & \textbf{Platform Aim Location} (lat, lon): $(35.5233^{\circ}, -83.0389^{\circ})$ \\[0.2cm]
\cline{2-2} \\[-0.1cm] & 
\textbf{Platform Position} (lat, lon): $(35.6463^{\circ}, -82.8889^{\circ})$ \\[0.2cm]
\cline{2-2} \\[-0.1cm] & 
\textbf{Folder}: \textbf{/CLUTTER/num79\_lat35.6463\_lon-82.8889/} \\[0.2cm]
\cline{2-2} \\[-0.1cm] &
\parbox[t][3.0cm][t]{0.62\textwidth}{\textbf{Description}: Scenario $q_i = 80 \ (i = 79)$ in RASPNet is in North Carolina's Blue Ridge Mountains. The radar platform is located at latitude $35.6463^{\circ}$ N and longitude $82.8889^{\circ}$ W. It aims towards a location at latitude $35.5233^{\circ}$ N and longitude $83.0389^{\circ}$ W. The radar is positioned near a mountain ridge, and the aim location has high terrain variability.} \\
\hline
\end{tabular}
}
\end{table}

\paragraph{\underline{Scenario $q_i = 81$}: Capitol Reef National Park, UT ($i = 38$)} \vphantom{ }
\begin{table}[h!]
\centering
\setlength{\fboxsep}{0pt}
\setlength{\fboxrule}{2pt}
\fbox{%
\begin{tabular}{c|>{\arraybackslash}m{0.615\textwidth}}
\hline
\multirow{3}{*}{\raisebox{-\totalheight}{\includegraphics[width=0.5\textwidth,height=5.3cm,keepaspectratio]{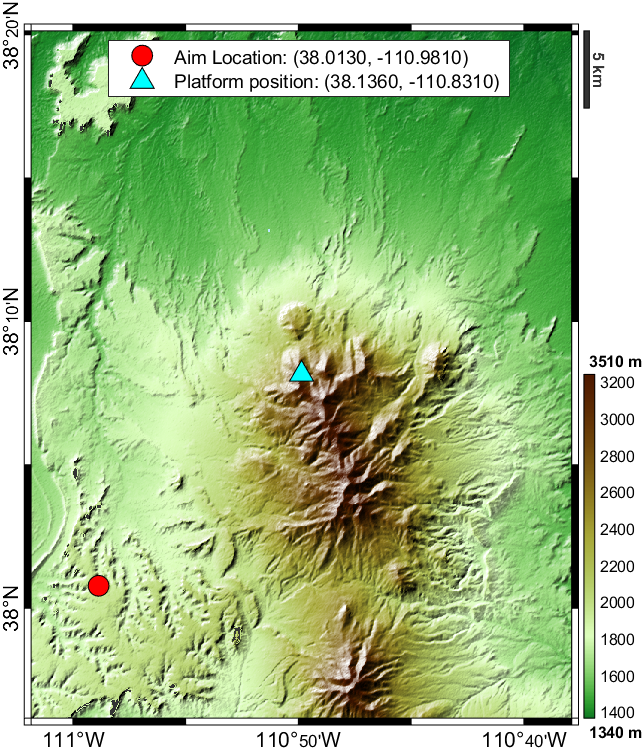}}} \\[-0.1cm] & \textbf{Platform Aim Location} (lat, lon): $(38.0130^{\circ}, -110.9810^{\circ})$ \\[0.2cm]
\cline{2-2} \\[-0.1cm] & 
\textbf{Platform Position} (lat, lon): $(38.1360^{\circ}, -110.8310^{\circ})$ \\[0.2cm]
\cline{2-2} \\[-0.1cm] & 
\textbf{Folder}: \textbf{/CLUTTER/num38\_lat38.136\_lon-110.831/} \\[0.2cm]
\cline{2-2} \\[-0.1cm] &
\parbox[t][3.0cm][t]{0.62\textwidth}{\textbf{Description}: Scenario $q_i = 81 \ (i = 38)$ in RASPNet is in Capitol Reef National Park in Central Utah. The radar platform is located at latitude $38.1360^{\circ}$ N and longitude $110.8310^{\circ}$ W. It aims towards a location at latitude $38.0130^{\circ}$ N and longitude $110.9810^{\circ}$ W. The radar is positioned near a mountain summit, and the aim location has high terrain variability.} \\
\hline
\end{tabular}
}
\end{table}

\paragraph{\underline{Scenario $q_i = 82$}: Sierra Nevada Range, CA ($i = 17$)} \vphantom{ }
\begin{table}[h!]
\centering
\setlength{\fboxsep}{0pt}
\setlength{\fboxrule}{2pt}
\fbox{%
\begin{tabular}{c|>{\arraybackslash}m{0.615\textwidth}}
\hline
\multirow{3}{*}{\raisebox{-\totalheight}{\includegraphics[width=0.5\textwidth,height=5.6cm,keepaspectratio]{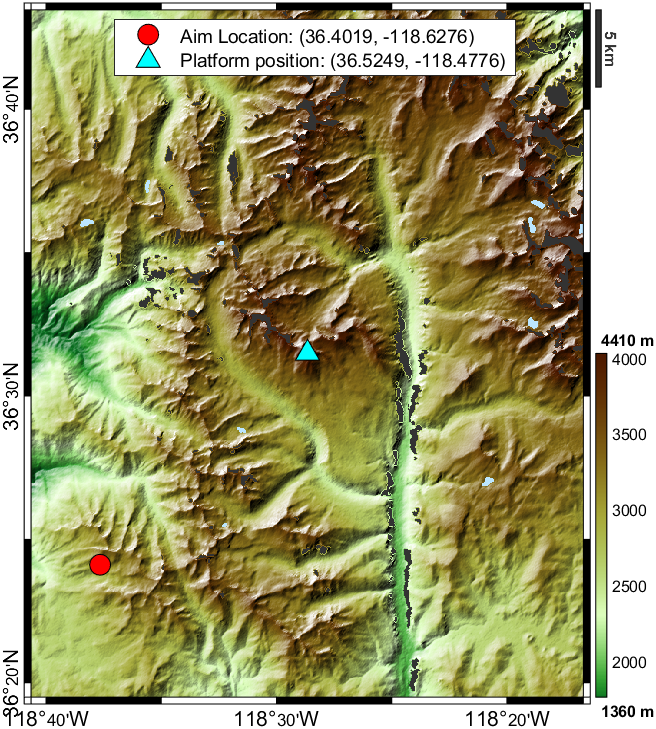}}} \\[-0.1cm] & \textbf{Platform Aim Location} (lat, lon): $(36.4019^{\circ}, -118.6276^{\circ})$ \\[0.2cm]
\cline{2-2} \\[-0.1cm] & 
\textbf{Platform Position} (lat, lon): $(36.5249^{\circ}, -118.4776^{\circ})$ \\[0.2cm]
\cline{2-2} \\[-0.1cm] & 
\textbf{Folder}: \textbf{/CLUTTER/num17\_lat36.5249\_lon-118.4776/} \\[0.2cm]
\cline{2-2} \\[-0.1cm] &
\parbox[t][3.0cm][t]{0.62\textwidth}{\textbf{Description}: Scenario $q_i = 82 \ (i = 17)$ in RASPNet is in the Sierra Nevada Range in California. The radar platform is located at latitude $36.5249^{\circ}$ N and longitude $118.4776^{\circ}$ W. It aims towards a location at latitude $36.4019^{\circ}$ N and longitude $118.6276^{\circ}$ W. The radar is positioned above a mountain summit, and the aim location has high terrain variability.} \\
\hline
\end{tabular}
}
\end{table}

\paragraph{\underline{Scenario $q_i = 83$}: Big Bend National Park, TX ($i = 54$)} \vphantom{ }
\begin{table}[h!]
\centering
\setlength{\fboxsep}{0pt}
\setlength{\fboxrule}{2pt}
\fbox{%
\begin{tabular}{c|>{\arraybackslash}m{0.615\textwidth}}
\hline
\multirow{3}{*}{\raisebox{-\totalheight}{\includegraphics[width=0.5\textwidth,height=5.3cm,keepaspectratio]{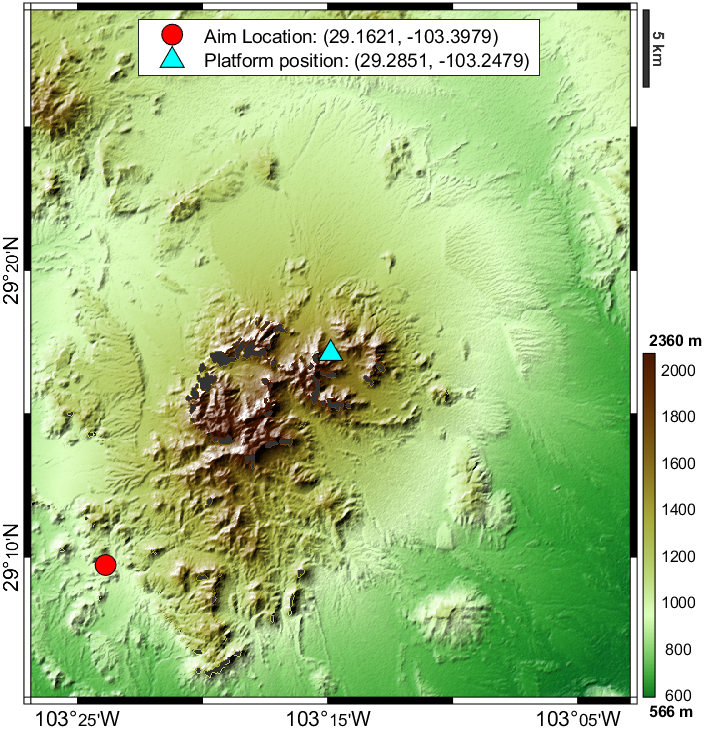}}} \\[-0.1cm] & \textbf{Platform Aim Location} (lat, lon): $(29.1621^{\circ}, -103.3979^{\circ})$ \\[0.2cm]
\cline{2-2} \\[-0.1cm] & 
\textbf{Platform Position} (lat, lon): $(29.2851^{\circ}, -103.2479^{\circ})$ \\[0.2cm]
\cline{2-2} \\[-0.1cm] & 
\textbf{Folder}: \textbf{/CLUTTER/num54\_lat29.2851\_lon-103.2479/} \\[0.2cm]
\cline{2-2} \\[-0.1cm] &
\parbox[t][3.0cm][t]{0.62\textwidth}{\textbf{Description}: Scenario $q_i = 83 \ (i = 54)$ in RASPNet is in Big Bend National Park in Western Texas. The radar platform is located at latitude $29.2851^{\circ}$ N and longitude $103.2479^{\circ}$ W. It aims towards a location at latitude $29.1621^{\circ}$ N and longitude $103.3979^{\circ}$ W. The radar is positioned above a mountain summit, and the aim location has high terrain variability.} \\
\hline
\end{tabular}
}
\end{table}

\paragraph{\underline{Scenario $q_i = 84$}: Lake Tahoe, CA ($i = 9$)} \vphantom{ }
\begin{table}[h!]
\centering
\setlength{\fboxsep}{0pt}
\setlength{\fboxrule}{2pt}
\fbox{%
\begin{tabular}{c|>{\arraybackslash}m{0.615\textwidth}}
\hline
\multirow{3}{*}{\raisebox{-\totalheight}{\includegraphics[width=0.5\textwidth,height=5.3cm,keepaspectratio]{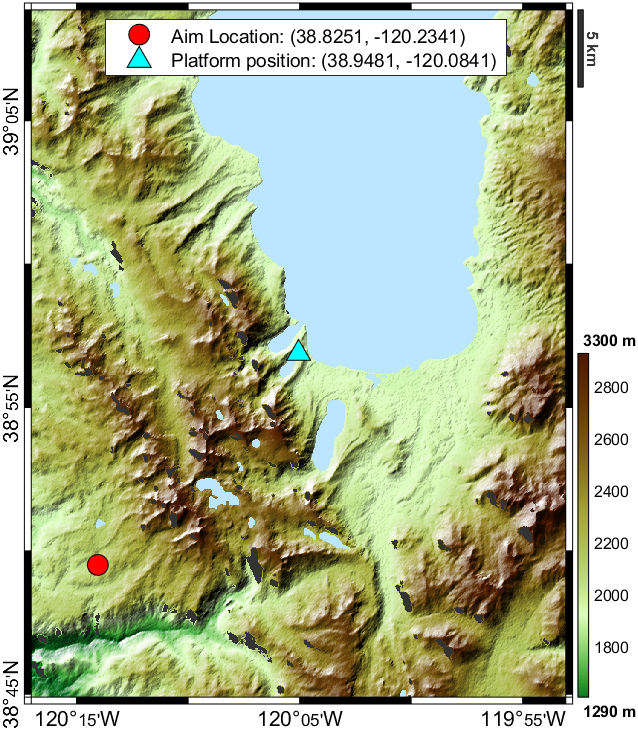}}} \\[-0.1cm] & \textbf{Platform Aim Location} (lat, lon): $(38.8251^{\circ}, -120.2341^{\circ})$ \\[0.2cm]
\cline{2-2} \\[-0.1cm] & 
\textbf{Platform Position} (lat, lon): $(38.9481^{\circ}, -120.0841^{\circ})$ \\[0.2cm]
\cline{2-2} \\[-0.1cm] & 
\textbf{Folder}: \textbf{/CLUTTER/num9\_lat38.9481\_lon-120.0841/} \\[0.2cm]
\cline{2-2} \\[-0.1cm] &
\parbox[t][3.0cm][t]{0.62\textwidth}{\textbf{Description}: Scenario $q_i = 84 \ (i = 9)$ in RASPNet is on the Lake Tahoe coastline in California. The radar platform is located at latitude $38.9481^{\circ}$ N and longitude $120.0841^{\circ}$ W. It aims towards a location at latitude $38.8251^{\circ}$ N and longitude $120.2341^{\circ}$ W. The radar is behind a mountain ridge, where the aim location has high terrain variability and is obscured.} \\
\hline
\end{tabular}
}
\end{table}

\paragraph{\underline{Scenario $q_i = 85$}: Mount Shasta, CA ($i = 4$)} \vphantom{ }
\begin{table}[h!]
\centering
\setlength{\fboxsep}{0pt}
\setlength{\fboxrule}{2pt}
\fbox{%
\begin{tabular}{c|>{\arraybackslash}m{0.615\textwidth}}
\hline
\multirow{3}{*}{\raisebox{-\totalheight}{\includegraphics[width=0.5\textwidth,height=5.3cm,keepaspectratio]{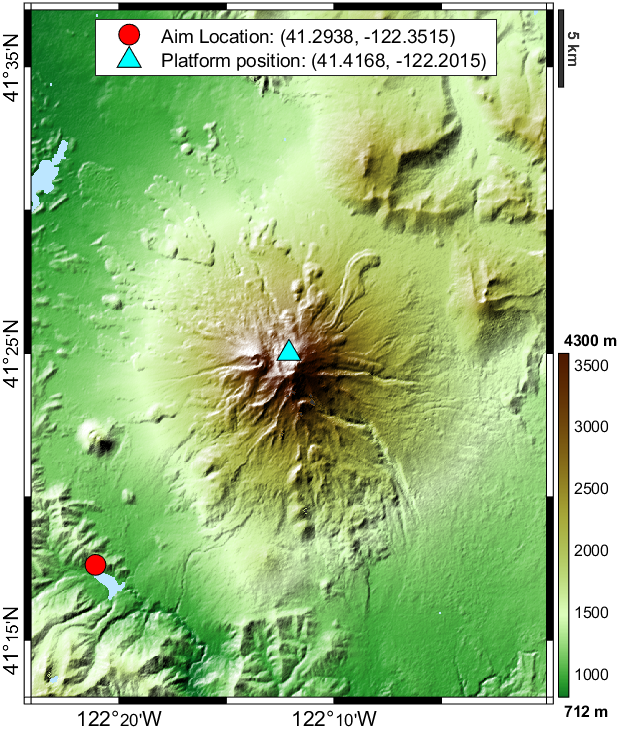}}} \\[-0.1cm] & \textbf{Platform Aim Location} (lat, lon): $(41.2938^{\circ}, -122.3515^{\circ})$ \\[0.2cm]
\cline{2-2} \\[-0.1cm] & 
\textbf{Platform Position} (lat, lon): $(41.4168^{\circ}, -122.2015^{\circ})$ \\[0.2cm]
\cline{2-2} \\[-0.1cm] & 
\textbf{Folder}: \textbf{/CLUTTER/num4\_lat41.4168\_lon-122.2015/} \\[0.2cm]
\cline{2-2} \\[-0.1cm] &
\parbox[t][3.0cm][t]{0.62\textwidth}{\textbf{Description}: Scenario $q_i = 85 \ (i = 4)$ in RASPNet is on Mount Shasta in Northern California. The radar platform is positioned at latitude $41.4168^{\circ}$ N and longitude $122.2015^{\circ}$ W. It aims towards a location at latitude $41.2938^{\circ}$ N and longitude $122.3515^{\circ}$ W. The radar is positioned above a mountain summit, and the aim location has high terrain variability.} \\
\hline
\end{tabular}
}
\end{table}

\paragraph{\underline{Scenario $q_i = 86$}: Saddleback Mountain, ME ($i = 99$)} \vphantom{ }
\begin{table}[h!]
\centering
\setlength{\fboxsep}{0pt}
\setlength{\fboxrule}{2pt}
\fbox{%
\begin{tabular}{c|>{\arraybackslash}m{0.615\textwidth}}
\hline
\multirow{3}{*}{\raisebox{-\totalheight}{\includegraphics[width=0.5\textwidth,height=5.3cm,keepaspectratio]{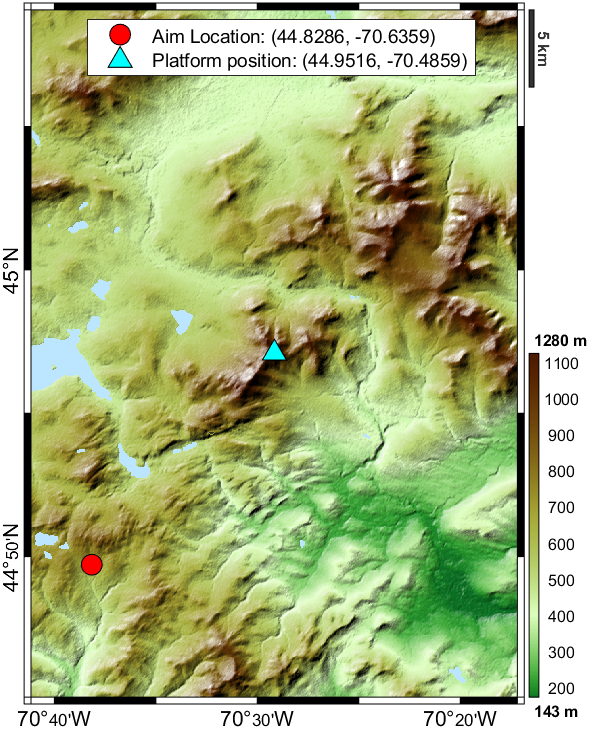}}} \\[-0.1cm] & \textbf{Platform Aim Location} (lat, lon): $(44.8286^{\circ}, -70.6359^{\circ})$ \\[0.2cm]
\cline{2-2} \\[-0.1cm] & 
\textbf{Platform Position} (lat, lon): $(44.9516^{\circ}, -70.4859^{\circ})$ \\[0.2cm]
\cline{2-2} \\[-0.1cm] & 
\textbf{Folder}: \textbf{/CLUTTER/num99\_lat44.9516\_lon-70.4859/} \\[0.2cm]
\cline{2-2} \\[-0.1cm] &
\parbox[t][3.0cm][t]{0.62\textwidth}{\textbf{Description}: Scenario $q_i = 86 \ (i = 99)$ in RASPNet is on Saddleback Mountain in Western Maine. The radar platform is located at latitude $44.9516^{\circ}$ N and longitude $70.4859^{\circ}$ W. It aims towards a location at latitude $44.8286^{\circ}$ N and longitude $70.6359^{\circ}$ W. The radar is positioned above a mountain ridge, and the aim location has high terrain variability.} \\
\hline
\end{tabular}
}
\end{table}

\paragraph{\underline{Scenario $q_i = 87$}: Mission Mountains, MT ($i = 28$)} \vphantom{ }
\begin{table}[h!]
\centering
\setlength{\fboxsep}{0pt}
\setlength{\fboxrule}{2pt}
\fbox{%
\begin{tabular}{c|>{\arraybackslash}m{0.615\textwidth}}
\hline
\multirow{3}{*}{\raisebox{-\totalheight}{\includegraphics[width=0.5\textwidth,height=5.3cm,keepaspectratio]{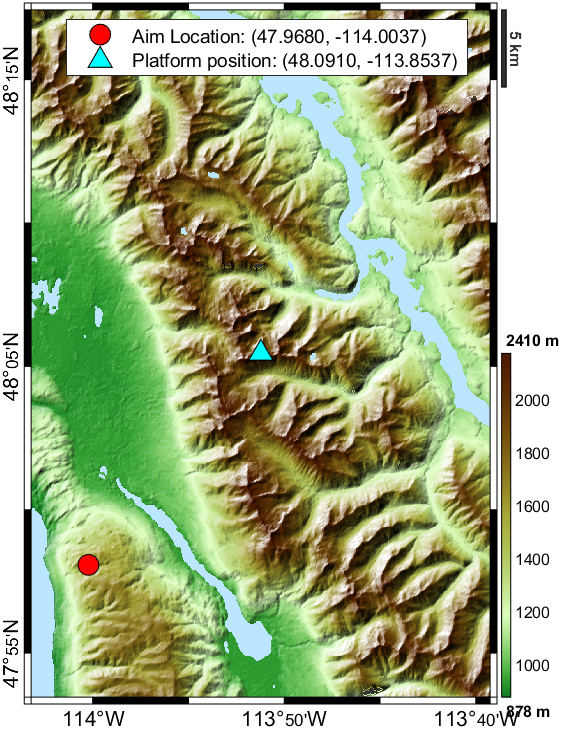}}} \\[-0.1cm] & \textbf{Platform Aim Location} (lat, lon): $(47.9680^{\circ}, -114.0037^{\circ})$ \\[0.2cm]
\cline{2-2} \\[-0.1cm] & 
\textbf{Platform Position} (lat, lon): $(48.0910^{\circ}, -113.8537^{\circ})$ \\[0.2cm]
\cline{2-2} \\[-0.1cm] & 
\textbf{Folder}: \textbf{/CLUTTER/num28\_lat48.091\_lon-113.8537/} \\[0.2cm]
\cline{2-2} \\[-0.1cm] &
\parbox[t][3.0cm][t]{0.62\textwidth}{\textbf{Description}: Scenario $q_i = 87 \ (i = 28)$ in RASPNet is in the Mission Mountains of Western Montana. The radar platform is positioned at latitude $48.0910^{\circ}$ N and longitude $113.8537^{\circ}$ W. It aims towards a location at latitude $47.9680^{\circ}$ N and longitude $114.0037^{\circ}$ W. The radar is positioned above a mountain ridge, and the aim location has high terrain variability.} \\
\hline
\end{tabular}
}
\end{table}

\paragraph{\underline{Scenario $q_i = 88$}: Ochoco Mountains, OR ($i = 10$)} \vphantom{ }
\begin{table}[h!]
\centering
\setlength{\fboxsep}{0pt}
\setlength{\fboxrule}{2pt}
\fbox{%
\begin{tabular}{c|>{\arraybackslash}m{0.615\textwidth}}
\hline
\multirow{3}{*}{\raisebox{-\totalheight}{\includegraphics[width=0.5\textwidth,height=5.3cm,keepaspectratio]{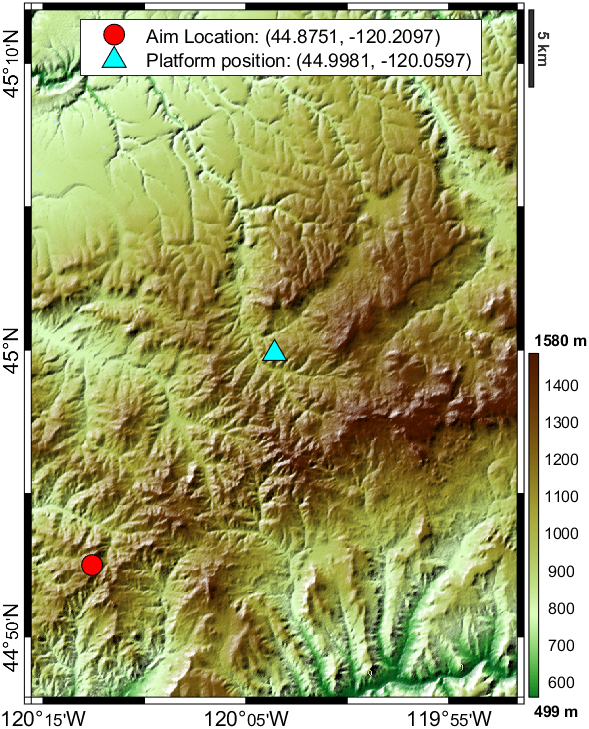}}} \\[-0.1cm] & \textbf{Platform Aim Location} (lat, lon): $(44.8751^{\circ}, -120.2097^{\circ})$ \\[0.2cm]
\cline{2-2} \\[-0.1cm] & 
\textbf{Platform Position} (lat, lon): $(44.9981^{\circ}, -120.0597^{\circ})$ \\[0.2cm]
\cline{2-2} \\[-0.1cm] & 
\textbf{Folder}: \textbf{/CLUTTER/num10\_lat44.9981\_lon-120.0597/} \\[0.2cm]
\cline{2-2} \\[-0.1cm] &
\parbox[t][3.0cm][t]{0.62\textwidth}{\textbf{Description}: Scenario $q_i = 88 \ (i = 10)$ in RASPNet is in the Ochoco Mountains in Central Oregon. The radar platform is located at latitude $44.9981^{\circ}$ N and longitude $120.0597^{\circ}$ W. It aims towards a location at latitude $44.8751^{\circ}$ N and longitude $120.2097^{\circ}$ W. The radar is positioned near a mountain ridge, and the aim location has high terrain variability.} \\
\hline
\end{tabular}
}
\end{table}

\paragraph{\underline{Scenario $q_i = 89$}: Canyonlands National Park, UT ($i = 41$)} \vphantom{ }
\begin{table}[h!]
\centering
\setlength{\fboxsep}{0pt}
\setlength{\fboxrule}{2pt}
\fbox{%
\begin{tabular}{c|>{\arraybackslash}m{0.615\textwidth}}
\hline
\multirow{3}{*}{\raisebox{-\totalheight}{\includegraphics[width=0.5\textwidth,height=5.3cm,keepaspectratio]{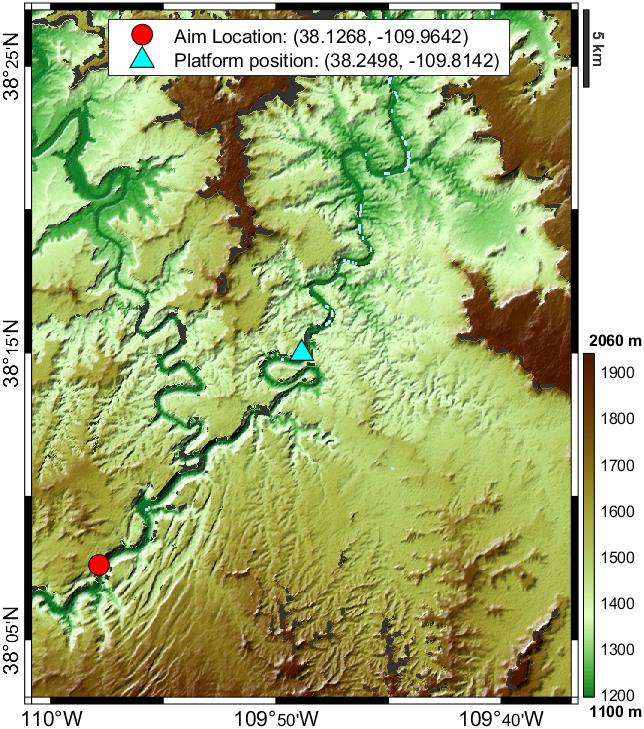}}} \\[-0.1cm] & \textbf{Platform Aim Location} (lat, lon): $(38.1268^{\circ}, -109.9642^{\circ})$ \\[0.2cm]
\cline{2-2} \\[-0.1cm] & 
\textbf{Platform Position} (lat, lon): $(38.2498^{\circ}, -109.8142^{\circ})$ \\[0.2cm]
\cline{2-2} \\[-0.1cm] & 
\textbf{Folder}: \textbf{/CLUTTER/num41\_lat38.2498\_lon-109.8142/} \\[0.2cm]
\cline{2-2} \\[-0.1cm] &
\parbox[t][3.0cm][t]{0.62\textwidth}{\textbf{Description}: Scenario $q_i = 89 \ (i = 41)$ in RASPNet is in Canyonlands National Park in Eastern Utah. The radar platform is located at latitude $38.2498^{\circ}$ N and longitude $109.8142^{\circ}$ W. It aims towards a location at latitude $38.1268^{\circ}$ N and longitude $109.9642^{\circ}$ W. This region is characterized by rugged canyonlands, and the aim location has high terrain variability.} \\
\hline
\end{tabular}
}
\end{table}

\paragraph{\underline{Scenario $q_i = 90$}: Lake Koocanusa, MT ($i = 25$)} \vphantom{ }
\begin{table}[h!]
\centering
\setlength{\fboxsep}{0pt}
\setlength{\fboxrule}{2pt}
\fbox{%
\begin{tabular}{c|>{\arraybackslash}m{0.615\textwidth}}
\hline
\multirow{3}{*}{\raisebox{-\totalheight}{\includegraphics[width=0.5\textwidth,height=5.3cm,keepaspectratio]{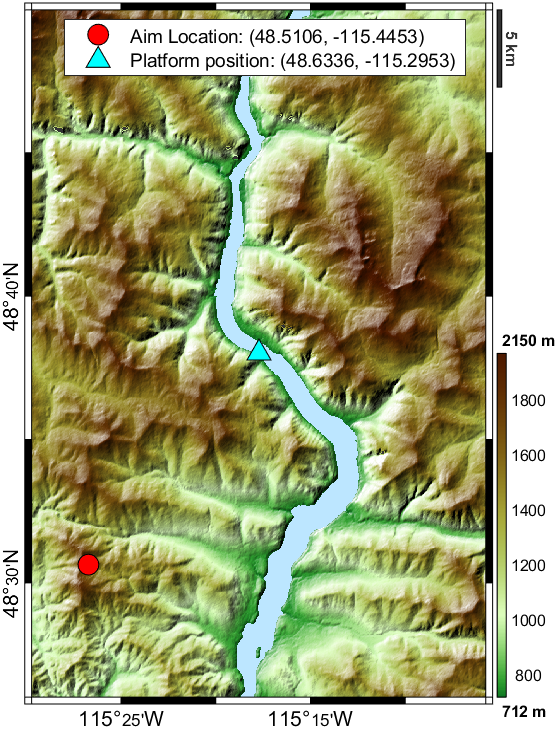}}} \\[-0.1cm] & \textbf{Platform Aim Location} (lat, lon): $(48.5106^{\circ}, -115.4453^{\circ})$ \\[0.2cm]
\cline{2-2} \\[-0.1cm] & 
\textbf{Platform Position} (lat, lon): $(48.6336^{\circ}, -115.2953^{\circ})$ \\[0.2cm]
\cline{2-2} \\[-0.1cm] & 
\textbf{Folder}: \textbf{/CLUTTER/num25\_lat48.6336\_lon-115.2953/} \\[0.2cm]
\cline{2-2} \\[-0.1cm] &
\parbox[t][3.0cm][t]{0.62\textwidth}{\textbf{Description}: Scenario $q_i = 90 \ (i = 25)$ in RASPNet is on Lake Koocanusa in Western Montana. The radar platform is positioned at latitude $48.6336^{\circ}$ N and longitude $115.2953^{\circ}$ W. It aims towards a location at latitude $48.5106^{\circ}$ N and longitude $115.4453^{\circ}$ W. The radar is positioned on the lake coastline, and the aim location is a mountain ridge.} \\
\hline
\end{tabular}
}
\end{table}

\paragraph{\underline{Scenario $q_i = 91$}: Yosemite Valley, CA ($i = 14$)} \vphantom{ }
\begin{table}[h!]
\centering
\setlength{\fboxsep}{0pt}
\setlength{\fboxrule}{2pt}
\fbox{%
\begin{tabular}{c|>{\arraybackslash}m{0.615\textwidth}}
\hline
\multirow{3}{*}{\raisebox{-\totalheight}{\includegraphics[width=0.5\textwidth,height=5.3cm,keepaspectratio]{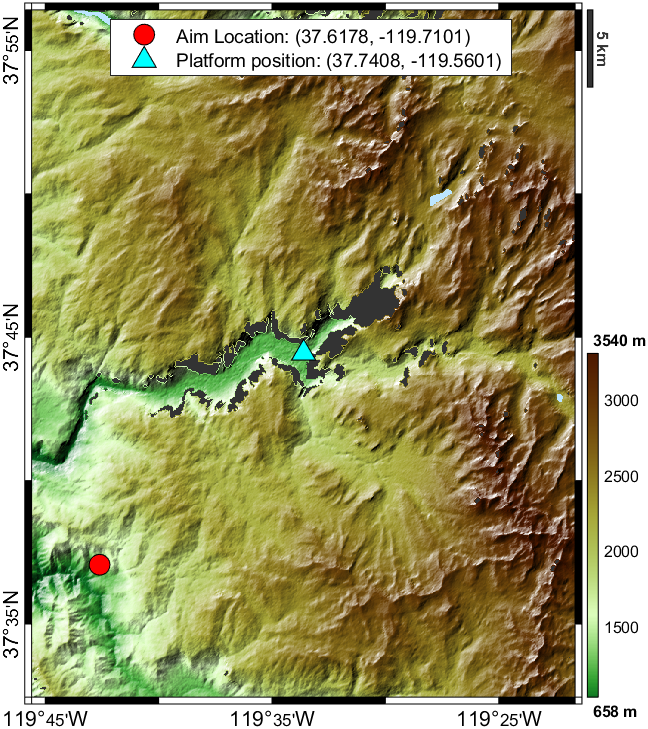}}} \\[-0.1cm] & \textbf{Platform Aim Location} (lat, lon): $(37.6178^{\circ}, -119.7101^{\circ})$ \\[0.2cm]
\cline{2-2} \\[-0.1cm] & 
\textbf{Platform Position} (lat, lon): $(37.7408^{\circ}, -119.5601^{\circ})$ \\[0.2cm]
\cline{2-2} \\[-0.1cm] & 
\textbf{Folder}: \textbf{/CLUTTER/num14\_lat37.7408\_lon-119.5601/} \\[0.2cm]
\cline{2-2} \\[-0.1cm] &
\parbox[t][3.0cm][t]{0.62\textwidth}{\textbf{Description}: Scenario $q_i = 91 \ (i = 14)$ in RASPNet is in Yosemite National Park in California. The radar platform is positioned at latitude $37.7408^{\circ}$ N and longitude $119.5601^{\circ}$ W. It aims towards a location at latitude $37.6178^{\circ}$ N and longitude $119.7101^{\circ}$ W. The radar platform is in a steep valley, and the aim location has high terrain variability and is obscured.} \\
\hline
\end{tabular}
}
\end{table}

\paragraph{\underline{Scenario $q_i = 92$}: Grand Canyon, AZ ($i = 35$)} \vphantom{ }
\begin{table}[h!]
\centering
\setlength{\fboxsep}{0pt}
\setlength{\fboxrule}{2pt}
\fbox{%
\begin{tabular}{c|>{\arraybackslash}m{0.615\textwidth}}
\hline
\multirow{3}{*}{\raisebox{-\totalheight}{\includegraphics[width=0.5\textwidth,height=5.3cm,keepaspectratio]{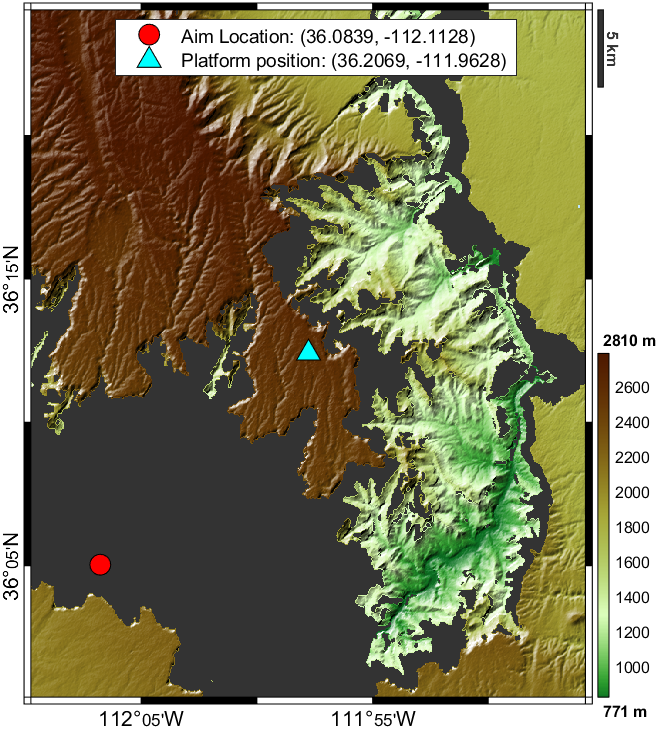}}} \\[-0.1cm] & \textbf{Platform Aim Location} (lat, lon): $(36.0839^{\circ}, -112.1128^{\circ})$ \\[0.2cm]
\cline{2-2} \\[-0.1cm] & 
\textbf{Platform Position} (lat, lon): $(36.2069^{\circ}, -111.9628^{\circ})$ \\[0.2cm]
\cline{2-2} \\[-0.1cm] & 
\textbf{Folder}: \textbf{/CLUTTER/num35\_lat36.2069\_lon-111.9628/} \\[0.2cm]
\cline{2-2} \\[-0.1cm] &
\parbox[t][3.0cm][t]{0.62\textwidth}{\textbf{Description}: Scenario $q_i = 92 \ (i = 35)$ in RASPNet is in Grand Canyon National Park in Arizona. The radar platform is located at latitude $36.2069^{\circ}$ N and longitude $111.9628^{\circ}$ W. It aims towards a location at latitude $36.0839^{\circ}$ N and longitude $112.1128^{\circ}$ W. This region is characterized by rugged canyonlands, and the aim location has high terrain variability.} \\
\hline
\end{tabular}
}
\end{table}

\paragraph{\underline{Scenario $q_i = 93$}: Adirondack Mountains, NY ($i = 92$)} \vphantom{ }
\begin{table}[h!]
\centering
\setlength{\fboxsep}{0pt}
\setlength{\fboxrule}{2pt}
\fbox{%
\begin{tabular}{c|>{\arraybackslash}m{0.615\textwidth}}
\hline
\multirow{3}{*}{\raisebox{-\totalheight}{\includegraphics[width=0.5\textwidth,height=5.3cm,keepaspectratio]{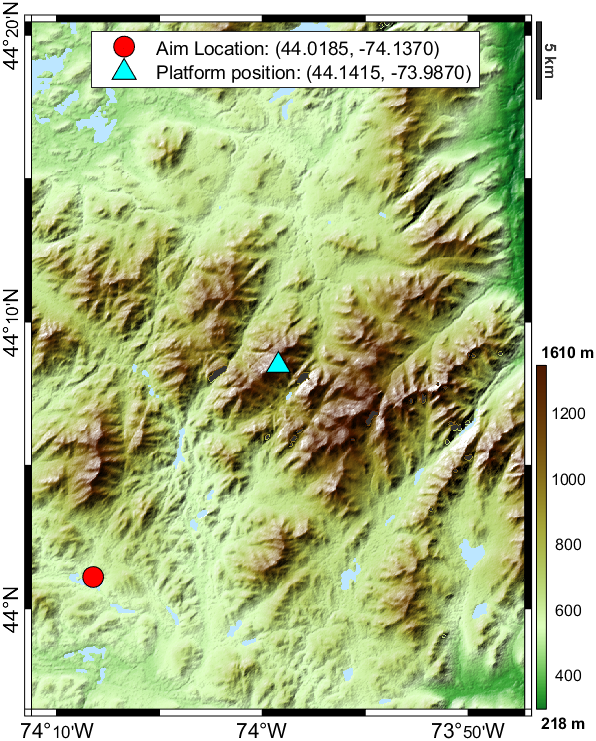}}} \\[-0.1cm] & \textbf{Platform Aim Location} (lat, lon): $(44.0185^{\circ}, -74.1370^{\circ})$ \\[0.2cm]
\cline{2-2} \\[-0.1cm] & 
\textbf{Platform Position} (lat, lon): $(44.1415^{\circ}, -73.9870^{\circ})$ \\[0.2cm]
\cline{2-2} \\[-0.1cm] & 
\textbf{Folder}: \textbf{/CLUTTER/num92\_lat44.1415\_lon-73.987/} \\[0.2cm]
\cline{2-2} \\[-0.1cm] &
\parbox[t][3.0cm][t]{0.62\textwidth}{\textbf{Description}: Scenario $q_i = 93 \ (i = 92)$ in RASPNet is in the Adirondack Mountains of New York. The radar platform is located at latitude $44.1415^{\circ}$ N and longitude $73.9870^{\circ}$ W. It aims towards a location at latitude $44.0185^{\circ}$ N and longitude $74.1370^{\circ}$ W. The radar is positioned above a mountain summit, and the aim location has high terrain variability.} \\
\hline
\end{tabular}
}
\end{table}

\paragraph{\underline{Scenario $q_i = 94$}: Presidential Range, NH ($i = 97$)} \vphantom{ }
\begin{table}[h!]
\centering
\setlength{\fboxsep}{0pt}
\setlength{\fboxrule}{2pt}
\fbox{%
\begin{tabular}{c|>{\arraybackslash}m{0.615\textwidth}}
\hline
\multirow{3}{*}{\raisebox{-\totalheight}{\includegraphics[width=0.5\textwidth,height=5.3cm,keepaspectratio]{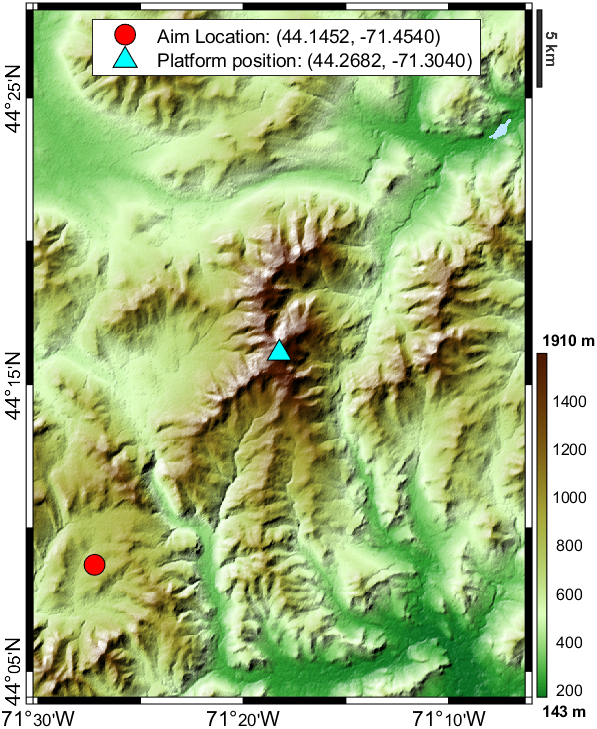}}} \\[-0.1cm] & \textbf{Platform Aim Location} (lat, lon): $(44.1452^{\circ}, -71.4540^{\circ})$ \\[0.2cm]
\cline{2-2} \\[-0.1cm] & 
\textbf{Platform Position} (lat, lon): $(44.2682^{\circ}, -71.3040^{\circ})$ \\[0.2cm]
\cline{2-2} \\[-0.1cm] & 
\textbf{Folder}: \textbf{/CLUTTER/num97\_lat44.2682\_lon-71.304/} \\[0.2cm]
\cline{2-2} \\[-0.1cm] &
\parbox[t][3.0cm][t]{0.62\textwidth}{\textbf{Description}: Scenario $q_i = 94 \ (i = 97)$ in RASPNet is in New Hampshire's Presidential Range. The radar platform is positioned at latitude $44.2682^{\circ}$ N and longitude $71.3040^{\circ}$ W. It aims towards a location at latitude $44.1452^{\circ}$ N and longitude $71.4540^{\circ}$ W. The radar is positioned above a mountain summit, and the aim location has high terrain variability.} \\
\hline
\end{tabular}
}
\end{table}

\paragraph{\underline{Scenario $q_i = 95$}: Idaho Panhandle, ID ($i = 22$)} \vphantom{ }
\begin{table}[h!]
\centering
\setlength{\fboxsep}{0pt}
\setlength{\fboxrule}{2pt}
\fbox{%
\begin{tabular}{c|>{\arraybackslash}m{0.615\textwidth}}
\hline
\multirow{3}{*}{\raisebox{-\totalheight}{\includegraphics[width=0.5\textwidth,height=5.3cm,keepaspectratio]{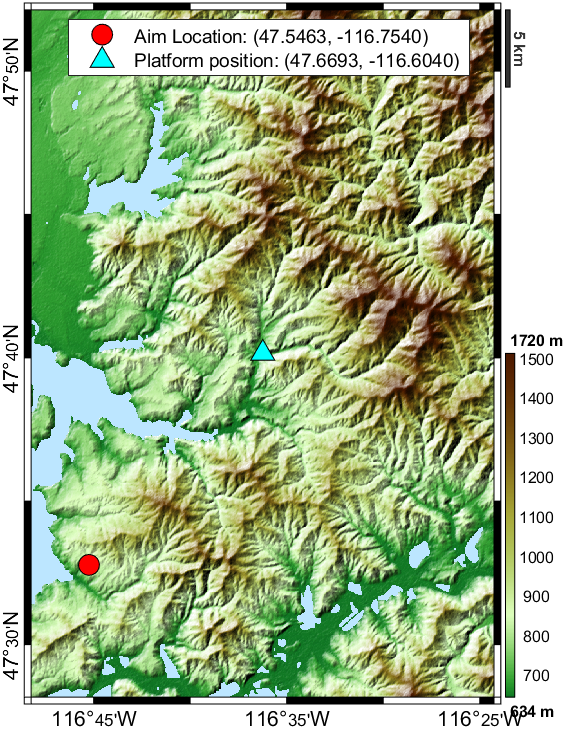}}} \\[-0.1cm] & \textbf{Platform Aim Location} (lat, lon): $(47.5463^{\circ}, -116.7540^{\circ})$ \\[0.2cm]
\cline{2-2} \\[-0.1cm] & 
\textbf{Platform Position} (lat, lon): $(47.6693^{\circ}, -116.6040^{\circ})$ \\[0.2cm]
\cline{2-2} \\[-0.1cm] & 
\textbf{Folder}: \textbf{/CLUTTER/num22\_lat47.6693\_lon-116.604/} \\[0.2cm]
\cline{2-2} \\[-0.1cm] &
\parbox[t][3.0cm][t]{0.62\textwidth}{\textbf{Description}: Scenario $q_i = 95 \ (i = 22)$ in RASPNet is in the mountains of the Idaho Panhandle. The radar platform is positioned at latitude $47.6693^{\circ}$ N and longitude $116.6040^{\circ}$ W. It aims towards a location at latitude $47.5463^{\circ}$ N and longitude $116.7540^{\circ}$ W. This region is characterized by rugged terrain, with highly nonuniform topography.} \\
\hline
\end{tabular}
}
\end{table}

\paragraph{\underline{Scenario $q_i = 96$}: Sawtooth Range, ID ($i = 27$)} \vphantom{ }
\begin{table}[h!]
\centering
\setlength{\fboxsep}{0pt}
\setlength{\fboxrule}{2pt}
\fbox{%
\begin{tabular}{c|>{\arraybackslash}m{0.615\textwidth}}
\hline
\multirow{3}{*}{\raisebox{-\totalheight}{\includegraphics[width=0.5\textwidth,height=5.3cm,keepaspectratio]{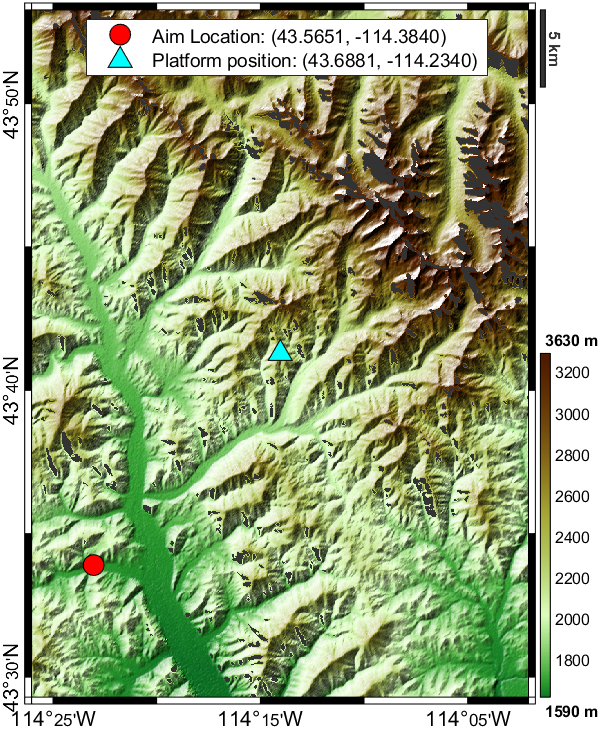}}} \\[-0.1cm] & \textbf{Platform Aim Location} (lat, lon): $(43.5651^{\circ}, -114.3840^{\circ})$ \\[0.2cm]
\cline{2-2} \\[-0.1cm] & 
\textbf{Platform Position} (lat, lon): $(43.6881^{\circ}, -114.2340^{\circ})$ \\[0.2cm]
\cline{2-2} \\[-0.1cm] & 
\textbf{Folder}: \textbf{/CLUTTER/num27\_lat43.6881\_lon-114.234/} \\[0.2cm]
\cline{2-2} \\[-0.1cm] &
\parbox[t][3.0cm][t]{0.62\textwidth}{\textbf{Description}: Scenario $q_i = 96 \ (i = 27)$ in RASPNet is in the Sawtooth Range of Central Idaho. The radar platform is positioned at latitude $43.6881^{\circ}$ N and longitude $114.2340^{\circ}$ W. It aims towards a location at latitude $43.5651^{\circ}$ N and longitude $114.3840^{\circ}$ W. The radar is positioned near a mountain ridge, and the aim location has high terrain variability.} \\
\hline
\end{tabular}
}
\end{table}

\paragraph{\underline{Scenario $q_i = 97$}: Mount Baker, WA ($i = 7$)} \vphantom{ }
\begin{table}[h!]
\centering
\setlength{\fboxsep}{0pt}
\setlength{\fboxrule}{2pt}
\fbox{%
\begin{tabular}{c|>{\arraybackslash}m{0.615\textwidth}}
\hline
\multirow{3}{*}{\raisebox{-\totalheight}{\includegraphics[width=0.5\textwidth,height=5.3cm,keepaspectratio]{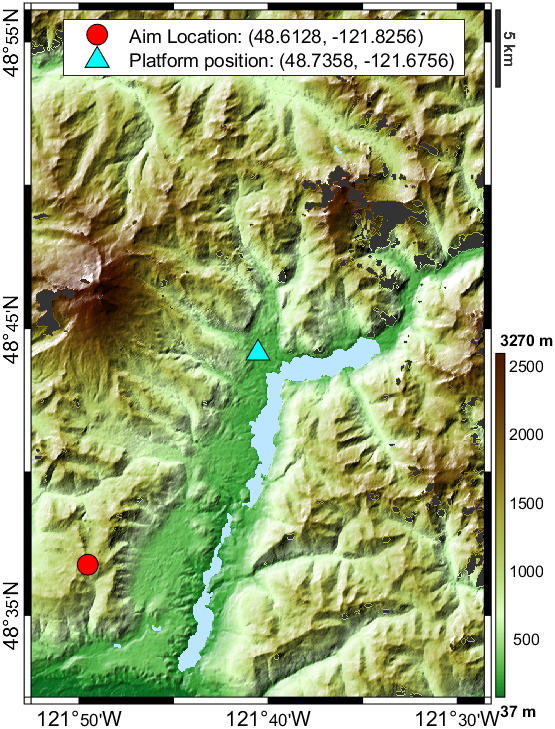}}} \\[-0.1cm] & \textbf{Platform Aim Location} (lat, lon): $(48.6128^{\circ}, -121.8256^{\circ})$ \\[0.2cm]
\cline{2-2} \\[-0.1cm] & 
\textbf{Platform Position} (lat, lon): $(48.7358^{\circ}, -121.6756^{\circ})$ \\[0.2cm]
\cline{2-2} \\[-0.1cm] & 
\textbf{Folder}: \textbf{/CLUTTER/num7\_lat48.7358\_lon-121.6756/} \\[0.2cm]
\cline{2-2} \\[-0.1cm] &
\parbox[t][3.0cm][t]{0.62\textwidth}{\textbf{Description}: Scenario $q_i = 97 \ (i = 7)$ in RASPNet is near Mount Baker in Western Washington. The radar platform is positioned at latitude $48.7358^{\circ}$ N and longitude $121.6756^{\circ}$ W. It aims towards a location at latitude $48.6128^{\circ}$ N and longitude $121.8256^{\circ}$ W. The radar is located above a mountain valley, and the aim location has high terrain variability.} \\
\hline
\end{tabular}
}
\end{table}

\paragraph{\underline{Scenario $q_i = 98$}: Wallowa Mountains, OR ($i = 19$)} \vphantom{ }
\begin{table}[h!]
\centering
\setlength{\fboxsep}{0pt}
\setlength{\fboxrule}{2pt}
\fbox{%
\begin{tabular}{c|>{\arraybackslash}m{0.615\textwidth}}
\hline
\multirow{3}{*}{\raisebox{-\totalheight}{\includegraphics[width=0.5\textwidth,height=5.3cm,keepaspectratio]{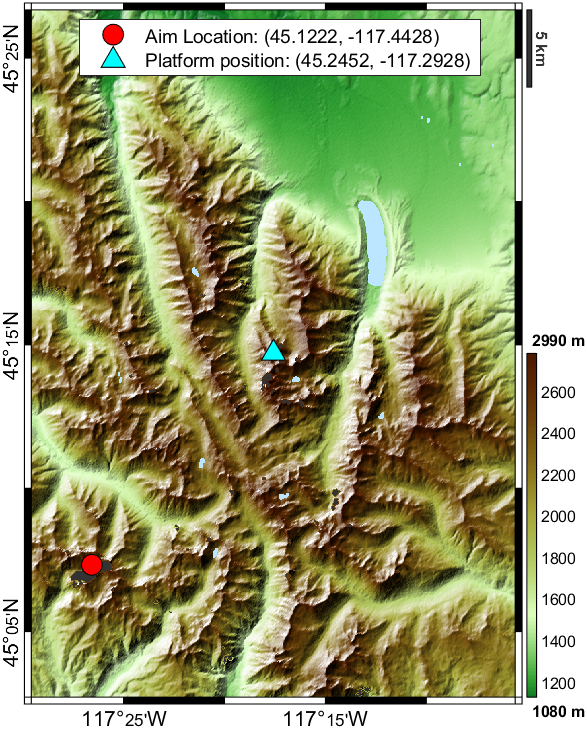}}} \\[-0.1cm] & \textbf{Platform Aim Location} (lat, lon): $(45.1222^{\circ}, -117.4428^{\circ})$ \\[0.2cm]
\cline{2-2} \\[-0.1cm] & 
\textbf{Platform Position} (lat, lon): $(45.2452^{\circ}, -117.2928^{\circ})$ \\[0.2cm]
\cline{2-2} \\[-0.1cm] & 
\textbf{Folder}: \textbf{/CLUTTER/num19\_lat45.2452\_lon-117.2928/} \\[0.2cm]
\cline{2-2} \\[-0.1cm] &
\parbox[t][3.0cm][t]{0.62\textwidth}{\textbf{Description}: Scenario $q_i = 98 \ (i = 19)$ in RASPNet is in Eastern Oregon's Wallowa Mountains. The radar platform is located at latitude $45.2452^{\circ}$ N and longitude $117.2928^{\circ}$ W. It aims towards a location at latitude $45.1222^{\circ}$ N and longitude $117.4428^{\circ}$ W. The radar is located above a mountain summit, and the aim location has high terrain variability.} \\
\hline
\end{tabular}
}
\end{table}

\paragraph{\underline{Scenario $q_i = 99$}: Zion National Park, UT ($i = 32$)} \vphantom{ }
\begin{table}[h!]
\centering
\setlength{\fboxsep}{0pt}
\setlength{\fboxrule}{2pt}
\fbox{%
\begin{tabular}{c|>{\arraybackslash}m{0.615\textwidth}}
\hline
\multirow{3}{*}{\raisebox{-\totalheight}{\includegraphics[width=0.5\textwidth,height=5.3cm,keepaspectratio]{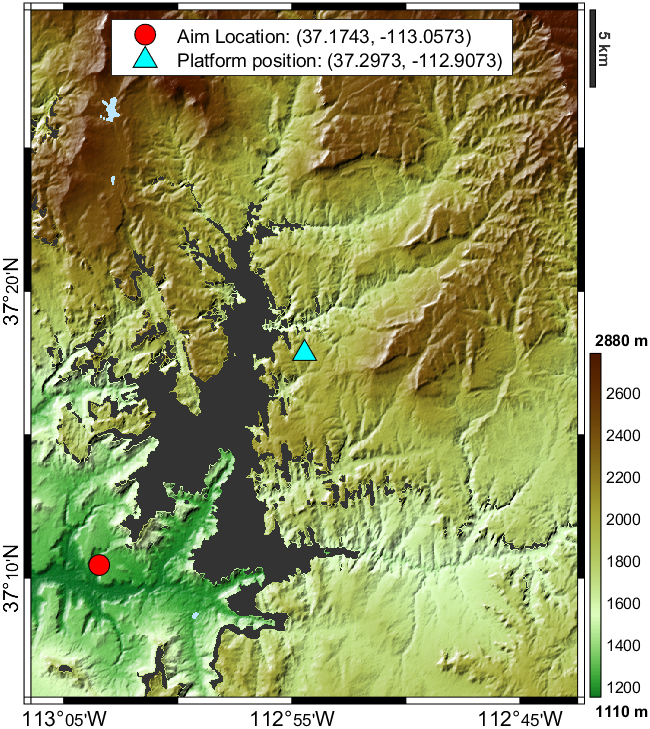}}} \\[-0.1cm] & \textbf{Platform Aim Location} (lat, lon): $(37.1743^{\circ}, -113.0573^{\circ})$ \\[0.2cm]
\cline{2-2} \\[-0.1cm] & 
\textbf{Platform Position} (lat, lon): $(37.2973^{\circ}, -112.9073^{\circ})$ \\[0.2cm]
\cline{2-2} \\[-0.1cm] & 
\textbf{Folder}: \textbf{/CLUTTER/num32\_lat37.2973\_lon-112.9073/} \\[0.2cm]
\cline{2-2} \\[-0.1cm] &
\parbox[t][3.0cm][t]{0.62\textwidth}{\textbf{Description}: Scenario $q_i = 99 \ (i = 32)$ in RASPNet is in Zion National Park in Southern Utah. The radar platform is located at latitude $37.2973^{\circ}$ N and longitude $112.9073^{\circ}$ W. It aims towards a location at latitude $37.1743^{\circ}$ N and longitude $113.0573^{\circ}$ W. This region is characterized by rugged canyonlands, and the aim location has high terrain variability.} \\
\hline
\end{tabular}
}
\end{table}

\paragraph{\underline{Scenario $q_i = 100$}: Mount Rainier, WA ($i = 5$)} \vphantom{ }
\begin{table}[h!]
\centering
\setlength{\fboxsep}{0pt}
\setlength{\fboxrule}{2pt}
\fbox{%
\begin{tabular}{c|>{\arraybackslash}m{0.615\textwidth}}
\hline
\multirow{3}{*}{\raisebox{-\totalheight}{\includegraphics[width=0.5\textwidth,height=5.3cm,keepaspectratio]{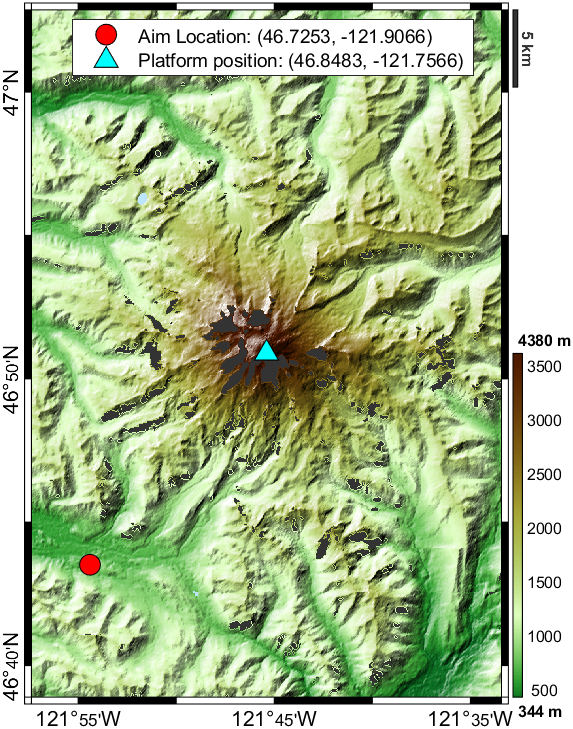}}} \\[-0.1cm] & \textbf{Platform Aim Location} (lat, lon): $(46.7253^{\circ}, -121.9066^{\circ})$ \\[0.2cm]
\cline{2-2} \\[-0.1cm] & 
\textbf{Platform Position} (lat, lon): $(46.8483^{\circ}, -121.7566^{\circ})$ \\[0.2cm]
\cline{2-2} \\[-0.1cm] & 
\textbf{Folder}: \textbf{/CLUTTER/num5\_lat46.8483\_lon-121.7566/} \\[0.2cm]
\cline{2-2} \\[-0.1cm] &
\parbox[t][3.0cm][t]{0.62\textwidth}{\textbf{Description}: Scenario $q_i = 100 \ (i = 5)$ in RASPNet is on Mount Rainier in Western Washington. The radar platform is located at latitude $46.8483^{\circ}$ N and longitude $121.7566^{\circ}$ W. It aims towards a location at latitude $46.7253^{\circ}$ N and longitude $121.9066^{\circ}$ W. The radar is located above a mountain summit, and the aim location has high terrain variability.} \\
\hline
\end{tabular}
}
\end{table}

\end{document}

\end{document}